\documentclass[]{llncs}
\usepackage{graphicx}
\usepackage{comment}
\usepackage{amsmath,amssymb} 
\usepackage{color}

\usepackage[width=122mm,left=12mm,paperwidth=146mm,height=193mm,top=12mm,paperheight=217mm]{geometry}

\usepackage{times}
\usepackage{epsfig}

\usepackage{url}            
\usepackage{booktabs}       
\usepackage{subfigure}
\usepackage{booktabs} 
\usepackage{amsfonts}       
\usepackage{nicefrac}       
\usepackage{microtype}      

\usepackage{array}
\usepackage{enumitem}

\usepackage[pagebackref=true,breaklinks=true,letterpaper=true,colorlinks,bookmarks=false]{hyperref}

\newtheorem{defn}{Definition}
\newcommand{\revise}[1]{#1}

\newcommand{\D}[1]{\frac{\partial \mathcal{L}}{\partial #1 }}
\newcommand{\hatD}[1]{\frac{\partial \widehat{\mathcal{L}}}{\partial #1 }}
\newcommand{\Dl}[1]{\frac{\partial \ell}{\partial #1}}

\newcommand{\x}[1]{\mathbf{x}_{#1}}
\newcommand{\h}[1]{\mathbf{h}_{#1}}
\newcommand{\s}[1]{\mathbf{s}_{#1}}
\newcommand{\z}[1]{\mathbf{z}_{#1}}
\newcommand{\y}[1]{\mathbf{y}_{#1}}
\newcommand{\W}[1]{\mathbf{W}_{#1}}

\newcommand{\hz}[1]{\hat{\mathbf{z}}_{#1}}
\newcommand{\hatx}[1]{\widehat{\mathbf{x}}_{#1}}
\newcommand{\hath}[1]{\widehat{\mathbf{h}}_{#1}}
\newcommand{\hats}[1]{\widehat{\mathbf{s}}_{#1}}
\newcommand{\hatW}[1]{\widehat{\mathbf{W}}_{#1}}

\newcommand{\Loss}{\mathcal{L}(\theta)}
\newcommand{\TLoss}{\widetilde{\mathcal{L}}(\theta,\mathbf{z})}

\def\eg{\emph{e.g.}}

\def\ie{\emph{i.e.}}

\def\wrt{w.r.t.}

\def\etal{\emph{et al.}}
\def\SM{Appendix}

\begin{document}
\def\ECCVSubNumber{3482}  


\title{Layer-wise Conditioning Analysis in Exploring the Learning Dynamics of DNNs}

\author{Lei Huang \quad Jie Qin \quad  Li Liu \quad Fan Zhu  \quad Ling Shao}


\institute{Inception Institute of Artificial Intelligence (IIAI), Abu Dhabi, UAE\\
\email{\{lei.huang, jie.qin, li.liu, fan.zhu,  ling.shao\}@inceptioniai.org}}
\maketitle

\begin{abstract}
	Conditioning analysis uncovers the landscape of an optimization objective by exploring the spectrum of its curvature matrix.
	This has been well explored theoretically for linear models.
	We extend this analysis to deep neural networks (DNNs) in order to investigate their learning dynamics.
	To this end,  we propose layer-wise conditioning analysis, which explores the optimization landscape with respect to each layer independently.
	Such an analysis is theoretically supported under mild assumptions that approximately hold in practice.
	Based on our analysis, we show that batch normalization (BN) can stabilize the training, but sometimes result in the false  impression of a local minimum, which has detrimental effects on the learning. Besides, we experimentally observe that BN can improve the layer-wise conditioning of the optimization problem.
	Finally, we find that the last linear layer of a very deep residual network displays ill-conditioned behavior. We solve this problem by only adding one BN layer before the last linear layer, which achieves improved performance over the original and pre-activation residual networks.
\end{abstract}

\section{Introduction}
\label{Sec-Intro}
Deep neural networks (DNNs) have been extensively used in various domains \cite{2015_nature_DeepLearning}. Their success depends heavily on the improvement of training techniques \cite{2006_science_Hinton,2015_ICML_Ioffe,2015_CVPR_He}, \eg, fine weight initialization  \cite{2006_science_Hinton,2010_AISTATS_Glorot,2013_CoRR_Saxe,2015_ICCV_He}, normalization of internal representations~\cite{2015_ICML_Ioffe,2018_ECCV_Wu}, and well-designed optimization methods \cite{2012_CoRR_Zeiler,2014_CoRR_Kingma}. It is believed that these techniques are well connected to the curvature of the loss \cite{2013_CoRR_Saxe,2018_NIPS_shibani,2018_arxiv_Kohler}. Analyzing this curvature is thus essential in determining various learning behaviors of DNNs.

In the interest of optimization, conditioning analysis uncovers the landscape of an optimization objective by exploring the spectrum of its curvature matrix. This has been well explored for linear models both in terms of regression \cite{1990_NIPS_LeCun} and classification \cite{2011_NIPS_Wiesler}, where the convergence condition of the optimization problem is controlled by the maximum eigenvalue of the curvature matrix \cite{1990_NIPS_LeCun,1998_NN_Yann}, and the learning time of the model is lower-bounded by its condition number~\cite{1990_NIPS_LeCun,1998_NN_Yann}. However, in the context of deep learning, the conditioning analysis suffers from several barriers: 1)  the model is over-parameterized and whether the direction with respect to small/zero eigenvalues contributes to the optimization progress is unclear~\cite{Sagun_CoRR_2017,2018_CoRR_vardan}; 2) the memory and computational costs are extremely high~\cite{Sagun_CoRR_2017,2019_ICML_Ghorbani}.

This paper aims to bridge the gap between the theoretical analyses developed by the optimization community and the empirical techniques used for training DNNs, in order to better understand the learning dynamics of DNNs. We propose a layer-wise conditioning analysis, where we analyze the optimization landscape with respect to each layer independently by exploring the spectra of their curvature matrices.
The motivation behind our layer-wise conditioning analysis is based on the recent success of second curvature approximation techniques in DNNs \cite{2012_ICML_Martens,2015_ICML_Martens,2017_ICLR_Ba,2017_ICML_RelNN,2018_NIPS_Alberto}. We show that the maximum eigenvalue and the condition number of the block-wise Fisher information matrix (FIM)  can be characterized based on the spectrum of the covariance matrix of the input and  output-gradient,  under mild assumptions, which makes evaluating optimization behavior practical in DNNs. Another theoretical base is the recently proposed proximal back-propagation \cite{2014_AISTATS_Miguel,2018_ICLR_Frerix,2018_NIPS_Zhang} where the original optimization problem can be approximately  decomposed into multiple independent sub-problems with respect to each layer \cite{2018_NIPS_Zhang}. We provide the connection between our analysis and the proximal back-propagation.

Based on our layer-wise conditioning analysis, we show that batch normalization (BN) \cite{2015_ICML_Ioffe} can  adjust the magnitude of the layer activations/gradients, and thus stabilizes the training. However, this kind of stabilization can drive certain layers into a particular state, referred to as \textbf{weight domination}, where the gradient update is feeble. This sometimes has detrimental effects on the learning (Section \ref{Sec:BN_stablization}).  We also experimentally observe that BN can improve the layer-wise conditioning of the optimization problem. Furthermore, we find that the unnormalized network has several small eigenvalues in the layer curvature matrix, which are mainly caused by the so-called \textbf{dying neurons} (Section \ref{Sec-BN-Accelerate}), while this behavior is almost entirely absent in batch normalized networks.


We further analyze the ignored difficulty in training very deep residual networks \cite{2015_CVPR_He}. Using our layer-wise conditioning analysis, we show that the difficulty mainly arises from the ill-conditioned behavior of the last linear layer. We solve this problem by only adding one BN layer before the last linear layer, which achieves improved performance over the original \cite{2015_CVPR_He} and pre-activation \cite{2016_CoRR_He} residual networks (Section \ref{Sec:veryDeep}).

\section{Preliminaries}
\label{Sec-Prelim}
\paragraph{\textbf{Optimization Objective}}
Consider a true data distribution $p_{*}(\mathbf{x}, \mathbf{y}) =p(\mathbf{x}) p(\mathbf{y}|\mathbf{x})$ and the sampled training sets $\mathcal{D} \sim p_{*}(\mathbf{x},\mathbf{y})$ of size $N$. We focus on a supervised learning task aiming to learn the conditional distribution $p(\mathbf{y}|\mathbf{x})$ using the model $q(\mathbf{y}|\mathbf{x})$, where $q(\mathbf{y}|\mathbf{x})$ is represented as a function $f_{\theta}(\mathbf{x})$ parameterized by $\theta$. From an optimization perspective, we aim to minimize the empirical risk,  averaged over the sample loss represented as $\ell(\mathbf{y}, f_{\theta}(\mathbf{x}))$ in training sets $\mathcal{D}$: $\mathcal{L}(\theta)=\frac{1}{N} \sum_{i=1}^{N} (\ell(\mathbf{y}^{(i)}, f_{\theta}(\mathbf{x}^{(i)})))$.
\vspace{-0.1in}
\paragraph{\textbf{Gradient Descent}}
In general, the gradient descent (GD) update seeks to iteratively reduce the loss $\mathcal{L}$  by $\mathbf{\theta}_{t+1} = \mathbf{\theta}_{t} - \eta \D{\mathbf{\theta}} $, where $\eta$ is the learning rate. For large-scale learning, stochastic gradient descent (SGD) is extensively used to approximate the gradients $\D{\mathbf{\theta}} $ with a mini-batch gradient.  In theory, the convergence behaviors (\eg, the number of iterations required for convergence to a stationary point) depend on the Lipschitz constant $C_L$  of the gradient function of $\mathcal{L}$\footnote{The loss' gradient function  $\D{\theta}$ is assumed Lipschitz continuous with Lipschitz constant $C_L$, i.e., $\| \D{\theta_1} - \D{\theta_2} \|_2 \leq C_L \| \theta_1 - \theta_2 \|$ for all $\theta_1$ and $\theta_2$.},
which characterizes the global smoothness of the optimization landscape. In practice, the Lipschitz constant is either unknown for complicated functions or too conservative to characterize the convergence behaviors \cite{2018_SAIM_Bottou}. Researchers thus turn to the local smoothness, characterized by the Hessian matrix  $\mathbf{H}=\frac{\partial \mathcal{L}^2}{\partial \theta \partial \theta}$ under the condition that $\mathcal{L}$ is twice differentiable.
\vspace{-0.1in}
\paragraph{\textbf{Approximate Curvature Matrices}}

The Hessian describes the local curvature of the optimization landscape.  Such curvature information intuitively guides the design of second-order optimization algorithms \cite{Sagun_CoRR_2017,2018_SAIM_Bottou}, where the update direction is adjusted by multiplying the inverse of a pre-conditioned matrix  $\mathbf{G}$ as: $\hatD{\mathbf{\theta}}=\mathbf{G}^{-1}\D{\mathbf{\theta}} $. $\mathbf{G}$ is a positive definite matrix that approximates the Hessian and is expect to sustain the its positive curvature. The second moment matrix of sample gradient: $\mathbf{M}= \mathbb{E}_{\mathcal{D}} (\Dl{\theta} \Dl{\theta}^T)$ is usually used as the pre-conditioned matrix \cite{2007_NIPS_Roux,2010_ICML_Martens}. Besides, Pascanu and Bengio \cite{2014_ICLR_Pascanu} showed that the FIM: $\mathbf{F}=\mathbb{E}_{ p(\mathbf{x}), ~ q(\mathbf{y}|\mathbf{x})}(\Dl{\theta} \Dl{\theta}^T)$ can be viewed as a pre-conditioned matrix when performing the natural gradient descent algorithm \cite{2014_ICLR_Pascanu}. Fore more analyses on the connections among $\mathbf{H}$,  $\mathbf{F}$, $\mathbf{M}$ please refer to \cite{2014_Martens_insights,2018_SAIM_Bottou}. In this paper, we refer to the analysis of the spectrum of the (approximate) curvature matrices as \emph{conditioning analysis}.
%
\vspace{-0.1in}
\paragraph{\textbf{Conditioning Analysis for Linear Models}}
Consider a linear regression model with a scalar output $f_{\mathbf{w}}(\mathbf{x})=\mathbf{w}^T \mathbf{x}$, and  mean square error loss $\ell=(y-f_{\theta}(\mathbf{x}))^2$.
As shown in \cite{1990_NIPS_LeCun,1998_NN_Yann}, the learning dynamics in such a quadratic surface are fully controlled by the spectrum of the Hessian matrix $\mathbf{H}=\mathbb{E}_\mathcal{D}(\mathbf{x} \mathbf{x}^T)$.
There are two statistical momentums that are essential for evaluating the convergence behaviors of the optimization problem. One is the maximum eigenvalue of the curvature matrix  $\lambda_{max}$, and the other is the condition number of the curvature matrix, denoted by $\kappa = \frac{\lambda_{max}}{\lambda_{min}}$,
where $\lambda_{min} $ is the  minimum nonzero eigenvalue of the curvature matrix.
Specifically, $\lambda_{max}$ controls the upper bound and the optimal learning rate (\eg, the optimal learning rate is $\eta=\frac{1}{\lambda_{max}(\mathbf{H})}$ and the training will diverge if $\eta \geq  \frac{2}{\lambda_{max}(\mathbf{H})}$).  $\kappa $ controls the iterations required for convergence (\eg, the lower bound of the iteration is $\kappa (\mathbf{H})$ \cite{1990_NIPS_LeCun}). If $\mathbf{H}$ is an identity matrix that can be obtained by whitening the input, the GD can converge within only one iteration.
It is easy to extend the solution of linear regression from a scalar output to a vectorial output $f_{\mathbf{W}}(\mathbf{x})=\mathbf{W}^T \mathbf{x}$. In this case, the Hessian is represented as
\begin{equation}
\label{eqn:Hessian_LR}
\mathbf{H}=\mathbb{E}_\mathcal{D}(\mathbf{x} \mathbf{x}^T) \otimes \mathbf{I},
\end{equation}
where 	$\otimes$ indicates the Kronecker product \cite{2016_ICML_Grosse} and $\mathbf{I}$ denotes the identity matrix.
For the linear classification model with cross entropy loss,  the Hessian is  approximated by \cite{2011_NIPS_Wiesler}:
\begin{equation}
\label{eqn:Hession_Logistic}
\mathbf{H}=\mathbb{E}_\mathcal{D}(\mathbf{x} \mathbf{x}^T) \otimes \mathbf{S}.
\end{equation}
$\mathbf{S} \in \mathbb{R}^{c \times c}$ is defined by $\mathbf{S}=\frac{1}{c}(\mathbf{I}_c- \frac{1}{c} \mathbf{1}_c)$, where $c$ is the number of classes and $\mathbf{1}_c \in \mathbb{R}^{c \times c}$ denotes a matrix in which all entries are 1. Eqn. \ref{eqn:Hession_Logistic} assumes  	
 the Hessian does not significantly change from the initial region to the optimal region \cite{2011_NIPS_Wiesler}.

\section{Layer-wise Conditioning Analysis for DNNs}
\label{Sec-Conditioning analysis}
\vspace{-0.05in}
Considering a multilayer perceptron (MLP), $f_{\theta}(\mathbf{x})$ can be represented as a layer-wise linear and nonlinear transformation, as follows:
\begin{equation}
\setlength\abovedisplayskip{0.05in}
\setlength\belowdisplayskip{0.05in}
\label{eqn:MLP}
\mathbf{h}_{k}= \mathbf{W}_{k} \mathbf{x}_{k-1},  ~~
\mathbf{x}_{k}= \phi(\mathbf{h}_k), ~~~ k=1,..., K,
\end{equation}
where $\mathbf{x}_0 = \mathbf{x}$, $\mathbf{W}_k \in \mathbb{R}^{d_{k} \times d_{k-1}}$ and the learnable parameters $\mathbf{\theta}=\{ \mathbf{W}_k, k=1,...,K \}$. To simplify the denotation, we set  $\mathbf{x}_K=\mathbf{h}_K $ as the output of the network $f_{\theta}(\mathbf{x})$.

A conditioning analysis on the full curvature matrix for DNNs is difficult due to the high memory and computational costs \cite{2019_ICML_Ghorbani,2018_CoRR_vardan}. We thus seek to analyze an approximation of the curvature matrix.
One successful example in second-order optimization over DNNs is approximating the FIM using the Kronecker product (K-FAC) \cite{2015_ICML_Martens,2017_ICLR_Ba,2017_ICML_RelNN,2018_NIPS_Alberto}.
In the K-FAC approach, there are two assumptions: 1) weight-gradients in different layers are assumed to be uncorrelated; 2) the input and output-gradient in each layer are approximated as independent, so the full FIM can be represented as a block diagonal matrix,  $\mathbf{F}=diag (F_1,..., F_K)$, where $F_k$ is the sub-FIM (the FIM with respect to the parameters in a  certain layer) and  computed as:
\begin{small}
	\setlength\abovedisplayskip{0.05in}
	\setlength\belowdisplayskip{0.05in}
	\begin{align}
	\label{eqn:DNN_layer_FIM}
	F_{k}=\mathbb{E}_{p(\mathbf{x}), ~ q(\mathbf{y}|\mathbf{x})}((\mathbf{x}_k \mathbf{x}_k^T) \otimes  (\Dl{\mathbf{h}_k}^T\Dl{\mathbf{h}_k}) )
	\approx \mathbb{E}_{p(\mathbf{x})}(\mathbf{x}_k \mathbf{x}_k^T) \otimes  \mathbb{E}_{q(\mathbf{y}|\mathbf{x})}( \Dl{\mathbf{h}_k}^T\Dl{\mathbf{h}_k}).
	\end{align}
\end{small}
\hspace{-0.08in}  $\mathbf{x}_k$ denotes the layer input, and  $\Dl{{\mathbf{h}_k}}$ denotes the layer output-gradient.
We note that Eqn.~\ref{eqn:DNN_layer_FIM} is similar to Eqn.~\ref{eqn:Hessian_LR} and~\ref{eqn:Hession_Logistic}, and all of them depend on the covariance matrix of the (layer) input. The main difference is that, in Eqn.~\ref{eqn:DNN_layer_FIM}, the covariance of output-gradient is considered and its value changes over different optimization regions, while in Eqn.~\ref{eqn:Hessian_LR} and~\ref{eqn:Hession_Logistic}, the covariance of output-gradient is  constant.

Based on this observation, we propose layer-wise conditioning analysis, \ie, we analyze each sub-FIM $F_k$'s spectrum $\lambda(F_k)$ independently. We expect the spectra of sub-FIMs: $\{\lambda(F_k)\}_{k=1}^K$ to effectively reveal that of the full FIM: $\lambda(\mathbf{F})$, at least in terms of analyzing the learning dynamics of the DNNs. Specifically, we analyze the maximum eigenvalue  $\lambda_{max}(F_k)$ and condition number $\kappa(F_k)$\footnote{Since DNNs are usually over-parameterized, we evaluate the general condition number with respect to the percentage: $\kappa_{p}=\frac{\lambda_{max}}{\lambda_{p}}$, where $\lambda_{p}$ is the $pd$-th eigenvalue (in descending order) and $d$ is the number of eigenvalues, \eg, $\kappa_{100\%}$ is the original definition of the condition number.}.
Based on the conclusion on the conditioning analysis of linear models shown in Section \ref{Sec-Prelim}, there are two remarkable properties that can be used to implicitly uncover the landscape of the optimization problem:
\vspace{-0.06in}
{\spaceskip=0.3em\relax
\begin{itemize}
	\item  \emph{Property 1}: $\lambda_{max}(F_k)$ indicates the magnitude of  the weight-gradient in each layer, which shows the steepness of the landscape \wrt different layers.
	\item  \emph{Property 2}: $\kappa(F_k)$ indicates how easy it is to optimize the corresponding layer.
\end{itemize}
}

\begin{figure}[t]
	\centering
	\vspace{-0.05in}
	\hspace{-0.2in}	\subfigure[full FIM]{
		\begin{minipage}[c]{.30\linewidth}
			\centering
			\includegraphics[width=3.8cm]{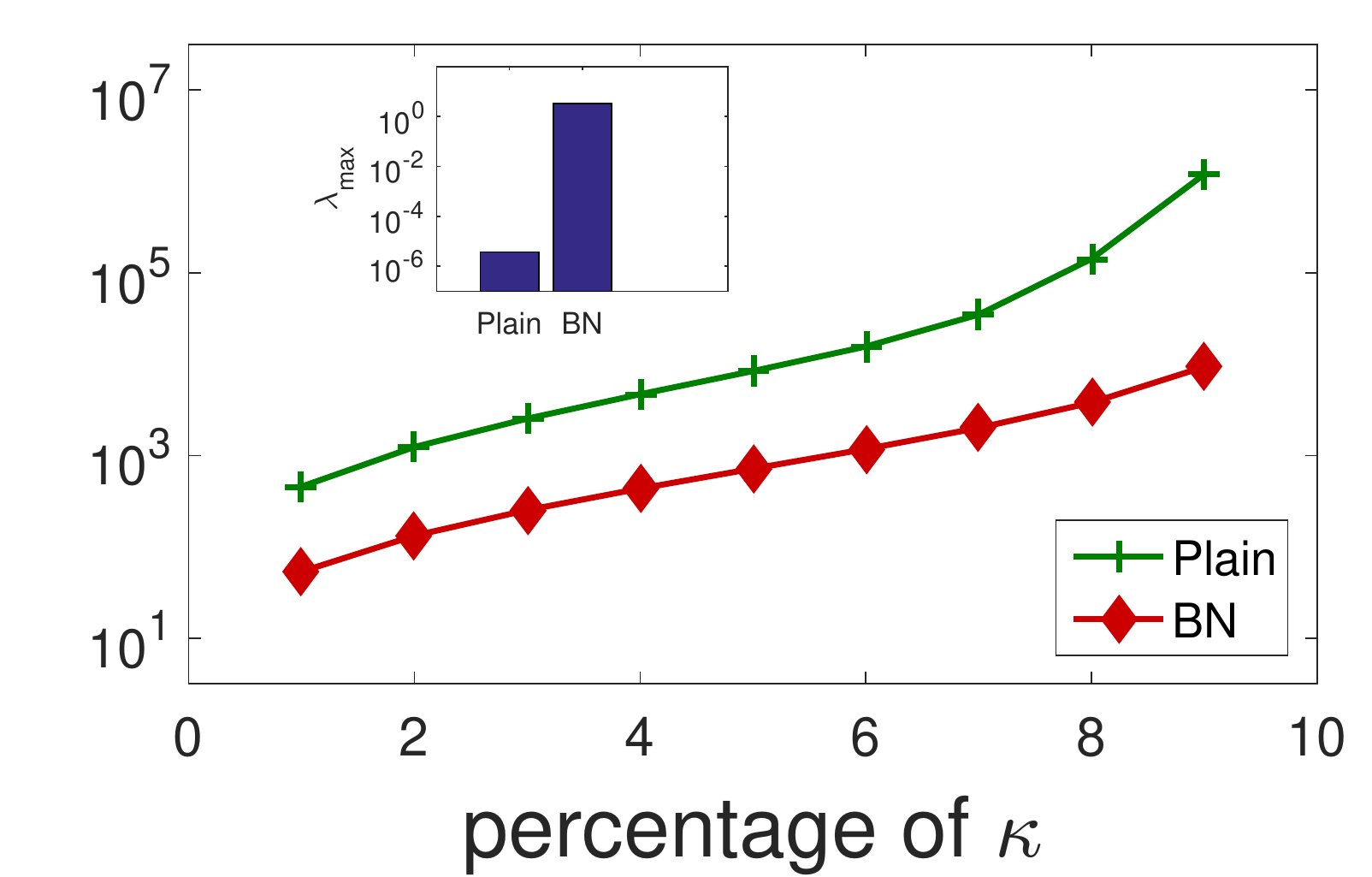}
		\end{minipage}
	}
	\hspace{0.05in}	\subfigure[sub-FIM (the 3rd layer)]{
		\begin{minipage}[c]{.30\linewidth}
			\centering
			\includegraphics[width=3.8cm]{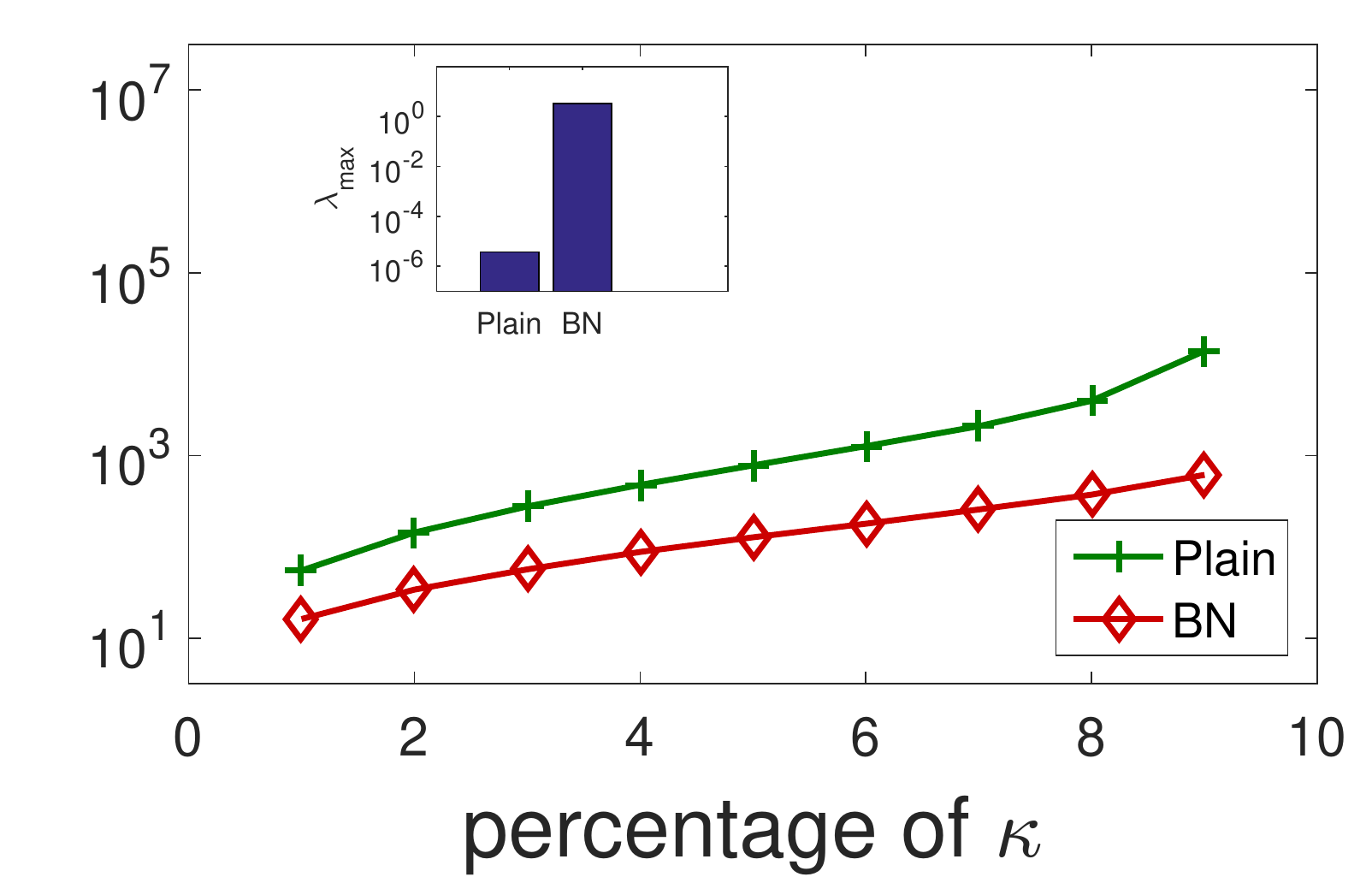}
		\end{minipage}
	}
	\hspace{0.05in}	\subfigure[sub-FIM (the 6th layer)]{
		\begin{minipage}[c]{.30\linewidth}
			\centering
			\includegraphics[width=3.8cm]{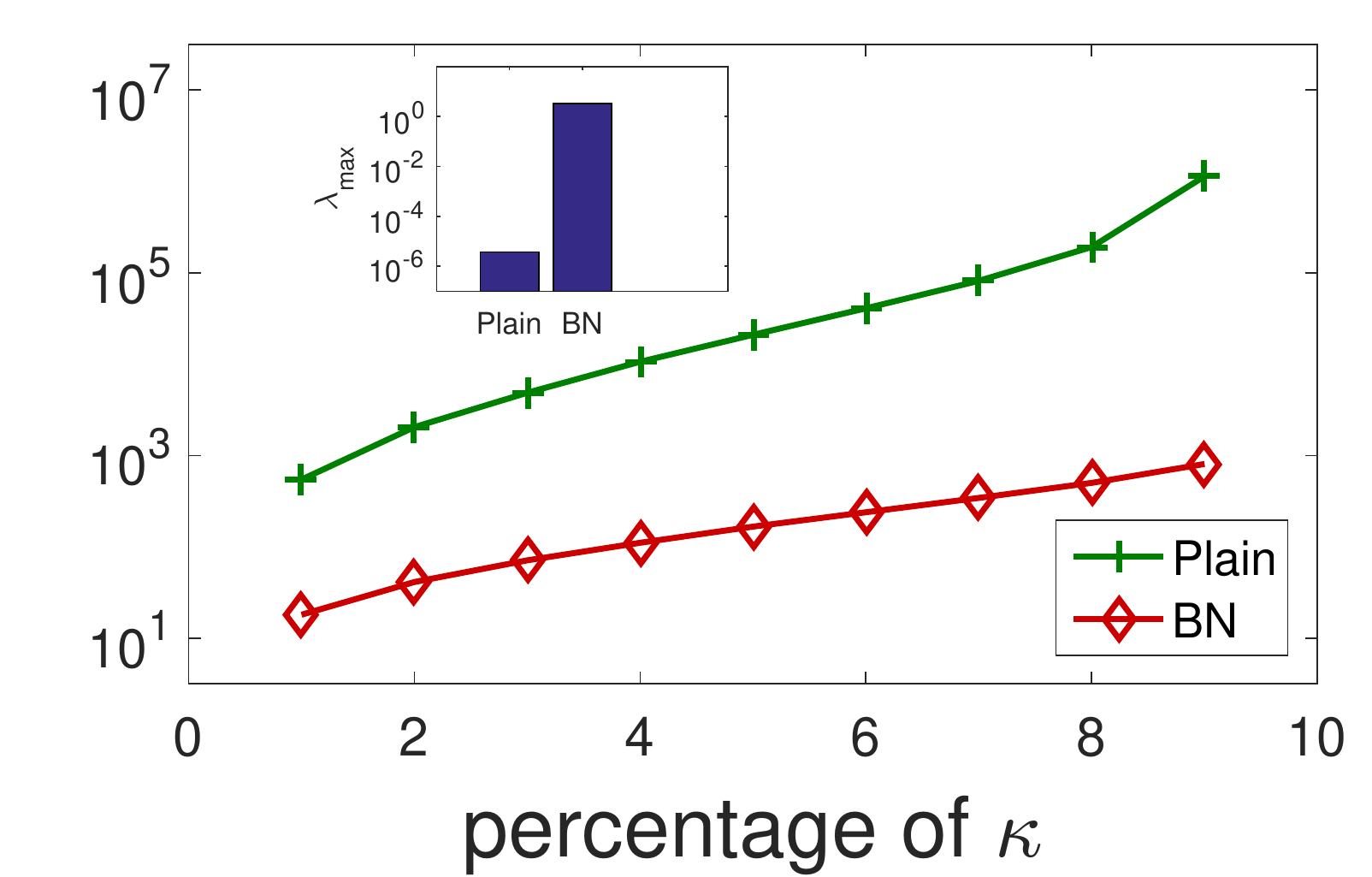}
		\end{minipage}
	}
	\vspace{-0.15in}
	\caption{Conditioning analysis for unnormalized (`Plain') and normalized networks (`BN'). We show the maximum eigenvalue $\lambda_{max}$ and the generalized condition number $\kappa_{p}$ for comparison between the full FIM $\mathbf{F}$ and sub-FIMs $\{F_k\}$.   The experiments are performed on an 8-layer MLP with 24 neurons in each layer, for MNIST classification. The input image is center-cropped and resized to $12 \times 12$ to remove uninformative pixels. We report the corresponding spectrum at random initialization \cite{1998_NN_Yann}. Here, we report the results of the 3rd and 6th layers in (b) and (c), respectively. We have similar observations for other layers (See \SM~\ref{Sec-sup-approximate}).}
	\label{fig:AP1-FIM}
	\vspace{-0.05in}
\end{figure}

\begin{figure}[t]
	\centering
	\vspace{-0.12in}
	\hspace{-0.2in}	\subfigure[$\lambda_{max}$]{
		\begin{minipage}[c]{.3\linewidth}
			\centering
			\includegraphics[width=3.8cm]{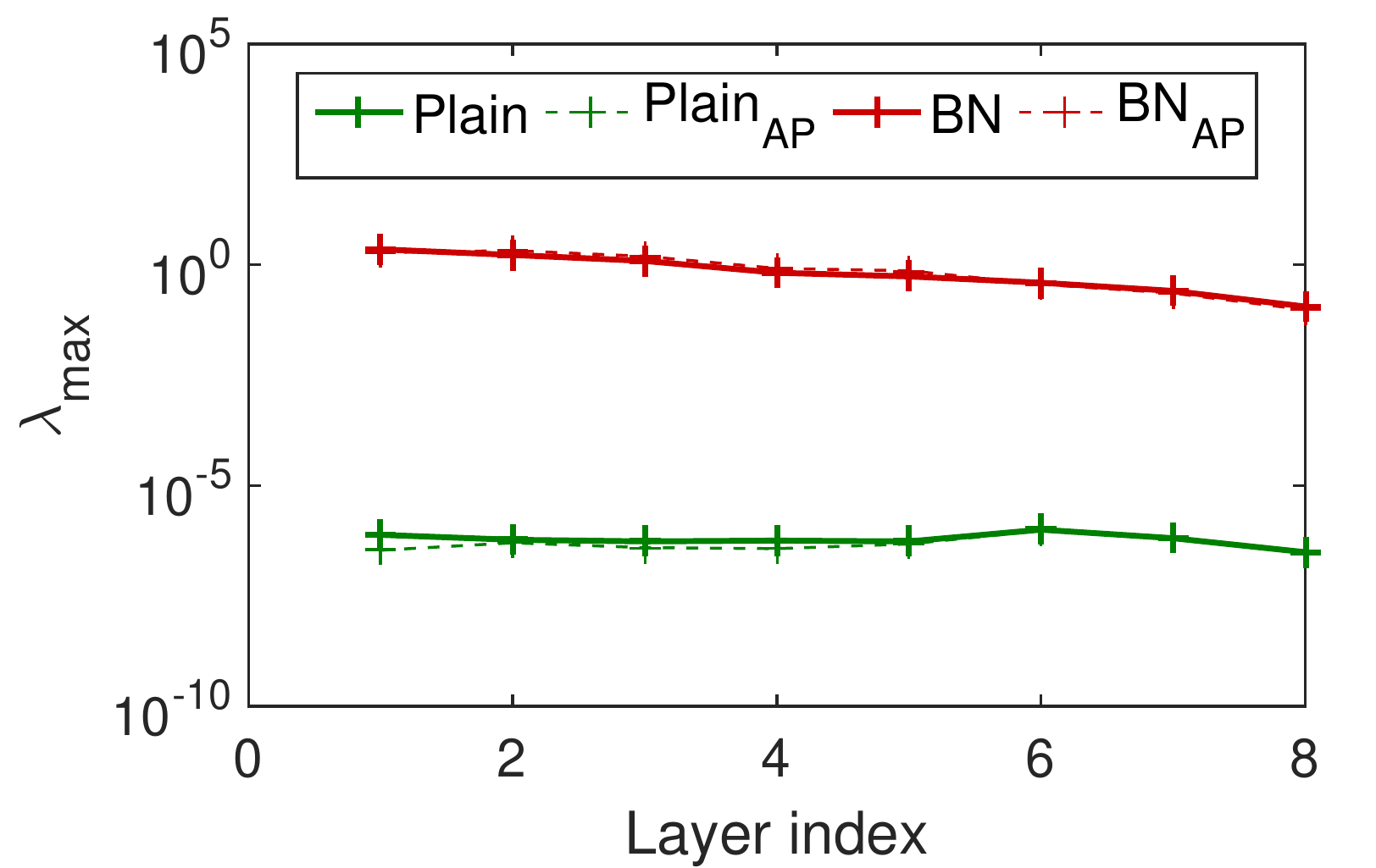}
		\end{minipage}
	}
	\hspace{0.05in}	\subfigure[$\kappa_{50\%}$]{
		\begin{minipage}[c]{.3\linewidth}
			\centering
			\includegraphics[width=3.8cm]{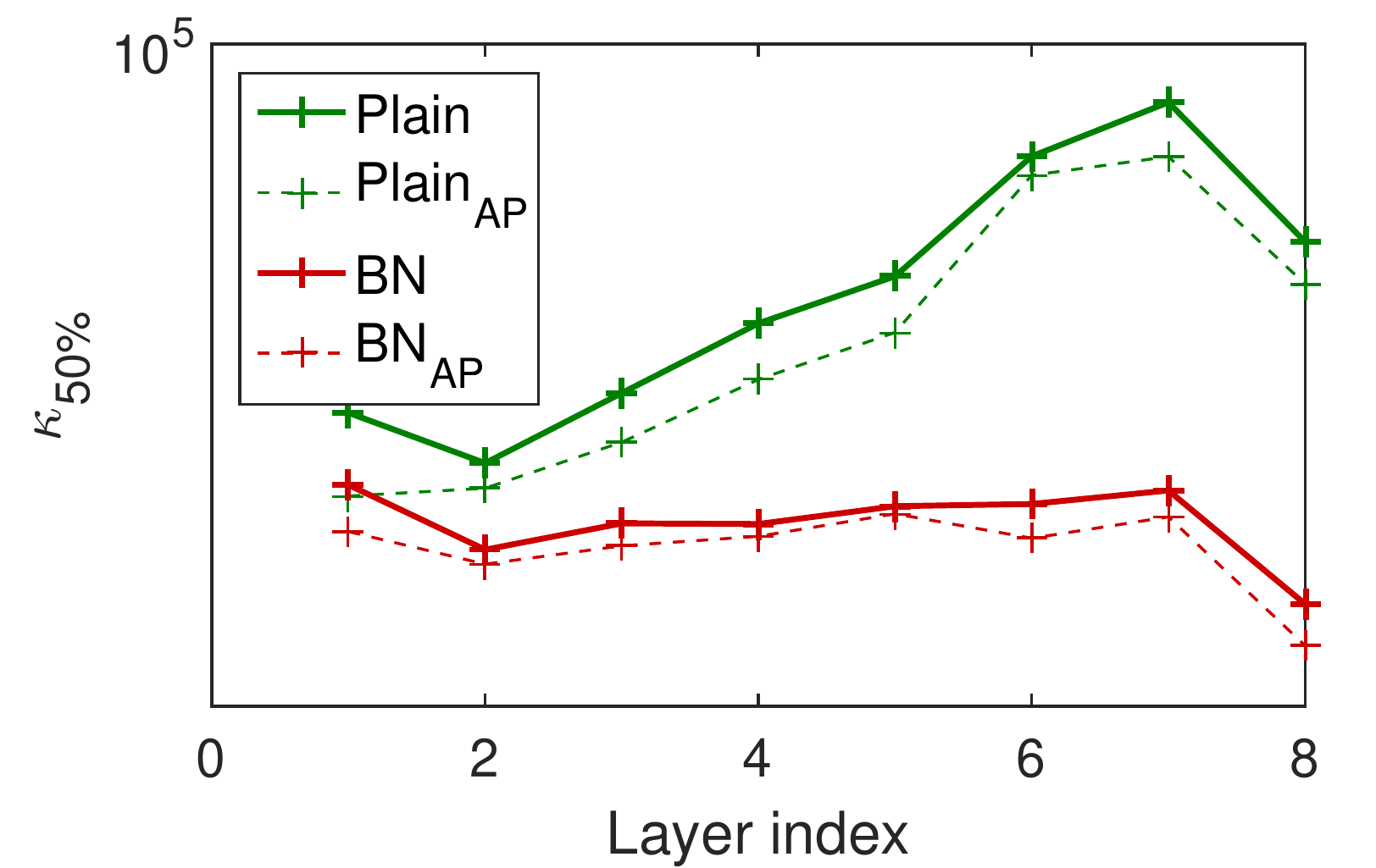}
		\end{minipage}
	}	
	\hspace{0.05in}	\subfigure[$\kappa_{80\%}$]{
		\begin{minipage}[c]{.3\linewidth}
			\centering
			\includegraphics[width=3.8cm]{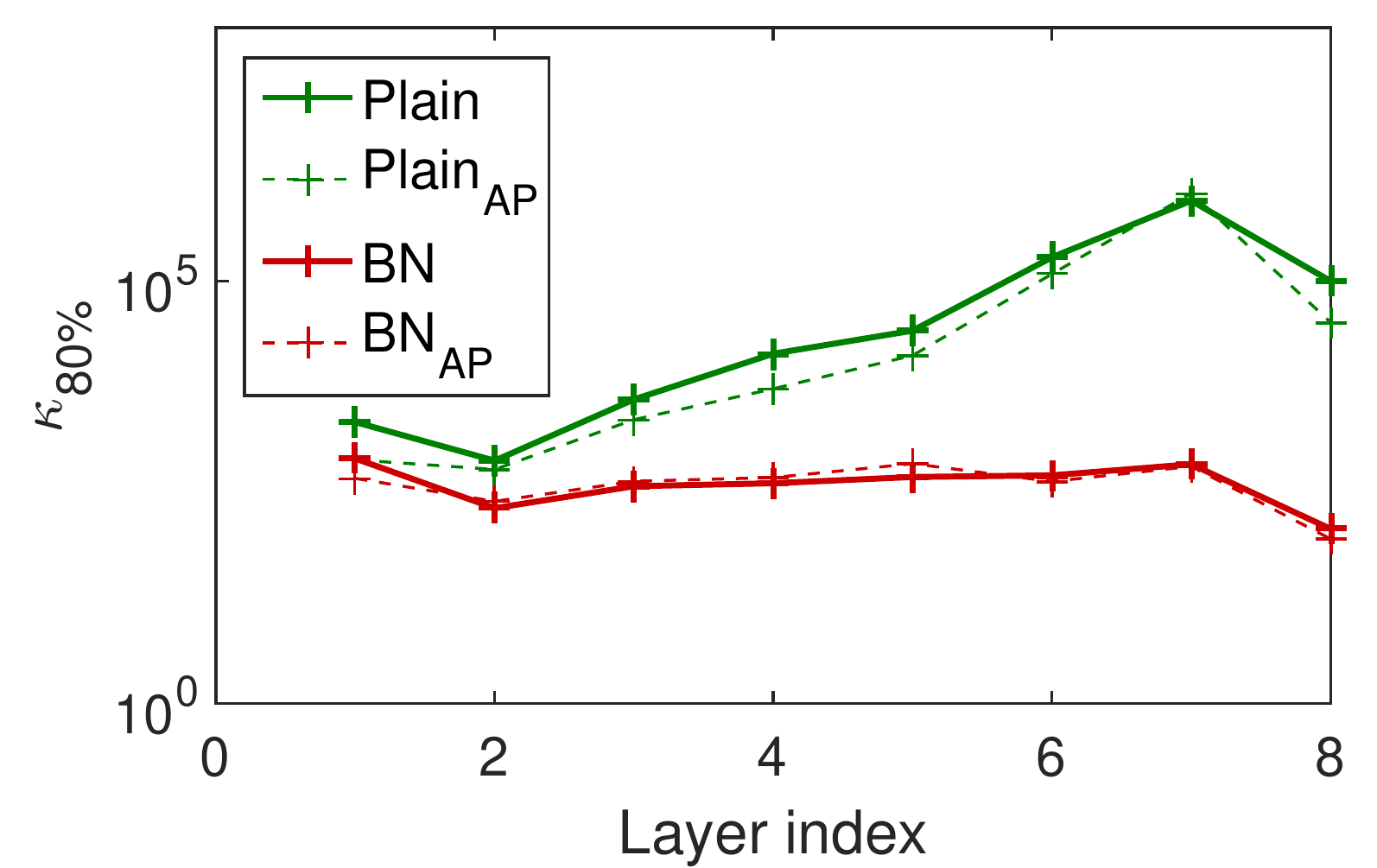}
		\end{minipage}
	}	
	\vspace{-0.15in}
	\caption{Validation in approximating the sub-FIMs. The experimental setups are the same as in Figure~\ref{fig:AP1-FIM}. We compare maximum eigenvalue $\lambda_{max}$ and generalized condition number $\kappa_{p}$ of the  sub-FIMs  (solid lines) and the approximated ones (dashed lines).}
	\label{fig:AP2}
\end{figure}

\vspace{-0.1in}
\paragraph{\textbf{Discussion}} 
One concern is the validity of the assumptions the K-FAC approximation is based on. Note that \cite{2014_Martens_insights,2015_ICML_Martens} have provided some empirical evidence to support their effectiveness in approximating the full FIM with block diagonal sub-FIMs. 
\cite{2018_arxiv_Ryo,2019_arxiv_Wei} also exploited similar assumptions to derive the mean$\&$variance of eigenvalues (and maximum eigenvalue) of the full FIM, which is calculated using information from layer inputs and output-gradients.  
Here, we argue that the assumptions required  for our analysis are weaker than those of the K-FAC approximation, since we only care about whether or not the spectra of sub-FIMs  can accurately reveal the spectrum of full FIM.  
We conduct experiments to analyze the training dynamics of  the unnormalized (`Plain') and batch normalized \cite{2015_ICML_Ioffe} (`BN') networks, by looking at the spectra of full curvature matrix and  sub-curvature matrices.
Figure \ref{fig:AP1-FIM} shows the results based on an 8-layer MLP with 24 neurons in each layer.
  By observing the results from the full FIM (Figure \ref{fig:AP1-FIM} (a)), we find that: 1) the unnormalized network suffers from gradient vanishing (the maximum eigenvalue is around $1e^{-5}$), while the batch normalized network has an appropriate magnitude of gradient (the maximum eigenvalue is around $1$); 2) `BN' has better conditioning than `Plain', which suggests batch normalization (BN) can improve the conditioning of the network, as observed in \cite{2018_NIPS_shibani,2019_ICML_Ghorbani}. We also obtain a similar conclusion when observing the results from the sub-FIMs (Figure~\ref{fig:AP1-FIM} (b), (c)). This experiment demonstrates that our layer-wise conditioning analysis  \revise{has the potentiality to}  uncover the training dynamics of the networks if the full conditioning analysis can. We also conduct experiments on MLPs with different layers and neurons, and further analyze the spectrum of the second moment matrix of sample gradient $\mathbf{M}$  (please refer to Appendix~\ref{Sec-sup-approximate} for details). We have the same observations as in the first experiment.


Furthermore, we find that investigating  $\{\lambda(F_k)\}_{k=1}^K$ is more  beneficial for diagnosing the problems behind training DNNs than investigating  $\lambda(\mathbf{F})$,  
 \eg, it enables the gradient
 vanishing/explosion to be located with respect to a specific layer from $\{\lambda_{max}(F_k)\}_{k=1}^K$, but not $\lambda_{max}(\mathbf{F})$.
For example, we know that the 8-layer unnormalized MLP described in Figure~\ref{fig:AP1-FIM} suffers from difficulty in training, but we cannot accurately diagnose the problem by  only investigating  the spectrum of the full FIM. However, by looking into the layer inputs and output-gradients, we find that this MLP suffers from exponentially decreased magnitudes of inputs (forward) and output-gradients (backward). This can be resolved this by using a better initialization with appropriate variance \cite{2015_ICCV_He} or using BN \cite{2015_ICML_Ioffe}.
We further elaborate on how to use the layer-wise conditioning analysis to `debug' the training of DNNs in the subsequent sections.



%
%

\subsection{Efficient Computation}
\vspace{-0.03in}
 We denote the covariance matrix of the layer input as $\Sigma_{\mathbf{x}}=\mathbb{E}_{p(\mathbf{x})}(\mathbf{x} \mathbf{x}^T)$ and the covariance matrix of the layer output-gradient as $\Sigma_{\nabla \mathbf{h}} =  \mathbb{E}_{q(\mathbf{y}|\mathbf{x})}( \Dl{\mathbf{h}}^T\Dl{\mathbf{h}})$.
The condition number and maximum eigenvalue of the sub-FIM $F$ can be derived based on the spectrum of $\Sigma_{\mathbf{x}}$ and $\Sigma_{\nabla \mathbf{h}}$, as shown in the following proposition.
\begin{small}
\begin{proposition}
	\label{th1:norm}
	Given $\Sigma_{\mathbf{x}}$, $\Sigma_{\nabla \mathbf{h}}$ and  $F= \Sigma_{\mathbf{x}} \otimes \Sigma_{\nabla \mathbf{h}}$, we have:
	1) $\lambda_{max}(F)=\lambda_{max}(\Sigma_{\mathbf{x}}) \cdot \lambda_{max}(\Sigma_{\nabla \mathbf{h}}) $; 	
	and 2) $\kappa(F)=\kappa(\Sigma_{\mathbf{x}}) \cdot \kappa(\Sigma_{\nabla \mathbf{h}}) $.
\end{proposition}
\end{small}
The proof is shown in the \SM~\ref{Sec-Proof-Pro1}. Proposition~\ref{th1:norm} provides an efficient way to calculate the maximum eigenvalue and condition number of sub-FIM $F$ by computing those of $ \Sigma_{\mathbf{x}}$ and $\Sigma_{\nabla \mathbf{h}}$.
In practice, we use the empirical distribution $\mathbb{D}$ to approximate the expected distribution $p(\mathbf{x}) $ and $q(\mathbf{y}|\mathbf{x})$ when calculating $\Sigma_{\mathbf{x}}$ and $\Sigma_{\nabla \mathbf{h}}$, since this is very efficient and can be performed with only one forward and backward pass, as has been shown in FIM approximation~\cite{2015_ICML_Martens,2017_ICLR_Ba}.

Note that Proposition~\ref{th1:norm}  depends on the second assumption of Eqn.~\ref{eqn:DNN_layer_FIM}.
We experimentally demonstrate the effectiveness of such an approximation  in Figure~\ref{fig:AP2}, finding that the maximum eigenvalue and the condition number of the sub-FIMs match well with the approximated ones.


\vspace{-0.13in}
\subsection{Connection to Proximal Back-propagation}
\vspace{-0.04in}
\label{Sec:Proximal}
Carreira-Perpinan and Wang \cite{2014_AISTATS_Miguel} proposed to use auxiliary coordinates to redefine the optimization object $\Loss$ with equality constraints imposed on each neuron. They solved the constrained optimization by adding a quadratic penalty as:
\begin{small}
	\setlength\abovedisplayskip{0.01in}
	\setlength\belowdisplayskip{0.01in}
	\begin{equation}
	\label{eqn:TLoss}
\TLoss=\mathcal{L}(\y, f_K(\W{K}, \z{K-1}))
	+ \sum_{k=1}^{K-1} \frac{\lambda}{2} \| \z{k} - f_b(\W{k}, \z{k-1})) \|^2,
	\end{equation}
\end{small}
\hspace{-0.05in}where $f_k(\cdot, \cdot)$ is a function with respect to each layer.
As shown in \cite{2014_AISTATS_Miguel}, the solution for minimizing $\TLoss$ converges to the solution for minimizing $\Loss$  as $\lambda \rightarrow \infty$, under mild conditions. 
\revise{Furthermore, the proximal propagation~\cite{2018_ICLR_Frerix} and the following back-matching propagation~\cite{2018_NIPS_Zhang} reformulate each sub-problem independently with a backward order, minimizing each layer object $	\mathcal{L}_k(\W{k},\z{k-1}; \hz{k} )$, given the target signal $\hz{k}$ from the upper layer, as follows:}
\begin{small}
	\begin{equation}
	\label{eqn:GradientMatch}
	\begin{cases}
	\mathcal{L}(\y, f_K(\W{K}, \z{K-1})),~~ for ~k=K \\
	\frac{1}{2}\| \hz{k} - f_k(\W{k}, \z{k-1})) \|^2, ~~for~  k=K-1,...,1.
	\end{cases}
	\end{equation}
\end{small}
\hspace{-0.05in}\revise{It has been shown that the produced $\W{k}$ using gradient update \wrt~ $\Loss$ equals to the $\W{k}$ produced by the back-matching propagation (Procedure 1 in~\cite{2018_NIPS_Zhang}) with one-step gradient update \wrt~Eqn. \ref{eqn:GradientMatch},  given an appropriate step size. Note that the target signal $\hz{k}$ is obtained by back-propagation, which means the loss $\Loss$ would be smaller if  $f_k(\W{k}, \z{k-1})$ is more close to $\hz{k}$. The loss $\Loss$ will be reduced more efficiently, if the sup-optimization problems in Eqn. \ref{eqn:GradientMatch} are well-conditioned. Please refer to \cite{2018_ICLR_Frerix,2018_NIPS_Zhang} for more details.}

Interestingly, if the target signal $\hz{k}$ is provided at the pre-activation\footnote{$\hz{k}$ can be obtained by updating the pre-activation $\h{k}$ with its gradient $\D{\h{k}}$~\cite{2018_NIPS_Zhang}.} in a specific layer, the sub-optimization problem in each layer is formulated as:
\begin{small}
	\begin{equation}
	\label{eqn:GradientMatch-Linear}
	\begin{cases}
	\mathcal{L}(\y, \W{K}\z{K-1}),~~ for ~k=K \\
	\frac{1}{2}\| \hz{k} - \W{k}\z{k-1} \|^2, ~~for~  k=K-1,...,1.
	\end{cases}
	\end{equation}
\end{small}
\hspace{-0.05in}It is clear that the sub-optimization problems with respect to $\W{k}$ are actually linear classification (for k=K) or regression  (for $k=1,...,K-1$) models. Their conditioning analysis is thoroughly characterized in Section \ref{Sec-Prelim}.  

This connection suggests: 1) the quality (conditioning) of the full optimization problem  $\Loss$ is well correlated to its sub-optimization problems shown in Eqn. \ref{eqn:GradientMatch-Linear}, whose local curvature matrix can be well explored; 2) We can    diagnose the ill behaviors of a DNN  by speculating its spectra with respect to certain layers. 

\vspace{-0.1in}
\section{Exploring Batch Normalized Networks}
\label{Sec:BN}
\vspace{-0.05in}
Let $x$ denote the input for a given neuron in one layer of a DNN. Batch normalization (BN) \cite{2015_ICML_Ioffe} standardizes the neuron  within $m$ mini-batch data by:
\begin{small}
	\setlength\abovedisplayskip{0.0in}
	\setlength\belowdisplayskip{0.0in}
	\begin{equation}
	\label{eqn:BN}
	BN(x^{(i)})= \gamma \frac{x^{(i)} -\mu}{\sqrt{\sigma^2 +\epsilon}} + \beta,	
	\end{equation}
\end{small}
\hspace{-0.04in}where $\mu=\frac{1}{m}  \sum_{i=1}^{m}  x^{(i)}$ and $\sigma^2 = \frac{1}{m}   \sum_{i=1}^{m} (x^{(i)}-\mu)^2 $ are the  mean and variance,  respectively. The learnable parameters $\gamma$ and $\beta$ are used to recover the representation capacity. BN is a ubiquitously employed technique in various architectures \cite{2015_ICML_Ioffe,2015_CVPR_He,2016_CoRR_Zagoruyko,2016_CoRR_Huang_a} due to its ability in stabilizing and accelerating training. Here, we explore how BN stabilizes and accelerates training based on our layer-wise conditioning analysis.

\vspace{-0.1in}
\subsection{Stabilizing Training}
\vspace{-0.05in}
\label{Sec:BN_stablization}
From the perspective of a practitioner, two phenomena relate to the instability in training a DNN: 1) the training loss first increases significantly and then diverges; or 2) the training loss hardly changes, compared to the initial condition. 
The former is mainly caused by weights with large updates (\eg, exploded gradients or optimization with a large learning rate). The latter is caused by weights with few updates (vanished gradients or optimization with a small learning rate).
In the following theorem, we show that the unnormalized rectifier neural network is very likely to encounter both phenomena.
\vspace{-0.06in}
\begin{small}
\begin{theorem}
	\label{th2:norm}
	Given a rectifier neural network (Eqn. \ref{eqn:MLP}) with nonlinearity $\phi(\alpha \mathbf{x})= \alpha \phi(\mathbf{x})$ ($\alpha >0$ ), if the weight in each layer is scaled by $\widehat{\mathbf{W}}_k = \alpha_k \mathbf{W}_k $ ($k=1,...,K$ and $\alpha_k >0$), we have the scaled layer input:
	$\widehat{\mathbf{x}}_k= (\prod\limits_{i=1}^{k} {\alpha_i}) \mathbf{x}_k$.
	Assuming that $\D{\hath{K}} = \mu \D{\h{K}}$, we have the output-gradient:
	$\D{\hath{k}} =  \mu ( \prod\limits_{i=k+1}^{K} {\alpha_i})  \D{\mathbf{h}_{k}}$,  and weight-gradient: $\D{\widehat{\mathbf{W}}_k } = (\mu \prod\limits_{i=1, i\neq k}^K \alpha_i ) \D{\mathbf{W}_k}$, for all $k=1,...,K$.
\end{theorem}
\end{small}
\vspace{-0.06in}
The proof is shown in the \SM~\ref{Sec-Proof-Th1}. From Theorem \ref{th2:norm}, we observe that the scaled factor $\alpha_k$ of the weight in layer $k$ will affect all other layers' weight-gradients. Specifically, if all $\alpha_k >1$ ($\alpha_k <1$), the weight-gradient will increase (decrease) exponentially for one iteration.
Moreover, such an exponentially increased weight-gradient will be sustained and amplified in the subsequent iteration, due to the increased magnitude of the weight caused by updating.  That is why the unnormalized neural network will diverge, once the training loss increases over a few continuous iterations.  
We show that such instability can be relieved by BN, based on the following theorem.
\vspace{-0.05in}
\begin{small}
\begin{theorem}
	\label{th3:norm}
	Under the same condition as Theorem \ref{th2:norm}, for the normalized network with $\h{k}=\W{k} \x{k-1}$ and $\mathbf{s}_k=BN(\mathbf{h}_k)$, we have:
	$\hatx{k}=\mathbf{x}_k$, $\D{\hath{k}} = \frac{1} {\alpha_k} \D{\mathbf{h}_k}$, $\D{\widehat{\mathbf{W}}_k } =\frac{1} {\alpha_k} \D{\mathbf{W}_k}$, for all $k=1,...,K$.
\end{theorem}
\end{small}
\vspace{-0.05in}
The proof is shown in the \SM~\ref{Sec-Proof-Th2}. From Theorem \ref{th3:norm}, the scaled factor $\alpha_k$ of the weight will not affect other layers' activations/gradients. The magnitude of the weight-gradient is inversely proportional to the scaled factor. Such a mechanism will stabilize the weight growth/reduction, as shown in \cite{2015_ICML_Ioffe,2019_ICLR_Yang}.
Note that the behaviors when stabilizing training (Theorem \ref{th3:norm}) also apply for other activation normalization methods \cite{2016_CoRR_Ba,2018_CoRR_Wu,2018_NIPS_Hoffer,2018_CVPR_Huang,2019_CVPR_Huang}. We note that the scale-invariance of BN in stabilizing training has been analyzed in previous work \cite{2016_CoRR_Ba}. Different to their analyses on the normalization layer itself, we provide an explicit formulation of weight-gradients and output-gradients in a network, which is more important when characterizing the learning dynamics of DNNs.
%

\begin{figure*}[t]
	\centering
	\vspace{-0.02in}
	\hspace{-0.3in}		\subfigure[$\lambda_{max}(\Sigma_x)$]{
		\begin{minipage}[c]{.3\linewidth}
			\centering
			\includegraphics[width=3.8cm]{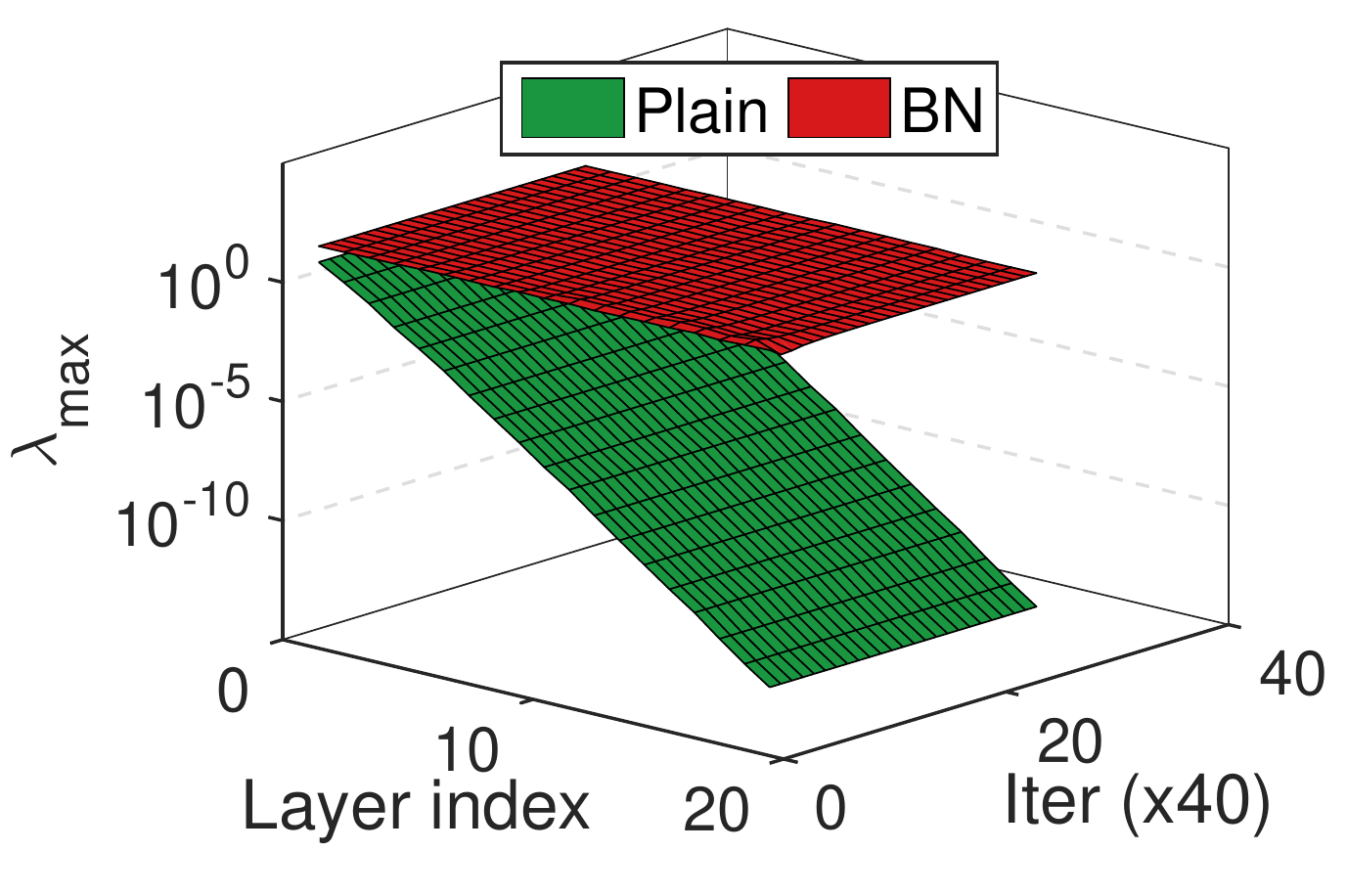}
		\end{minipage}
	}
	\subfigure[$\lambda_{max}(\Sigma_{\nabla \mathbf{h}})$]{
		\begin{minipage}[c]{.3\linewidth}
			\centering
			\includegraphics[width=3.8cm]{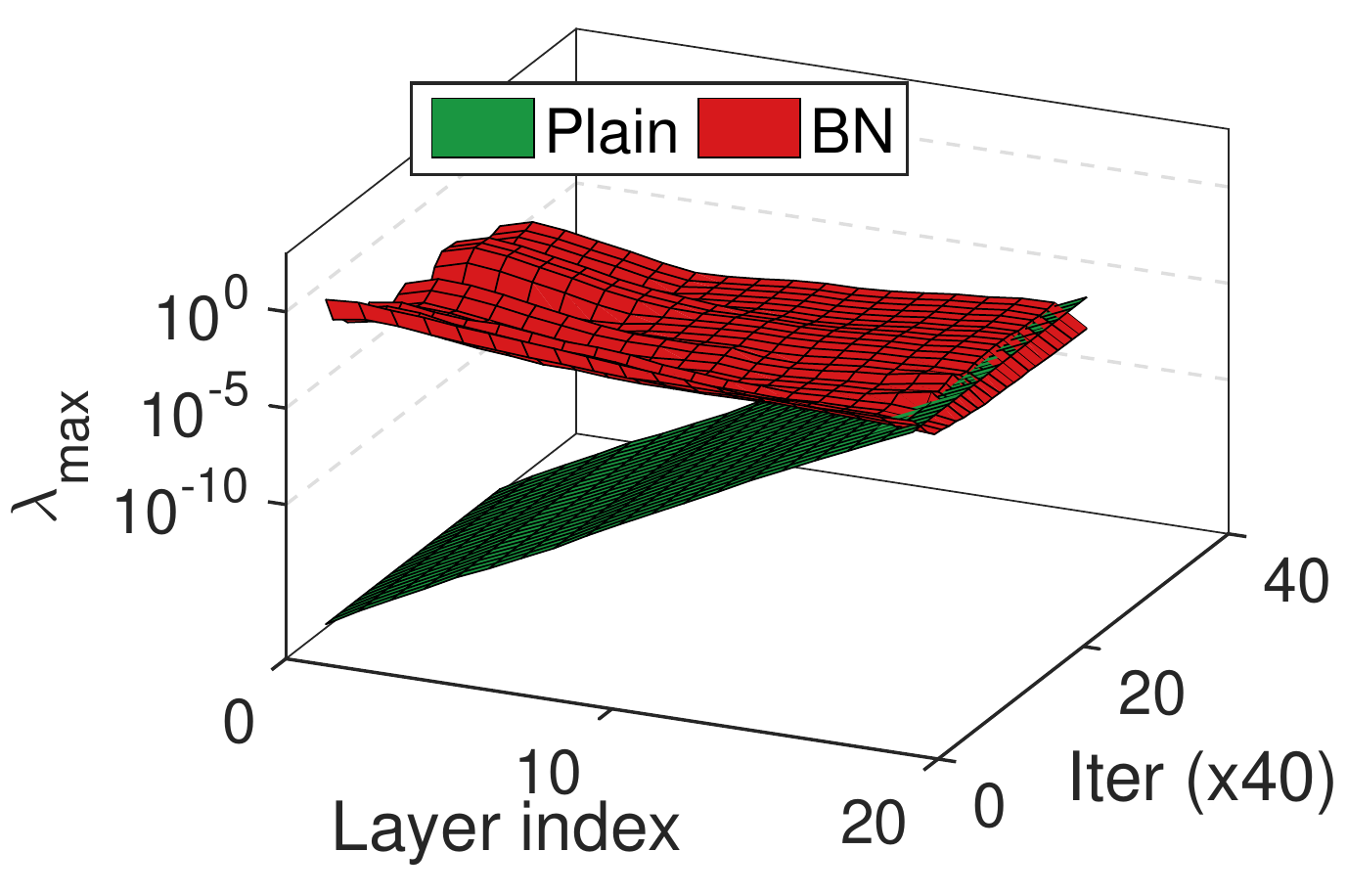}
		\end{minipage}
	}
	\subfigure[Training loss]{
		\begin{minipage}[c]{.3\linewidth}
			\centering
			\includegraphics[width=3.8cm]{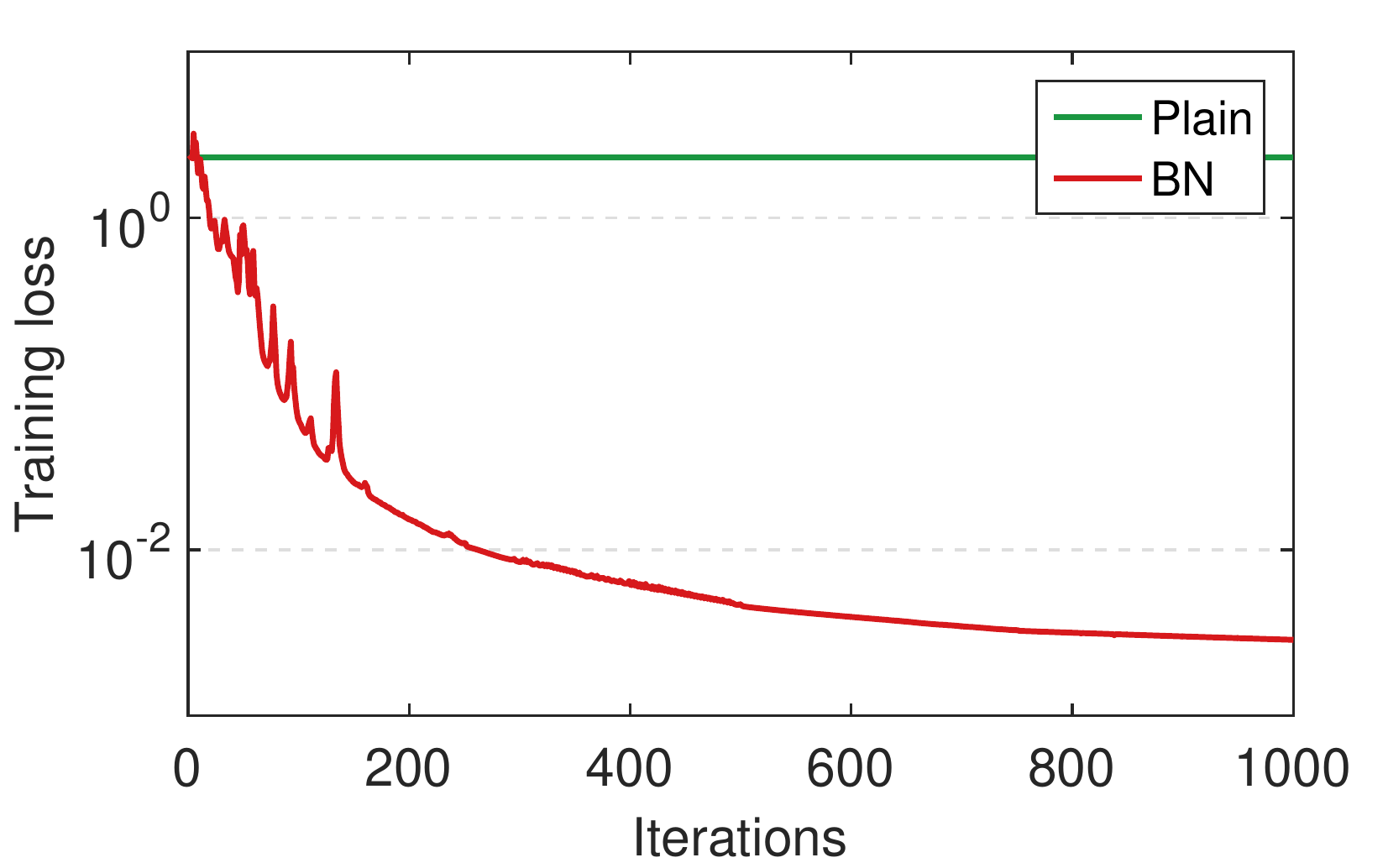}
		\end{minipage}
	}
	\\
	\vspace{-0.12in}
	
	\hspace{-0.3in}		\subfigure[$\lambda_{max}(\Sigma_x)$]{
		\begin{minipage}[c]{.3\linewidth}
			\centering
			\includegraphics[width=3.8cm]{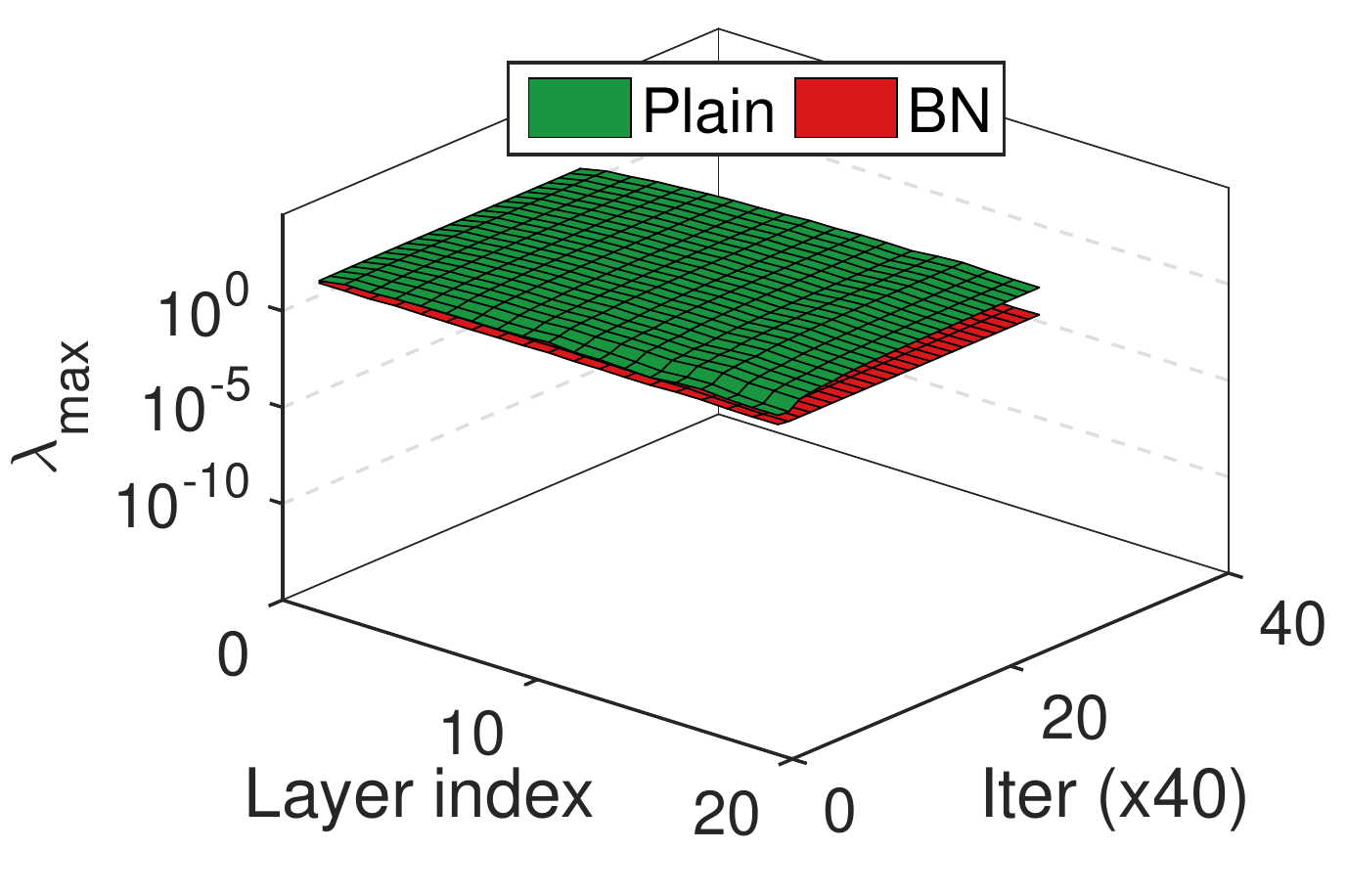}
		\end{minipage}
	}
	\subfigure[$\lambda_{max}(\Sigma_{\nabla \mathbf{h}})$]{
		\begin{minipage}[c]{.3\linewidth}
			\centering
			\includegraphics[width=3.8cm]{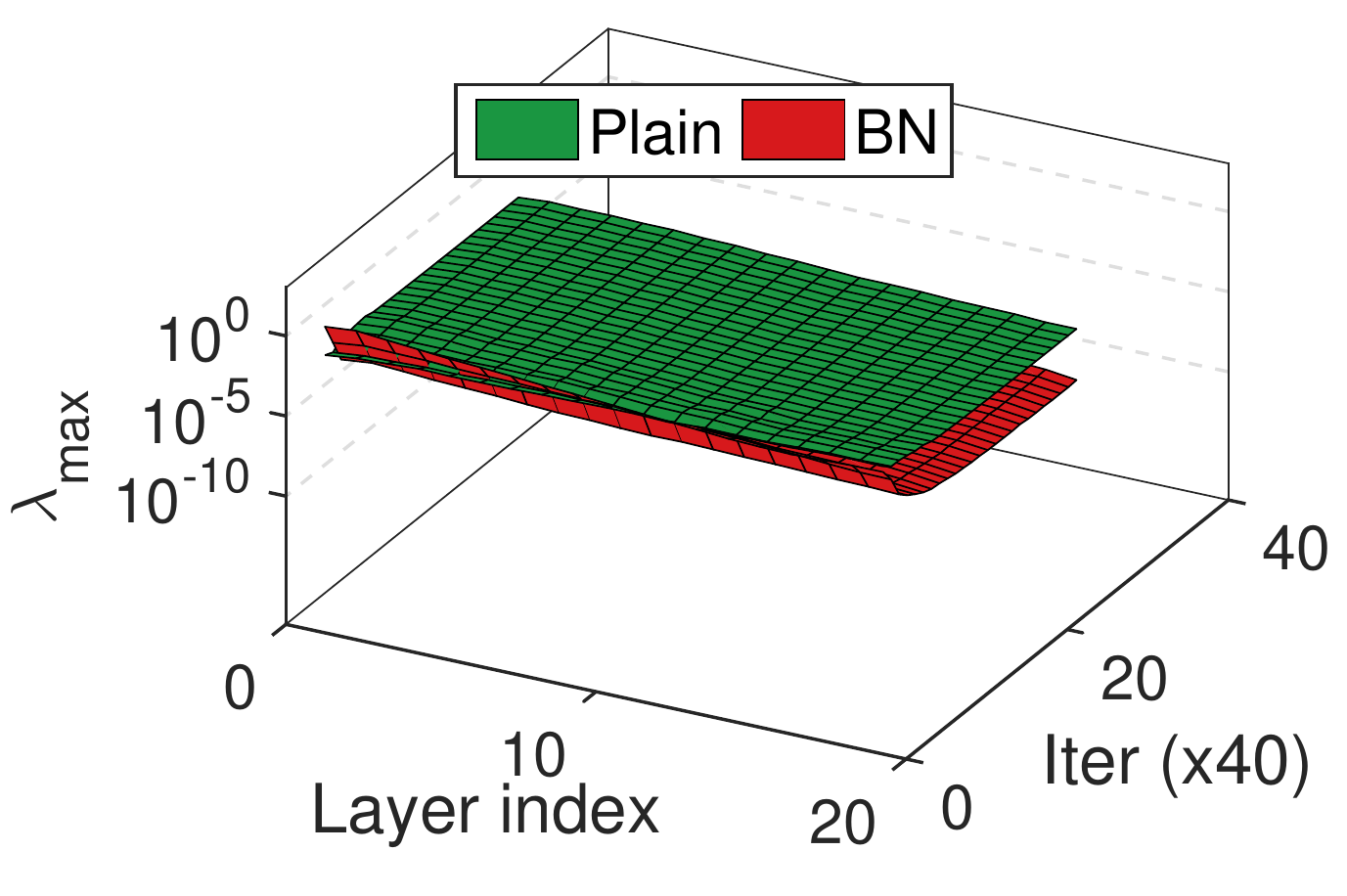}
		\end{minipage}
	}
	\subfigure[Training loss]{
		\begin{minipage}[c]{.3\linewidth}
			\centering
			\includegraphics[width=3.8cm]{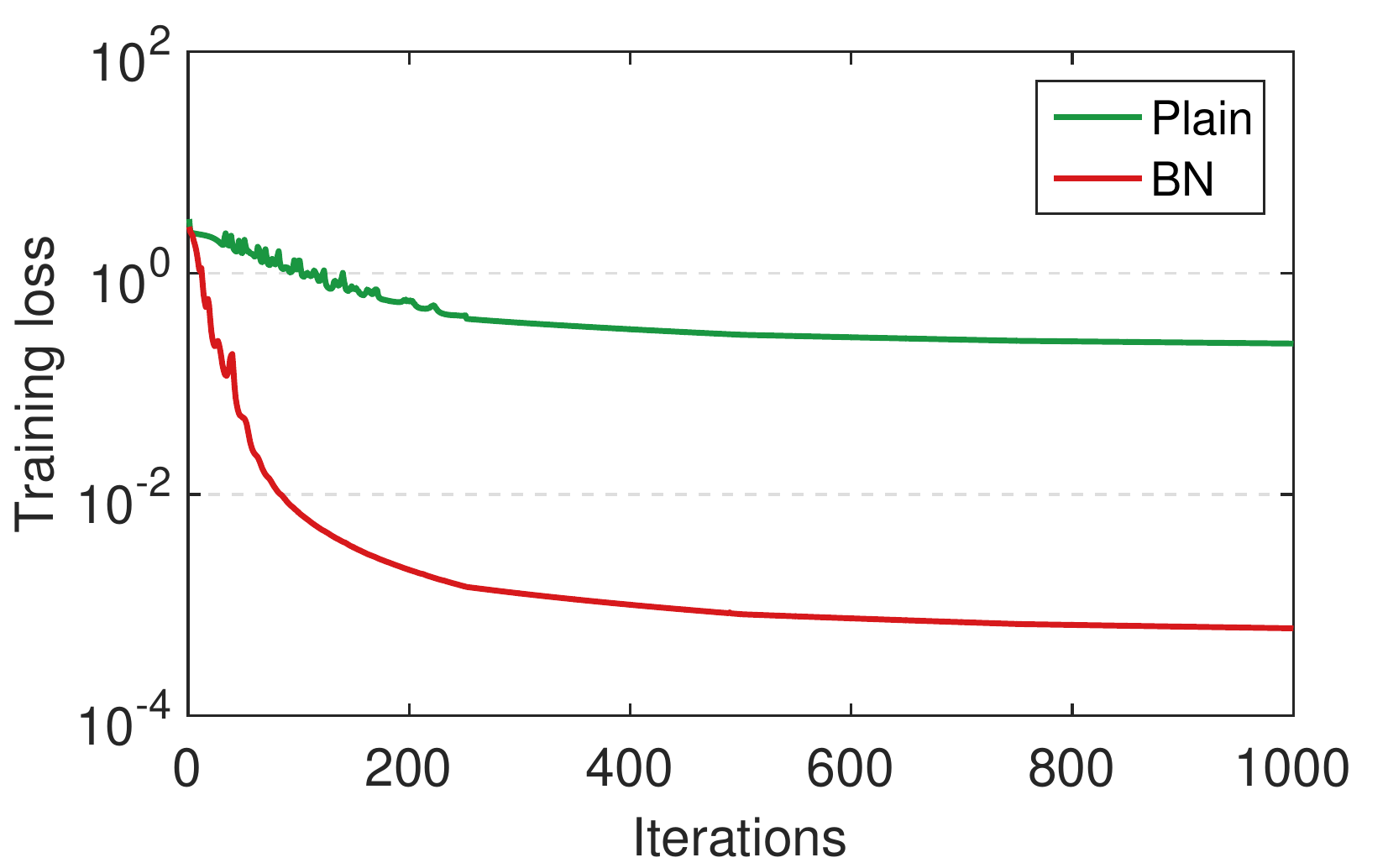}
		\end{minipage}
	}	
	\vspace{-0.14in}
	\caption{Analysis of the magnitude of the layer input (indicated by $\lambda_{max}(\Sigma_x)$) and layer output-gradient (indicated by $\lambda_{max}(\Sigma_{\nabla \mathbf{h}})$ ). The experiments are performed on a 20-layer MLP with 256 neurons in each layer, for MNIST classification. The results of (a)(b)(c) are under random initialization \cite{1998_NN_Yann}, while (d)(e)(f) He-initialization \cite{2015_ICCV_He}.}
	\label{fig:stablizeBN}
\end{figure*}
\paragraph{\textbf{Empirical Analysis}}
We further conduct experiments to show how the activation/gradient is affected by initialization in unnormalized DNNs (indicated as `Plain') and batch normalized DNNs (indicated as `BN'). 
We train a 20-layer MLP, with 256 neurons in each layer, for MNIST classification. The nonlinearity is ReLU.  We use the full gradient descent\footnote{We also perform SGD with a batch size of 1024, and further perform experiments on convolutional neural networks (CNNs) for CIFAR-10 \revise{and ImageNet} classification. The results are shown in \SM~\ref{Sec-sup-ExpBNSGD}, in which we have the same observation as the full gradient descent.}, and report the best training loss among learning rates in $\{0.05, 0.1, 0.5, 1\}$. In Figure~\ref{fig:stablizeBN} (a) and (b), we observe that the magnitude of the layer input (output-gradient) of `Plain' for random initialization \cite{1998_NN_Yann} suffers from exponential decrease during forward pass (backward pass). The main reason for this is that the weight has a small magnitude, based on Theorem \ref{th2:norm}.
This problem can be relieved by He-initialization \cite{2015_ICCV_He}, where the magnitude of the input/output-gradient is stable across layers  (Figure \ref{fig:stablizeBN} (d) and (e)).
We observe that BN  can well preserve the magnitude of the input/output-gradient across different layers for both initialization methods.

\begin{figure}[t]
	\centering
	\hspace{-0.3in}	\subfigure[Training loss]{
		\begin{minipage}[c]{.3\linewidth}
			\centering
			\includegraphics[width=3.8cm]{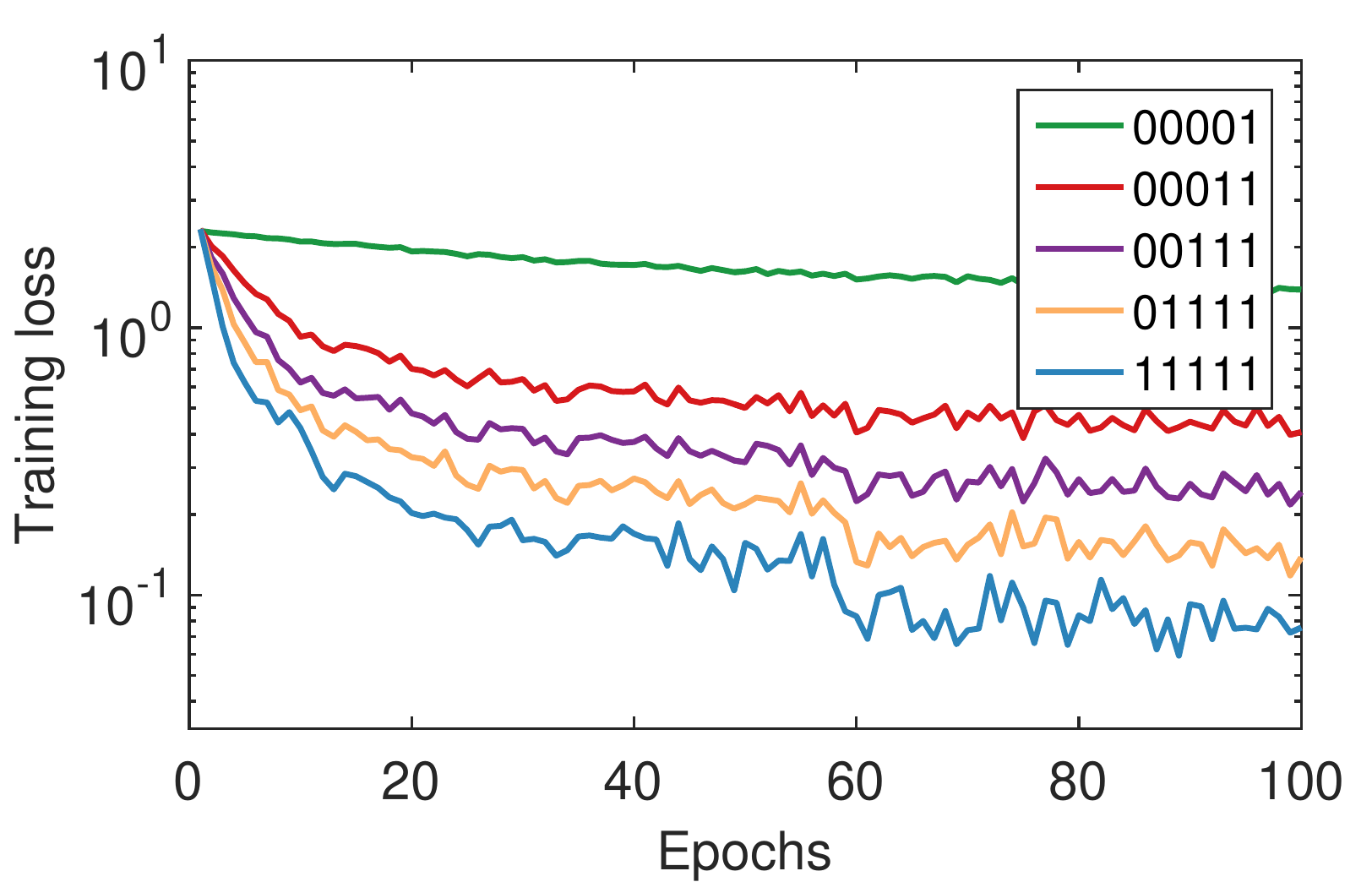}
		\end{minipage}
	}
	\hspace{0.03in}	\subfigure[Training error]{
		\begin{minipage}[c]{.3\linewidth}
			\centering
			\includegraphics[width=3.8cm]{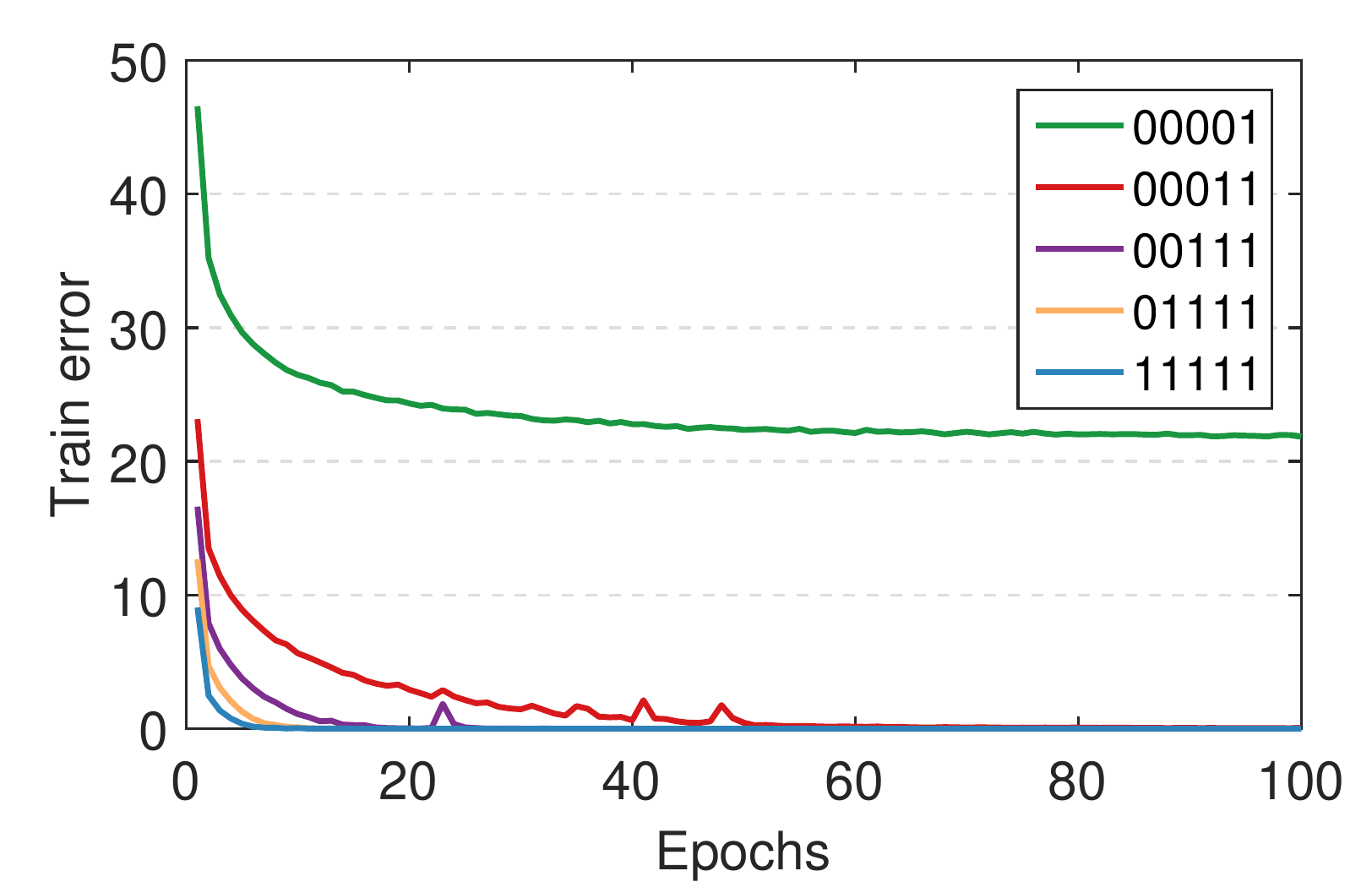}
		\end{minipage}
	}
	\hspace{0.03in}	\subfigure[Test error]{
		\begin{minipage}[c]{.3\linewidth}
			\centering
			\includegraphics[width=3.8cm]{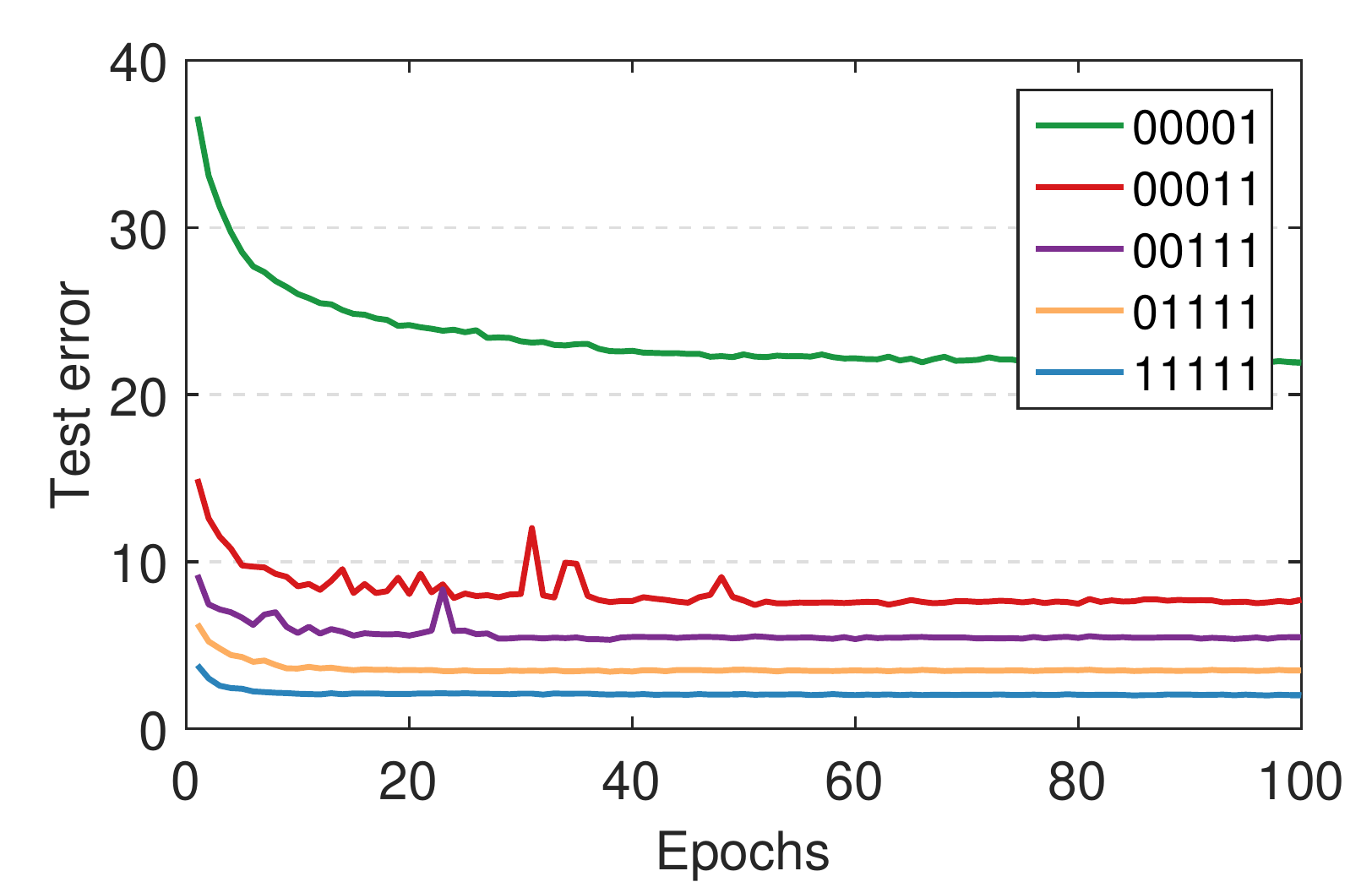}
		\end{minipage}
	}
	
	\vspace{-0.14in}
	\caption{Exploring the effectiveness of weight domination. We run the experiments on a 5-layer MLP with BN and the number of neuron in each layer is 256. We simulate weight domination in a given layer by blocking its weight updates. We denote `0' in the legend as the state of weight domination (the first digit represents  the first layer).}
	\label{fig:control}
\end{figure}
\vspace{-0.1in}
\paragraph{\textbf{Weight Domination}}
It was shown  the scale-invariant property of BN has an implicit early stopping effect on the weight matrices ~\cite{2016_CoRR_Ba} ,  helping to stabilize learning towards
convergence.
\revise{Here, we show that this layer-wise `early stopping' sometimes results in the false impression of a local minimum, which has detrimental effects on the learning, since the network does not well learn the representation in the corresponding layer.}
For illustration, we provide a rough definition termed \textit{weight domination}, with respect to a given layer.
	\vspace{-0.02in}
\begin{small}
\begin{defn}
	\label{def1:wd}
	Let $\W{k}$ and $\D{\W{k}}$ be the weight matrix and its gradient in layer $k$. If $\lambda_{max}(\D{\W{k}}) \ll \lambda_{max}(\W{k}) $, where $\lambda_{max}(\cdot)$ indicates the maximum singular value of a matrix,
	we refer to layer $k$ has a state of \textbf{weight domination}.
\end{defn}
\end{small}
	\vspace{-0.02in}
Weight domination implies a smoother gradient with respect to the given layer. This is a desirable property for linear models (the distribution of the input is fixed), where the optimization objective targets to arrive the stationary points with smooth (zero) gradient. However, weight domination is not always desirable for a given layer of a DNN, since such a state of one layer is possibly caused by the increased magnitude of the weight matrix or decreased magnitude of the layer input (the non-convex optimization in Eqn.~\ref{eqn:GradientMatch-Linear}), not necessary driven by the optimization objective itself.
 Although BN ensures a stable distribution of layer inputs, a network with BN still has the possibility that the   magnitude of the weight in a certain layer is significantly increased. We experimentally observe this phenomenon, as shown in the \SM~\ref{Sec-sup-ExpBNWD}. A similar phenomenon is also observed in \cite{2019_ICLR_Yang}, where BN results in large updates of the corresponding weights.

Weight domination sometimes harms the learning of the network, because this state limits its ability to \revise{learn the representation in} the corresponding layer. To investigate this, we conduct experiments on a 5-layer MLP and show the results in Figure \ref{fig:control}.
We observe that the network with weight domination in certain layers, can still decrease the loss, but has degenerated performance.
We also conduct experiments on CNNs for CIFAR-10 datasets, shown in the \SM~\ref{Sec-sup-ExpBNWD}.

\begin{figure*}[t]
		\vspace{-0.02in}
	\centering
	\hspace{-0.2in}		\subfigure[$\kappa_{50\%}(\Sigma_x)$]{
		\begin{minipage}[c]{.22\linewidth}
			\centering
			\includegraphics[width=3.0cm]{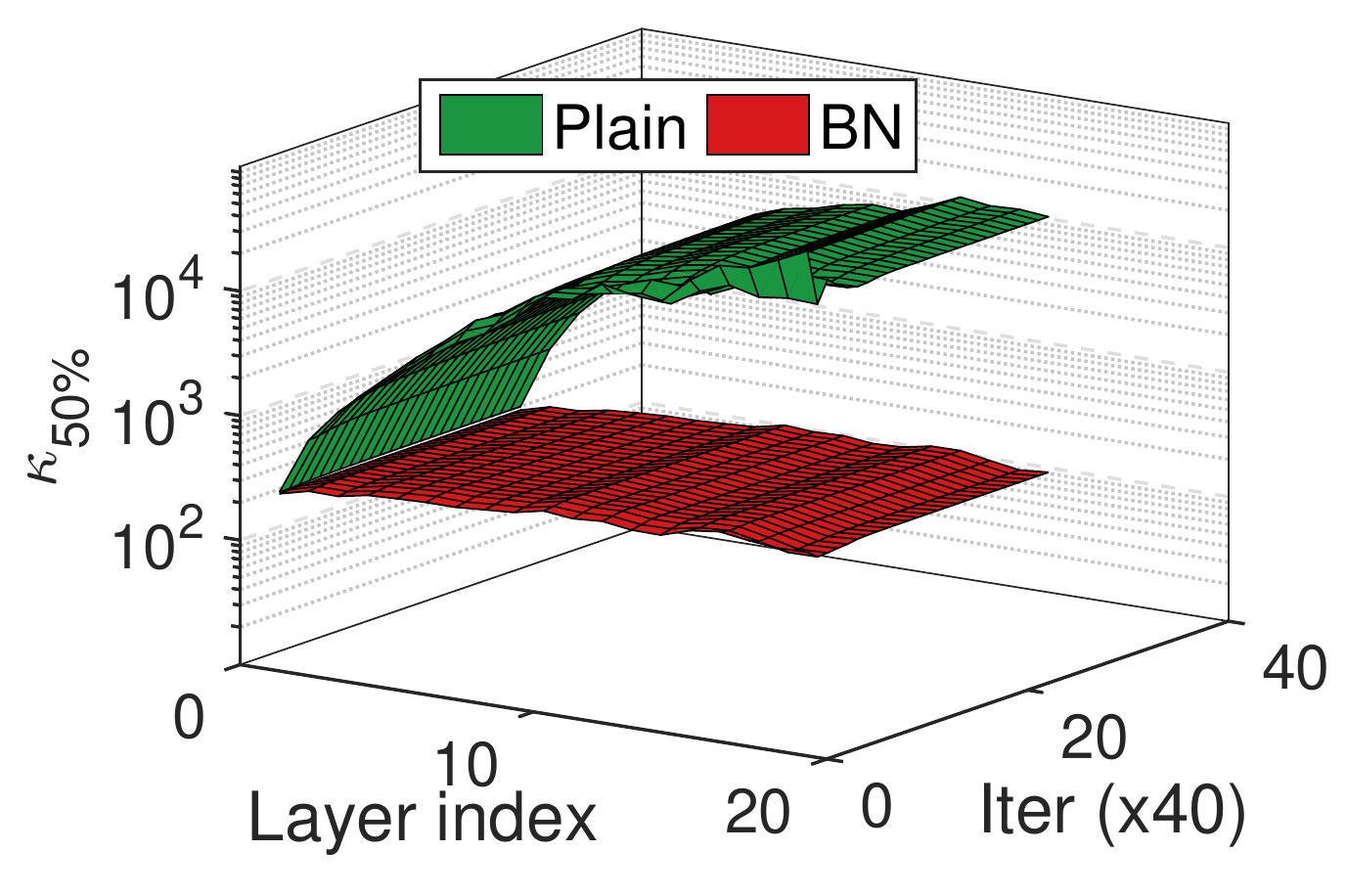}
		\end{minipage}
	}
	\hspace{0.03in}	\subfigure[$\kappa_{50\%}(\Sigma_{\nabla \mathbf{h}})$]{
		\begin{minipage}[c]{.22\linewidth}
			\centering
			\includegraphics[width=3.0cm]{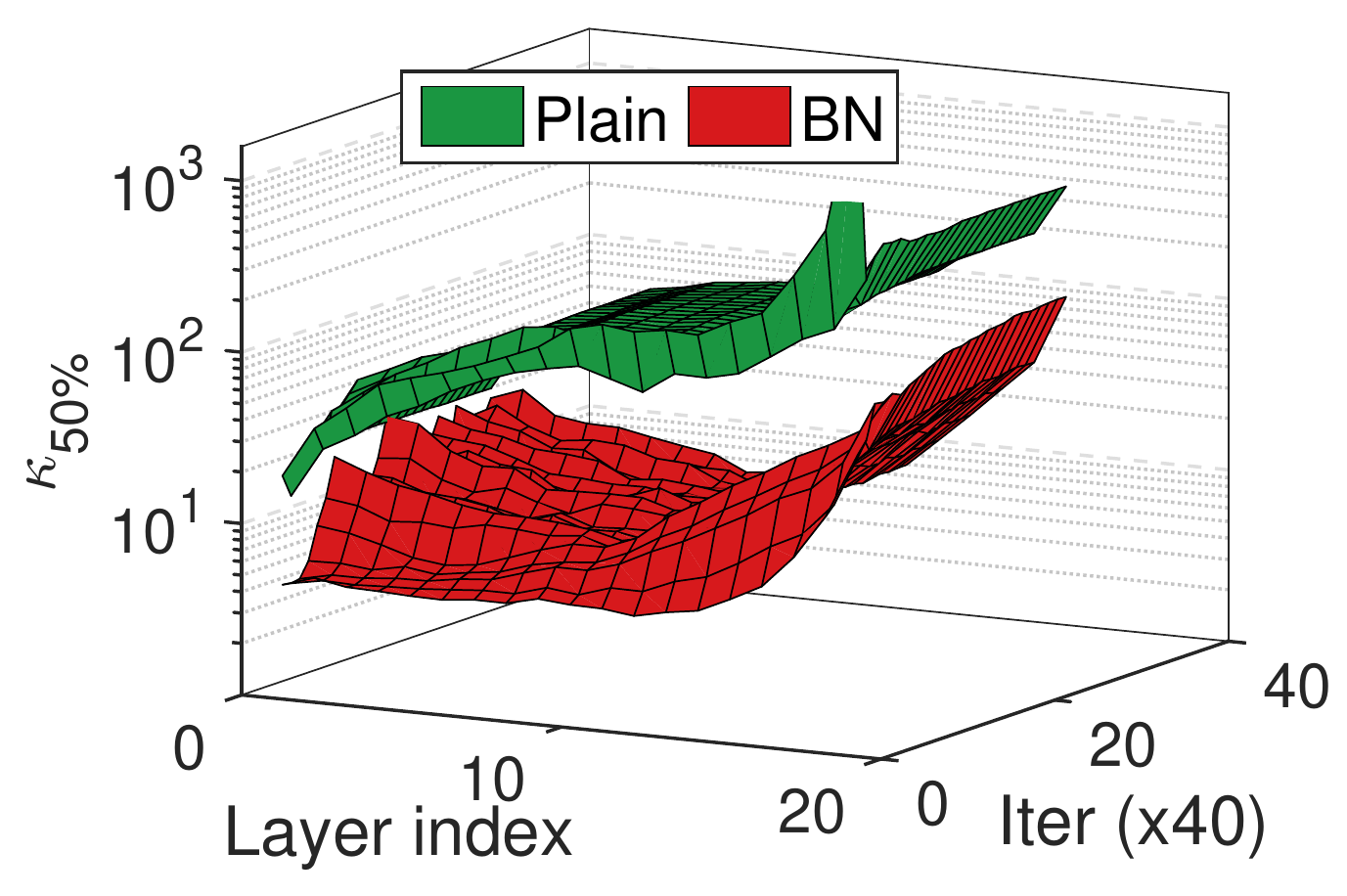}
		\end{minipage}
	}
	\hspace{0.03in}	\subfigure[$\kappa_{80\%}(\Sigma_x)$]{
		\begin{minipage}[c]{.22\linewidth}
			\centering
			\includegraphics[width=3.0cm]{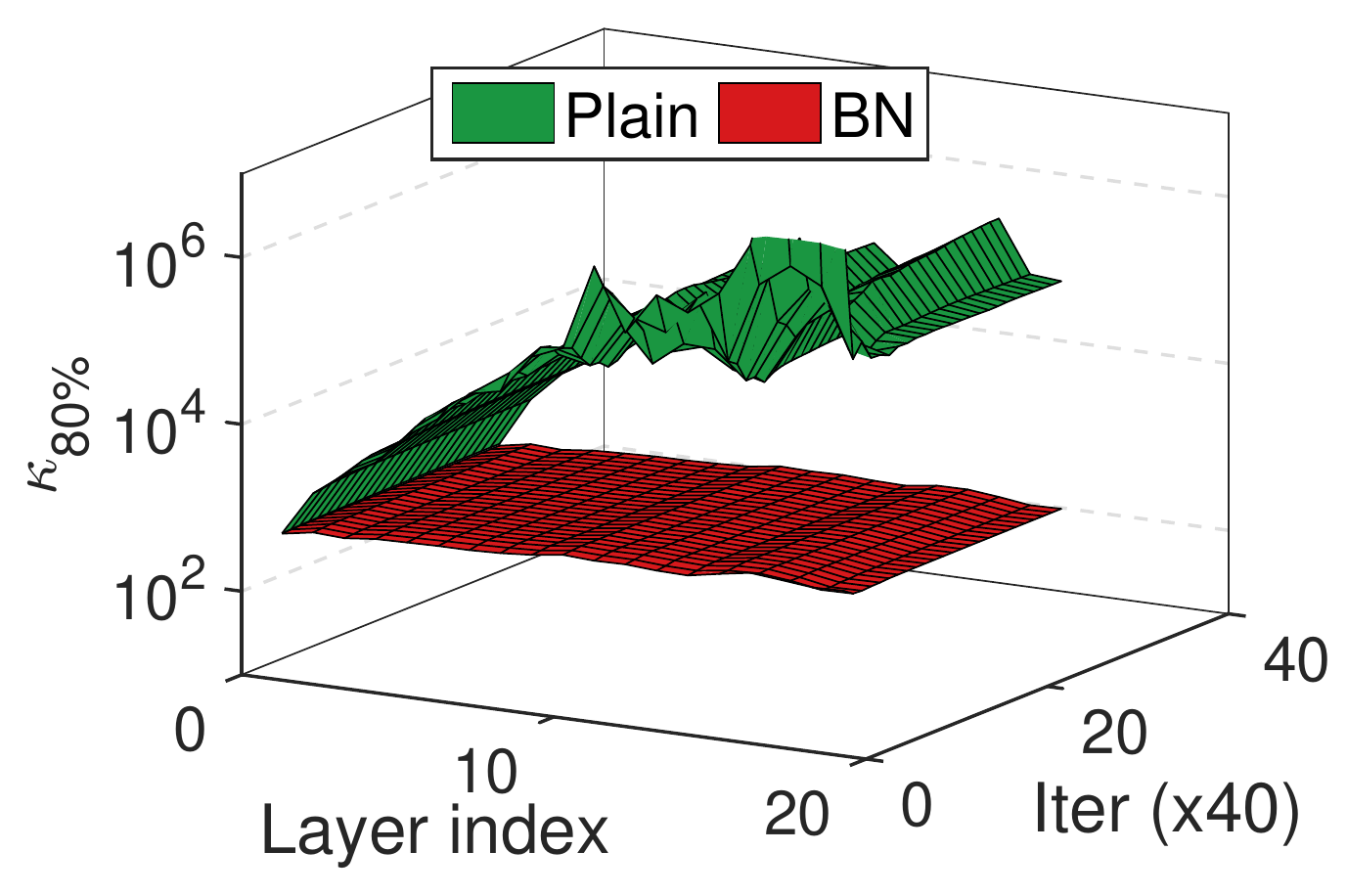}
		\end{minipage}
	}
	\hspace{0.03in}	\subfigure[$\kappa_{80\%}(\Sigma_{\nabla \mathbf{h}})$]{
		\begin{minipage}[c]{.22\linewidth}
			\centering
			\includegraphics[width=3.0cm]{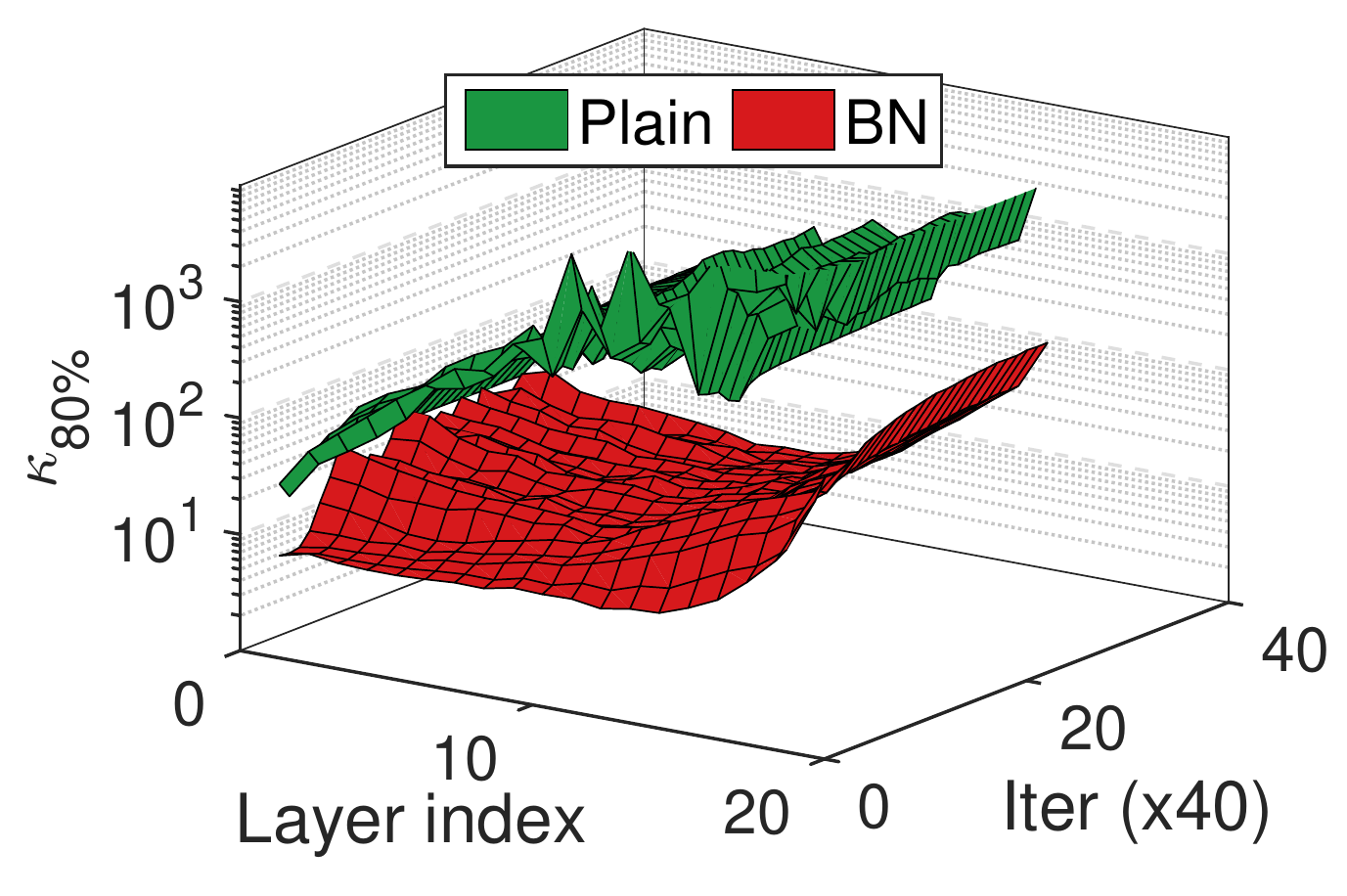}
		\end{minipage}
	}
	\vspace{-0.14in}
	\caption{Analysis on the condition number of the layer input ( $\kappa_{p}(\Sigma_x)$) and layer output-gradient ( $\kappa_{p}(\Sigma_{\nabla \mathbf{h}})$ ). The experimental setups are the same as in Figure \ref{fig:stablizeBN}.}
	\label{fig:conditionNumber}
\end{figure*}

\vspace{-0.12in}
\subsection{Improved Conditioning}
\label{Sec-BN-Accelerate}
\vspace{-0.05in}
One motivation behind BN is that whitening the input can improve the conditioning of the optimization \cite{2015_ICML_Ioffe}  (\eg, the Hessian will be an identity matrix under the condition that $\mathbb{E}_{\mathcal{D}}(\mathbf{x} \mathbf{x}^T)= \mathbf{I}  $ for a linear model, based on Eqn.~\ref{eqn:Hessian_LR}, and thus can accelerate training \cite{2015_NIPS_Desjardins,2018_CVPR_Huang}. However, such a motivation is seldom validated  by either theoretical or empirical analysis on the context of DNNs \cite{2015_NIPS_Desjardins,2018_NIPS_shibani}. Furthermore, it only holds under the condition that BN is placed before the linear layer, while, in practice, BN is typically placed after the linear layer, as recommended in \cite{2015_ICML_Ioffe}.
In this section, we will empirically explore this motivation
using our layer-wise conditioning analysis for the scenario of training DNNs.


We first experimentally observe that BN not only improves the conditioning of the layer input's covariance matrix, but also improves the conditioning of the \revise{output-gradient's covariation}, as shown in Figure \ref{fig:conditionNumber}.
It has been shown that centered data is more likely to be well-conditioned \cite{1990_NIPS_LeCun,1998_Schraudolph,2012_NN_Gregoire,2017_ICCV_Huang}. This suggests that placing BN after the linear layer can improve the conditioning of the output-gradient, because centering the activation, with the gradient back-propagating through such a transformation \cite{2015_ICML_Ioffe},  also centers the gradient.

We also observe that the unnormalized network (`Plain') has several small eigenvalues.
For further exploration, we monitor the output of each neuron in each layer, and find that `Plain' has some neurons that are not activated (zero output of ReLU) for \revise{all training examples}. We refer to these neurons as \textit{dying neurons}. We also observe that  `Plain' has some neurons that are always activated for every training example, which we refer to  as \textit{full neurons}.
This observation is most obvious in the initial iterations.  The number of dying/full neurons increases as the layer number increases (Please refer to \SM~\ref{Sec-sup-dyingNode} for details). We conjecture that the dying neurons causes `Plain' to have numerous small/zero eigenvalues. In contrast, batch normalized networks have no dying/full neurons, because the centering operation ensures that half the examples get activated. This further suggests that placing BN before the nonlinear activation can improve the conditioning.

\section{Training Very Deep Residual Networks}
\label{Sec:veryDeep}
\vspace{-0.05in}

Residual networks \cite{2015_CVPR_He} have significantly relieved the  difficulty of training deep networks by their introduction of the residual connection, which makes training networks with hundreds or even thousands of layers possible.
However, residual networks also suffer from degenerated performance when the model is extremely deep (\eg, the 1202-layer residual network has worse performance than the 110-layer one), as shown in \cite{2015_CVPR_He}. He \etal \cite{2015_CVPR_He} argued that this is from over-fitting, not optimization difficulty. Here, we show that a very deep residual network may also suffer difficulty in optimization.

\begin{figure*}[t]
	\centering
	\hspace{-0.2in}		\subfigure[Training loss]{
		\begin{minipage}[c]{.22\linewidth}
			\centering
			\includegraphics[width=3.0cm]{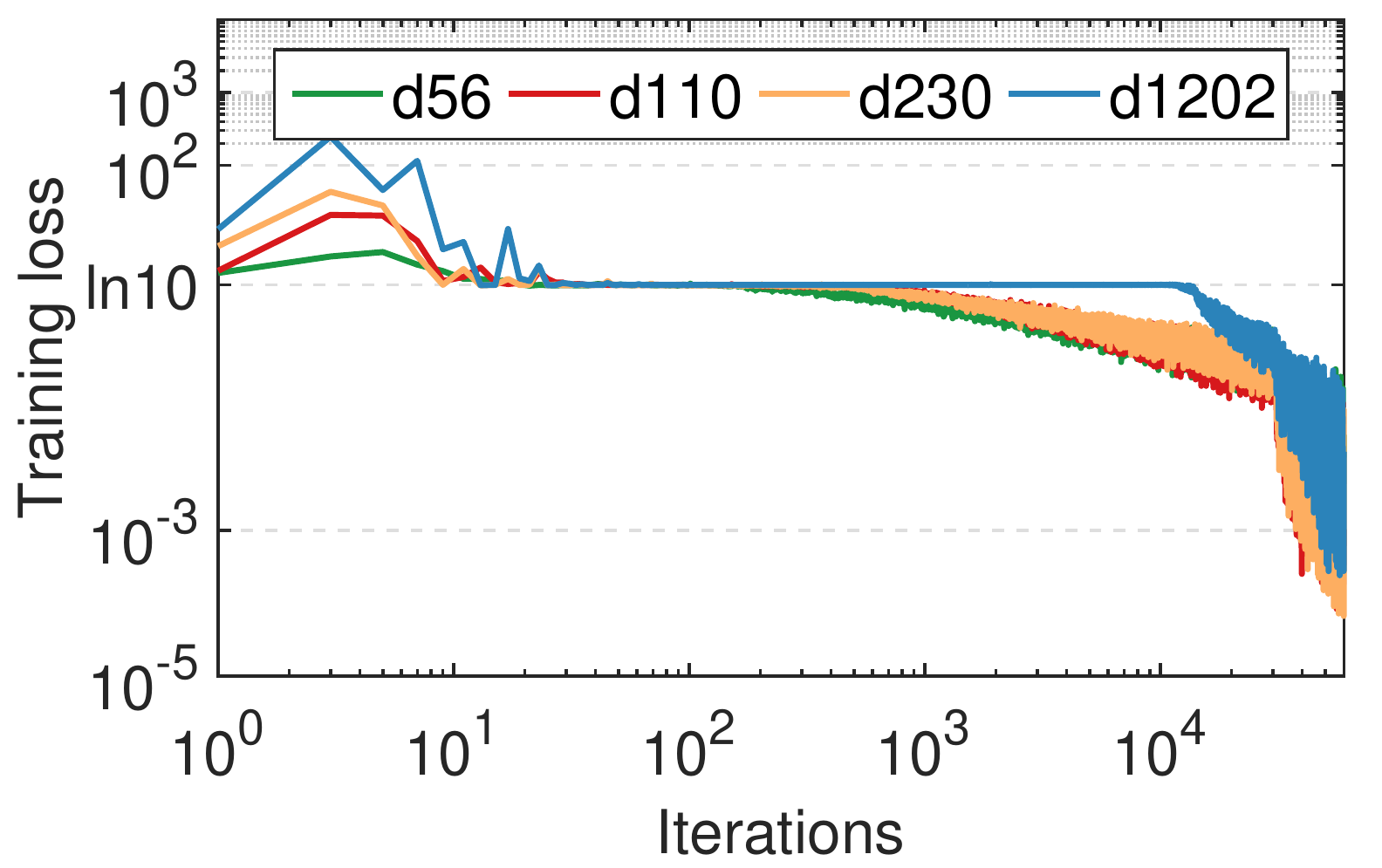}
		\end{minipage}
	}
	\hspace{0.03in}	\subfigure[$\lambda_{\Sigma_{\mathbf{x}}}$]{
		\begin{minipage}[c]{.22\linewidth}
			\centering
			\includegraphics[width=3.0cm]{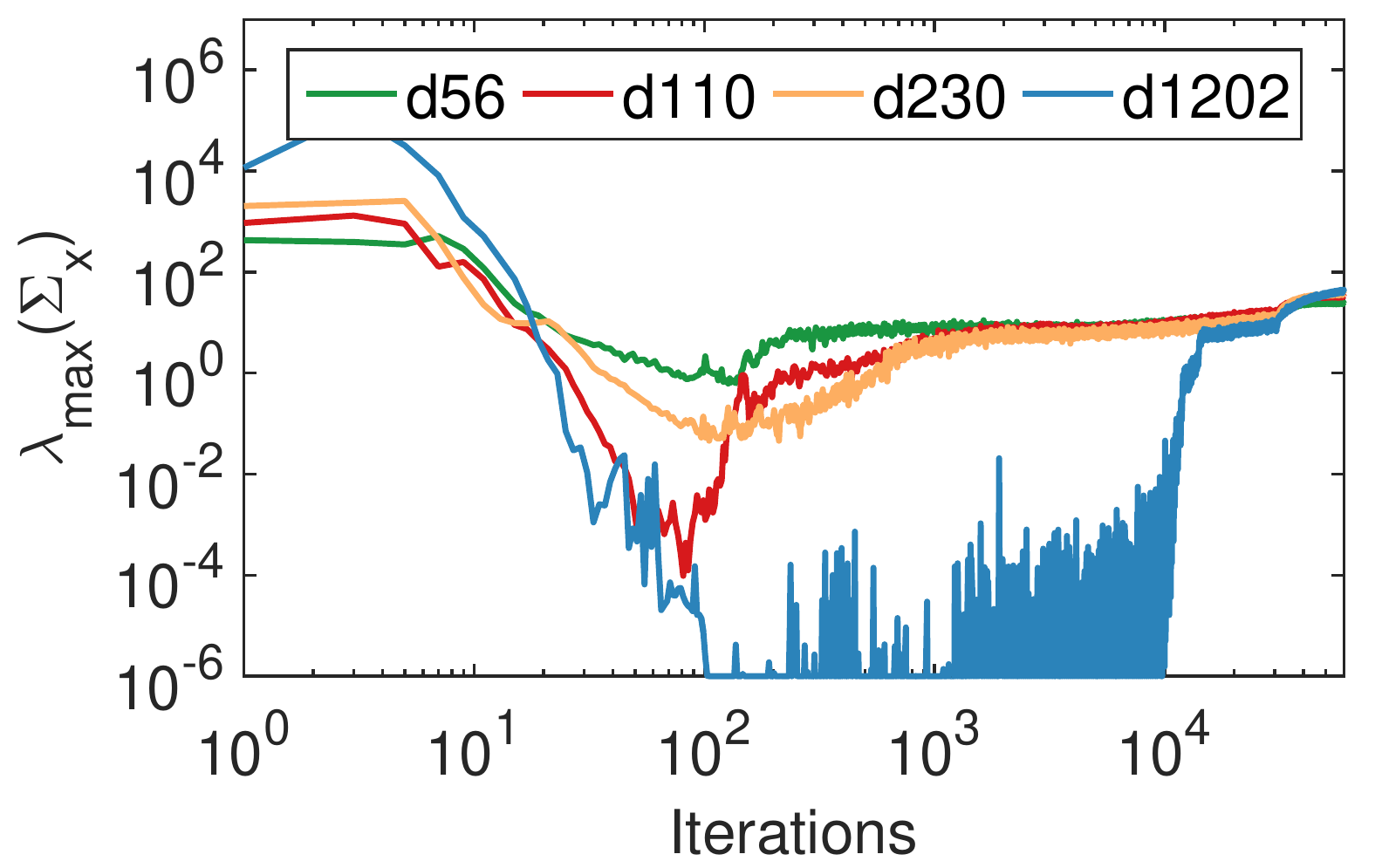}
		\end{minipage}
	}
	\hspace{0.03in}	\subfigure[$\lambda_{\Sigma_{\D{\W{}}}}$]{
		\begin{minipage}[c]{.22\linewidth}
			\centering
			\includegraphics[width=3.0cm]{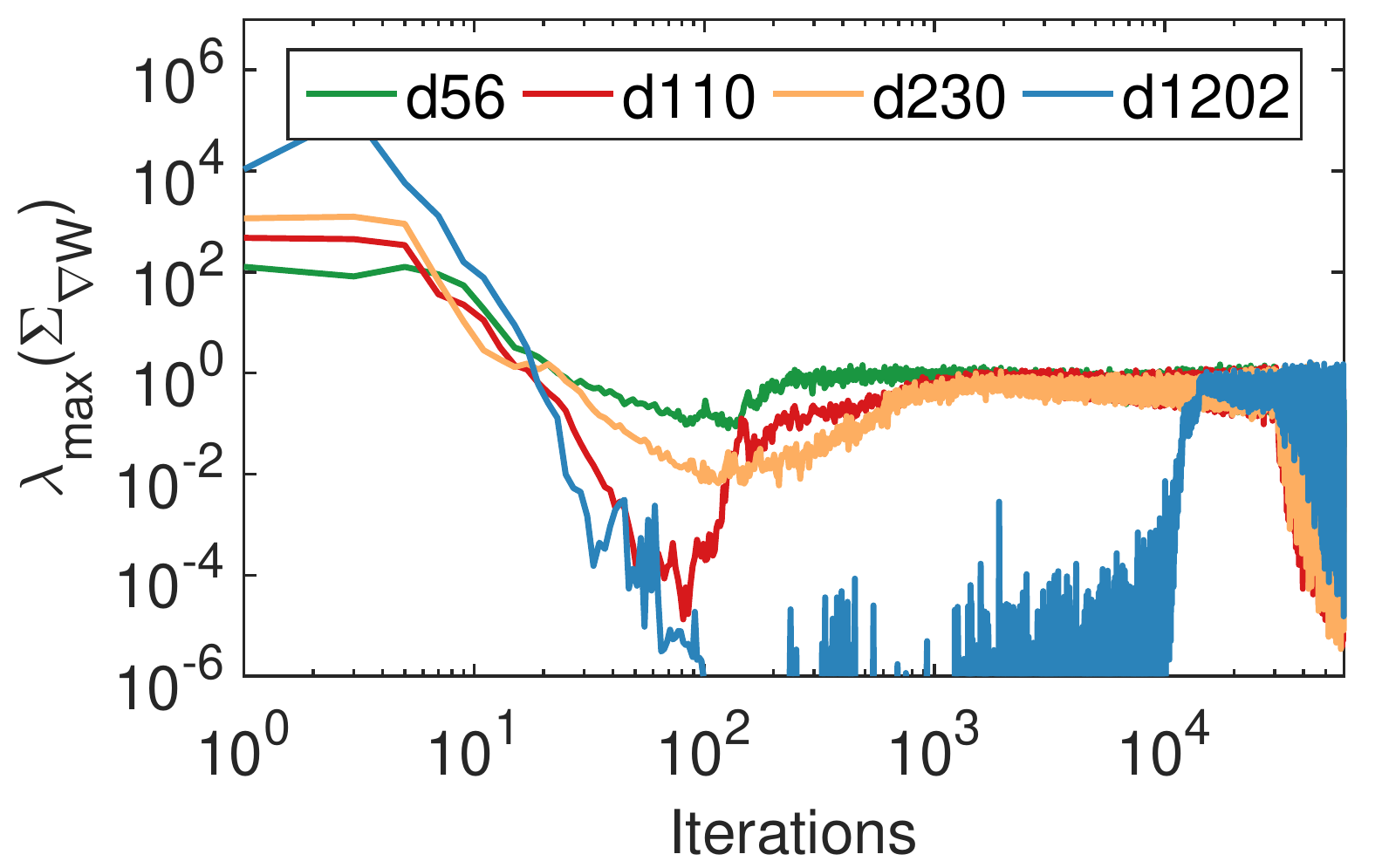}
		\end{minipage}
	}
	\hspace{0.03in}	\subfigure[$\|\mathbf{W} \|_2$]{
		\begin{minipage}[c]{.22\linewidth}
			\centering
			\includegraphics[width=3.0cm]{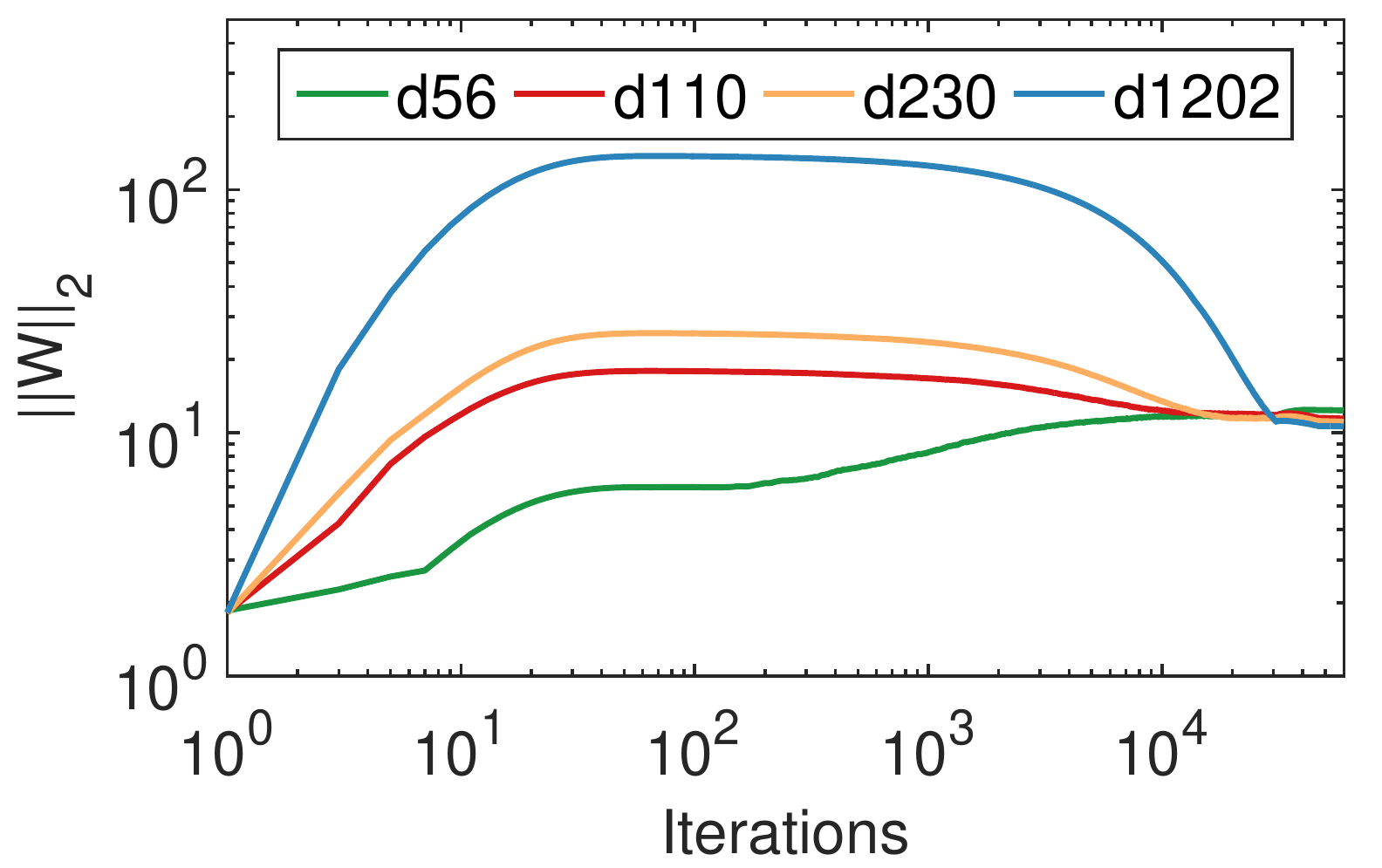}
		\end{minipage}
	}
	\vspace{-0.2in}
	\caption{Analysis on the last linear layer in residual networks for  CIFAR-10 classification. We vary the depth ranging in $\{56, 110, 230, 1202\}$ and analyze the results over the course of training. We show (a) the training loss; (b) the maximum eigenvalue of the input's covariance matrix; (c) the maximum eigenvalue of the  second moment matrix of the weight-gradient; and (d) the F2-norm of the weight.  Note that both the x- and y-axes are in log scale.}
	\label{fig:problem}
\end{figure*}

We perform experiments on CIFAR-10 with residual networks, following the same experimental setup as in \cite{2015_CVPR_He}\footnote{We use the Torch implementation: https://github.com/facebook/fb.resnet.torch}, except that we run the experiments on one GPU.
We vary the network depth, ranging in $\{56, 110, 230, 1202\}$,  and show the training loss in Figure \ref{fig:problem} (a).
We observe the residual networks have an increased loss in the initial iterations, which is amplified for deeper networks. Later, the training gets stuck in a state of randomly guessing (the loss stays at $\ln 10$). Although the networks can escape such a state with enough iterations, they suffer from degenerated training performance, especially if they are very deep.

\begin{figure*}[t]
	\centering
	\hspace{-0.2in}		\subfigure[Training loss]{
		\begin{minipage}[c]{.22\linewidth}
			\centering
			\includegraphics[width=3.0cm]{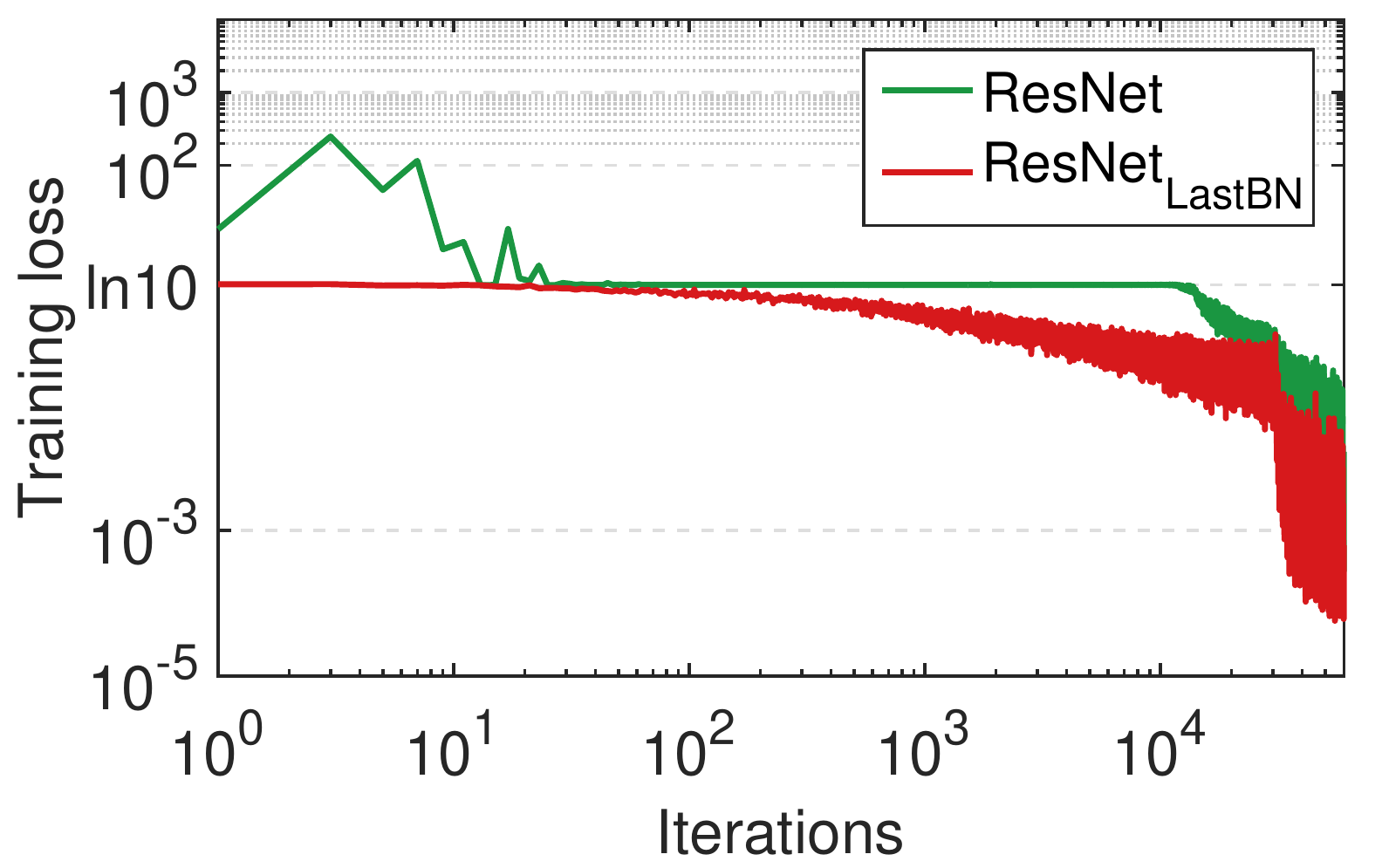}
		\end{minipage}
	}
	\hspace{0.03in}	\subfigure[$\lambda_{\Sigma_{\mathbf{x}}}$]{
		\begin{minipage}[c]{.22\linewidth}
			\centering
			\includegraphics[width=3.0cm]{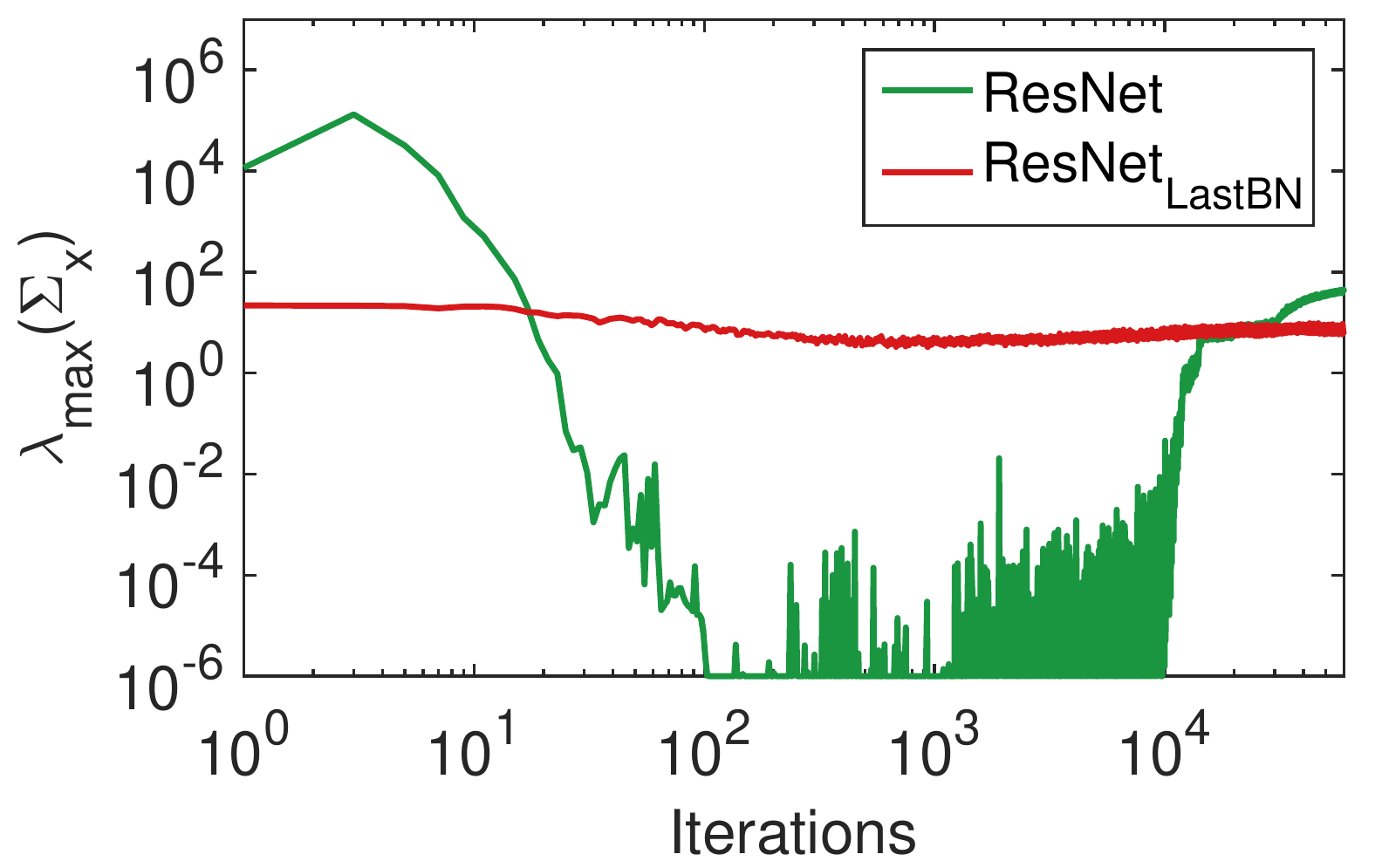}
		\end{minipage}
	}
	\hspace{0.03in}	\subfigure[$\lambda_{\Sigma_{\D{\W{}}}}$]{
		\begin{minipage}[c]{.22\linewidth}
			\centering
			\includegraphics[width=3.0cm]{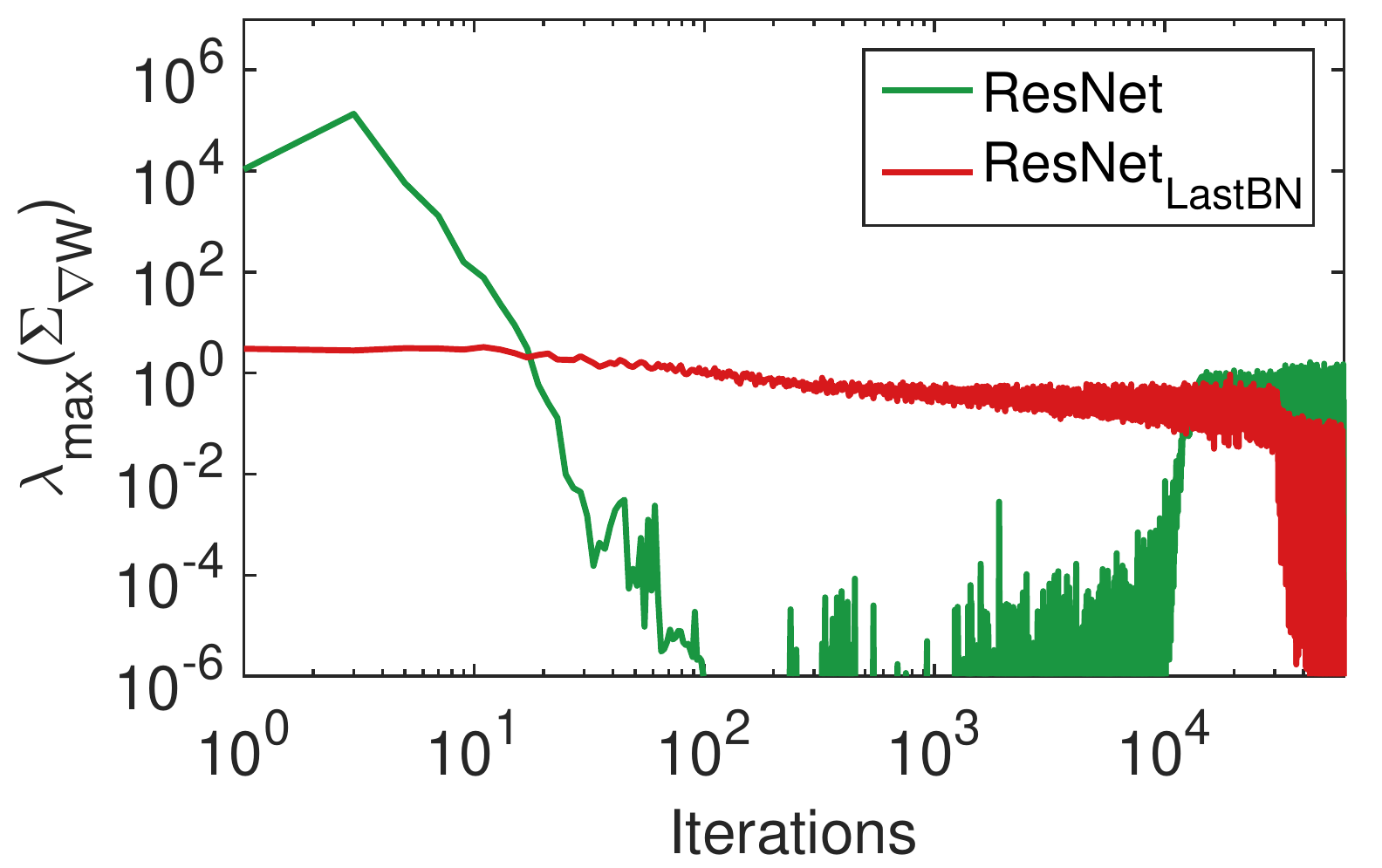}
		\end{minipage}
	}
	\hspace{0.03in}	\subfigure[$\|\mathbf{W} \|_2$]{
		\begin{minipage}[c]{.22\linewidth}
			\centering
			\includegraphics[width=3.0cm]{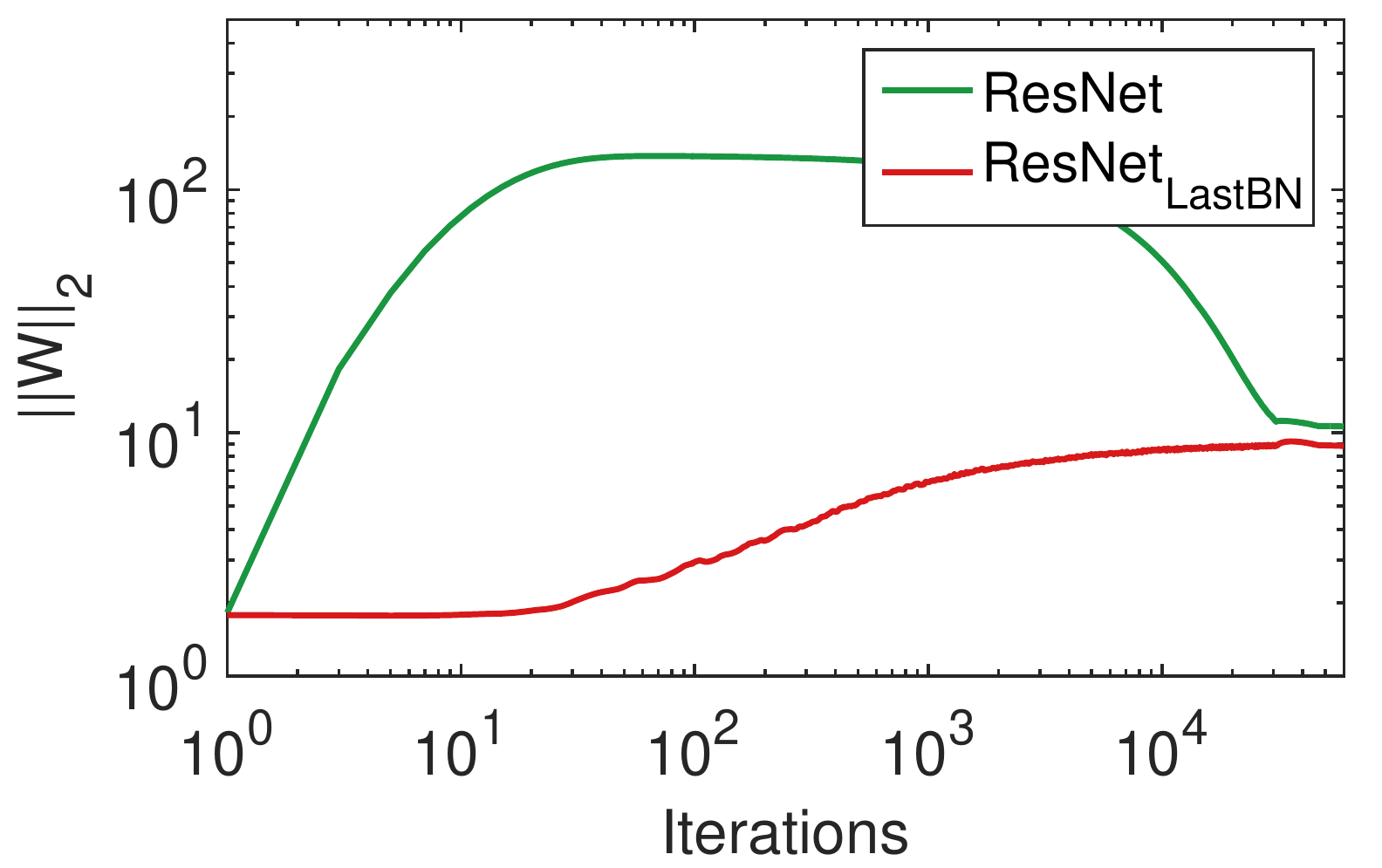}
		\end{minipage}
	}
	\vspace{-0.2in}
	\caption{Analysis of how ResNet$_{LastBN}$ solves the ill-conditioned problem of its last linear layer on the 1202-layer network for CIFAR-10 classification.}
	\label{fig:problem-LastBN}
\end{figure*}

\vspace{-0.08in}
\paragraph{\textbf{Analysis of Learning Dynamics}} 
To explore why residual networks have such a mysterious behavior, we perform the layer-wise conditioning analysis on the last linear layer (before the cross entropy loss).
We monitor  the maximum eigenvalue of the covariance matrix $\lambda_{\Sigma_{\mathbf{x}}}$, the maximum eigenvalue of the second moment matrix of the weight-gradient  $\lambda_{\Sigma_{\D{\W{}}}}$, and the norm of the weight ($\|\mathbf{W} \|_2$).

We observe that the initial increase in loss is mainly caused by the large magnitude of $\lambda_{\Sigma_{\mathbf{x}}}$\footnote{The large magnitude of $\lambda_{\Sigma_{\mathbf{x}}}$ is caused mainly by the addition of multiple residual connections from the previous layers with ReLU output.} (Figure \ref{fig:problem} (b)), which results in a large magnitude for  $\lambda_{\Sigma_{\D{\W{}}}}$ \revise{(Figure \ref{fig:problem} (c))}, and thus a large magnitude for $\|\mathbf{W} \|_2$ (Figure \ref{fig:problem} (d)).
The increased $\|\mathbf{W} \|_2$ further facilities the increase of the loss.
However, the learning objective is to decrease the loss, and thus it should decrease the magnitude of $\W{}$ or $\x{}$ (based on Eqn.~\ref{eqn:GradientMatch-Linear}) in this case. Apparently, $\W{}$ is harder to adjust, because the landscape of its loss surface is controlled by $\x{}$, and all the values of $\x{}$ are non-negative with large magnitude. The network thus tries to decrease $\x{}$ based on the given learning objective. We experimentally find that the learnable parameters of BN have a large number of negative values, which causes the ReLUs (positioned after the residual adding operation) deactivated. Such a dynamic results in a significant reduction in the magnitude of  $\lambda_{\Sigma_{\mathbf{x}}}$. The small $\x{}$ and large $\W{}$ drive the last linear layer of the network into the state of weight domination, and make the network display a random guess behavior.  Although the residual network can escape such a state with enough iterations, the weight domination hinders optimization and results in degenerated training performance.

	\vspace{-0.14in}
\subsection{Proposed Solution}
	\vspace{-0.03in}
Based on the above analysis, it is essential to reduce the large magnitude of $\lambda(\Sigma_{\x{}})$.
We propose  a simple solution and add one BN layer before the last linear layer to normalize its input.  We refer to this residual network as `ResNet$_{LastBN}$', and the original one as `ResNet'.
We also conduct an analysis on the last linear layer of ResNet$_{LastBN}$, providing a comparison between ResNet and  ResNet$_{LastBN}$ on the 1202-layer  in Figure \ref{fig:problem-LastBN}. We observe that ResNet$_{LastBN}$ can be steadily trained. It does not reach the state of weight domination or encounter a  large magnitude of $\x{}$ in the last linear layer.

We try a similar solution where a constant is divided before the linear layer, and we find it also benefits the training. However, the main disadvantage of this solution is that the value of the constant has to be finely tuned on networks with different depths.
We also try putting one BN before the average pooling, which has similar effects as putting it before the last linear layer. We  note that Bjorck \etal \cite{2018_NIPS_Bjorck} proposed to train a 110-layer residual network with only one BN layer, which is placed before the average pooling. They showed that this achieves good results. However, we argue that this does not hold for very deep networks. We perform an experiment on the 1202-layer residual network, and  find that the model always fails in training with various hyper-parameters.

\begin{figure}[t]
	\centering
	\vspace{-0.05in}
	\hspace{-0.25in}	\subfigure[ResNet]{
		\begin{minipage}[c]{.35\linewidth}
			\centering
			\includegraphics[width=4.5cm]{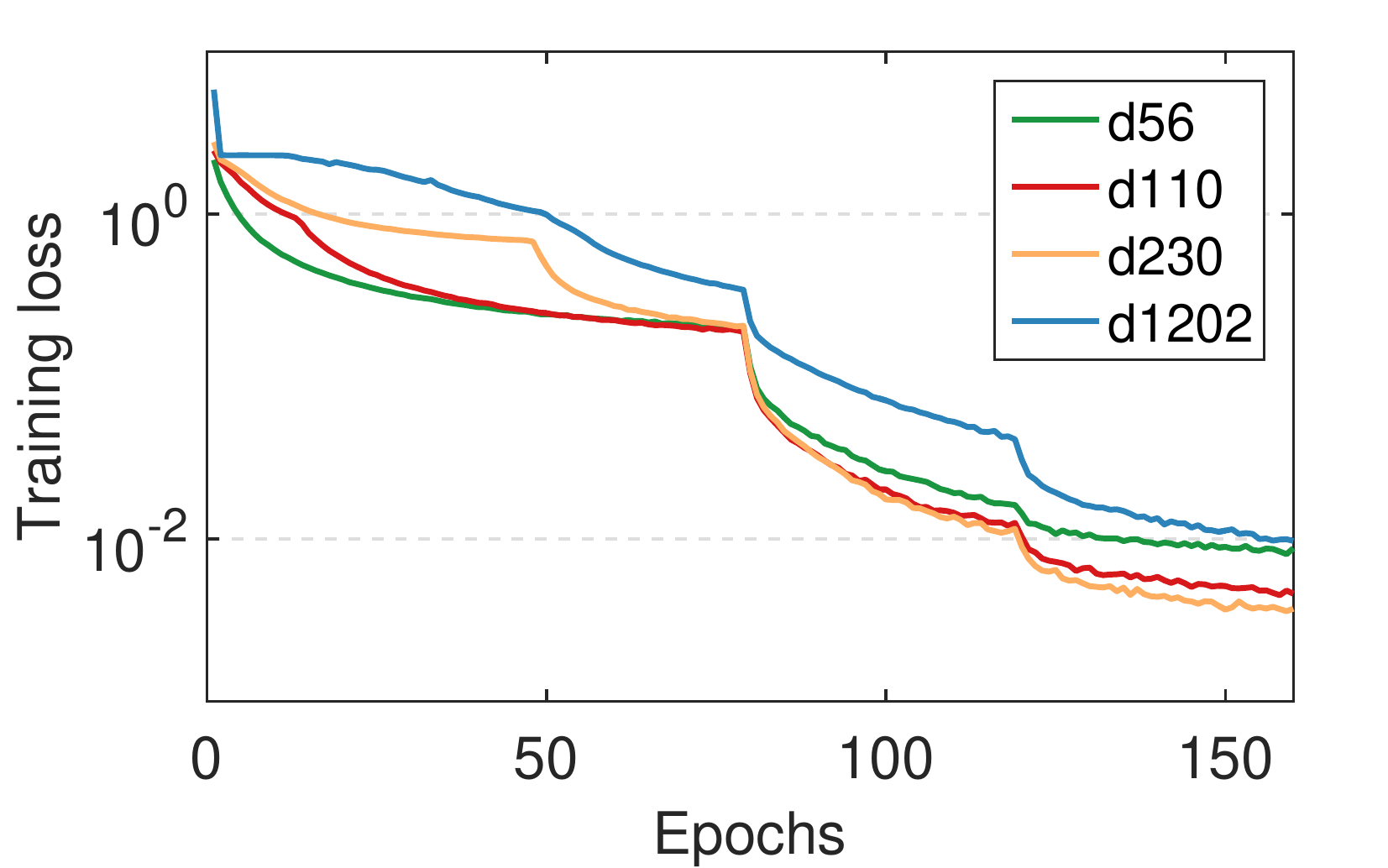}
		\end{minipage}
	}
	\hspace{0.15in}		\subfigure[ResNet$_{LastBN}$]{
		\begin{minipage}[c]{.35\linewidth}
			\centering
			\includegraphics[width=4.5cm]{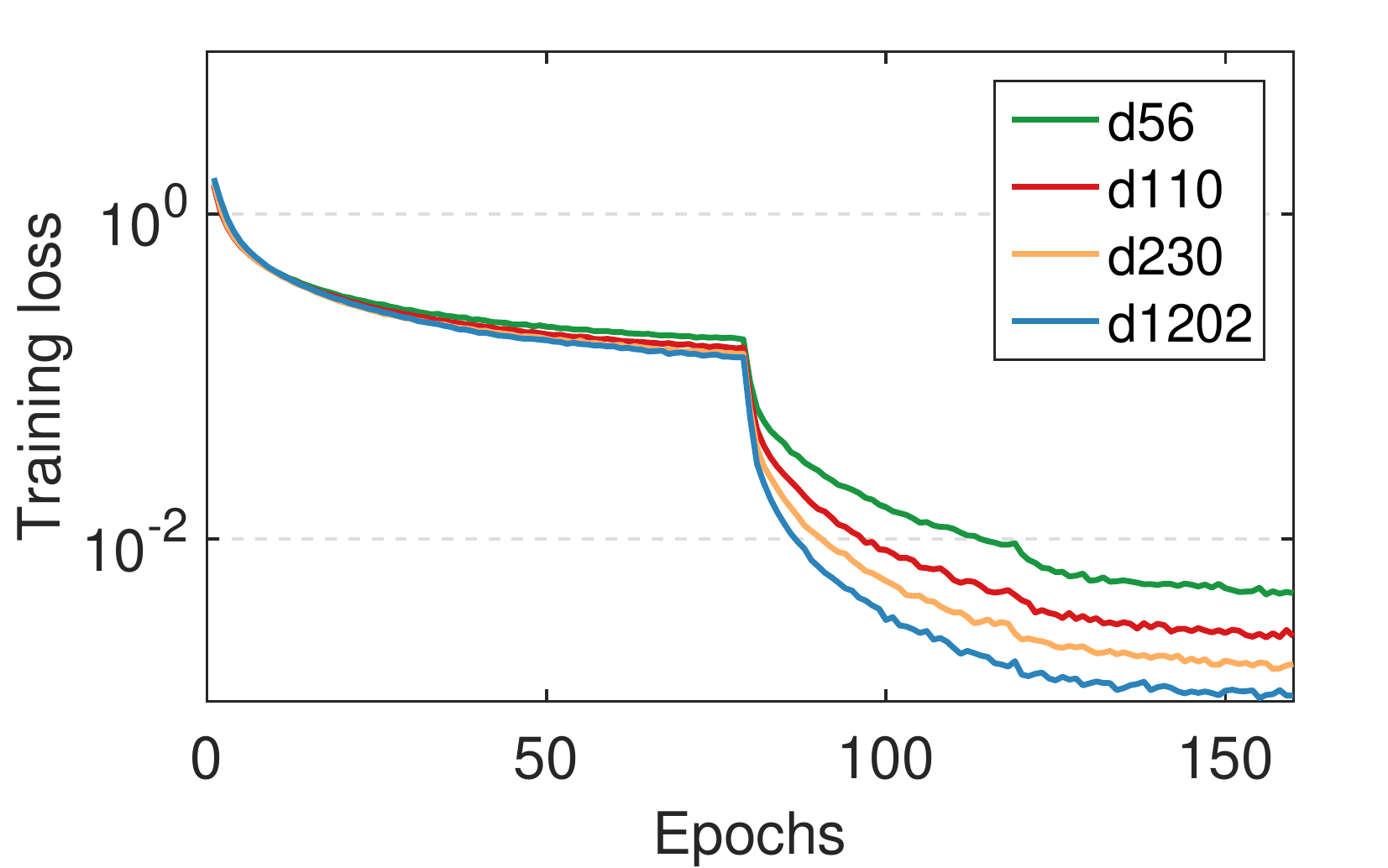}
		\end{minipage}
	}	
	\vspace{-0.1in}
	\caption{Training loss comparison between  (a) ResNet and (b) ResNet$_{LastBN}$  with different depth on CIFAR-10. We evaluate the  training loss  with respect to the epochs. }
	\label{fig:loss}
\end{figure}

ResNet$_{LastBN}$, a simple revision of ResNet,  achieves significant improvement in performance for very deep residual networks. 
Figure \ref{fig:loss} (a) and (b)  show the training loss of ResNet and ResNet$_{LastBN}$, respectively, on the CIFAR-10 dataset. We observe that ResNet, with a depth of 1202, appears to have degenerated training performance, especially in the initial phase. Note that, as the depth increases, ResNet obtains worse training performance in the first 80 epochs (before the learning rate is reduced), which coincides with our previous analysis.
ResNet$_{LastBN}$ obtains nearly the same training loss for the networks with different depths in the  first 80 epochs. Moreover, ResNet$_{LastBN}$ shows lower training loss with  increasing depth. Comparing  Figure \ref{fig:loss} (b) to (a), we  observe that ResNet$_{LastBN}$ has better training loss than  ResNet for all depths (\eg, at a depth of 56, the loss of ResNet is 0.081, while for ResNet$_{LastBN}$ it is 0.043.).

Table \ref{Tabe:C10} shows the test errors. We observe that ResNet$_{LastBN}$ achieves better test performance with increasing depth, while ResNet has degenerated performance. Compared to ResNet, ResNet$_{LastBN}$ has consistently improved performance over different depths. Particularly, ResNet$_{LastBN}$ reduces the absolute test error of ResNet by $1.02 \%$, $0.79\%$, $1.41 \%$ and $3.74 \%$  at depths of 56, 110, 230 and 1202, respectively.
Due to ResNet$_{LastBN}$'s optimization efficiency, the training performance is likely improved if we amplify the regularization of the training. Thus, we set the weight decay to 0.0002 and double the training iterations, finding that the 1202-layer ResNet$_{LastBN}$ achieves a test error  of $4.79 \pm 0.12$. We also train a 2402-layer network. We observe that $ResNet$ cannot converge, while ResNet$_{LastBN}$ achieves a test error of $5.04 \pm0.30$.

We further perform the experiment on CIFAR-100 and use the same experimental setup as CIFAR-10.
Table \ref{Table:C100} shows the test errors. ResNet$_{LastBN}$ reduces the absolute test error of ResNet by $0.78 \%$, $1.25\%$, $3.45 \%$ and $4.98 \%$  at depths of 56, 110, 230 and 1202, respectively. We also validate the effectiveness of ResNet$_{LastBN}$ on the large-scale ImageNet classification, with 1000 classes \cite{2009_ImageNet}. ResNet$_{LastBN}$ has better optimization efficiency and achieves better test performance, compared to ResNet. Please refer to the \SM~\ref{Sec-sup-ExpRes} for more details.

\begin{figure*}[t]
	\centering
	\hspace{-0.2in}		\subfigure[Training loss]{
		\begin{minipage}[c]{.22\linewidth}
			\centering
			\includegraphics[width=3.0cm]{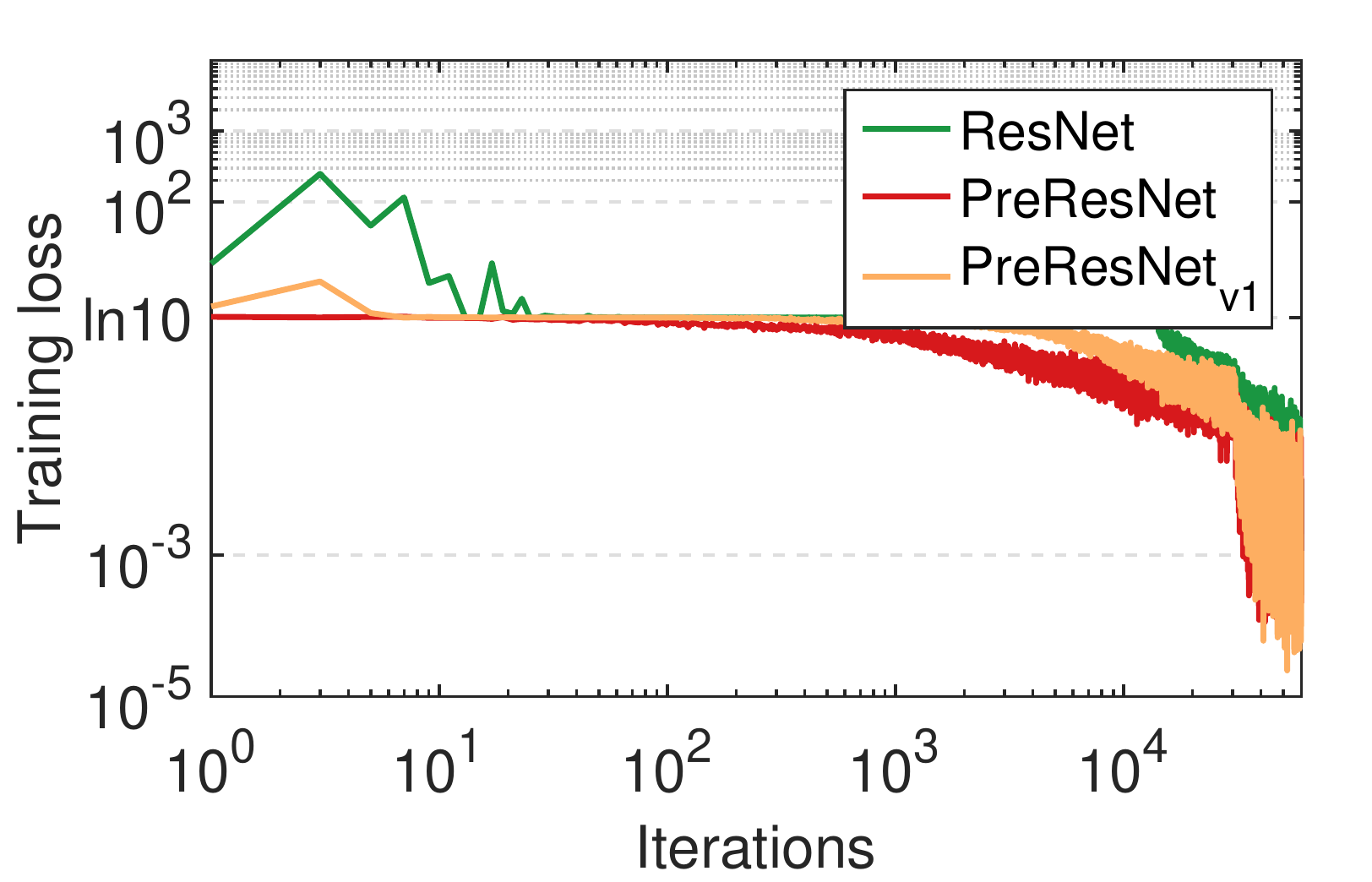}
		\end{minipage}
	}
	\hspace{0.03in}	\subfigure[$\lambda_{\Sigma_{\mathbf{x}}}$]{
		\begin{minipage}[c]{.22\linewidth}
			\centering
			\includegraphics[width=3.0cm]{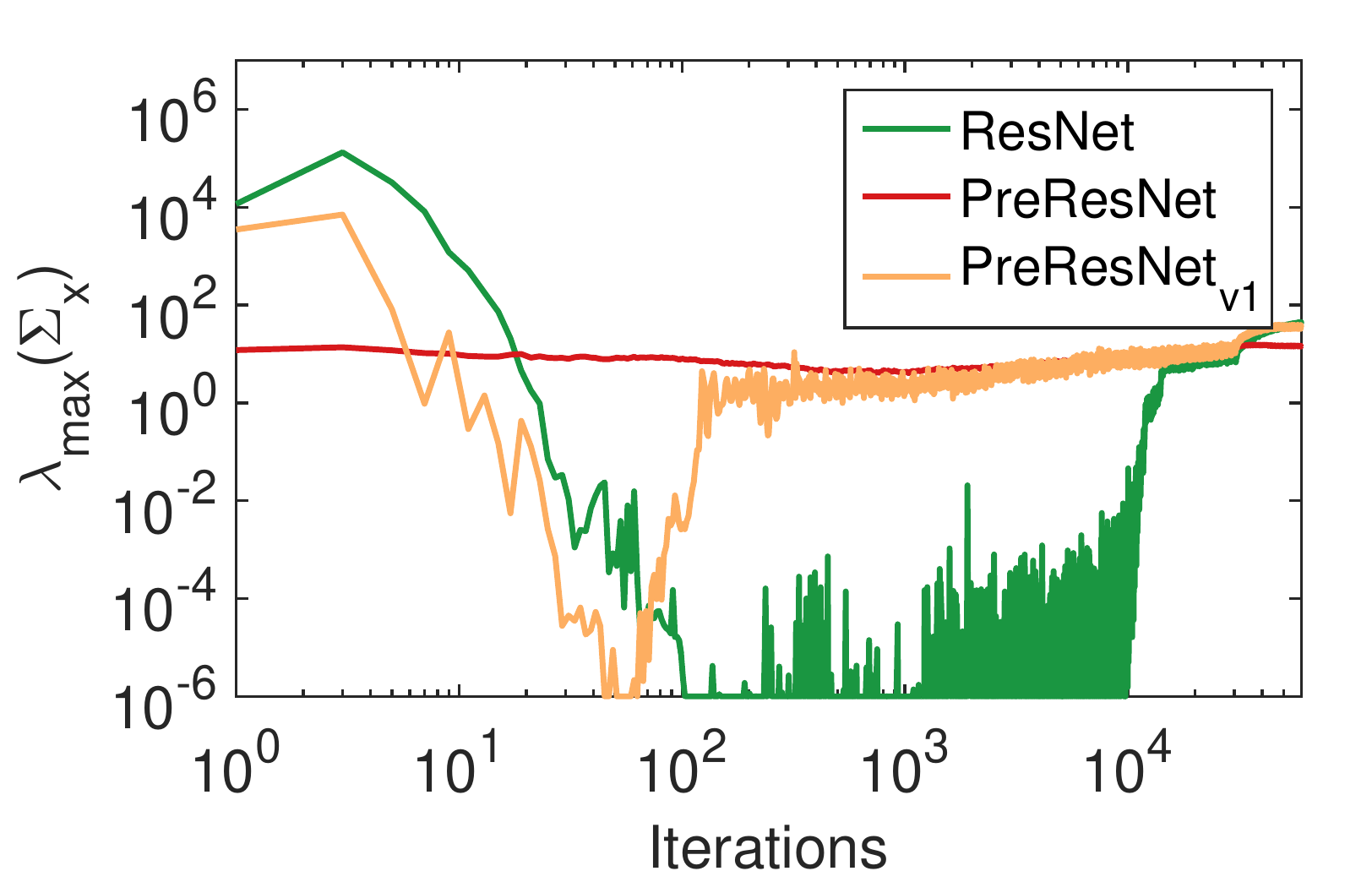}
		\end{minipage}
	}
	\hspace{0.03in}	\subfigure[$\lambda_{\Sigma_{\D{\W{}}}}$]{
		\begin{minipage}[c]{.22\linewidth}
			\centering
			\includegraphics[width=3.0cm]{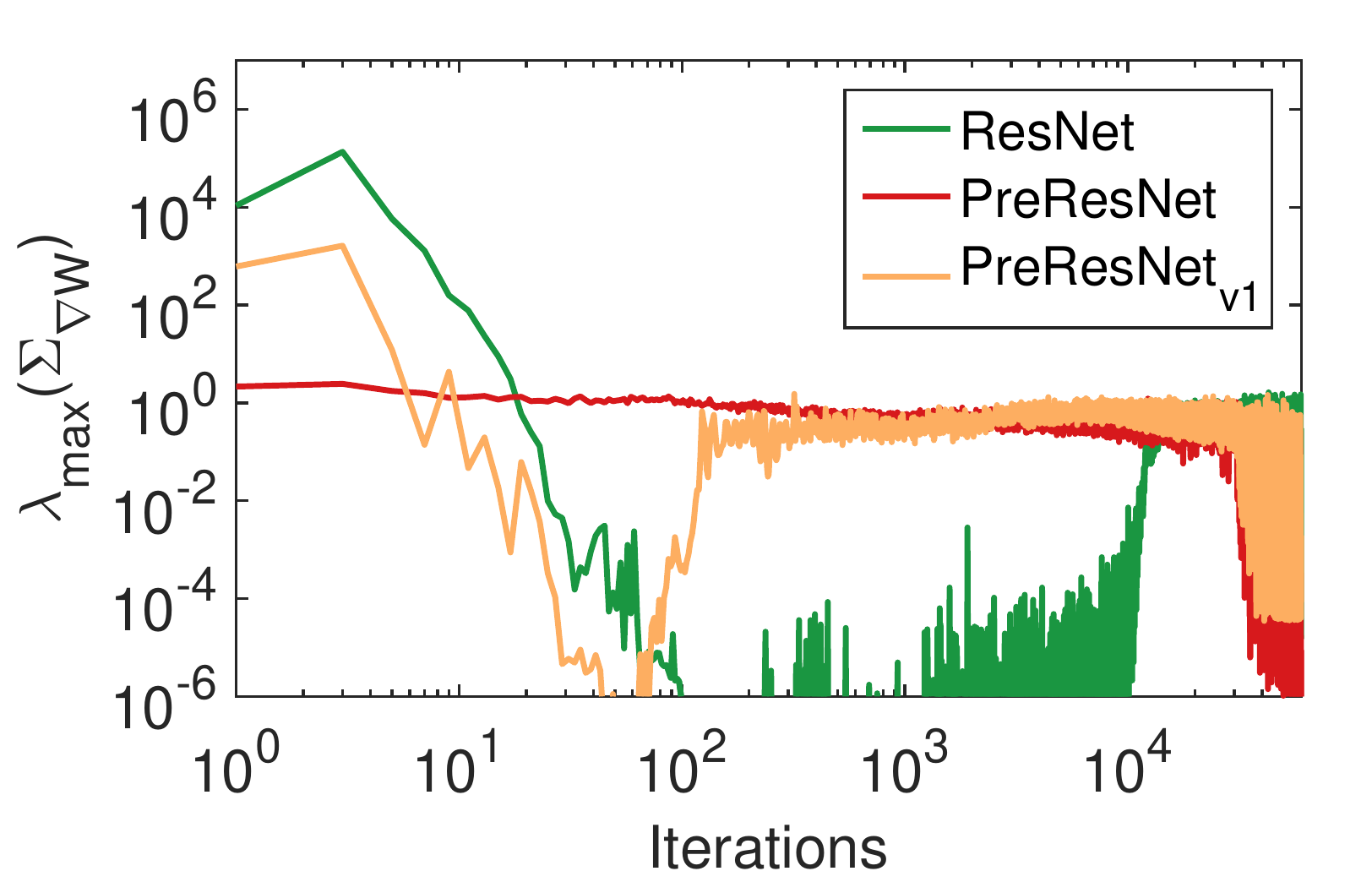}
		\end{minipage}
	}
	\hspace{0.03in}	\subfigure[$\|\mathbf{W} \|_2$]{
		\begin{minipage}[c]{.22\linewidth}
			\centering
			\includegraphics[width=3.0cm]{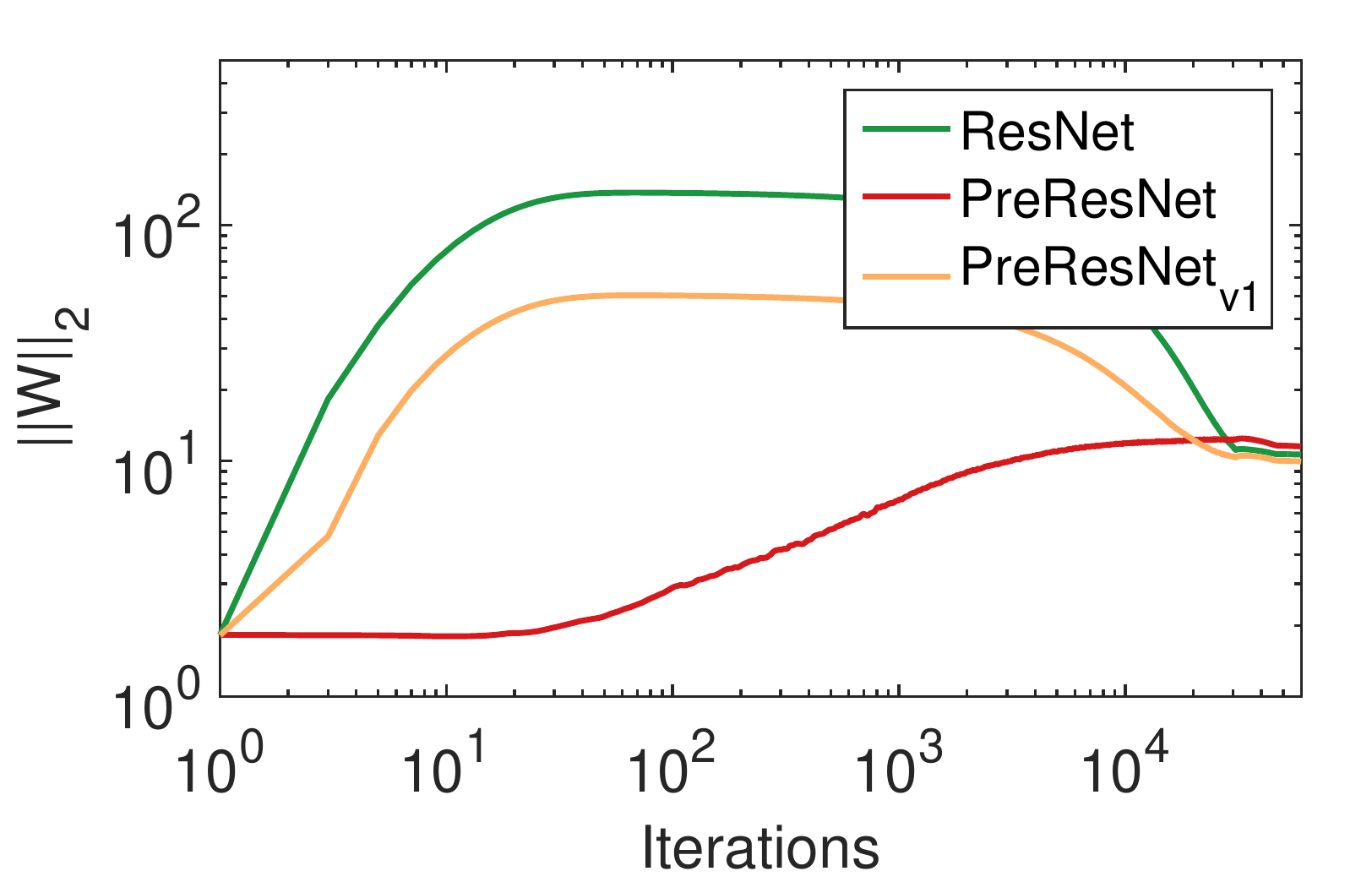}
		\end{minipage}
	}
	\vspace{-0.2in}
	\caption{Analysis on the last linear layer of 1202-layer `ResNet', `PreResNet' and `PreResNet$_{v1}$' for CIFAR-10 classification. }
	\label{fig:problem-PreResNet}
	\vspace{-0.2in}
\end{figure*}

	\vspace{-0.12in}
\subsection{Revisiting the Pre-activation Residual Network}
	\vspace{-0.03in}
We note that He \etal \cite{2016_CoRR_He} tried to improve the optimization and generalization of the original residual network \cite{2015_CVPR_He} by re-arranging the activation functions (using the pre-activation).  By looking into the implementation of \cite{2016_CoRR_He}, we find that it also uses an extra BN layer before the last average pooling. It is interesting to investigate which component in \cite{2016_CoRR_He} (\eg, the  pre-activation or the extra BN layer)  benefits the optimization behaviors, using our analysis. Here, we denote `PreResNet' as the pre-activation residual network \cite{2016_CoRR_He}, and denote `PreResNet$_{v1}$' as the PreResNet without the extra BN layer. 
We use the exact same setups as the previous experiments on ResNet. Figure~\ref{fig:problem-PreResNet} shows the conditioning analysis on the last linear layer of the 1202-layer network. We observe that: 1) PreResNet$_{v1}$ also gets stuck in the weight domination state with its last linear layer, even though it escapes this states faster than ResNet; 2) PreResNet, like our proposed ResNet$_{LastBN}$, does not suffer the ill-conditioned problem in its last linear layer. These observations suggest that the pre-activation can relieve the ill-conditioned problem to some degree, but more importantly, the extra BN layer is  key to improving the optimization efficiency of PreResNet \cite{2016_CoRR_He} for very deep networks.

We report the test errors of PreResNet and PreResNet$_{v1}$ in Table~\ref{Tabe:C10} and ~\ref{Table:C100}. We find that `PreResNet' generally has better test performance than PreResNet$_{v1}$, especially for very deep networks (\eg, the 1202-layer one). This supports our arguments that the extra BN layer is the key component of PreResNet \cite{2016_CoRR_He} for very deep networks. Interestingly, 
we further  observe that our proposed ResNet$_{LastBN}$ is consistently better than PreResNet \cite{2016_CoRR_He} over different layers and datasets. This  demonstrates the effectiveness of our proposed architecture. We believe that our analysis method can be further used to improve  residual architectures by looking into the intermediate (inner) layers of networks.

\begin{table*}[t]
	\caption{Comparison of test error ($\%$)  on CIFAR-10. The results are shown in the format of `mean $\pm$ std' computed over five random seeds.}
	\label{Tabe:C10}
	\centering
	\begin{tabular}{p{1in} p{0.8in}<{\centering}  p{0.8in}<{\centering}  p{0.8in}<{\centering} p{0.8in}<{\centering} p{0.8in}<{\centering}}
		\toprule
		method     & depth-56     & depth-110  & depth-230 & depth-1202 \\
		\midrule
		ResNet~\cite{2015_CVPR_He}     &  7.52 $\pm$ 0.30  & 6.89 $\pm$ 0.52 & 7.35 $\pm$ 0.64 & 9.42 $\pm$ 3.10  \\
		PreResNet~\cite{2016_CoRR_He}    &  6.89 $\pm$ 0.09  & 6.25 $\pm$ 0.08 & 6.12 $\pm$ 0.21 & 6.07 $\pm$ 0.10  \\
		PreResNet$_{v1}$  &  6.75 $\pm$ 0.26  & 6.37 $\pm$ 0.24 & 6.32 $\pm$ 0.21 & 7.89 $\pm$ 0.58  \\
		ResNet$_{LastBN}$  &  \textbf{6.50 $\pm$ 0.22} &  \textbf{6.10 $\pm$ 0.09} & \textbf{5.94 $\pm$ 0.18}&  \textbf{5.68 $\pm$ 0.14}  \\
			\toprule[1pt]
	\end{tabular}
\end{table*}

\begin{table*}[t]
	\caption{Comparison of test error ($\%$) on CIFAR-100. The results are shown in the format of `mean $\pm$ std', computed over five random seeds.}
	\vspace{-0.14in}
	\label{Table:C100}
	\centering
	\begin{tabular}{p{1in} p{0.8in}<{\centering}  p{0.8in}<{\centering}  p{0.8in}<{\centering} p{0.8in}<{\centering} p{0.8in}<{\centering}}
		\toprule
		method     & depth-56     & depth-110  & depth-230 & depth-1202 \\
		\midrule
		ResNet~\cite{2015_CVPR_He}     &  29.60 $\pm$ 0.41  & 28.3 $\pm$ 1.09 & 29.25 $\pm$ 0.44 & 30.49 $\pm$ 4.44  \\
		PreResNet~\cite{2016_CoRR_He}      &  29.29 $\pm$ 0.44  & 27.58 $\pm$ 0.12  & 26.72 $\pm$ 0.33 & 26.23 $\pm$ 0.26  \\
		PreResNet$_{v1}$     &  29.60 $\pm$ 0.21  & 28.54 $\pm$ 0.26 & 27.92 $\pm$ 0.34 & 30.07 $\pm$ 2.04  \\
		ResNet$_{LastBN}$  &  \textbf{28.82 $\pm$ 0.38} &  \textbf{27.05 $\pm$ 0.23} & \textbf{25.80 $\pm$ 0.10} &  \textbf{25.51 $\pm$ 0.27}  \\
		\toprule[1pt]
	\end{tabular}
	\vspace{-0.12in}
\end{table*}
	\vspace{-0.1in}
	\section{Conclusion and Future Work}
	\vspace{-0.08in}
	We proposed a layer-wise conditioning analysis to investigate the learning dynamics of DNNs. Such an analysis is theoretically derived under mild assumptions that  approximately hold in practice.
	Based on our layer-wise conditioning analysis, we showed how batch normalization stabilizes training and improves the conditioning of the optimization problem. We further found that very deep residual networks still suffer difficulty in optimization, which is caused by the ill-conditioned state of the last linear layer. We remedied this by adding only one BN layer before the last linear layer. 
	
	We believe there are many potential applications of our method, \eg, investigating the training dynamics of other normalization methods (layer normalization \cite{2016_CoRR_Ba} and instance normalization \cite{2016_arxiv_Ulyanov}) and comparing them to BN. We also believe it would be interesting to analyze the training dynamics of GANs \cite{2019_ICLR_Brock} using our method. We expect our method to provide new insights for analyzing and understanding training techniques for DNNs.

\clearpage
%
%
\bibliographystyle{splncs04}
\bibliography{conditioning}

\clearpage

\appendix

\renewcommand{\thetable}{A\arabic{table}}
\setcounter{table}{0}

\renewcommand{\thefigure}{A\arabic{figure}}
\setcounter{figure}{0}

\section{Proof of Theorems}
\label{Sec-Proof}

Here, we provide proofs for the three proposition/theorems in the paper. 

\subsection{Proof of Proposition 1}
\label{Sec-Proof-Pro1}
\textbf{Proposition 1.}	Given $\Sigma_{\mathbf{x}}$, $\Sigma_{\nabla \mathbf{h}}$ and  $F= \Sigma_{\mathbf{x}} \otimes \Sigma_{\nabla \mathbf{h}}$, we have:
	1) $\lambda_{max}(F)=\lambda_{max}(\Sigma_{\mathbf{x}}) \cdot \lambda_{max}(\Sigma_{\nabla \mathbf{h}}) $; 	
	2) $\kappa(F)=\kappa(\Sigma_{\mathbf{x}}) \cdot \kappa(\Sigma_{\nabla \mathbf{h}}) $.

\begin{proof}
	The proof is mainly based on the conclusion from Theorem 4.2.12 in \cite{1991_Matrix}, which is restated as follows:
	\begin{lemma}
		\label{lemma1}
		Let $\mathbf{A} \in \mathbb{R}^{m \times m}$ and $\mathbf{B} \in \mathbb{R}^{n \times n}$. Furthermore, let $\lambda_{a}$ be the arbitrary eigenvalue of $\mathbf{A}$ and  $\lambda_{b}$ be the arbitrary eigenvalue of $\mathbf{B}$. We have $\lambda_{a} \cdot \lambda_{b}$ as an eigenvalue of $\mathbf{A} \otimes \mathbf{B}$. Furthermore, any eigenvalue of  $\mathbf{A} \otimes \mathbf{B}$ arises as a product of the eigenvalues of $\mathbf{A}$ and $\mathbf{B}$.
	\end{lemma}	
	
	Based on the definitions of $\Sigma_{\mathbf{x}}$ and $\Sigma_{\nabla \mathbf{h}}$, we have that $\Sigma_{\mathbf{x}}$ and $\Sigma_{\nabla \mathbf{h}}$ are positive semidefinite. Therefore, all the eigenvalues of $\Sigma_{\mathbf{x}}$/$\Sigma_{\nabla \mathbf{h}}$	are non-negative.  Let $\lambda(\mathbf{A})$ denote the eigenvalue of matrix $\mathbf{A}$.
	Based on Lemma \ref{lemma1}, we have $\lambda(F)=\lambda(\Sigma_{\mathbf{x}})  \lambda(\Sigma_{\nabla \mathbf{h}}) $. Since $\lambda(\Sigma_{\mathbf{x}})$ and  $\lambda(\Sigma_{\nabla \mathbf{h}}) $ are non-negative, we thus have  $\lambda_{max}(F)=\lambda_{max}(\Sigma_{\mathbf{x}}) \cdot \lambda_{max}(\Sigma_{\nabla \mathbf{h}}) $.  Similarly, we can prove that  $\lambda_{min}(F)=\lambda_{min}(\Sigma_{\mathbf{x}}) \cdot \lambda_{min}(\Sigma_{\nabla \mathbf{h}}) $. We thus have $\kappa(F)=\kappa(\Sigma_{\mathbf{x}}) \cdot \kappa(\Sigma_{\nabla \mathbf{h}}) $.	
	
\end{proof}

\subsection{Proof of Theorem 1}
\label{Sec-Proof-Th1}

\textbf{Theorem 1.}
	Given a rectifier neural network (Eqn. \ref{eqn:MLP}) with nonlinearity $\phi(\alpha \mathbf{x})= \alpha \phi(\mathbf{x})$ ($\alpha >0$ ), if the weight in each layer is scaled by $\widehat{\mathbf{W}}_k = \alpha_k \mathbf{W}_k $ ($k=1,...,K$ and $\alpha_k >0$), we have the scaled layer input:
	$\widehat{\mathbf{x}}_k= (\prod\limits_{i=1}^{k} {\alpha_i}) \mathbf{x}_k$.
	Assuming that $\D{\hath{K}} = \mu \D{\h{K}}$, we have the output-gradient:
	$\D{\hath{k}} =  \mu ( \prod\limits_{i=k+1}^{K} {\alpha_i})  \D{\mathbf{h}_{k}}$,  and weight-gradient: $\D{\widehat{\mathbf{W}}_k } = (\mu \prod\limits_{i=1, i\neq k}^K \alpha_i ) \D{\mathbf{W}_k}$, for all $k=1,...,K$.

\begin{proof}
	(1) We first demonstrate that the scaled layer input $\widehat{\mathbf{x}}_k= (\prod\limits_{i=1}^{k} {\alpha_i}) \mathbf{x}_k$ ($k=1,...,K$),
	using mathematical induction. It is easy to validate that $\widehat{\mathbf{h}}_1 = \alpha_1 \mathbf{h}_1 $ and $\widehat{\mathbf{x}}_1 = \alpha_1 \mathbf{x}_1 $. We assume that  $\widehat{\mathbf{h}}_t = (\prod\limits_{i=1}^{t} {\alpha_i})  \mathbf{h}_t $ and $\widehat{\mathbf{x}}_t = (\prod\limits_{i=1}^{t} {\alpha_i})  \mathbf{x}_t $ hold, for $t=1,...,k$.
	When $t=k+1$, we have
	\begin{eqnarray}
	\hath{k+1}=\hatW{k+1} \hatx{k+1}=\alpha_{k+1} \W{k+1}  (\prod\limits_{i=1}^{k} {\alpha_i}) \x{k}=(\prod\limits_{i=1}^{k+1} {\alpha_i}) \h{k+1}.
	\end{eqnarray}
	We thus have	
	\begin{eqnarray}
	\hatx{k+1} = \phi(\hath{k+1}) = \phi((\prod\limits_{i=1}^{k+1} {\alpha_i}) \h{k+1})= (\prod\limits_{i=1}^{k+1} {\alpha_i})  \phi(\h{k+1})= (\prod\limits_{i=1}^{k+1} {\alpha_i})  \x{k+1}.
	\end{eqnarray}
	
	By induction, we have $\hatx{k} = (\prod\limits_{i=1}^{k} {\alpha_i}) \x{k}$, for $k=1,...,K$.  We also have $\hath{k} = (\prod\limits_{i=1}^{k} {\alpha_i})  \h{k}$ for $k=1,...,K$.
	
	(2) We then demonstrate that the scaled output-gradient $\D{\hath{k}} = \mu (\prod\limits_{i=k+1}^{K} {\alpha_i}) \D{\h{k}} $ for $k=1,...,K$.
	We also provide this using mathematical induction.
	Based on back-propagation, we have
	\begin{eqnarray}
	\D{\x{k-1}} = \D{\h{k}} \W{k}, ~~~ \D{\h{k-1}} = \D{\x{k-1}} \frac{\partial \x{k-1}}{\partial \h{k-1}},
	\end{eqnarray}
	
	and
	\begin{eqnarray}
	\label{eqn:bequ}
	\frac{\partial \hatx{k-1}}{\partial \hath{k-1}}
	=   \frac{\partial (\prod\limits_{i=1}^{k-1} {\alpha_i}) \x{k-1}}{\partial (\prod\limits_{i=1}^{k-1} {\alpha_i}) \h{k-1} }
	=   \frac{ (\prod\limits_{i=1}^{k-1} {\alpha_i})  \partial  \x{k-1}}{ (\prod\limits_{i=1}^{k-1} {\alpha_i}) \partial \h{k-1}}
	=  \frac{\partial \x{k-1}}{\partial \h{k-1}} , ~~ k=2, ..., K.
	\end{eqnarray}
	
	Based on the assumption that $\D{\hath{K}} = \mu \D{\h{K}}$, we have $\D{\hath{K}} = \mu (\prod\limits_{i=K+1}^{K} {\alpha_i}) \D{\h{K}} $\footnote{We denote $\prod\limits_{i=a}^{b}  {\alpha_i} =1$ if $a>b$.}.
	
	We assume that  $\D{\hath{t}} = \mu (\prod\limits_{i=t+1}^{K} {\alpha_i}) \D{\h{t}} $  holds, for $t=K,...k$.
	When $t=k-1$, we have
	\begin{eqnarray}
	\label{eqn:33}
	\D{\hatx{k-1}} = \D{\hath{k}} \hatW{k}
	= \mu (\prod\limits_{i=k+1}^{K} {\alpha_i})  \D{\h{k}} \alpha_k \W{k}
	=\mu  (\prod\limits_{i=k}^{K} {\alpha_i}) \D{\x{k-1}}.
	\end{eqnarray}
	We also have
	\begin{eqnarray}
	\label{eqn:44}
	\D{\hath{k-1}}=\D{\hatx{k-1}}. \frac{\partial \hatx{k-1}}{\partial \hath{k-1}}
	=\mu  (\prod\limits_{i=k}^{K} {\alpha_i})  \D{\x{k-1}} .  \frac{\partial \x{k-1}}{\partial \h{k-1}} =  \mu  (\prod\limits_{i=k}^{K} {\alpha_i})  \D{\h{k-1}}.
	\end{eqnarray}
	By induction, we thus have  $\D{\hath{k}} = \mu (\prod\limits_{i=k+1}^{K} {\alpha_i}) \D{\h{k}} $, for $k=1,...,K$.
	
	(3) Based on $\D{\W{k}} = \D{\h{k}}^T \x{k-1}^T$, $\widehat{\mathbf{x}}_k= (\prod\limits_{i=1}^{k} {\alpha_i}) \mathbf{x}_k$ and $\D{\hath{k}} = \mu (\prod\limits_{i=k+1}^{K} {\alpha_i}) \D{\h{k}} $, it is easy to prove that $\D{\widehat{\mathbf{W}}_k } = (\mu \prod\limits_{i=1,i\neq k}^K \alpha_i ) \D{\mathbf{W}_k}$ for $k=1,...,K$.
\end{proof}

\subsection{Proof of Theorem 2}
\label{Sec-Proof-Th2}
\textbf{Theorem 2.}
	Under the same condition of Theorem \ref{th2:norm}, for the normalized network with $\h{k}=\W{k} \x{k-1}$ and $\mathbf{s}_k=BN(\mathbf{h}_k)$, we have:
	$\hatx{k}=\mathbf{x}_k$, $\D{\hath{k}} = \frac{1} {\alpha_k} \D{\mathbf{h}_k}$, $\D{\widehat{\mathbf{W}}_k } =\frac{1} {\alpha_k} \D{\mathbf{W}_k}$, for all $k=1,...,K$.

\begin{proof}
	(1) Following the proof in Theorem \ref{th2:norm}, by mathematical induction, it is easy to demonstrate that  $\hath{k}= \alpha_k \h{k}$, $\hats{k}=\s{k} $	and $\hatx{k}=\x{k}$, for all $k=1, ...,K$.
	
	(2) We also use mathematical induction to demonstrate $\D{\hath{k}}=\frac{1}{\alpha_k}\D{\h{k}}$ for all $k=1,...,K.$
	
	We first show the formulation of the gradient back-propagating through  each neuron of the BN layer as:
	\begin{eqnarray}
	\label{eq:BNBack}
	\D{h} =\frac{1}{\mathbf{\sigma}} (\D{s} - \mathbb{E}_B(\D{s}) - \mathbb{E}_B(\D{s} s) s),	
	\end{eqnarray}
	where $\sigma $ is the standard deviation and $\mathbb{E}_B$ denotes the expectation over mini-batch examples. We have $\hat{\sigma}_{K} = \alpha_K \sigma_K$ based on $\hath{K}=\alpha_K \h{K} $.
	Since $\hats{K} = \s{K}$, we have $\D{\hats{K}} = \D{\s{K}}$.  Therefore, we have $\D{\hath{K}}= \frac{\sigma_K}{\hat{\sigma}_K} \D{\h{K}} = \frac{1} {\alpha_K} \D{\h{K}}$ from Eqn. \ref{eq:BNBack}.
	
	Assume that  $\D{\hath{t}} = \frac{1}{\alpha_t} \D{\h{t}}$ for $t=K,..., k+1$.  When $t=k$, we have:
	\begin{eqnarray}
	\label{eqn:BN33}
	\D{\hatx{k}} = \D{\hath{k+1}} \hatW{k+1}
	= \frac{1}{\alpha_{k+1}}  \D{\h{k+1}}   \alpha_{k+1} \W{k+1}
	=\D{\x{k}}.
	\end{eqnarray}
	Following the proof for Theorem \ref{th2:norm} , it is easy to get $\D{\hats{k}}=\D{\s{k}}$. Based on $\D{\hats{k}}=\D{\s{k}}$ and $\hats{k}=\s{k}$, we have
	$\D{\hath{k}}= \frac{\sigma_k}{\hat{\sigma}_k} \D{\h{k}} = \frac{1}{\alpha_k} \D{\h{k}}$ from Eqn. \ref{eq:BNBack}.
	
	By induction, we have  $\D{\hath{k}}=\frac{1}{\alpha_k}\D{\h{k}}$, for all $k=1,...,K.$
	
	(3) Based on $\D{\W{k}} = \D{\h{k}}^T \x{k-1}^T$, $\hatx{k}=\x{k}$ and $\D{\hath{k}}=\frac{1}{\alpha_k}\D{\h{k}}$, we have that $\D{\widehat{\mathbf{W}}_k } = \frac{1}{\alpha_{k}} \D{\mathbf{W}_k}$, for all $k=1,...,K.$
	
\end{proof}

\begin{figure}[]
	\centering
	\hspace{-0.2in}	\subfigure[full FIM]{
		\begin{minipage}[c]{.32\linewidth}
			\centering
			\includegraphics[width=4.0cm]{./figures_sup/MLP-F/Figure1_FIM_all.pdf}
		\end{minipage}
	}
	\subfigure[sub-FIM  (the 1st layer)]{
		\begin{minipage}[c]{.32\linewidth}
			\centering
			\includegraphics[width=4.0cm]{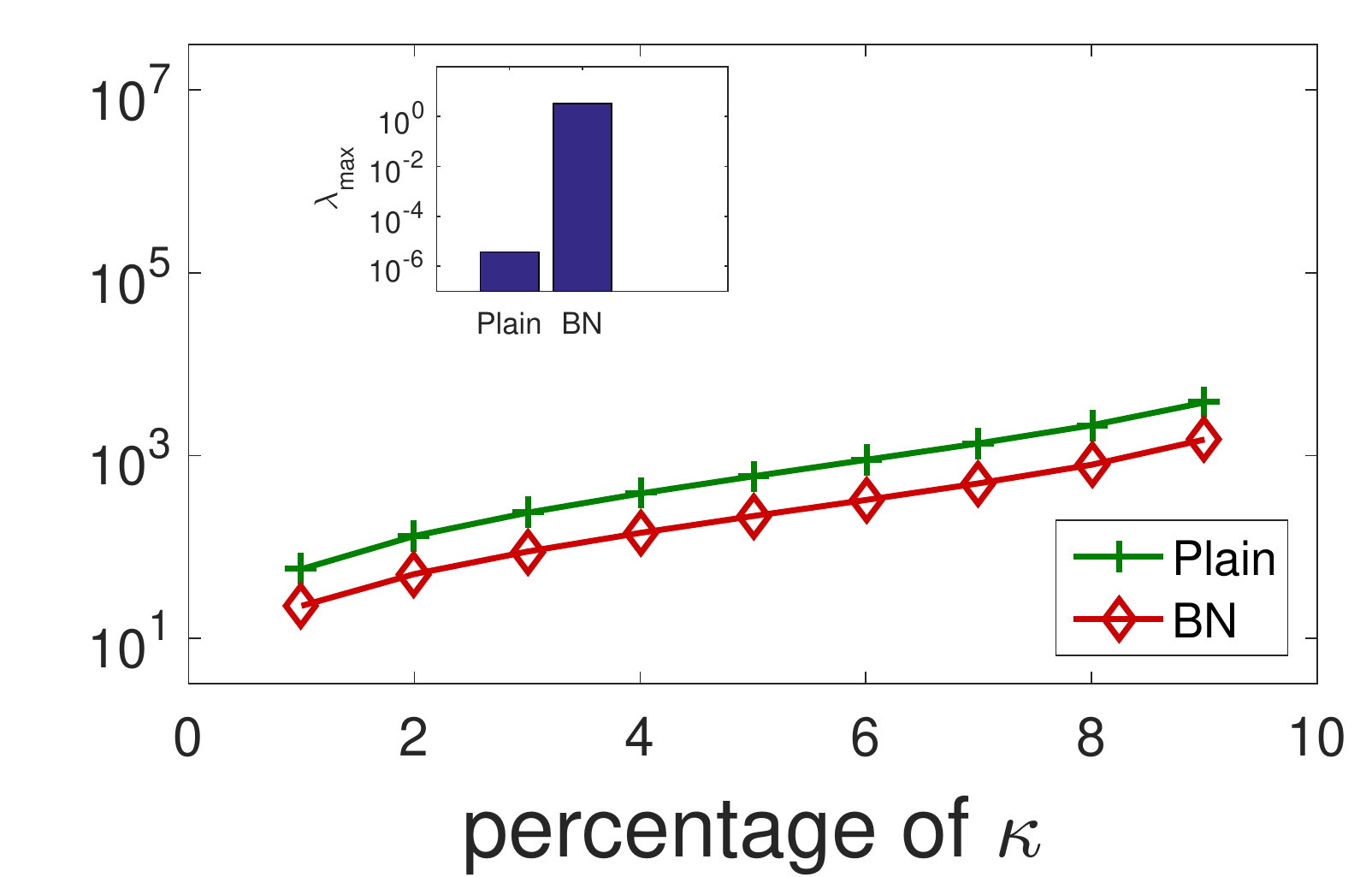}
		\end{minipage}
	}	
	\subfigure[sub-FIM  (the 2nd layer)]{
		\begin{minipage}[c]{.32\linewidth}
			\centering
			\includegraphics[width=4.0cm]{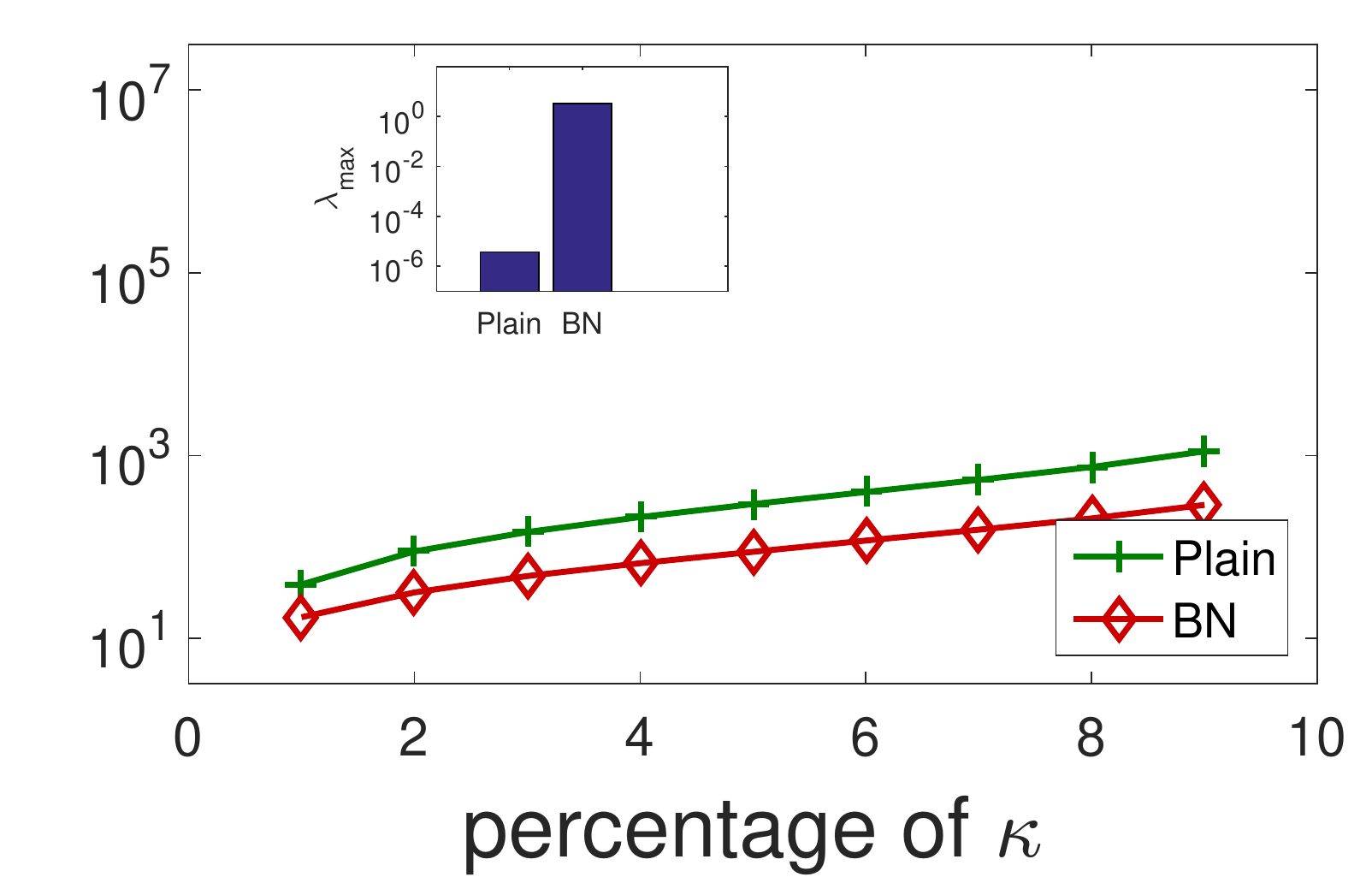}
		\end{minipage}
	}	\\
	\hspace{-0.2in}	\subfigure[sub-FIM  (the 3rd layer)]{
		\begin{minipage}[c]{.32\linewidth}
			\centering
			\includegraphics[width=4.0cm]{./figures_sup/MLP-F/Figure1_FIM_layer3.pdf}
		\end{minipage}
	}
	\subfigure[sub-FIM  (the 4th layer)]{
		\begin{minipage}[c]{.32\linewidth}
			\centering
			\includegraphics[width=4.0cm]{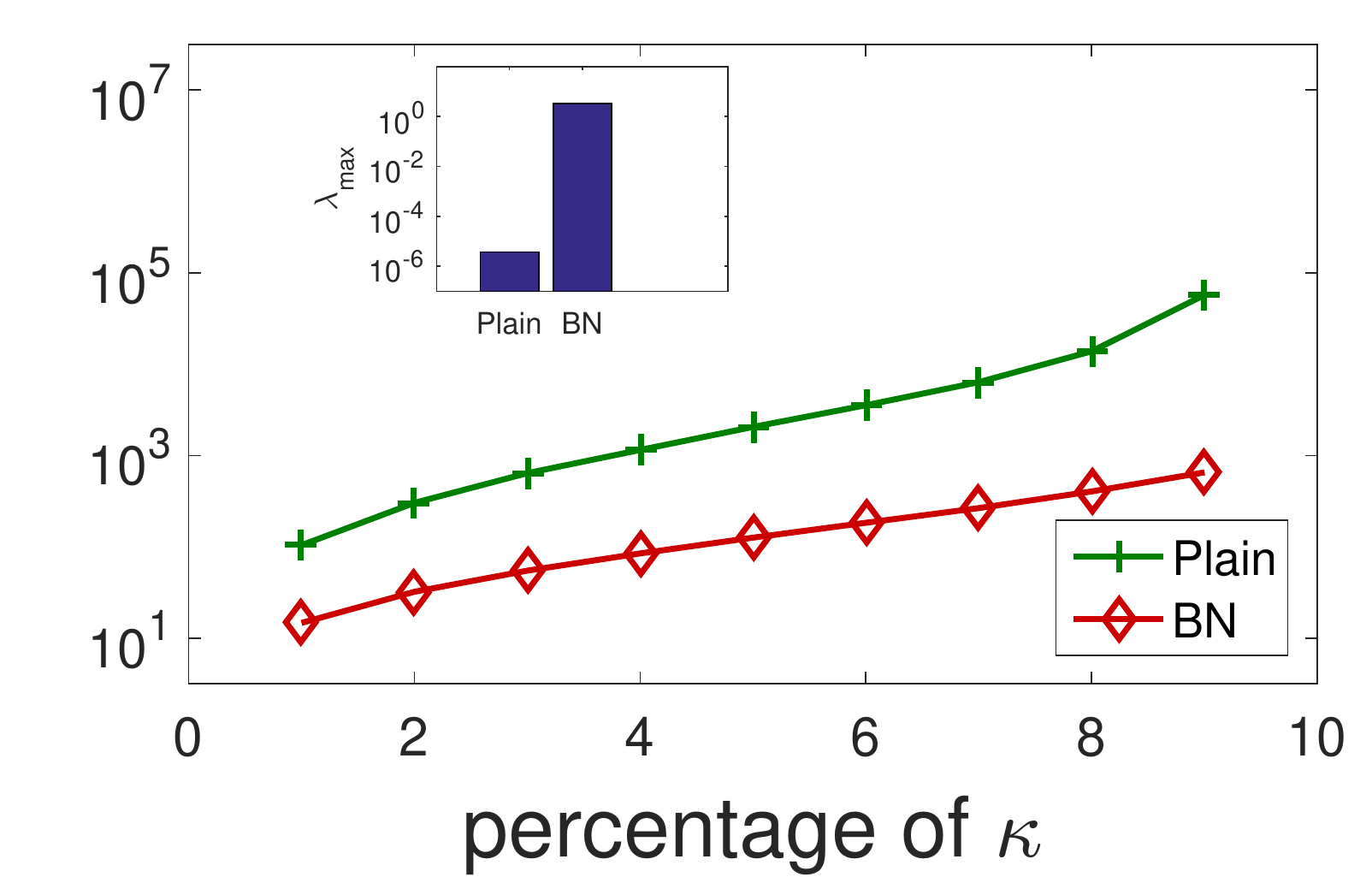}
		\end{minipage}
	}	
	\subfigure[sub-FIM  (the 5th layer)]{
		\begin{minipage}[c]{.32\linewidth}
			\centering
			\includegraphics[width=4.0cm]{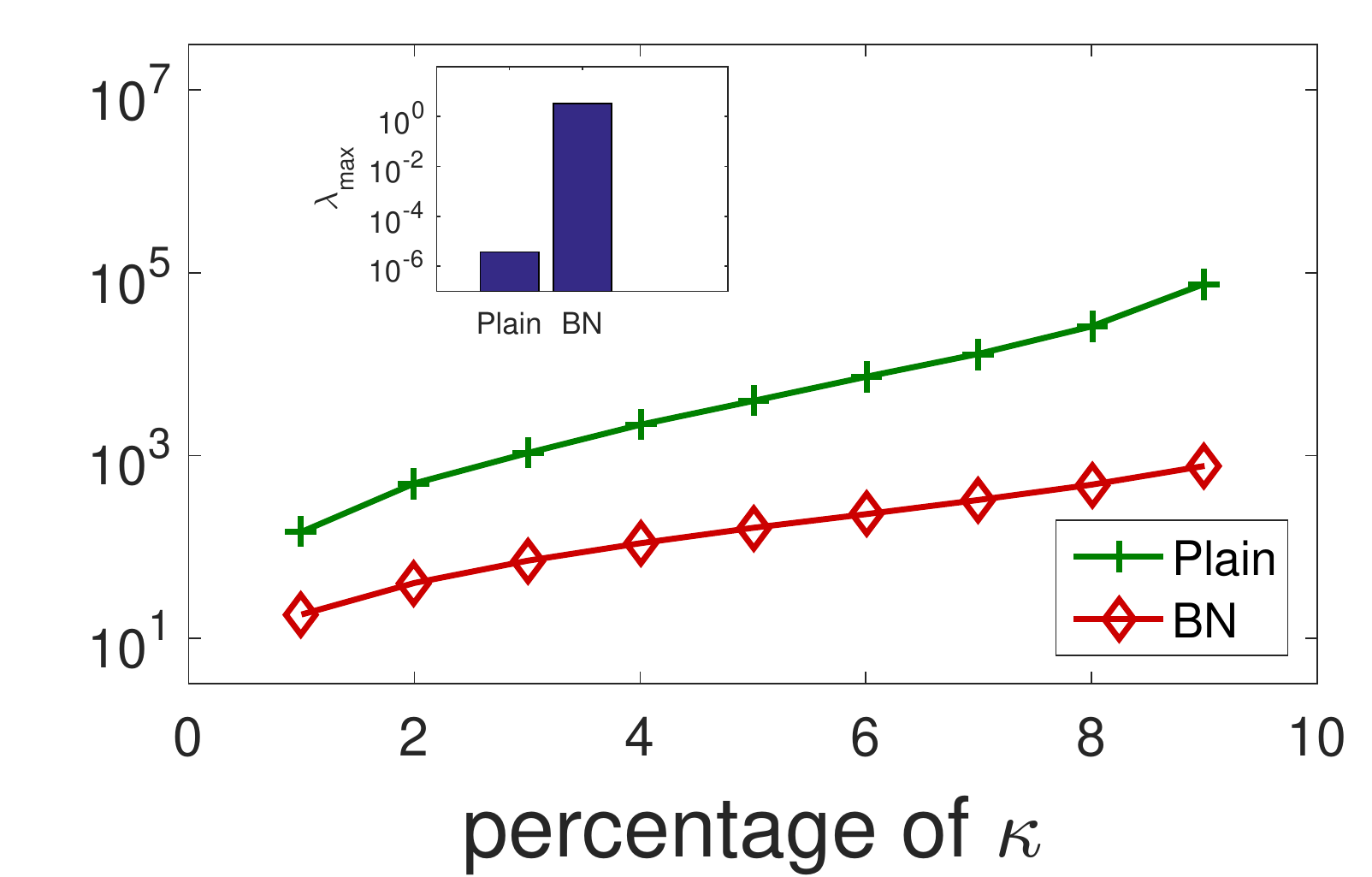}
		\end{minipage}
	}\\
	\hspace{-0.2in}	\subfigure[sub-FIM  (the 6th layer)]{
		\begin{minipage}[c]{.32\linewidth}
			\centering
			\includegraphics[width=4.0cm]{./figures_sup/MLP-F/Figure1_FIM_layer6.pdf}
		\end{minipage}
	}
	\subfigure[sub-FIM  (the 7th layer)]{
		\begin{minipage}[c]{.32\linewidth}
			\centering
			\includegraphics[width=4.0cm]{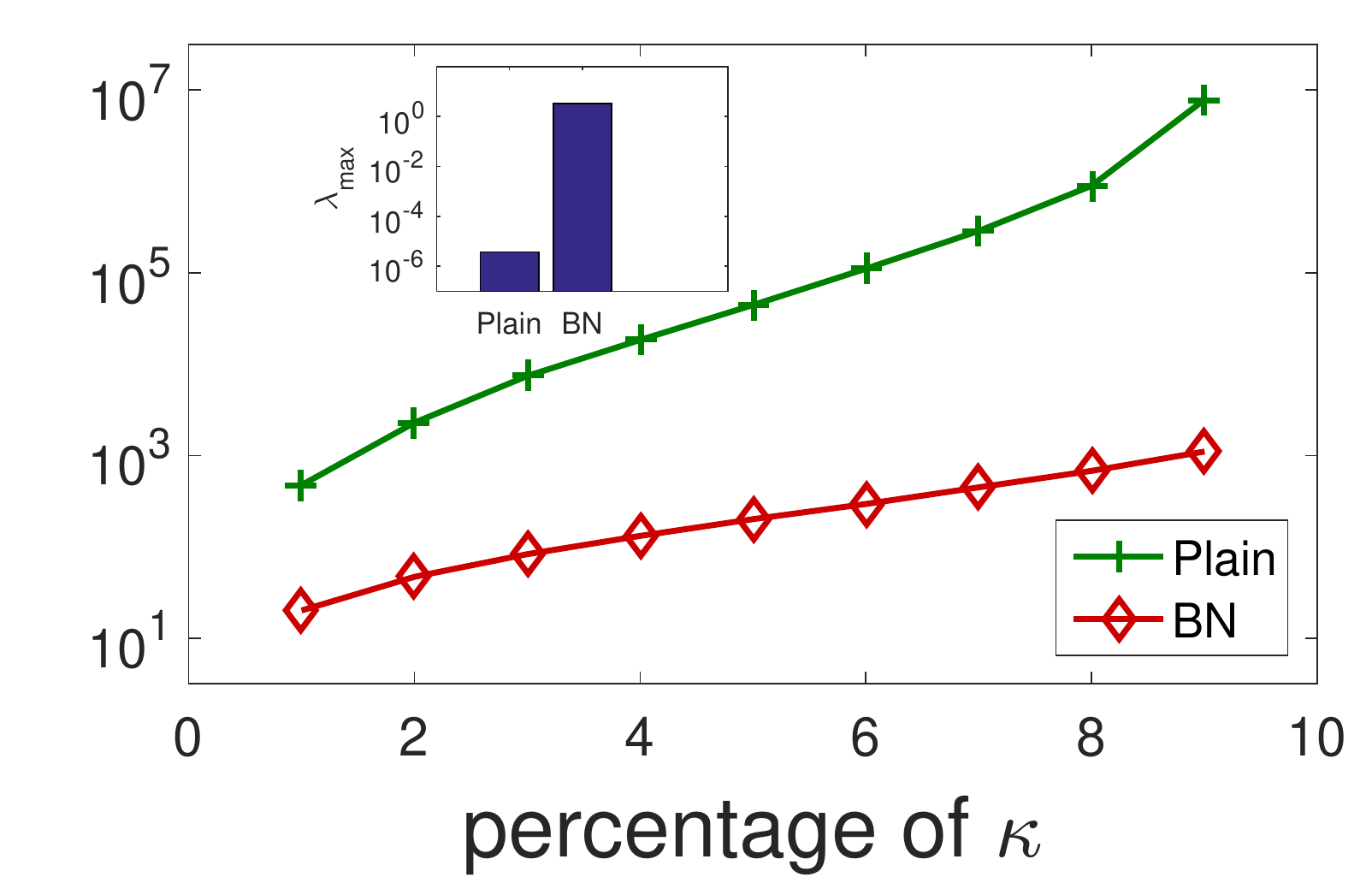}
		\end{minipage}
	}	
	\subfigure[sub-FIM  (the 8th layer)]{
		\begin{minipage}[c]{.32\linewidth}
			\centering
			\includegraphics[width=4.0cm]{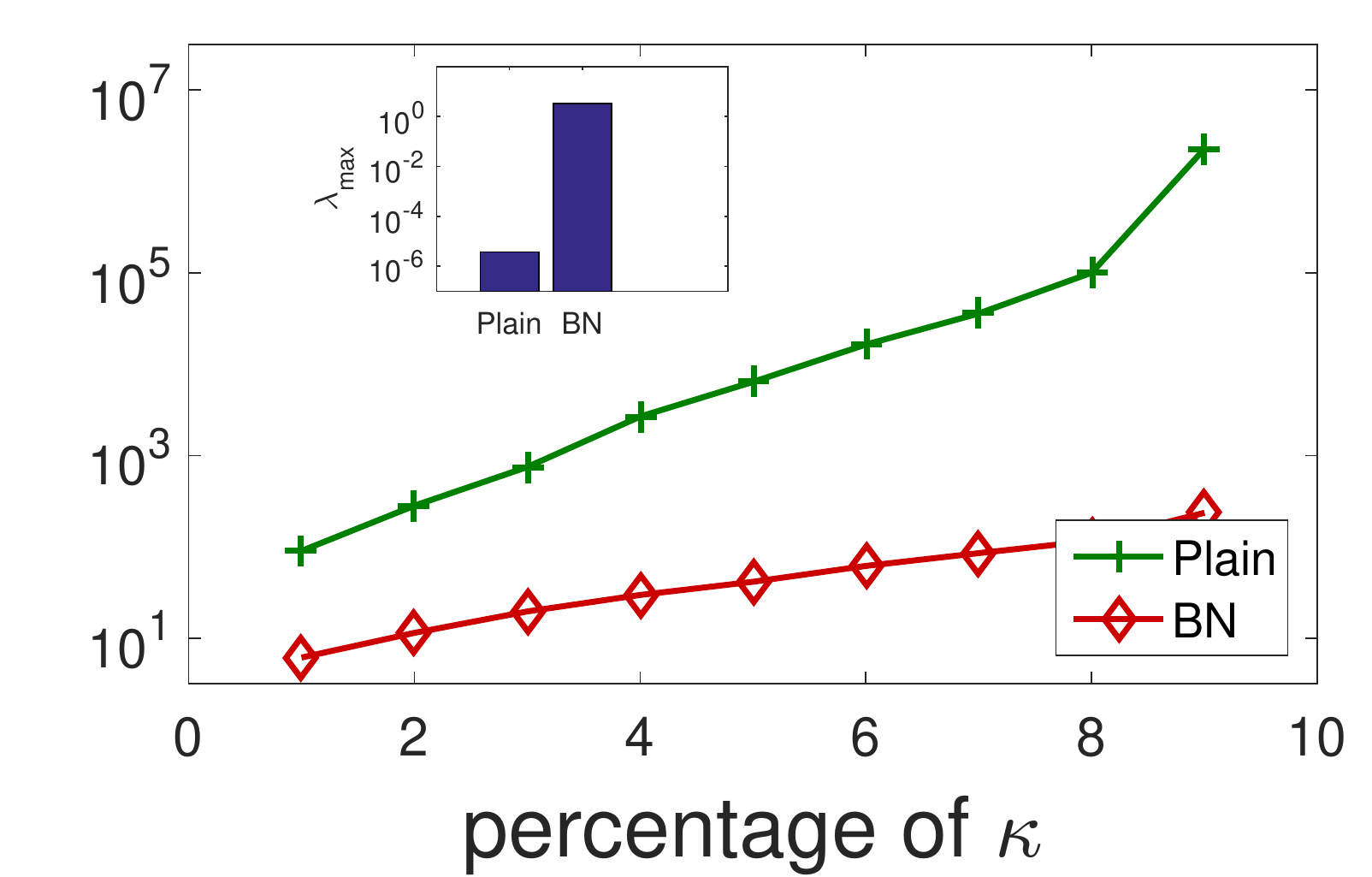}
		\end{minipage}
	}
	\caption{Conditioning analysis for unnormalized (`Plain') and normalized (`BN') networks. We show the maximum eigenvalue $\lambda_{max}$ and the generalized condition number $\kappa_{p}$ for comparison between the full FIM $\mathbf{F}$ and sub-FIMs $\{F_k\}$.   The experiments are performed on an 8-layer MLP with 24 neurons in each layer, for MNIST classification. The input image is center-cropped and resized to $12 \times 12$ to remove uninformative pixels. We report the corresponding spectrum at random initialization \cite{1998_NN_Yann}.  }
	\label{fig:AP1-FIM-full}
\end{figure}

\begin{figure}[]
	\centering
	\hspace{-0.2in}	\subfigure[full $\mathbf{M}$]{
		\begin{minipage}[c]{.32\linewidth}
			\centering
			\includegraphics[width=4.0cm]{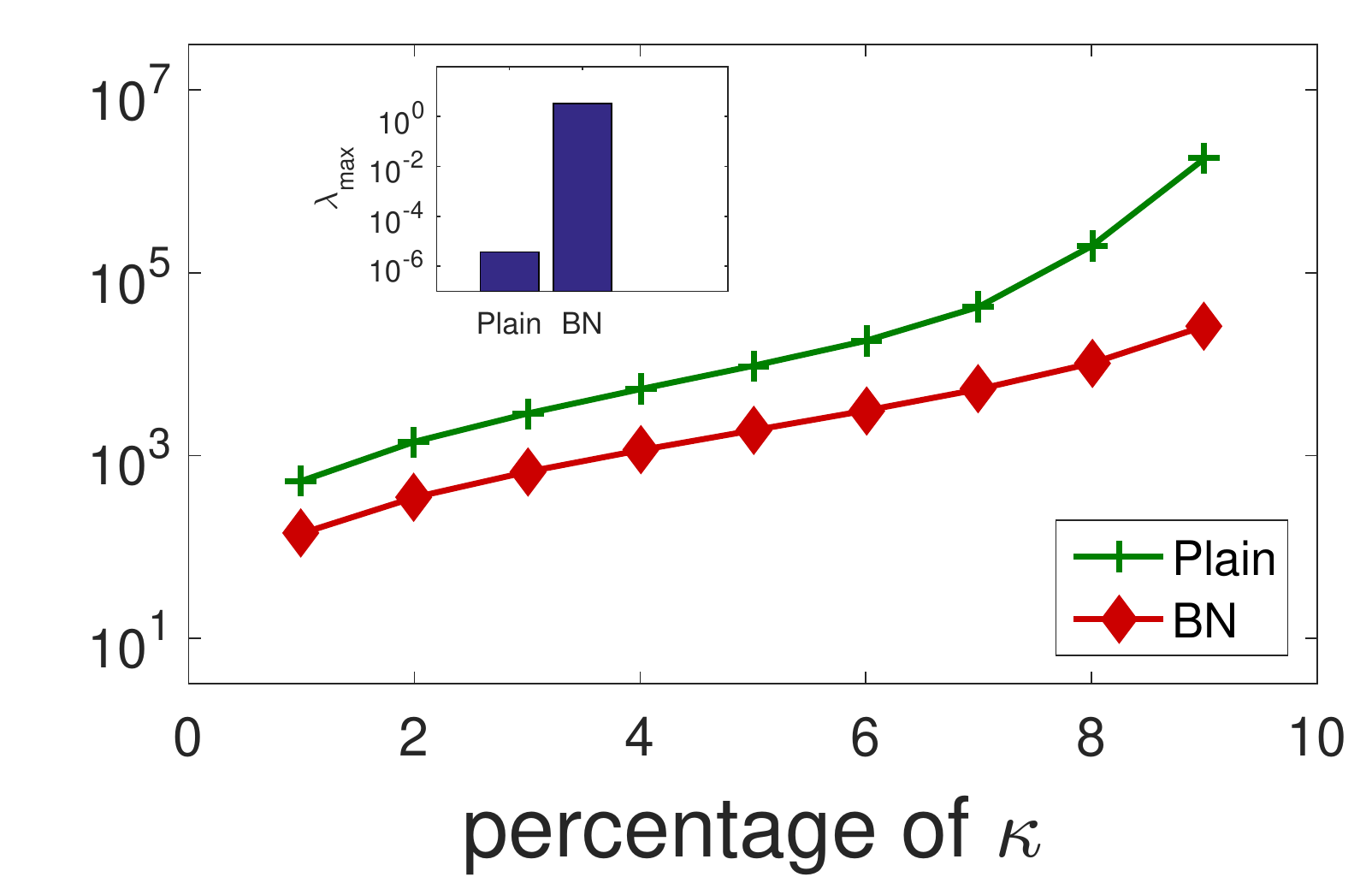}
		\end{minipage}
	}
	\subfigure[sub-$\mathbf{M}$  (the 1st layer)]{
		\begin{minipage}[c]{.32\linewidth}
			\centering
			\includegraphics[width=4.0cm]{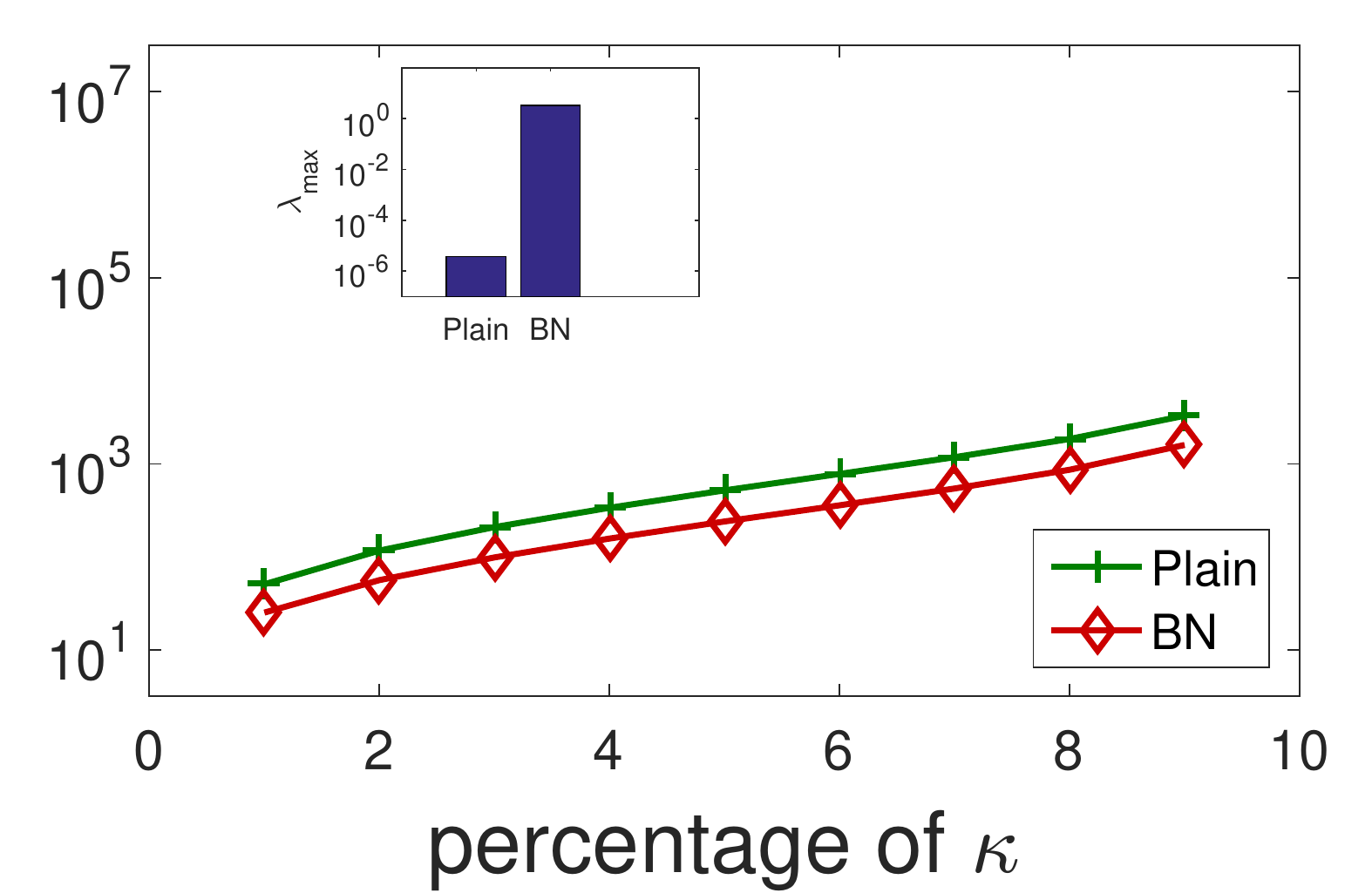}
		\end{minipage}
	}	
	\subfigure[sub-$\mathbf{M}$  (the 2nd layer)]{
		\begin{minipage}[c]{.32\linewidth}
			\centering
			\includegraphics[width=4.0cm]{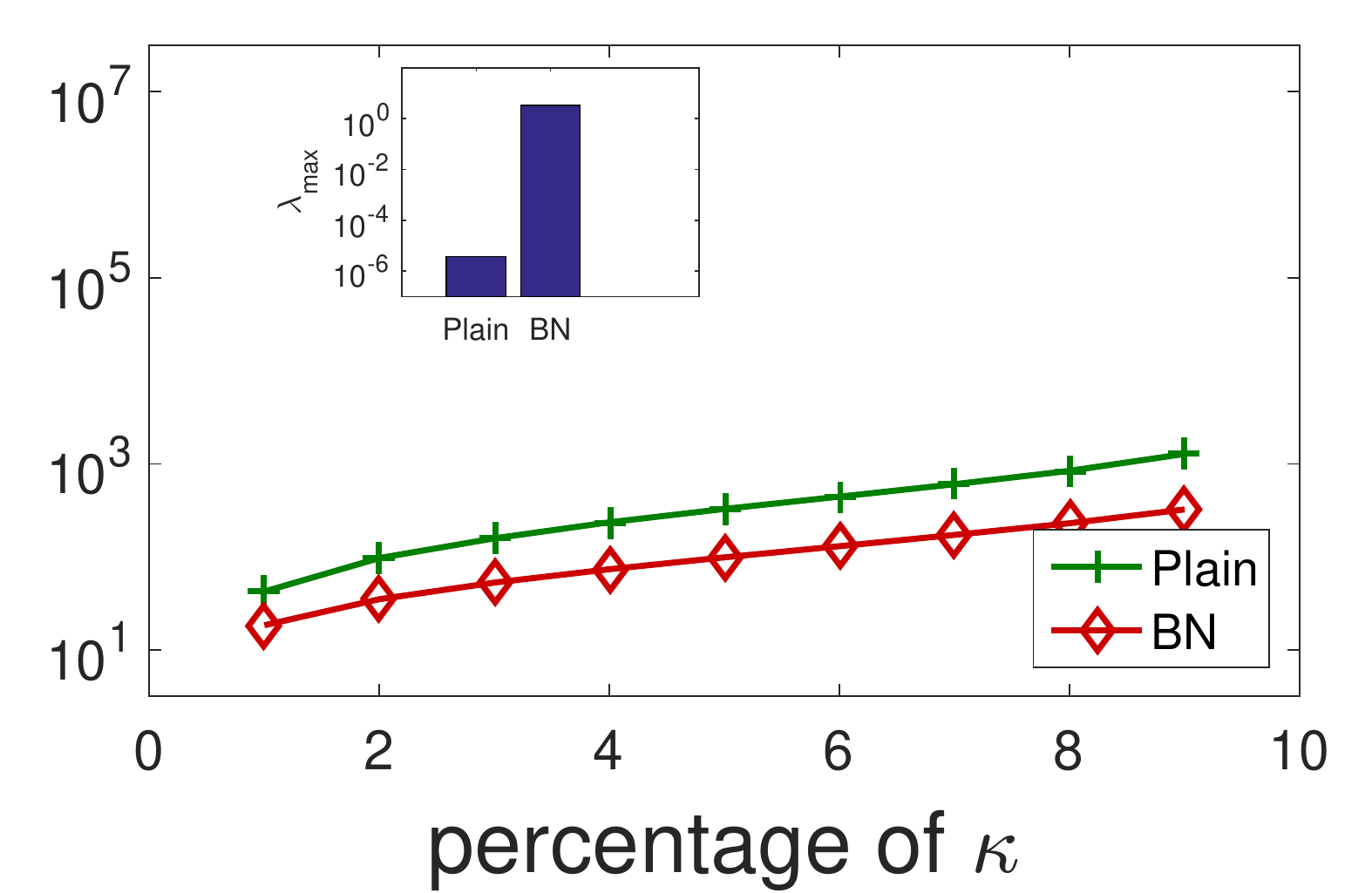}
		\end{minipage}
	}	\\
	\hspace{-0.2in}	\subfigure[sub-$\mathbf{M}$  (the 3rd layer)]{
		\begin{minipage}[c]{.32\linewidth}
			\centering
			\includegraphics[width=4.0cm]{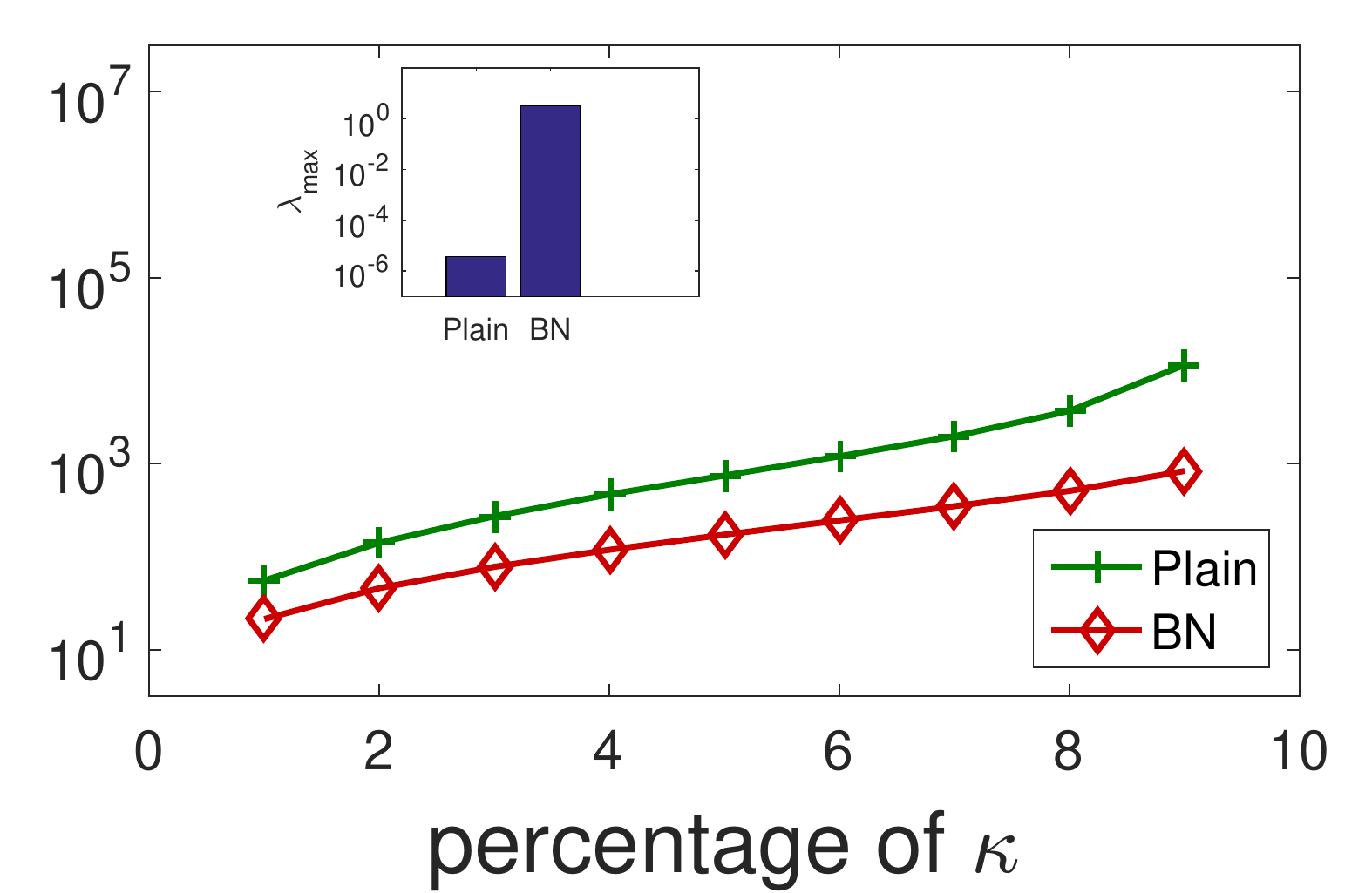}
		\end{minipage}
	}
	\subfigure[sub-$\mathbf{M}$  (the 4th layer)]{
		\begin{minipage}[c]{.32\linewidth}
			\centering
			\includegraphics[width=4.0cm]{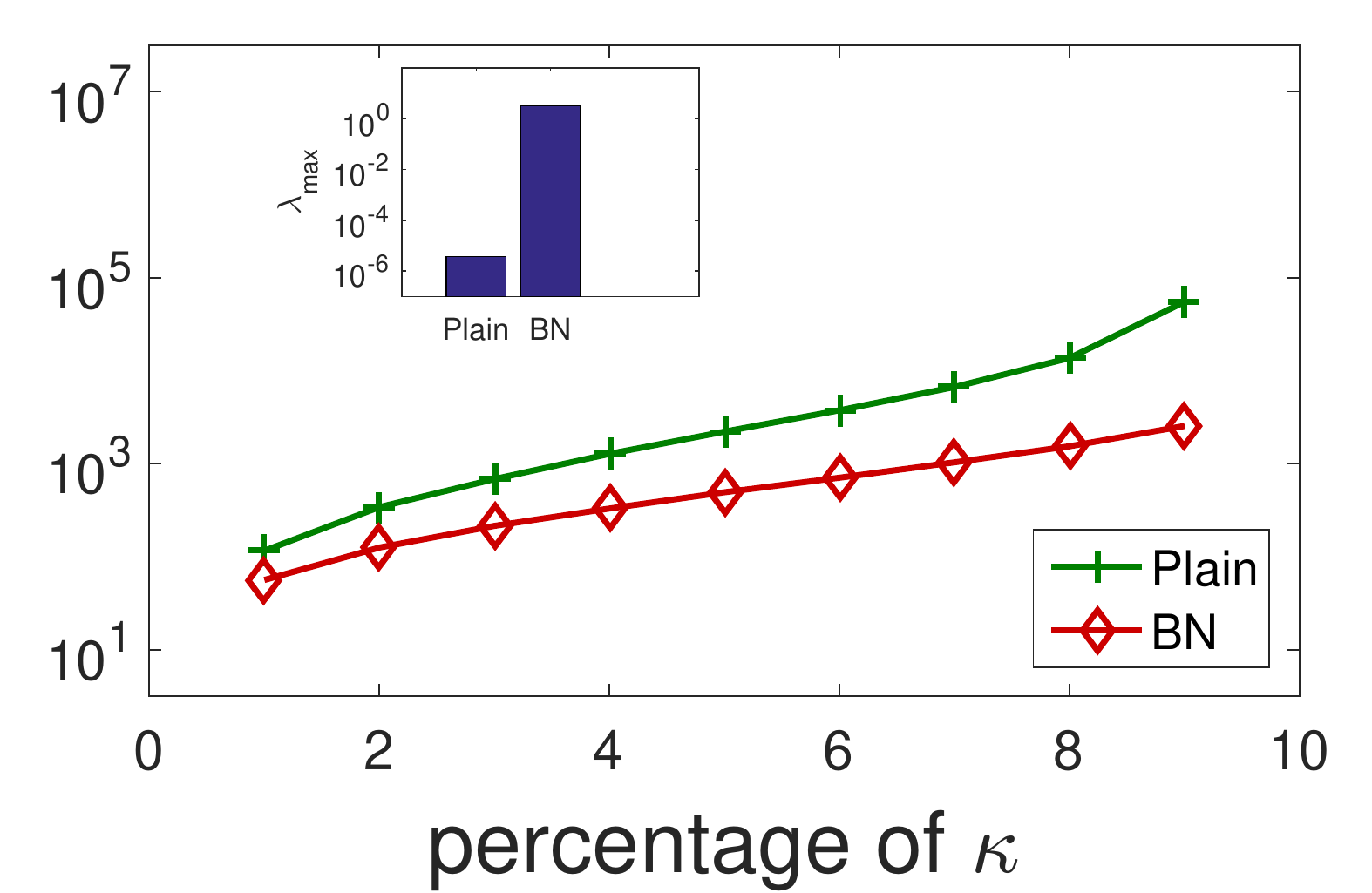}
		\end{minipage}
	}	
	\subfigure[sub-$\mathbf{M}$  (the 5th layer)]{
		\begin{minipage}[c]{.32\linewidth}
			\centering
			\includegraphics[width=4.0cm]{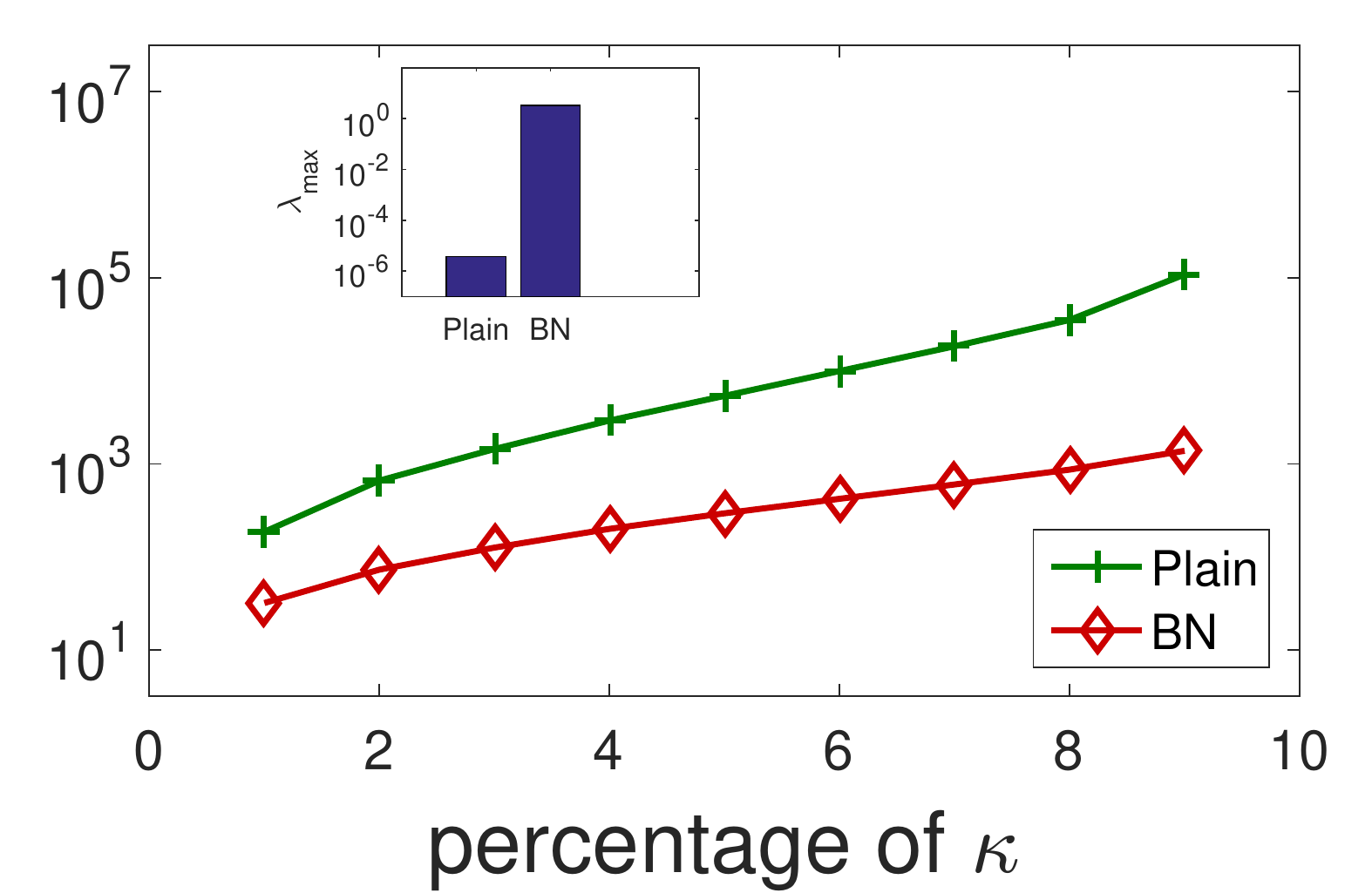}
		\end{minipage}
	}\\
	\hspace{-0.2in}	\subfigure[sub-$\mathbf{M}$  (the 6th layer)]{
		\begin{minipage}[c]{.32\linewidth}
			\centering
			\includegraphics[width=4.0cm]{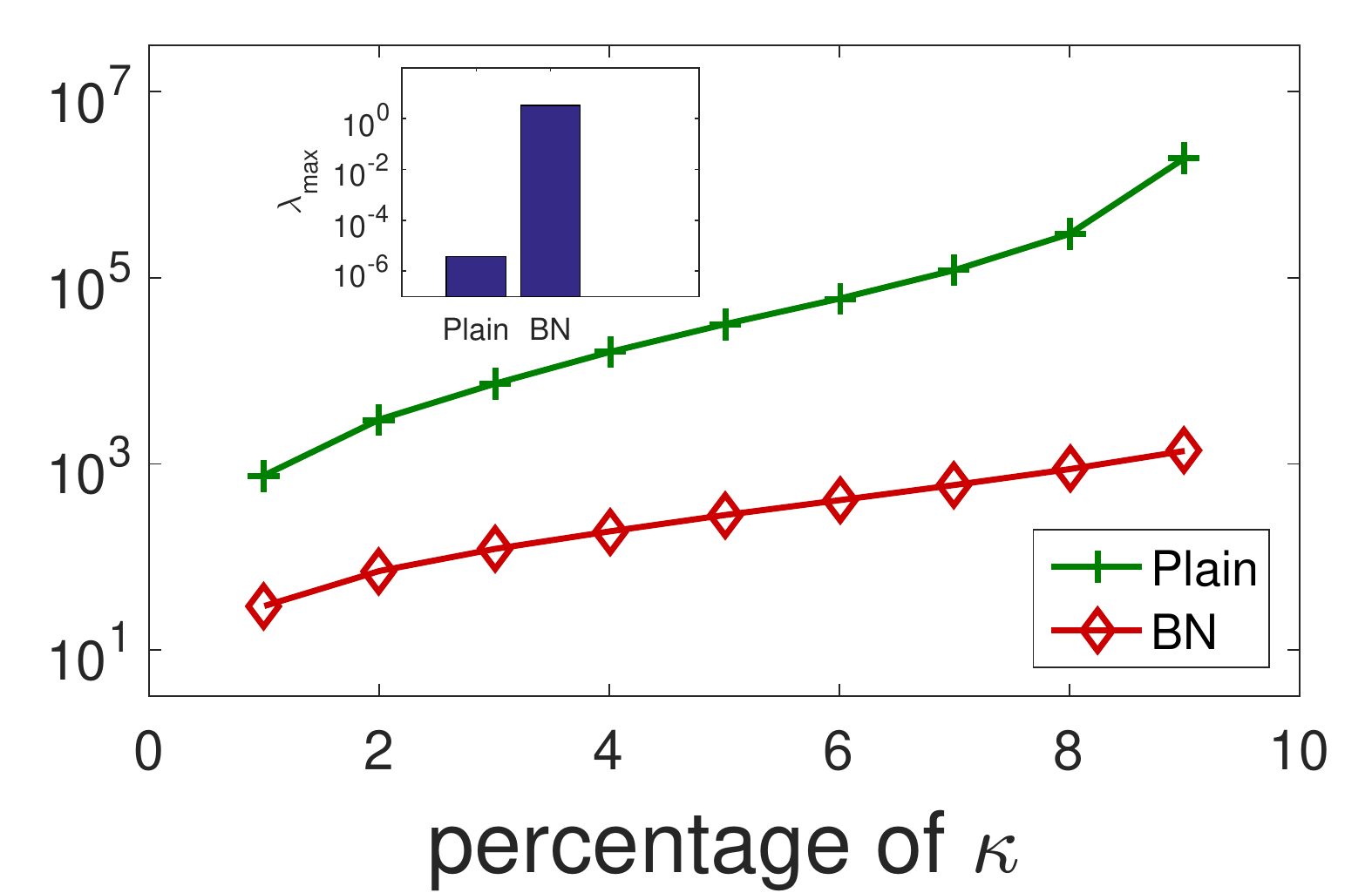}
		\end{minipage}
	}
	\subfigure[sub-$\mathbf{M}$  (the 7th layer)]{
		\begin{minipage}[c]{.32\linewidth}
			\centering
			\includegraphics[width=4.0cm]{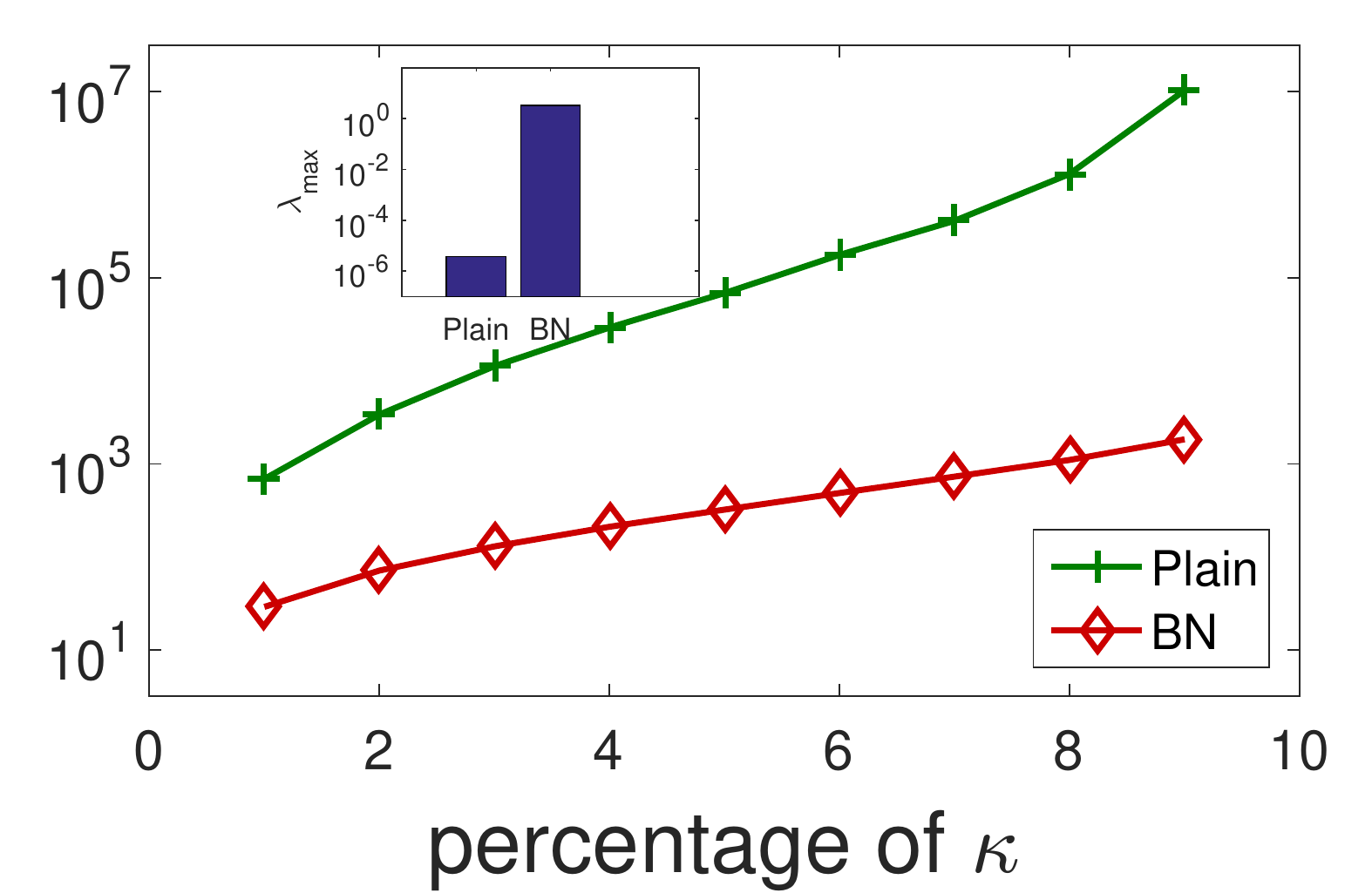}
		\end{minipage}
	}	
	\subfigure[sub-$\mathbf{M}$  (the 8th layer)]{
		\begin{minipage}[c]{.32\linewidth}
			\centering
			\includegraphics[width=4.0cm]{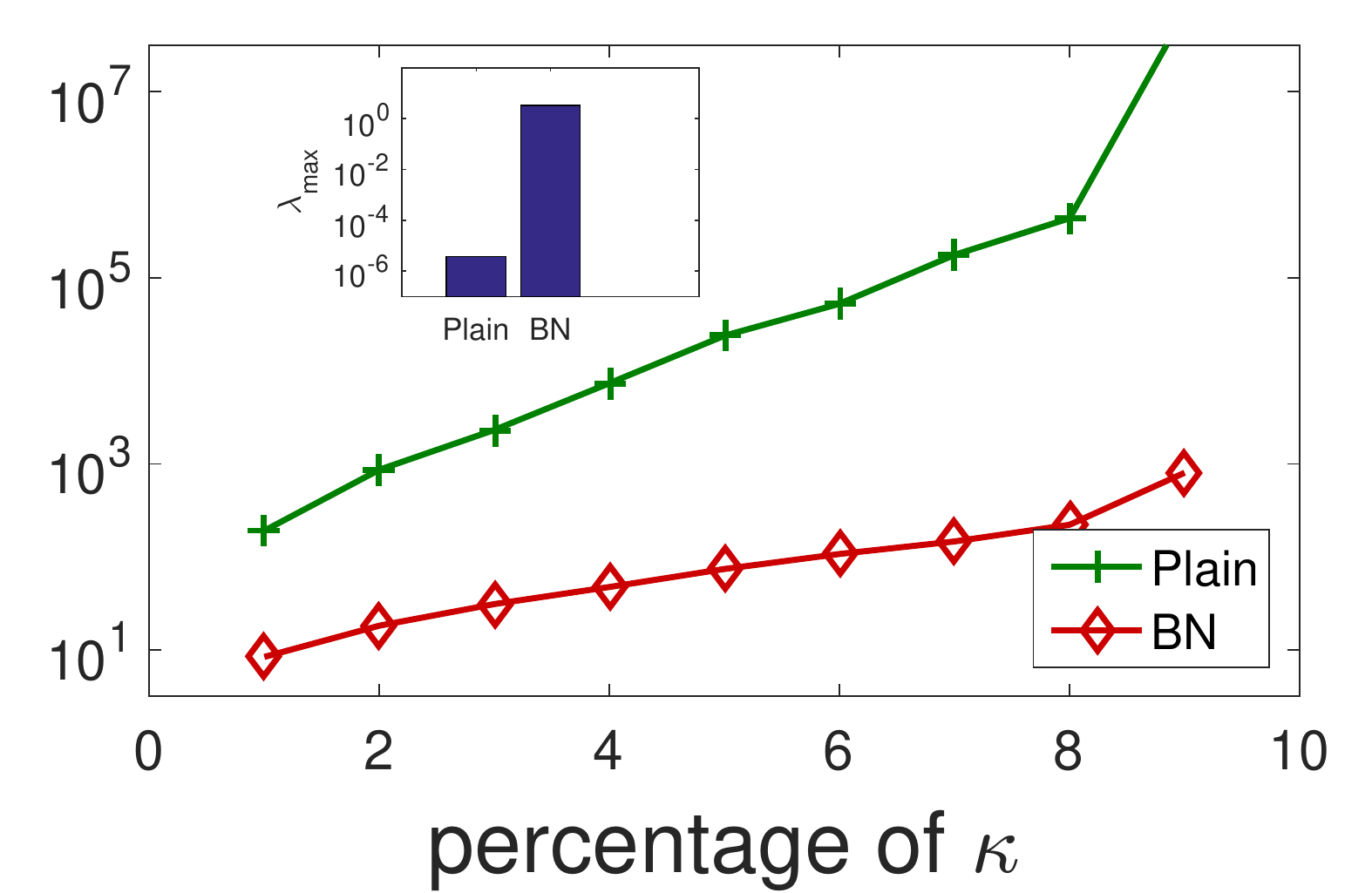}
		\end{minipage}
	}
	\caption{Conditioning analysis for unnormalized (`Plain') and normalized (`BN') networks. We show  the maximum eigenvalue $\lambda_{max}$ and the generalized condition number $\kappa_{p}$ for comparison between the full second moment matrix of sample gradient $\mathbf{M}$ and sub-$\{\mathbf{M}_k\}$.   The experiments are performed on an 8-layer MLP with 24 neurons in each layer, for MNIST classification. The input image is center-cropped and resized to $12 \times 12$ to remove uninformative pixels. We report the corresponding spectrum at random initialization \cite{1998_NN_Yann}.  }
	\label{fig:AP1-AH}
\end{figure}

\begin{figure}[]
	\centering
	\hspace{-0.2in}	\subfigure[full FIM]{
		\begin{minipage}[c]{.32\linewidth}
			\centering
			\includegraphics[width=4.0cm]{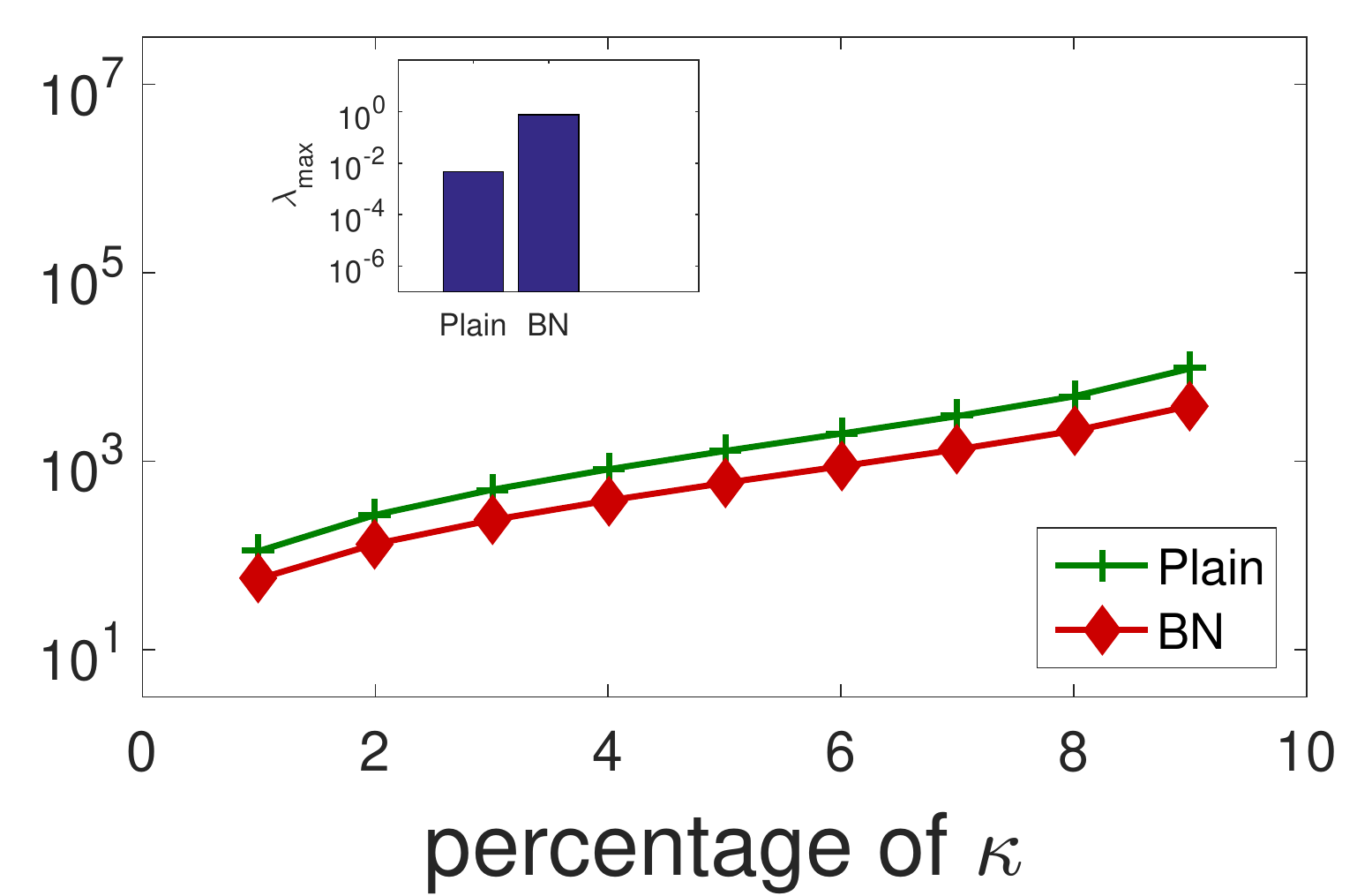}
		\end{minipage}
	}
	\subfigure[sub-FIM (the 2nd layer)]{
		\begin{minipage}[c]{.32\linewidth}
			\centering
			\includegraphics[width=4.0cm]{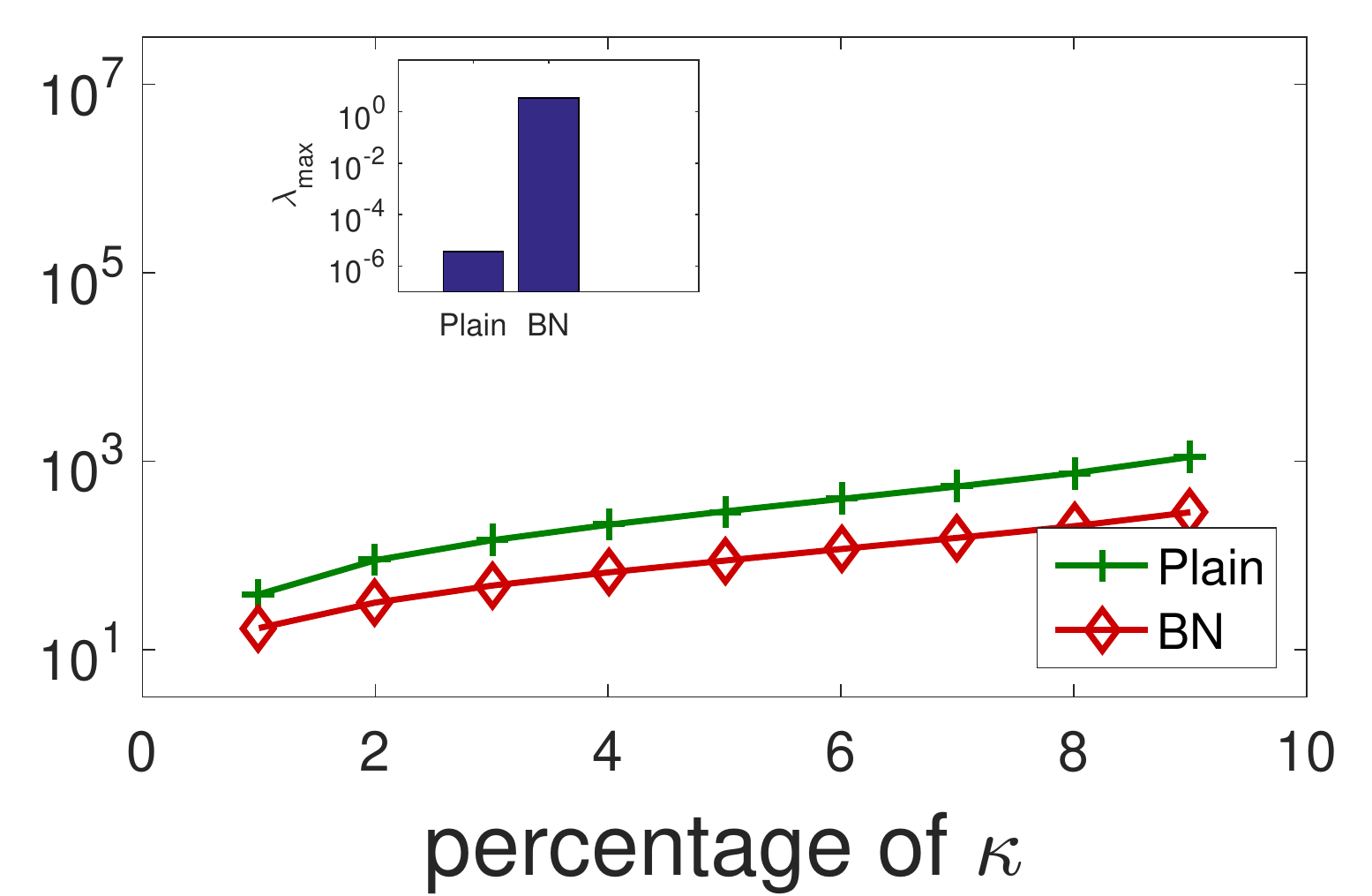}
		\end{minipage}
	}	
	\subfigure[sub-FIM (the 3rd layer)]{
		\begin{minipage}[c]{.32\linewidth}
			\centering
			\includegraphics[width=4.0cm]{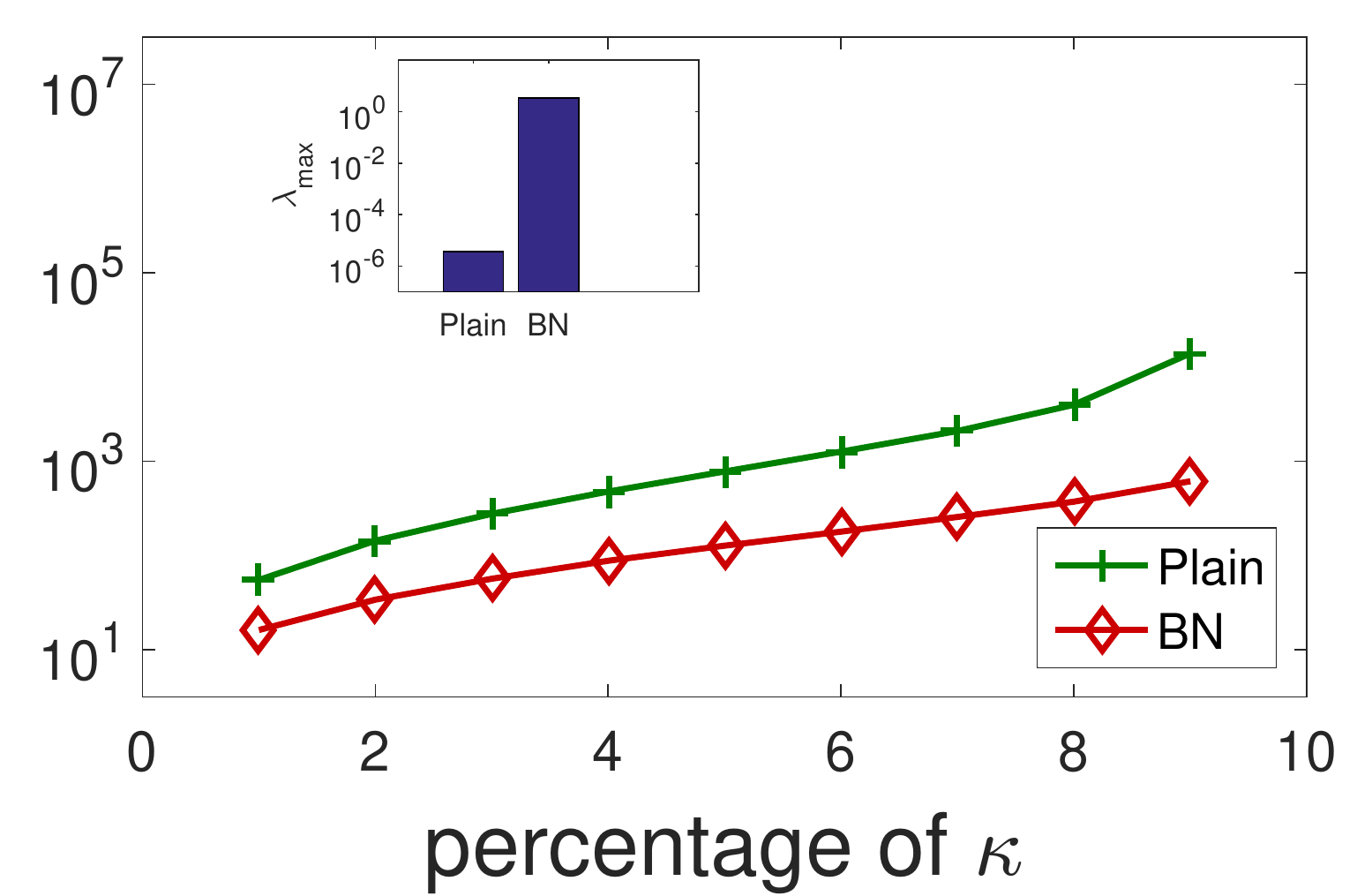}
		\end{minipage}
	}	\\
	\hspace{-0.2in}	\subfigure[full $\mathbf{M}$]{
		\begin{minipage}[c]{.32\linewidth}
			\centering
			\includegraphics[width=4.0cm]{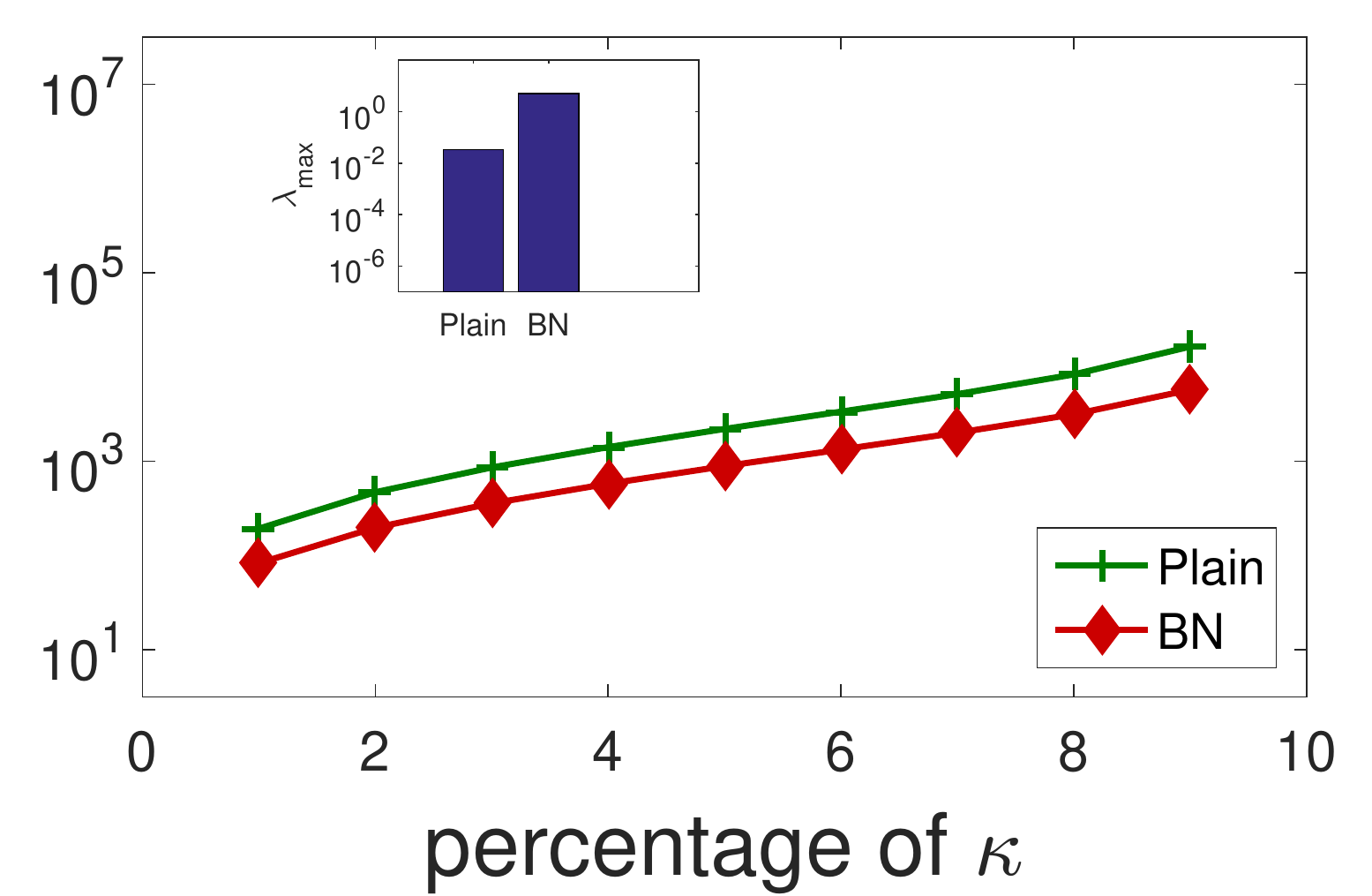}
		\end{minipage}
	}
	\subfigure[sub-$\mathbf{M}$  (the 2nd layer)]{
		\begin{minipage}[c]{.32\linewidth}
			\centering
			\includegraphics[width=4.0cm]{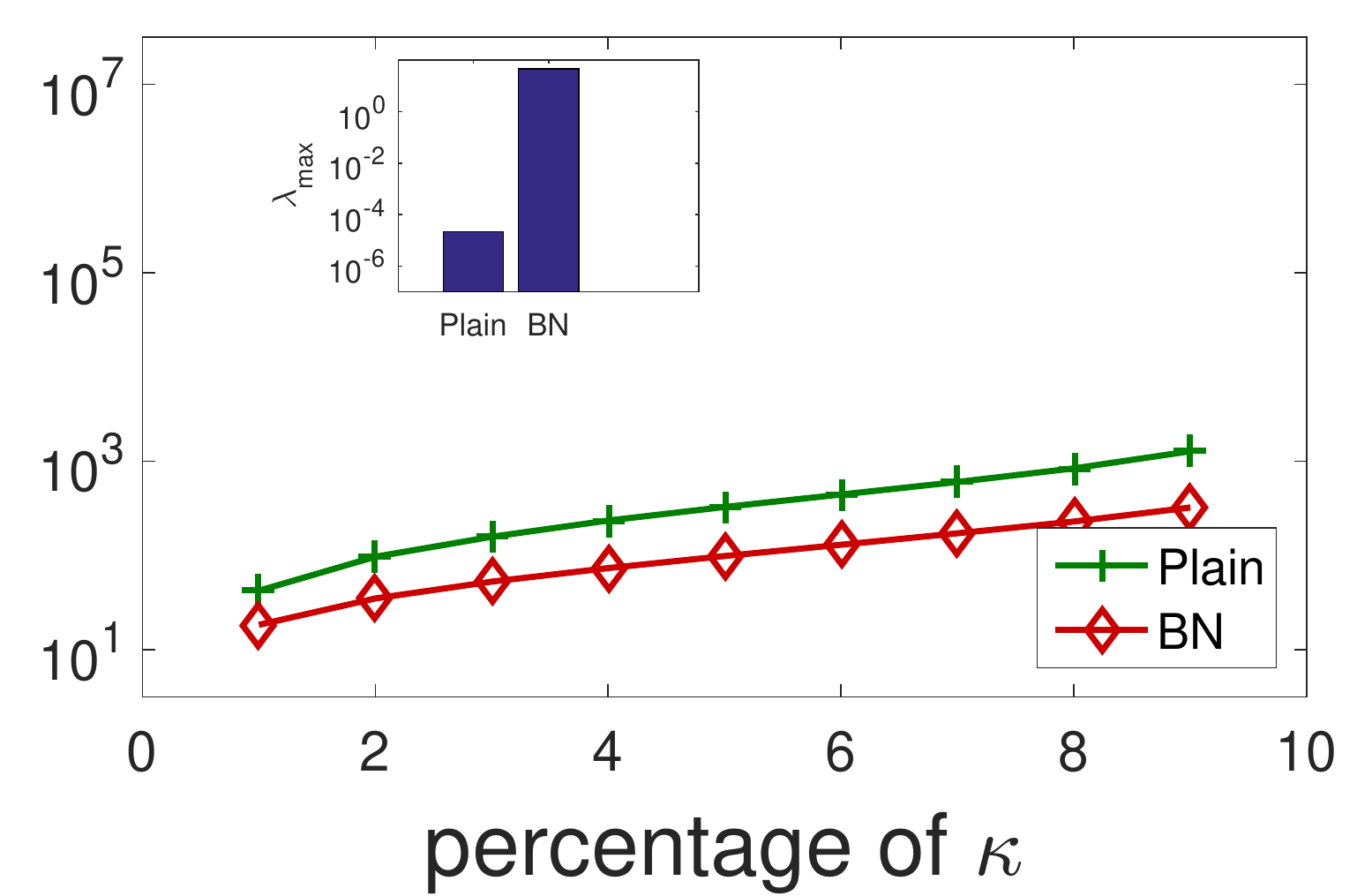}
		\end{minipage}
	}	
	\subfigure[sub-$\mathbf{M}$  (the 3rd layer)]{
		\begin{minipage}[c]{.32\linewidth}
			\centering
			\includegraphics[width=4.0cm]{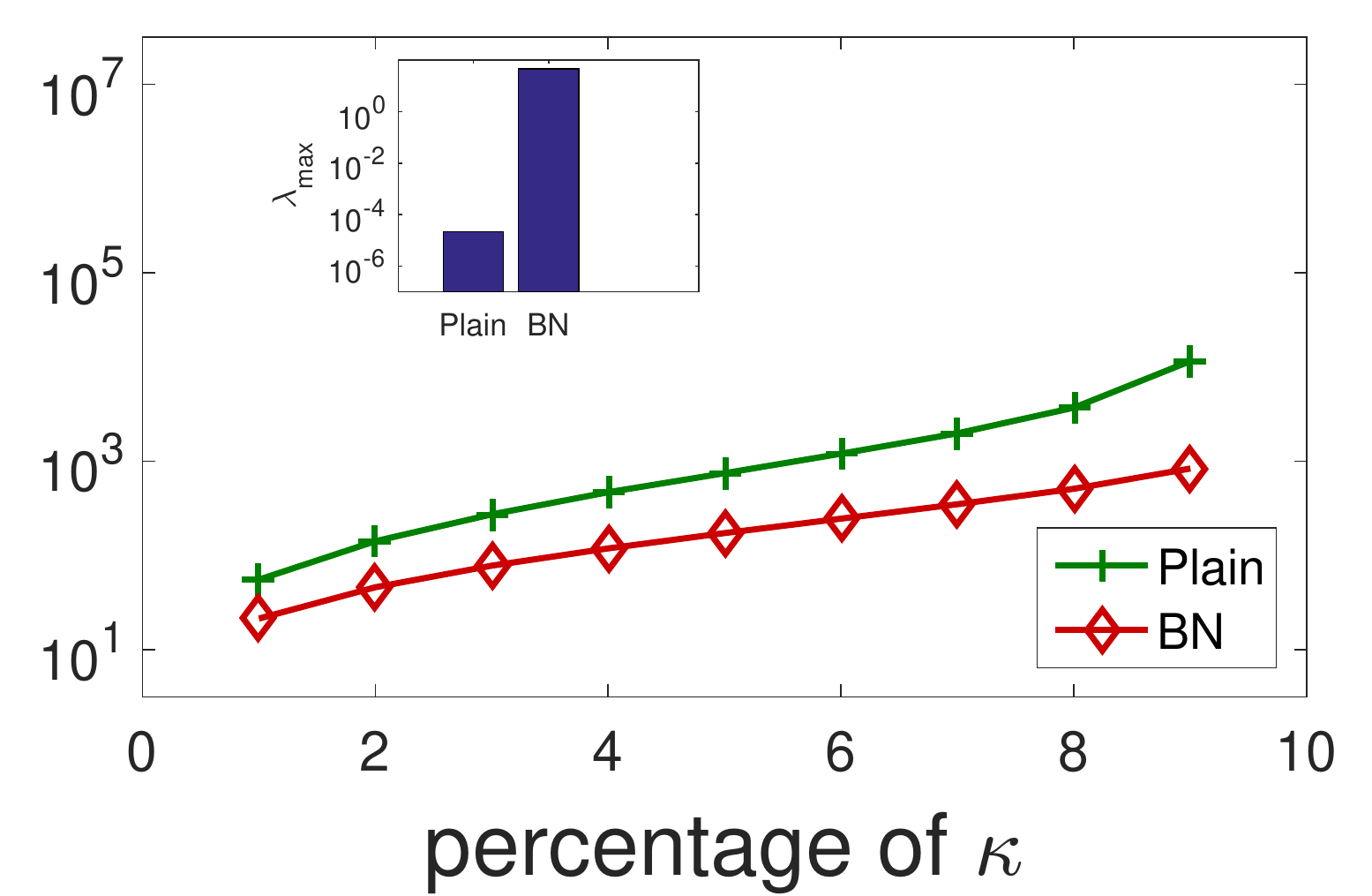}
		\end{minipage}
	}\\
	\caption{Conditioning analysis for unnormalized (`Plain') and normalized  (`BN') networks.  The experiments are performed on a 4-layer MLP with 24 neurons in each layer, for MNIST classification. The input image is center-cropped and resized to $12 \times 12$ to remove the uninformative pixels. We report the corresponding spectrum at random initialization \cite{1998_NN_Yann}.  }
	\label{fig:AP1-l4-FIM}
\end{figure}

\begin{figure}[]
	\centering
	\hspace{-0.2in}	\subfigure[full FIM]{
		\begin{minipage}[c]{.32\linewidth}
			\centering
			\includegraphics[width=4.0cm]{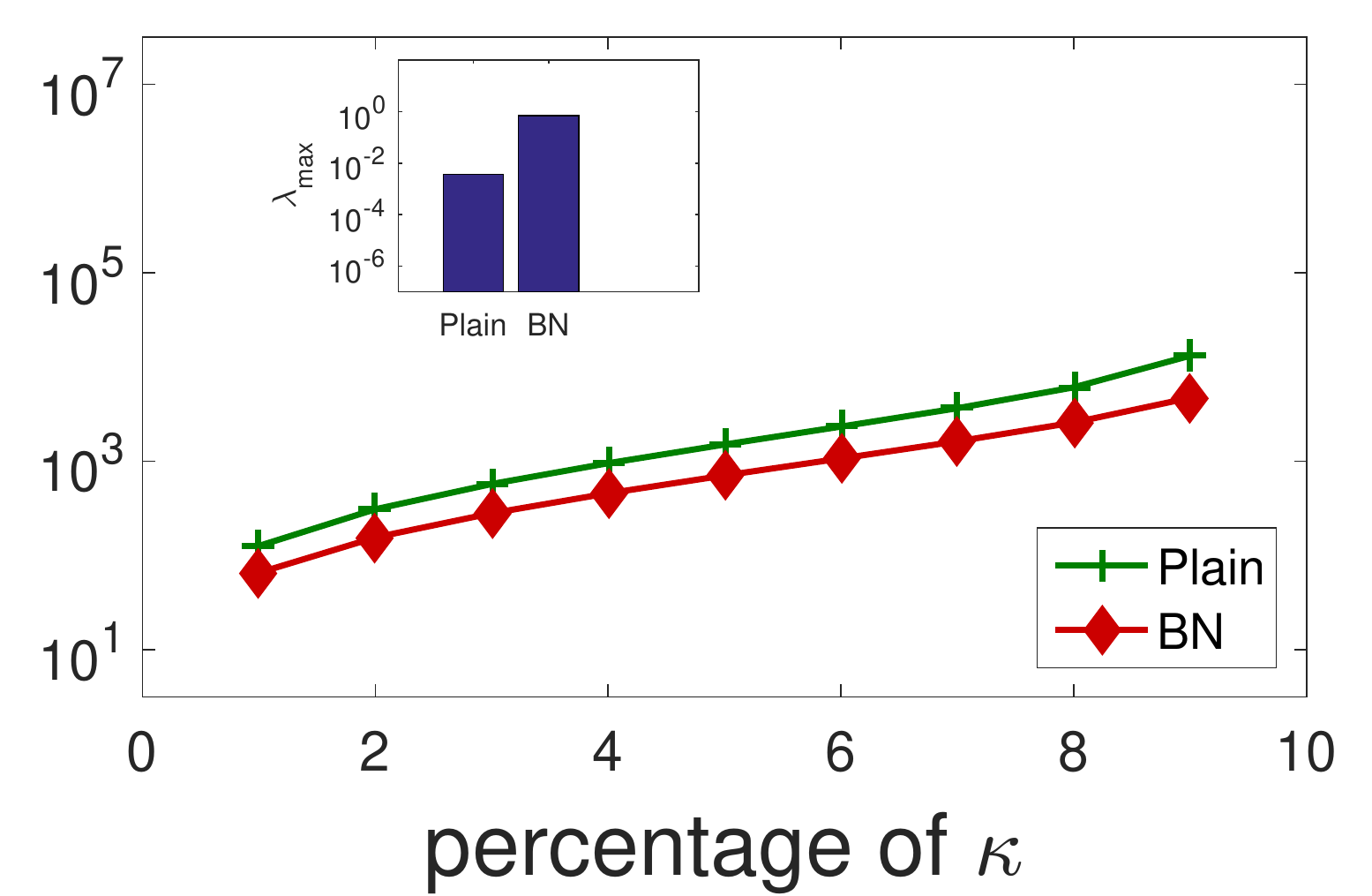}
		\end{minipage}
	}
	\subfigure[sub-FIM (the 2nd layer)]{
		\begin{minipage}[c]{.32\linewidth}
			\centering
			\includegraphics[width=4.0cm]{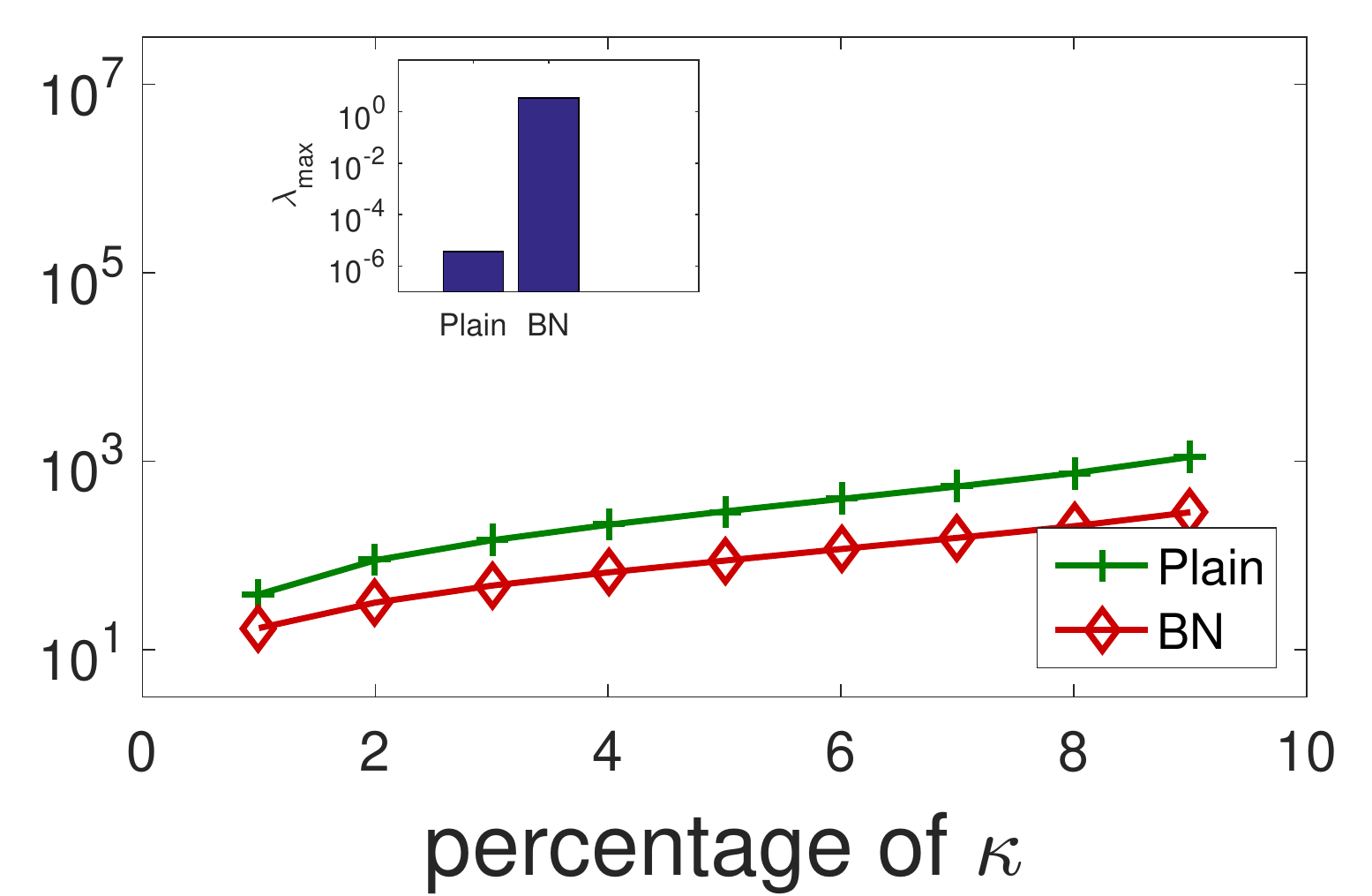}
		\end{minipage}
	}	
	\subfigure[sub-FIM (the 3rd layer)]{
		\begin{minipage}[c]{.32\linewidth}
			\centering
			\includegraphics[width=4.0cm]{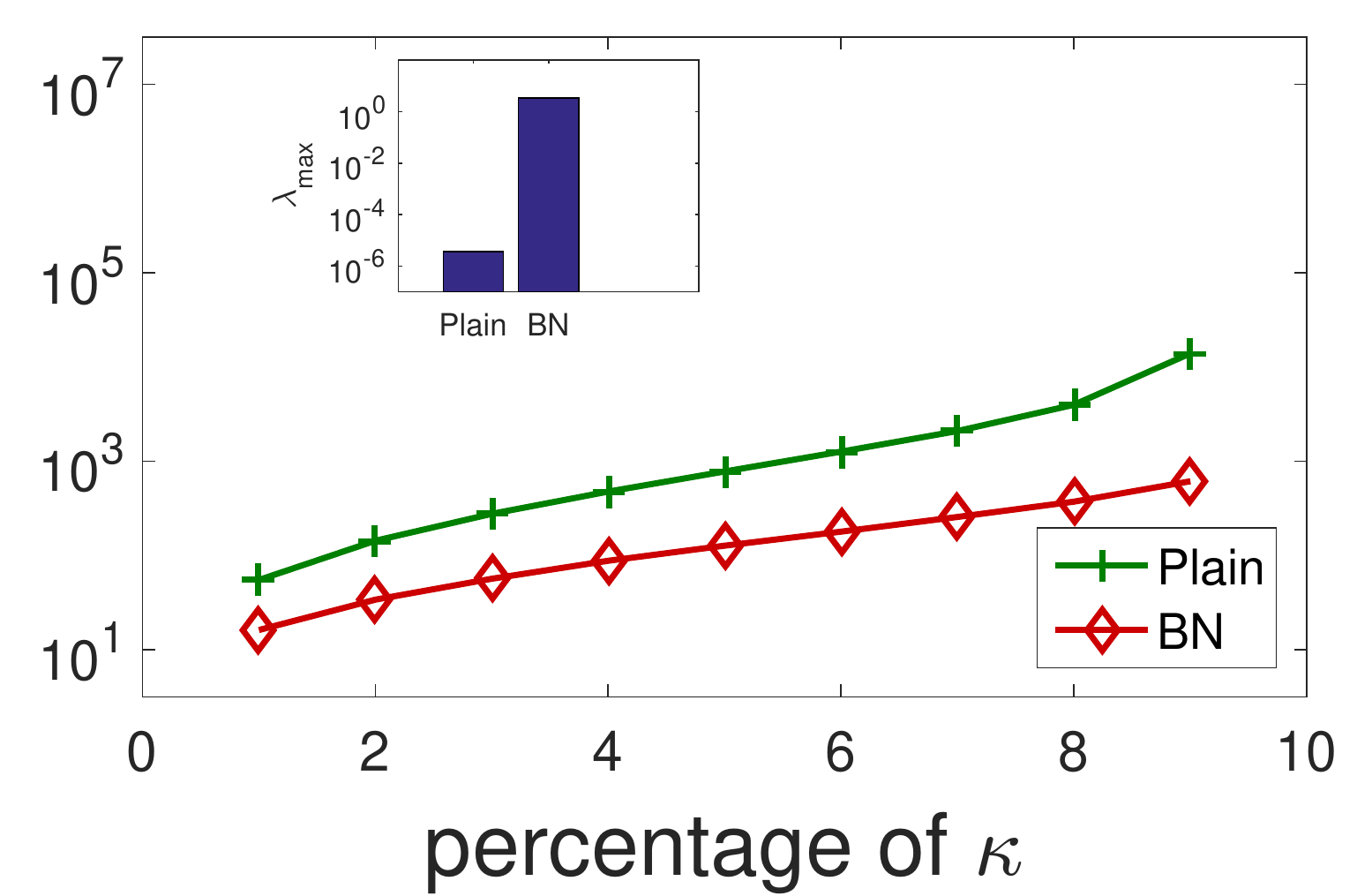}
		\end{minipage}
	}	\\
	\hspace{-0.2in}	\subfigure[full $\mathbf{M}$ ]{
		\begin{minipage}[c]{.32\linewidth}
			\centering
			\includegraphics[width=4.0cm]{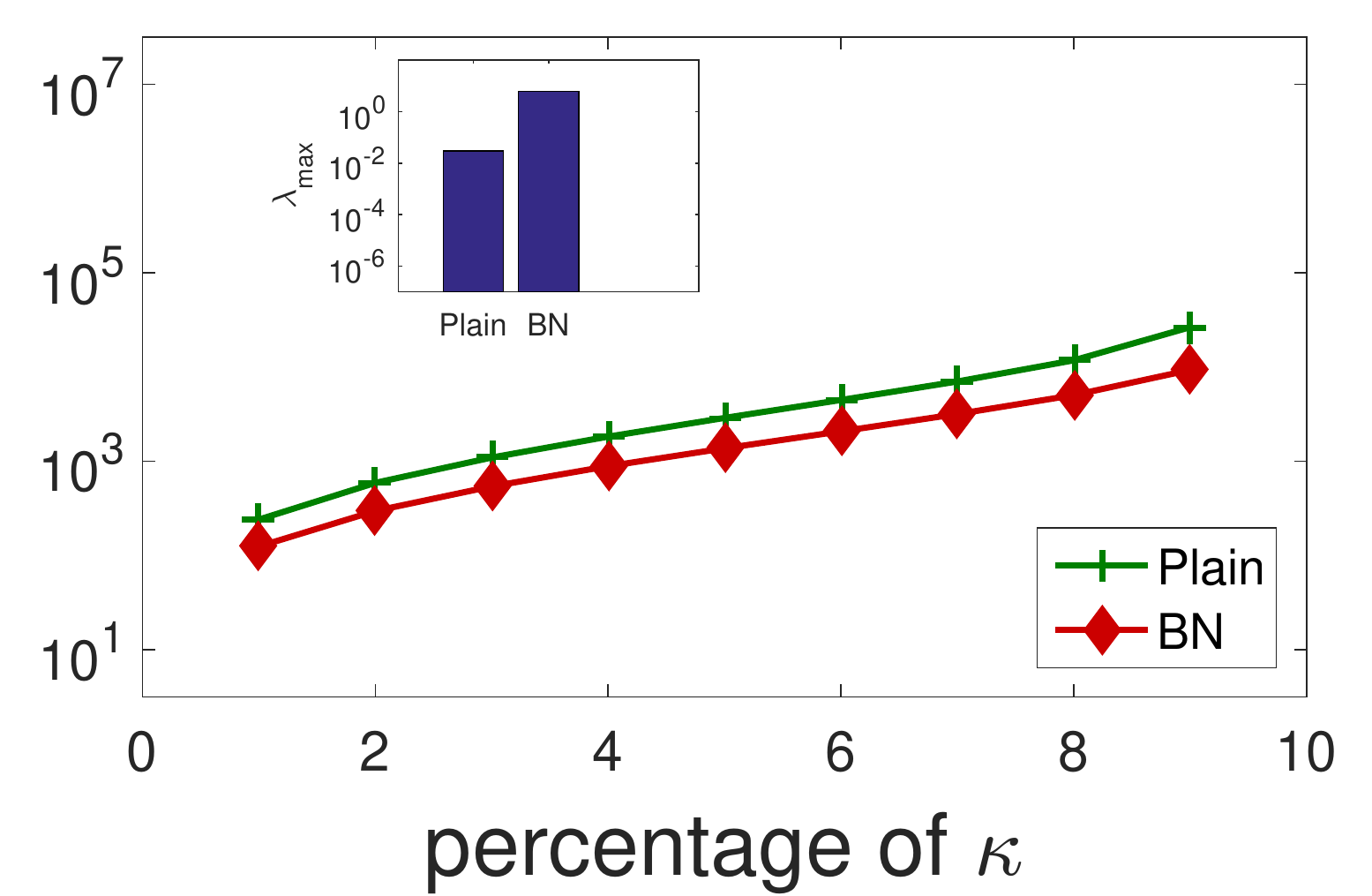}
		\end{minipage}
	}
	\subfigure[sub-$\mathbf{M}$  (the 2nd layer)]{
		\begin{minipage}[c]{.32\linewidth}
			\centering
			\includegraphics[width=4.0cm]{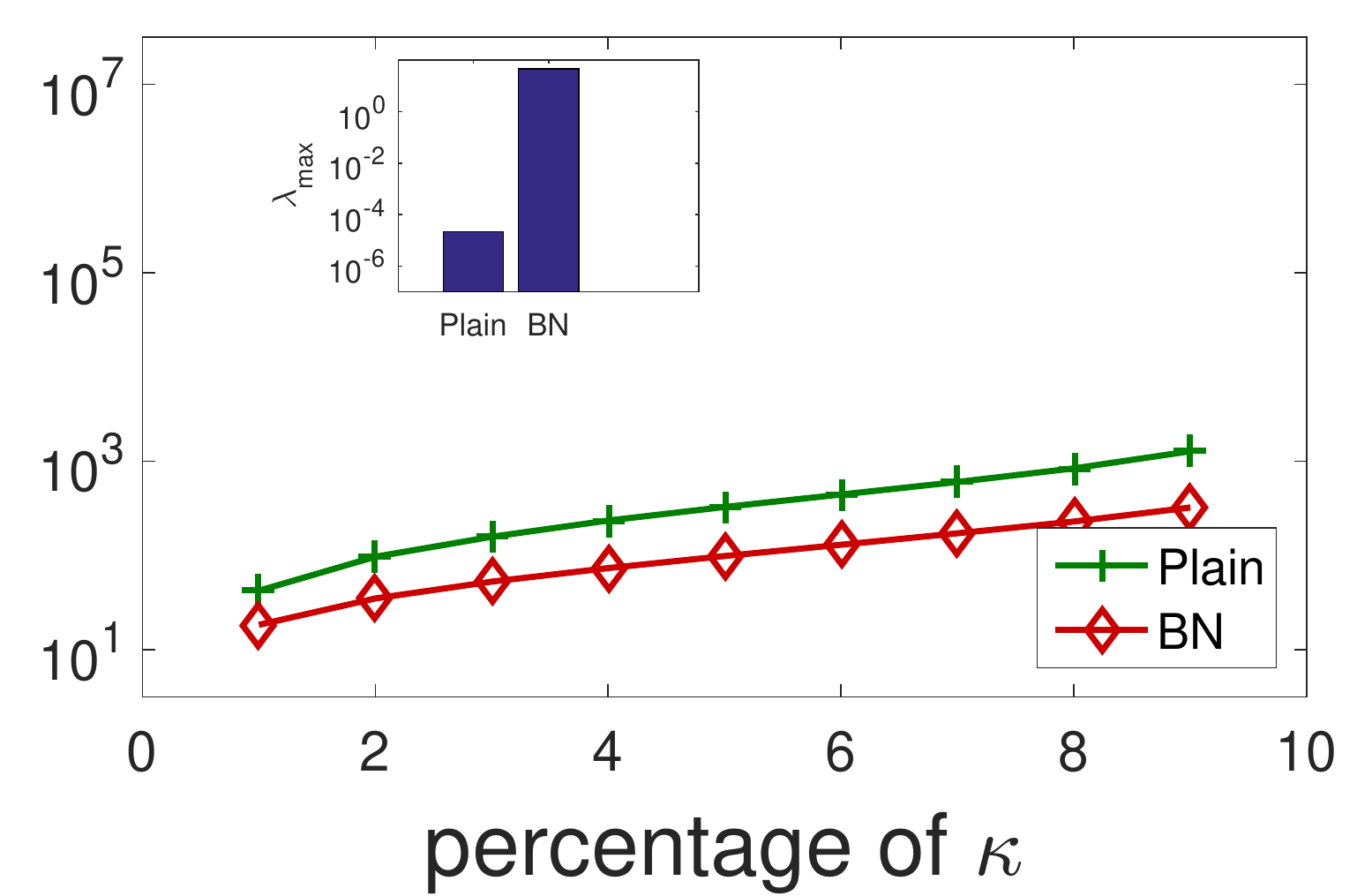}
		\end{minipage}
	}	
	\subfigure[sub-$\mathbf{M}$ (the 3rd layer)]{
		\begin{minipage}[c]{.32\linewidth}
			\centering
			\includegraphics[width=4.0cm]{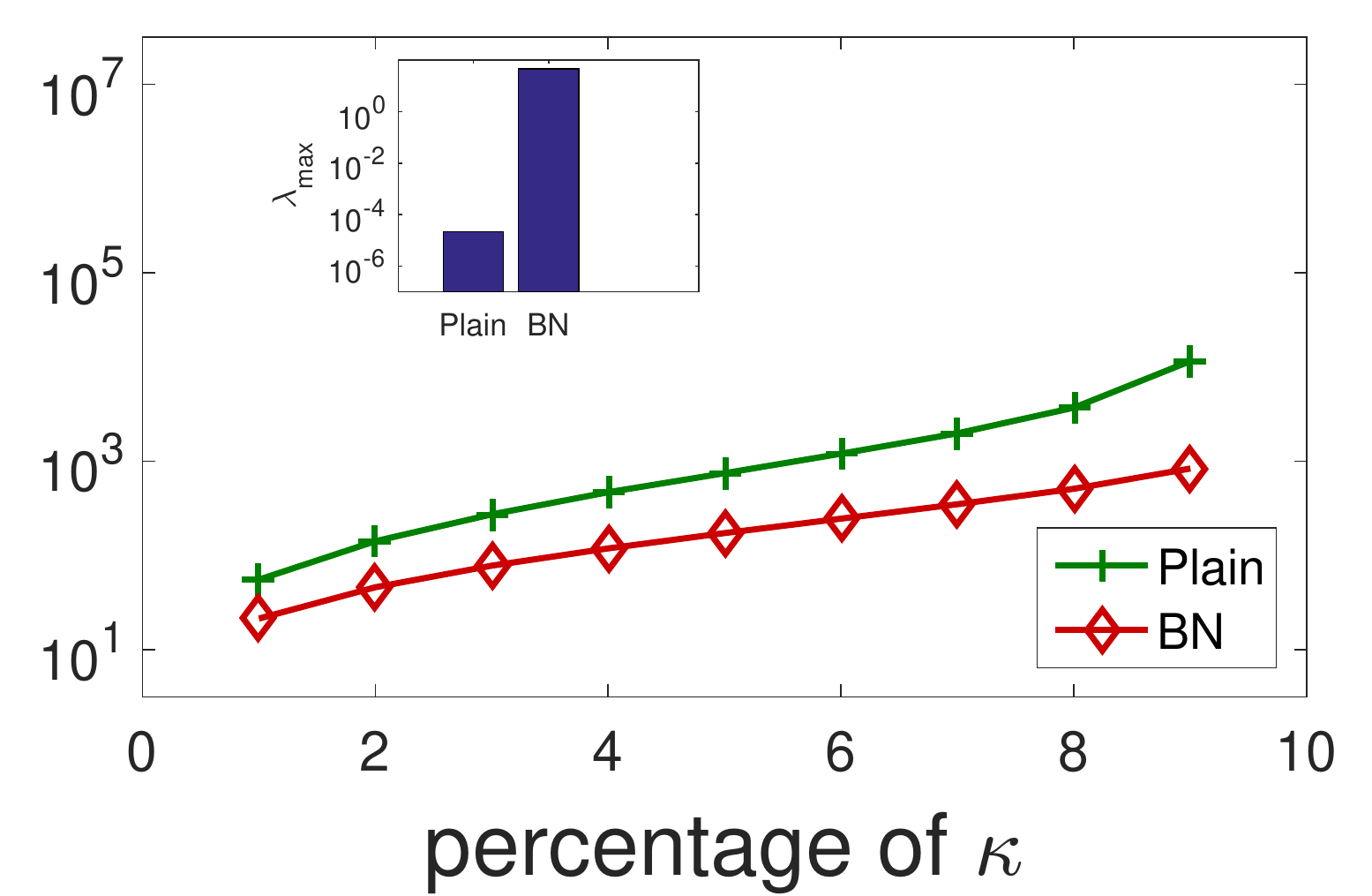}
		\end{minipage}
	}\\
	\caption{Conditioning analysis for unnormalized (`Plain') and normalized (`BN') networks.   The experiments are performed on a 4-layer MLP with 36 neurons in each layer, for MNIST classification. The input image is center-cropped and resized to $12 \times 12$ to remove the uninformative pixels. We report the corresponding spectrum at random initialization \cite{1998_NN_Yann}.  }
	\label{fig:AP1-l4-h36-FIM}
\end{figure}

\section{Comparison Between the Analyses with Full Curvature and Sub-curvature Matrices}
\label{Sec-sup-approximate}
In Section~\ref{Sec-Conditioning analysis} of the  paper, we conduct experiments to analyze the training dynamics of  the unnormalized (`Plain') and batch normalized \cite{2015_ICML_Ioffe} (`BN') networks, on an 8-layer MLP, by analyzing the spectrum of the full Fisher Information Matrix (FIM)  and  sub-FIMs. We only report the sub-FIMs with respect to the 3rd and 6th layer. Here, we 
show the corresponding results for all sum-FIMs in Figure \ref{fig:AP1-FIM-full}.

We also conduct experiments to analyze `Plain' and `BN', using the second moment matrix of sample gradient $\mathbf{M}$. The results are shown in   Figure \ref{fig:AP1-AH}. We have the same observation as when using the full FIM.
\revise{It is interesting to use sub-Hessian for the conditioning analysis, since Hessian provides a good characterization of the landscapes~\cite{2018_NIPS_Li}. 
	We believe that the max eigenvalue of sub-Hessian can also indicate the magnitude of the weight-gradient in each layer, like the sub-FIM/sub-M. However, the general condition number shown in this paper is not well defined for sub-Hessian, since it has negative eigenvalues.
}

We further conduct experiments on a 4-layer MLP with 24 neurons in each layer, and a 4-layer MLP with 36 neurons in each layer. The corresponding results with different curvature matrices are shown in Figures~\ref{fig:AP1-l4-FIM} and~\ref{fig:AP1-l4-h36-FIM}, respectively. 

\vspace{-0.15in}
\subsubsection{Complexity Analysis}	
\revise{
	Here, we provide the complexity analysis. For simplifying  notation, we consider a $L$ layer MLP with $d$ neurons in each layer. The example number is $N$.  For full conditioning analysis,  the cost  includes computing the curvature matrix  ($O(N(Ld^2)^2) $) and the eigen-decomposition ($ O((Ld^2)^3)$). 
	The computation cost is reduced to $O(NL(d^2)^2)+ O(L (d^2)^3)$ for layer-wise conditioning analysis, and further reduced to  $O(NL d^2) + O(Ld^3)$ using our efficient approximation. Note that we only consider naive implementation without any tricks in acceleration (e.g., using implicitly restarted Lanczos method~\cite{2018_NIPS_Li}).	
}

\begin{figure*}[h]
	\centering
	\hspace{-0.2in}		\subfigure[ $\lambda_{max}(\Sigma_x)$]{
		\begin{minipage}[c]{.32\linewidth}
			\centering
			\includegraphics[width=4.0cm]{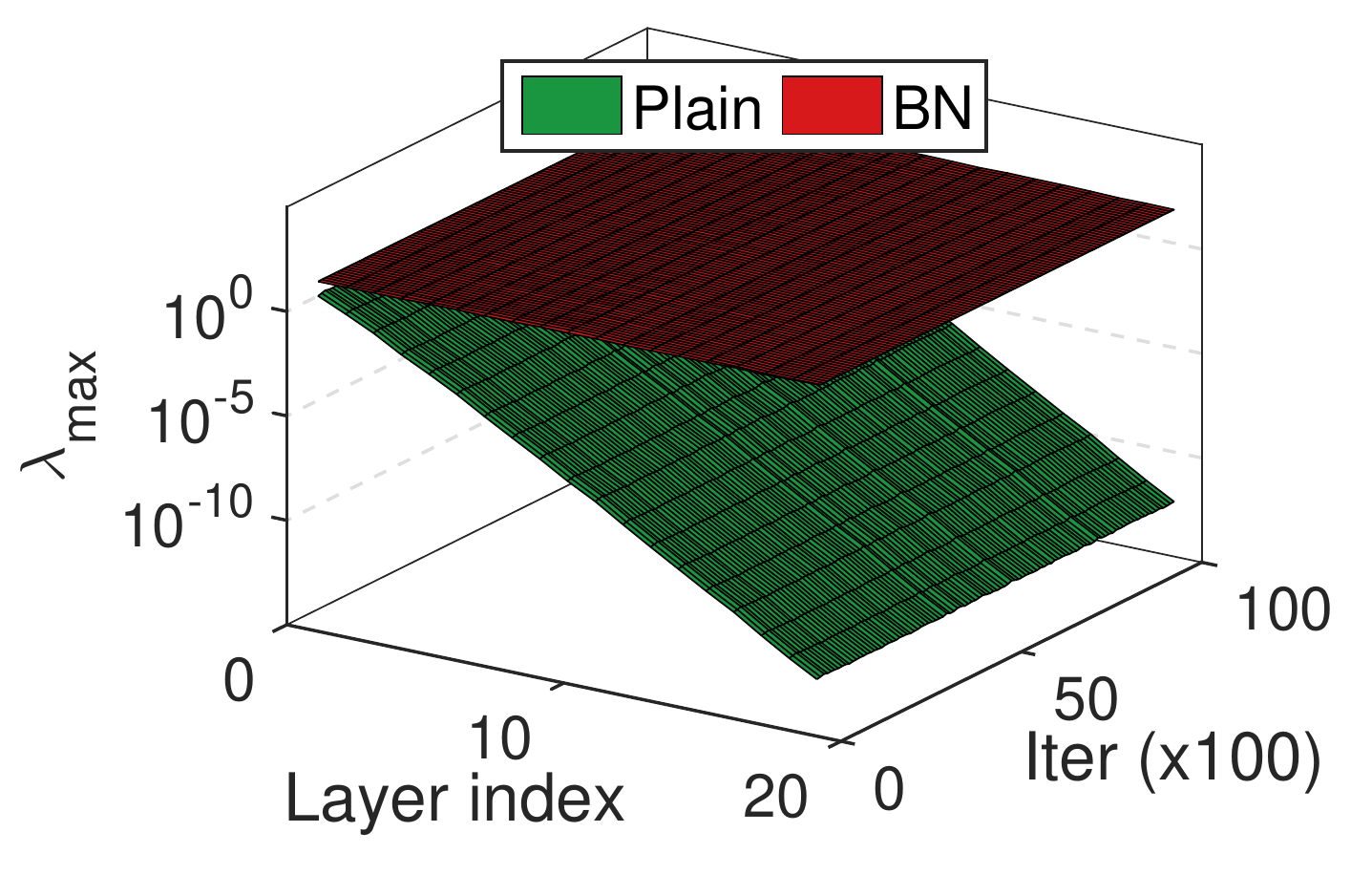}
		\end{minipage}
	}
	\subfigure[$\lambda_{max}(\Sigma_{\nabla \mathbf{h}})$]{
		\begin{minipage}[c]{.32\linewidth}
			\centering
			\includegraphics[width=4.0cm]{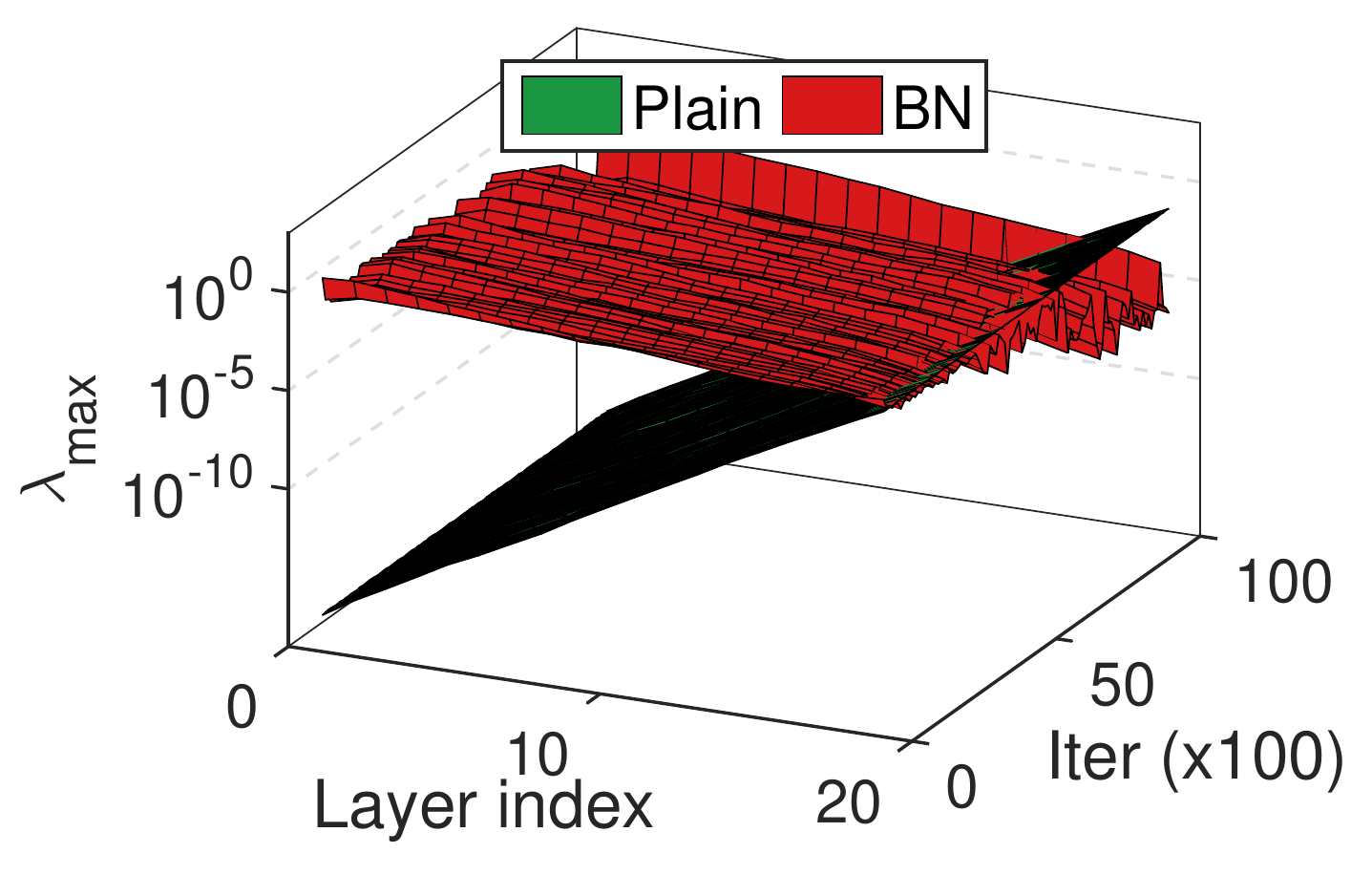}
		\end{minipage}
	}
	\subfigure[Training loss]{
		\begin{minipage}[c]{.32\linewidth}
			\centering
			\includegraphics[width=4.0cm]{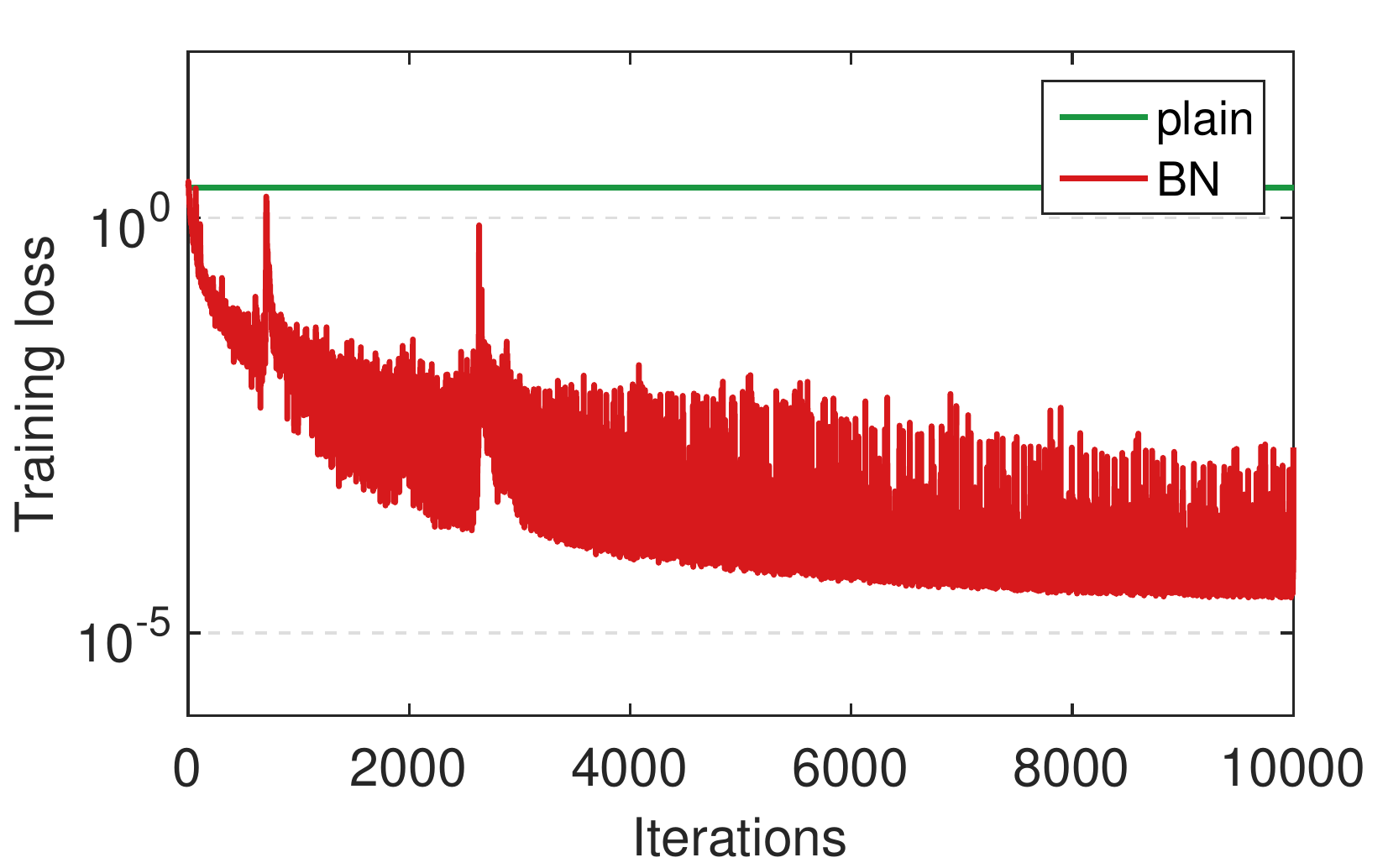}
		\end{minipage}
	}
	\\
	\vspace{-0.12in}
	
	\hspace{-0.2in}		\subfigure[ $\lambda_{max}(\Sigma_x)$]{
		\begin{minipage}[c]{.32\linewidth}
			\centering
			\includegraphics[width=4.0cm]{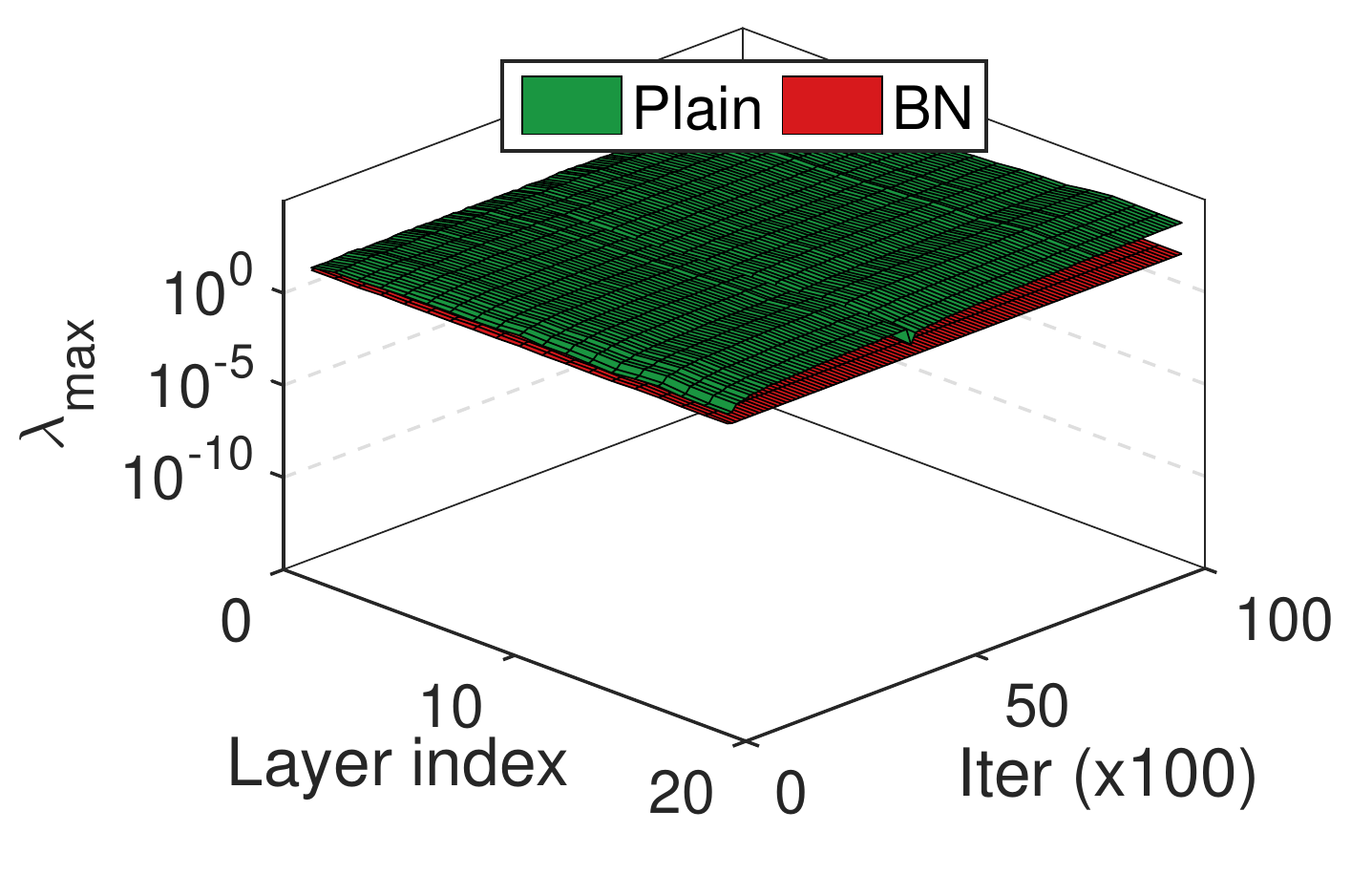}
		\end{minipage}
	}
	\subfigure[$\lambda_{max}(\Sigma_{\nabla \mathbf{h}})$]{
		\begin{minipage}[c]{.32\linewidth}
			\centering
			\includegraphics[width=4.0cm]{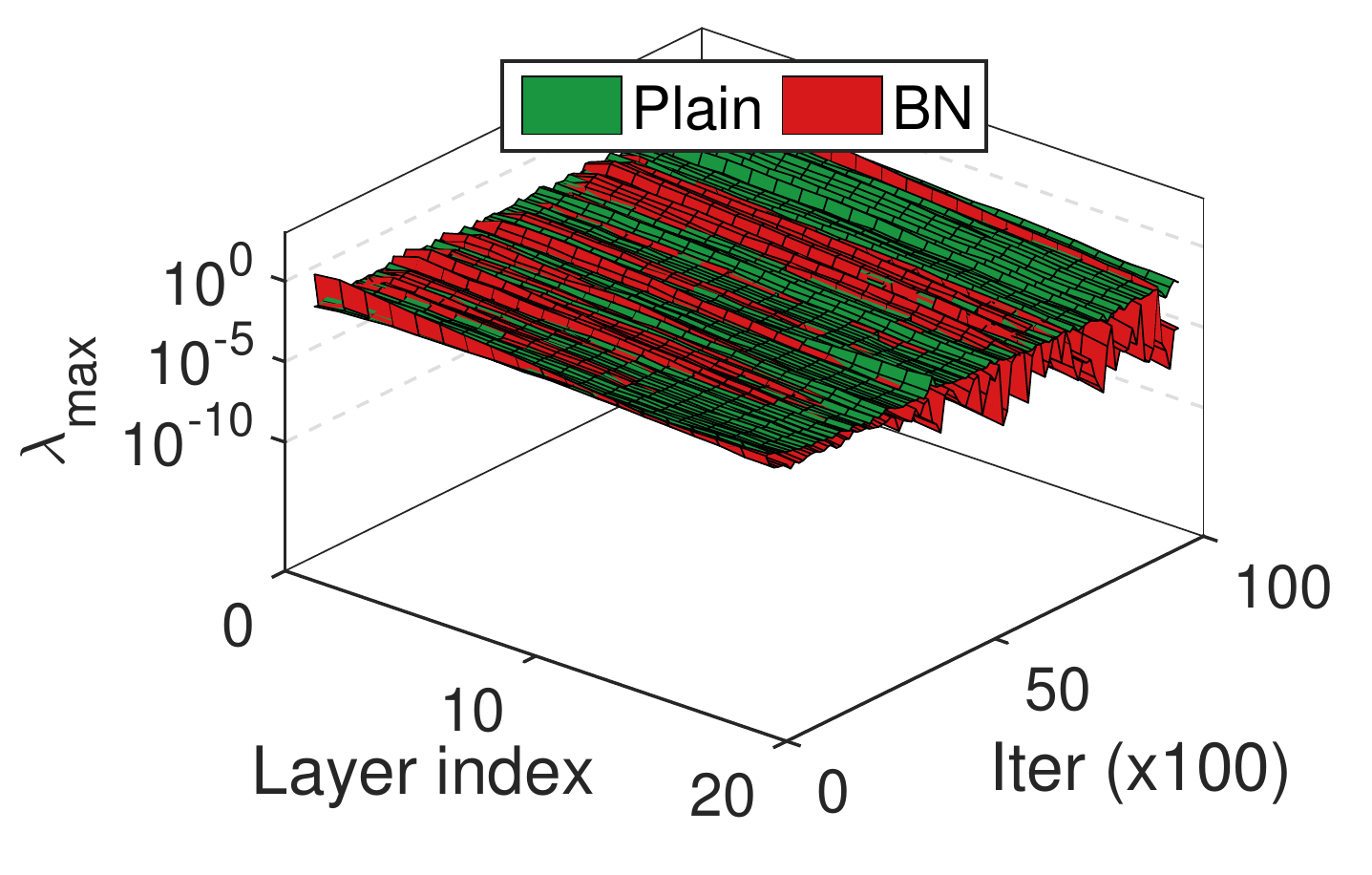}
		\end{minipage}
	}
	\subfigure[Training loss]{
		\begin{minipage}[c]{.32\linewidth}
			\centering
			\includegraphics[width=4.0cm]{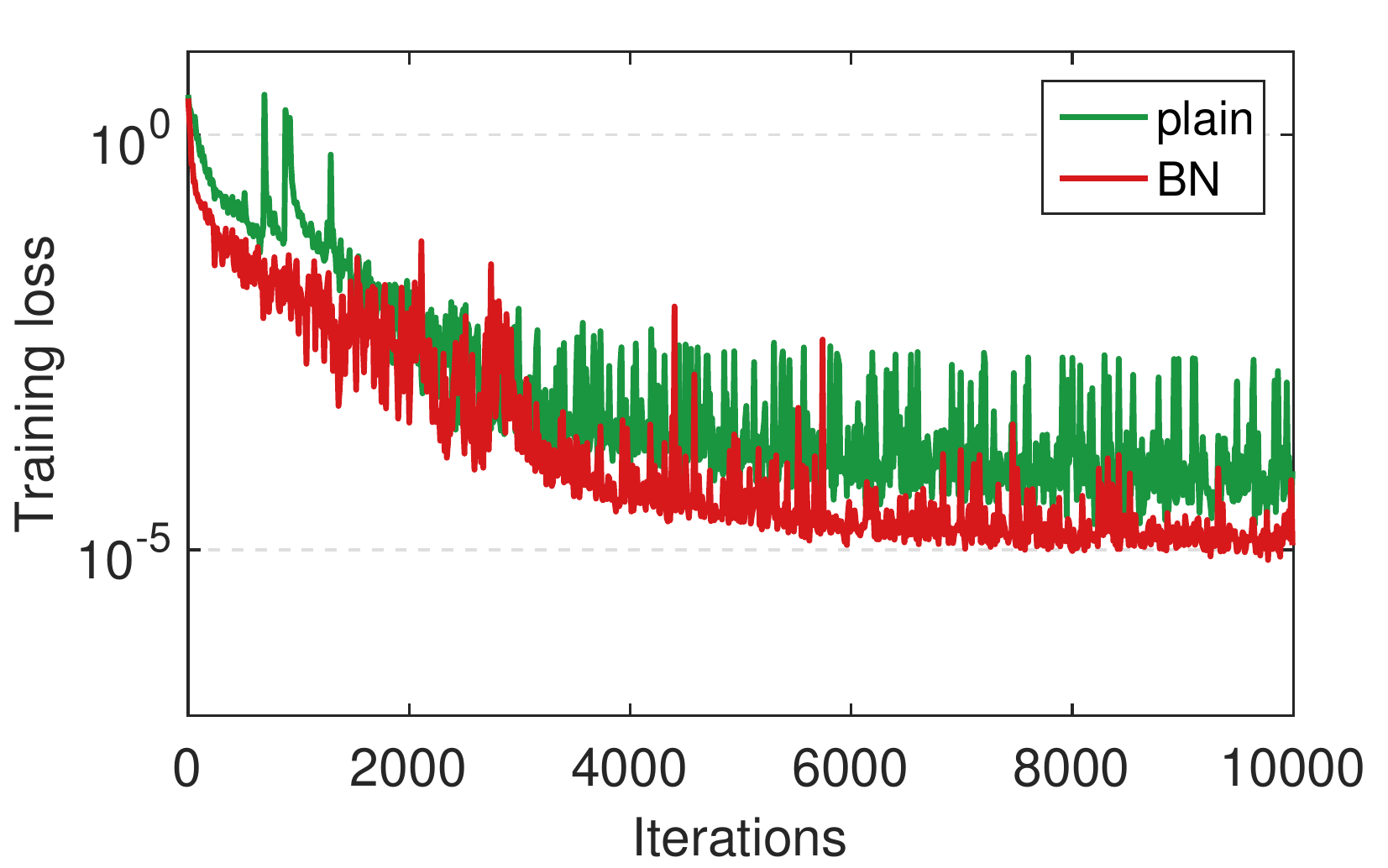}
		\end{minipage}
	}	
	\vspace{-0.15in}
	\caption{Analysis of the magnitude of the layer input (indicated by $\lambda_{max}(\Sigma_x)$) and layer output-gradient (indicated by $\lambda_{max}(\Sigma_{\nabla \mathbf{h}})$ ). We use the SGD with a batch size of 1024 to train the 20-layer MLP for classification. The results of (a)(b)(c) are under random initialization \cite{1998_NN_Yann}, while (d)(e)(f)  use He-initialization \cite{2015_ICCV_He}.  }
	\label{fig-sup:stablizeBN}
	\vspace{-0.16in}
\end{figure*}

\begin{figure*}[h]
	\centering
	\hspace{-0.2in}		\subfigure[$\kappa_{80\%}(\Sigma_x)$]{
		\begin{minipage}[c]{.22\linewidth}
			\centering
			\includegraphics[width=3.0cm]{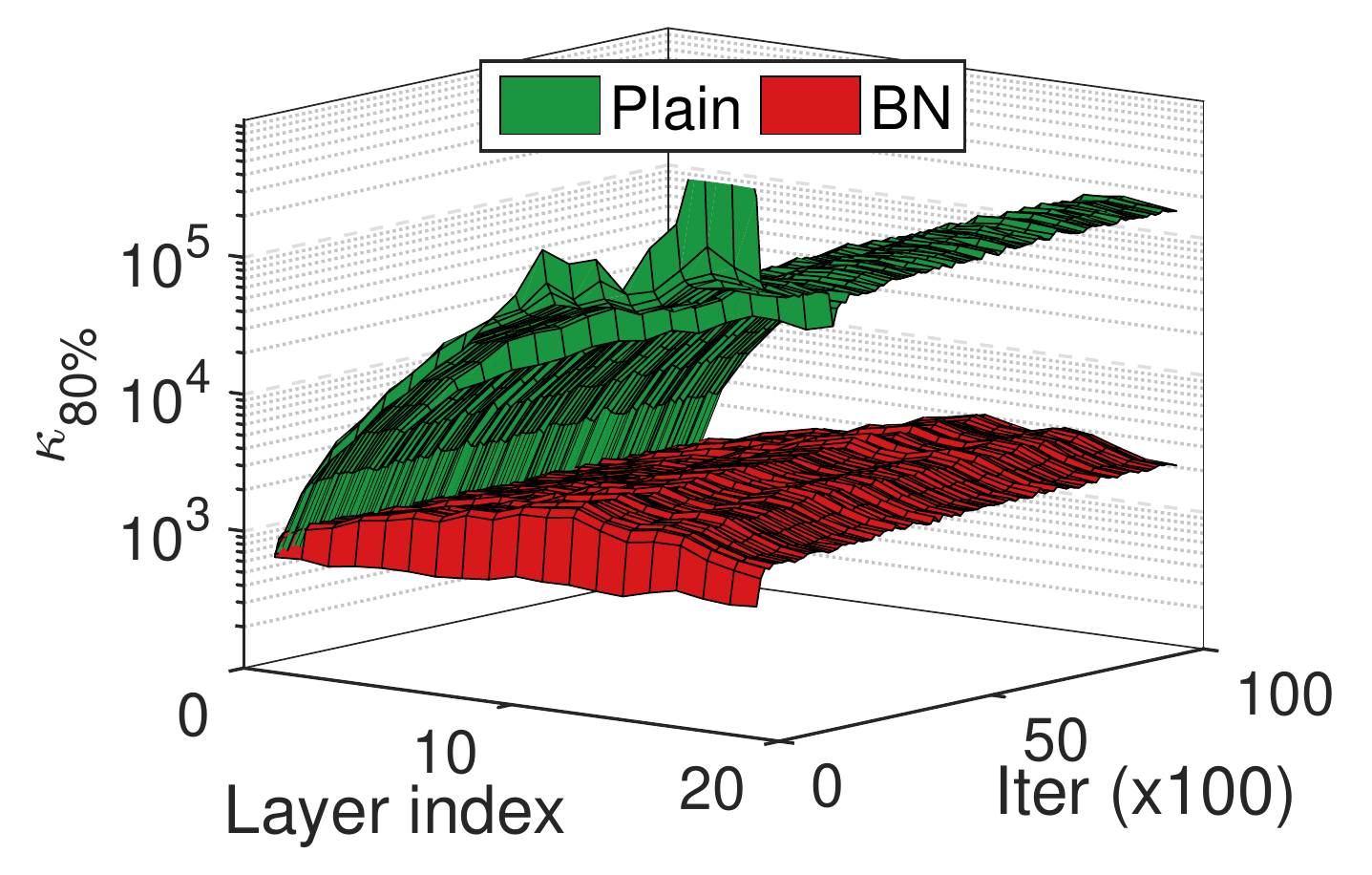}
		\end{minipage}
	}
	\hspace{0.03in}	\subfigure[$\kappa_{80\%}(\Sigma_{\nabla \mathbf{h}})$]{
		\begin{minipage}[c]{.22\linewidth}
			\centering
			\includegraphics[width=3.0cm]{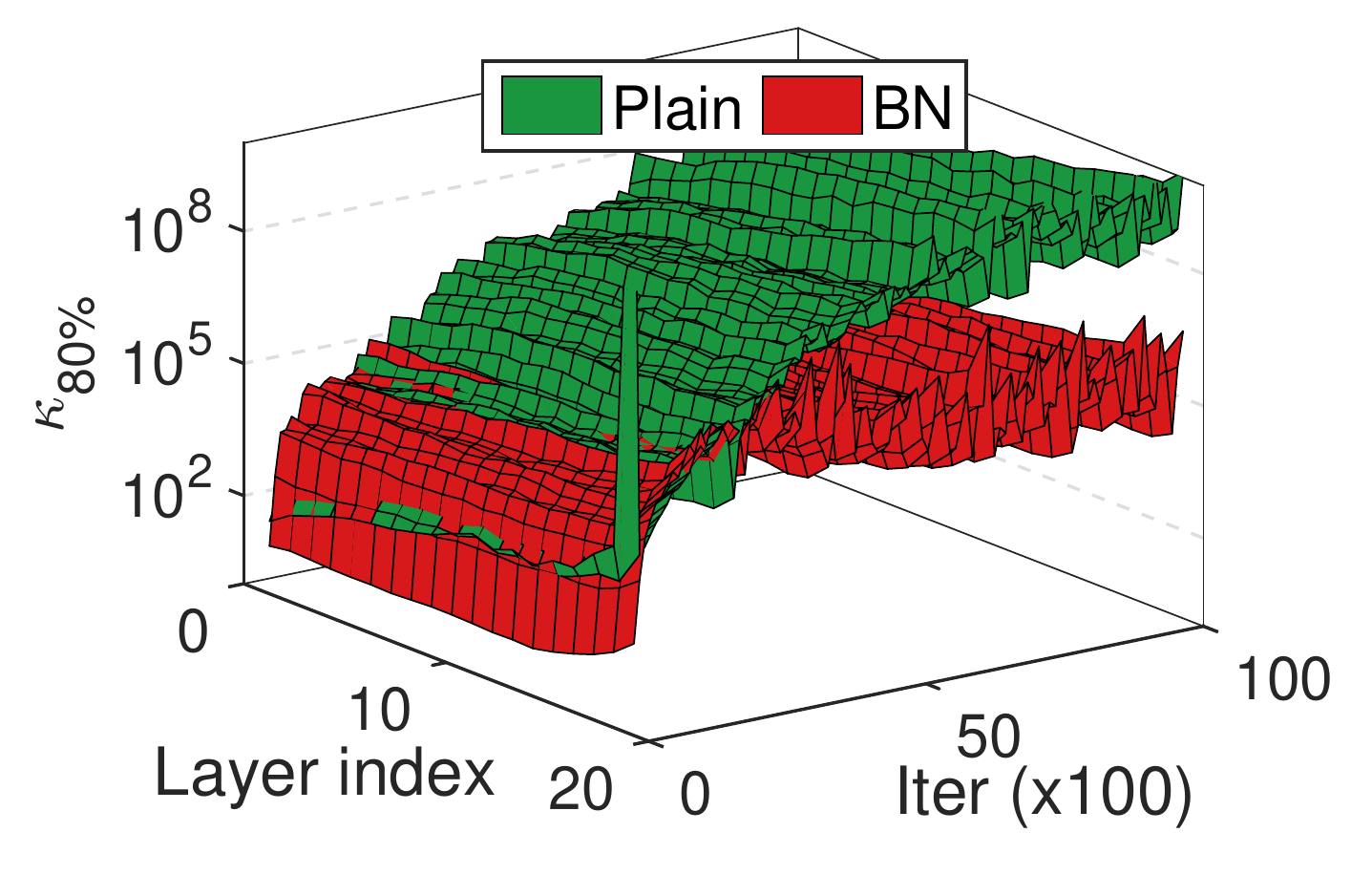}
		\end{minipage}
	}
	\hspace{0.03in}	\subfigure[$\kappa_{100\%}(\Sigma_x)$]{
		\begin{minipage}[c]{.22\linewidth}
			\centering
			\includegraphics[width=3.0cm]{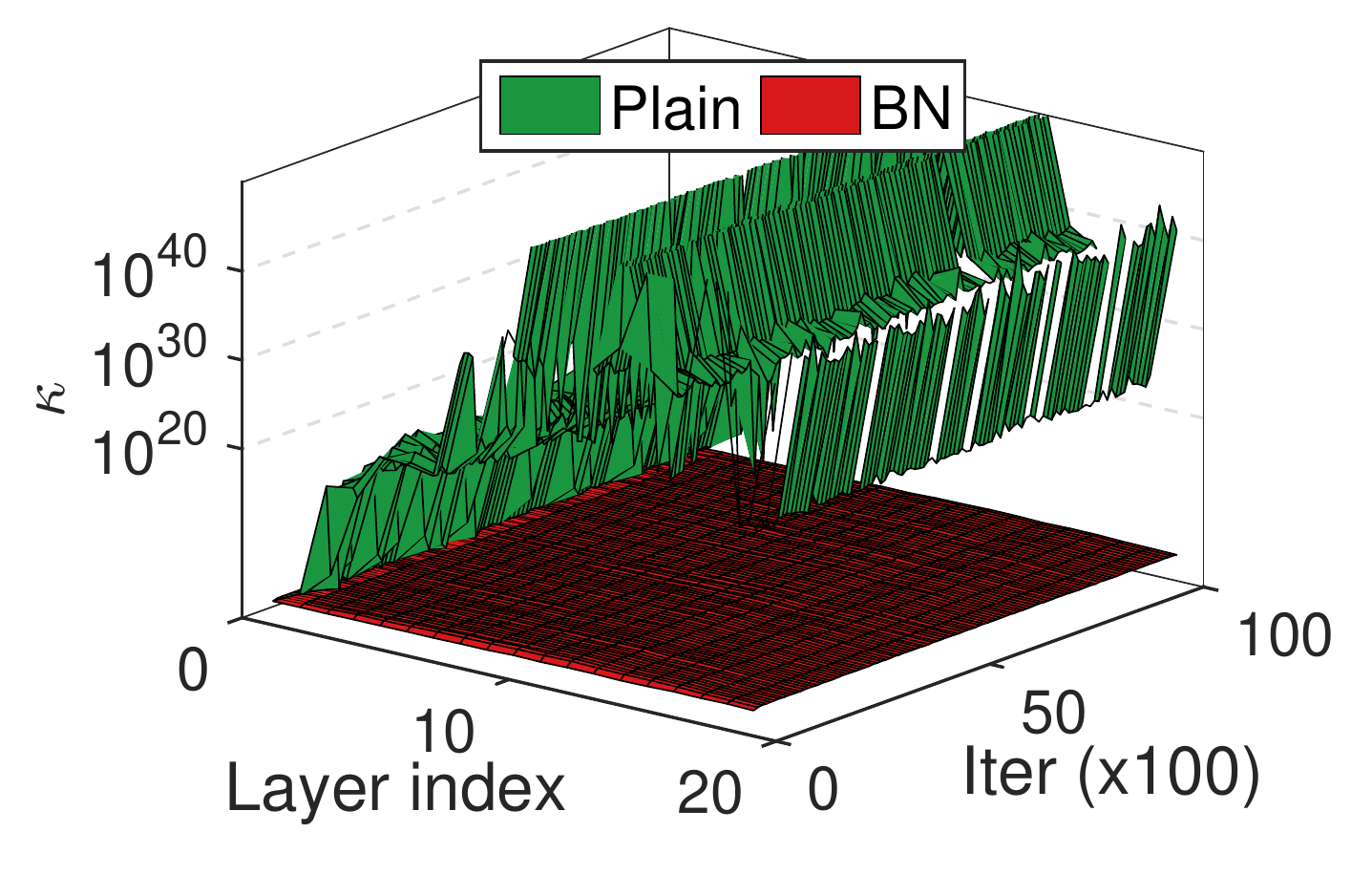}
		\end{minipage}
	}
	\hspace{0.03in}	\subfigure[$\kappa_{100\%}(\Sigma_{\nabla \mathbf{h}})$]{
		\begin{minipage}[c]{.22\linewidth}
			\centering
			\includegraphics[width=3.0cm]{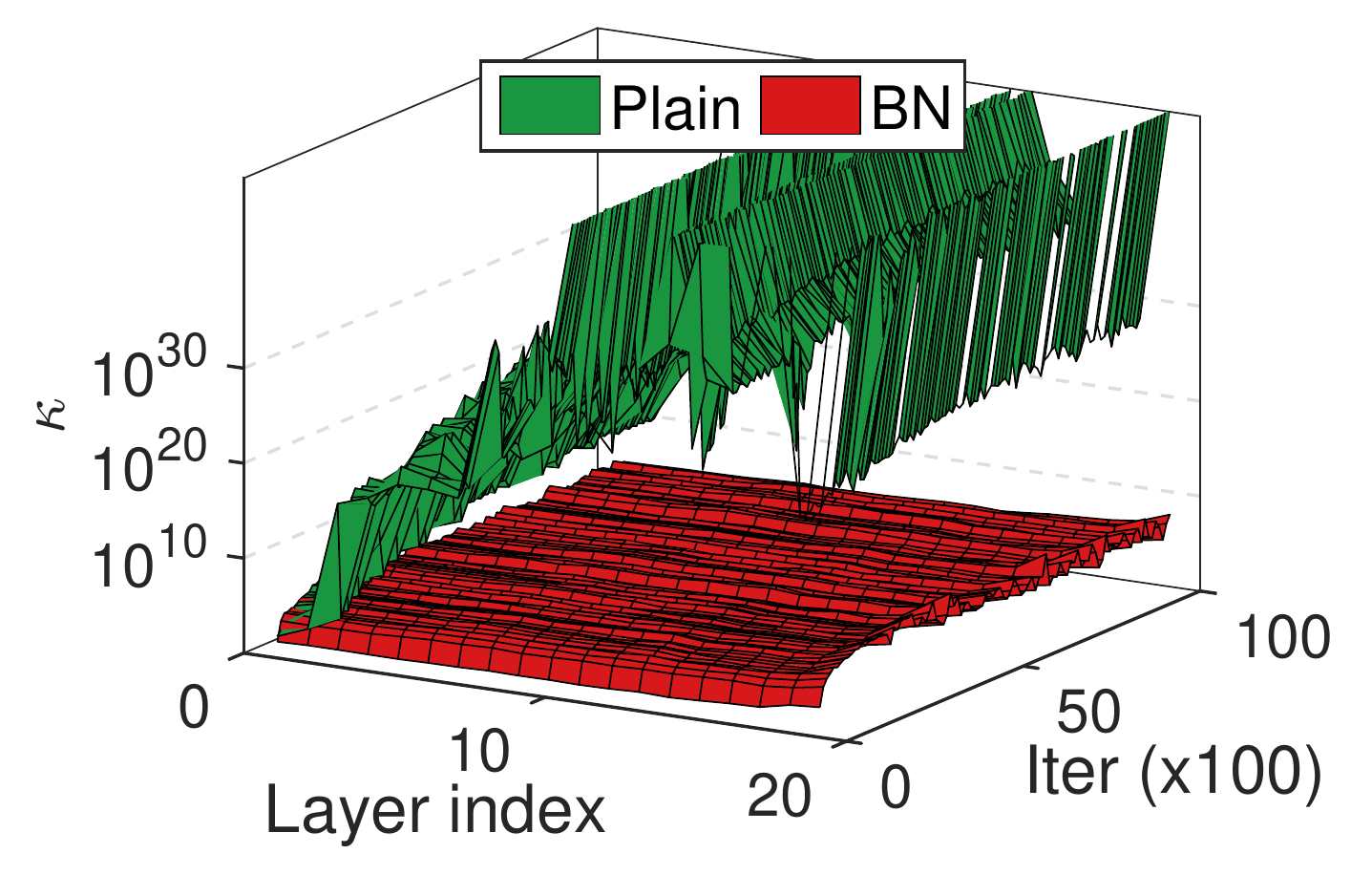}
		\end{minipage}
	}
	\vspace{-0.14in}
	\caption{Analysis on the condition number of the layer input (indicated by $\kappa_{p}(\Sigma_x)$) and layer output-gradient (indicated by $\kappa_{p}(\Sigma_{\nabla \mathbf{h}})$ ). The experimental setups are the same as in Figure \ref{fig-sup:stablizeBN}.}
	\label{fig-sup:conditionNumber}
	\vspace{-0.22in}
\end{figure*}

	\vspace{-0.1in}
\section{More Experiments in Exploring Batch Normalized Networks}
\label{Sec-sup-ExpBN}
In this section, we provide more experimental results relating to the exploration of batch normalization (BN) \cite{2015_ICML_Ioffe} by layer-wise conditioning analysis, which is discussed in Section~\ref{Sec:BN} of the paper. We include  experiments that train neural networks with Stochastic Gradient Descent (SGD),  experiments relating to weight domination and experiments relating to dying/full neurons.
	\vspace{-0.1in}
\subsection{Experiments with SGD}
	\vspace{-0.1in}
\label{Sec-sup-ExpBNSGD}
Here, we perform  experiments  on the Multiple Layer Perceptron (MLP) for MNIST classification and Convolutional Neural Networks (CNNs) for CIFAR-10  and \revise{ImageNet~\cite{2009_ImageNet}} classification.

\vspace{-0.1in}
\subsubsection{MLP for MNIST Classification}
	\vspace{-0.1in}
Here, we use the same experimental setup as the experiments described in Section~\ref{Sec:BN_stablization} of the  paper for MNIST classification, except that we use SGD with a batch size of 1024. The results are shown in Figures \ref{fig-sup:stablizeBN} and \ref{fig-sup:conditionNumber}. We obtain similar results as those obtained using full gradient descent, as described in Section~\ref{Sec:BN} of the  paper.

\begin{figure*}[]
	\centering
	\hspace{-0.2in}		\subfigure[ $\lambda_{max}(\Sigma_x)$]{
		\begin{minipage}[c]{.32\linewidth}
			\centering
			\includegraphics[width=4.0cm]{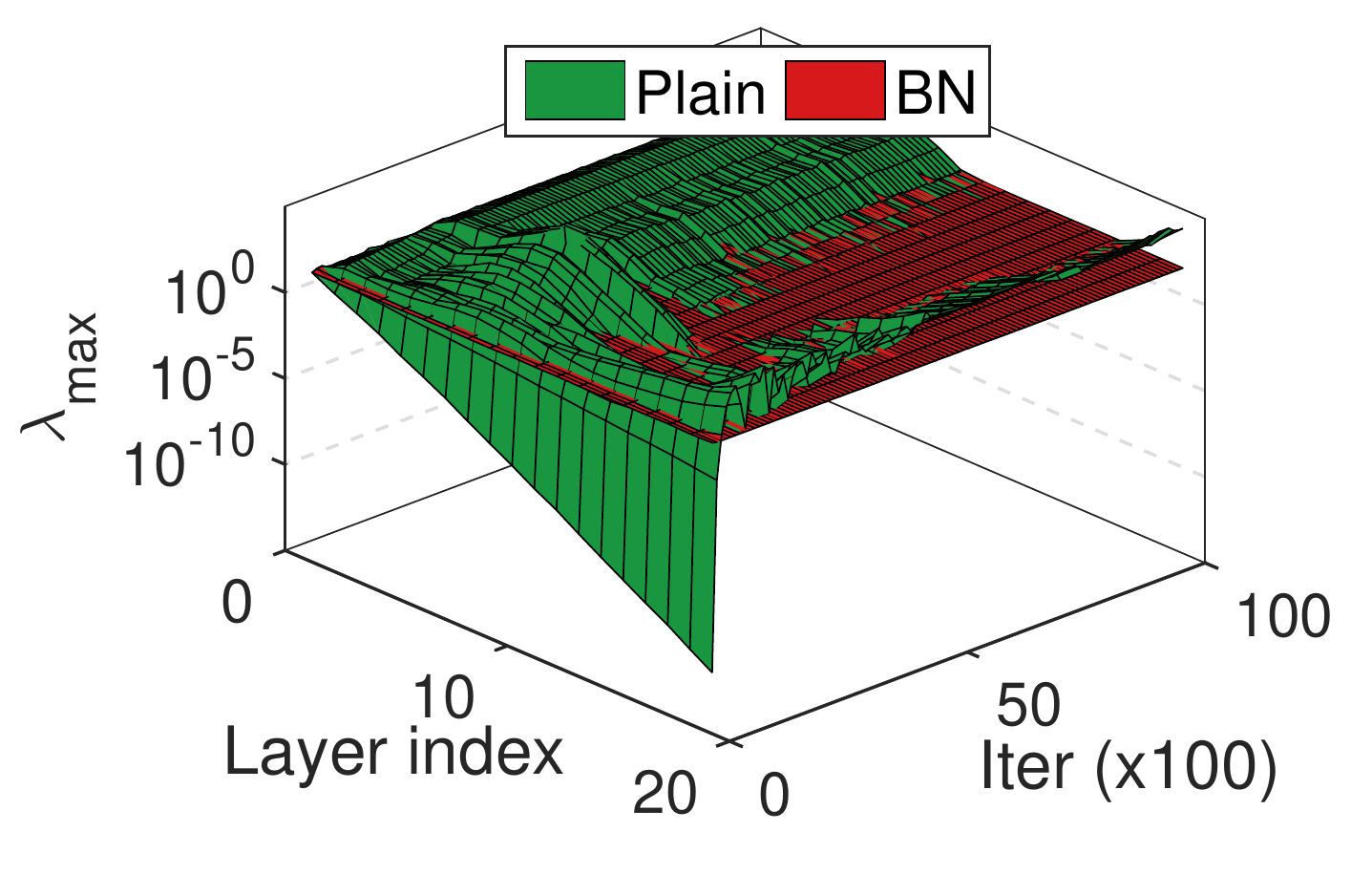}
		\end{minipage}
	}
	\subfigure[$\lambda_{max}(\Sigma_{\nabla \mathbf{h}})$]{
		\begin{minipage}[c]{.32\linewidth}
			\centering
			\includegraphics[width=4.0cm]{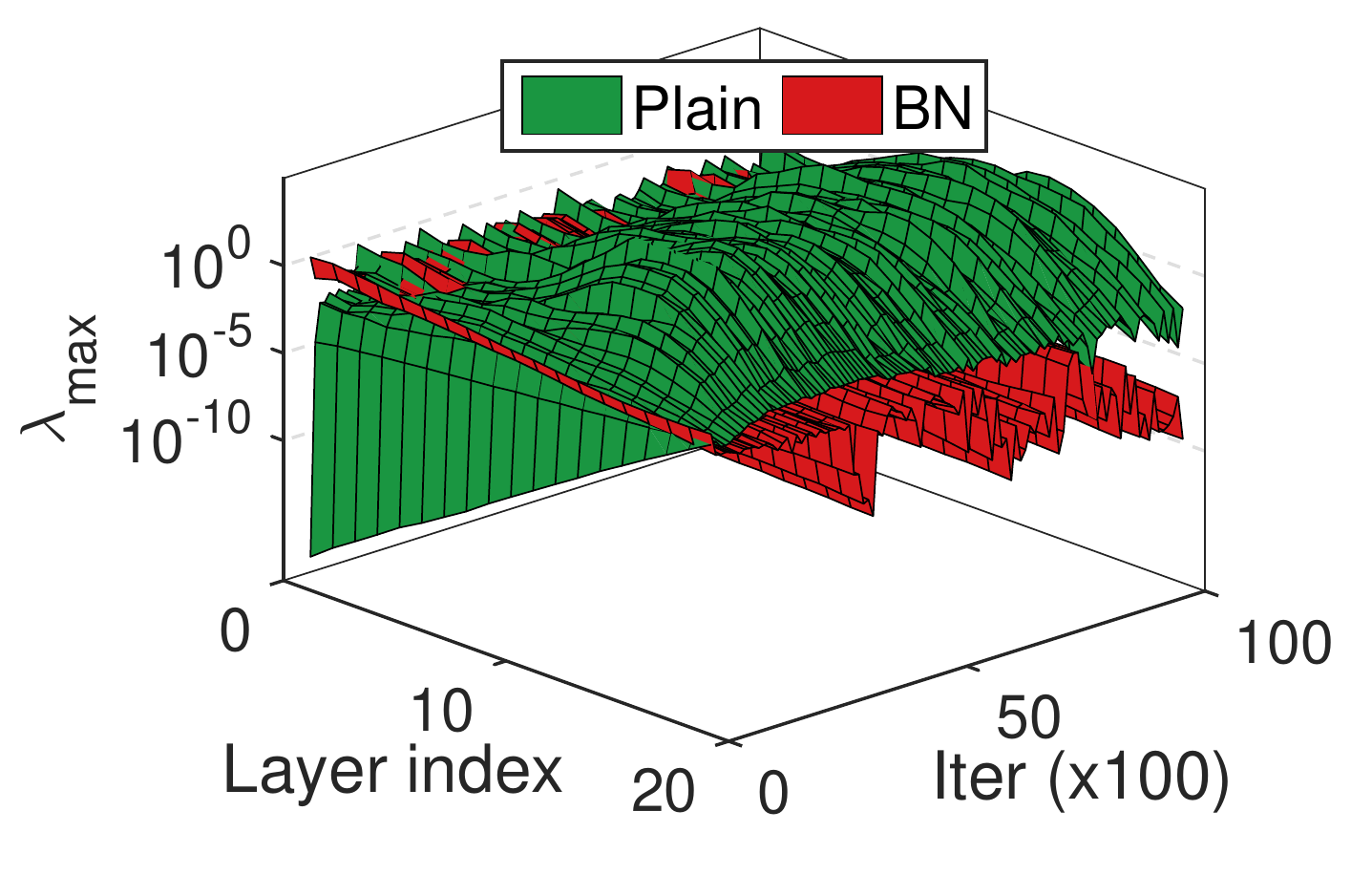}
		\end{minipage}
	}
	\subfigure[Training loss]{
		\begin{minipage}[c]{.32\linewidth}
			\centering
			\includegraphics[width=4.0cm]{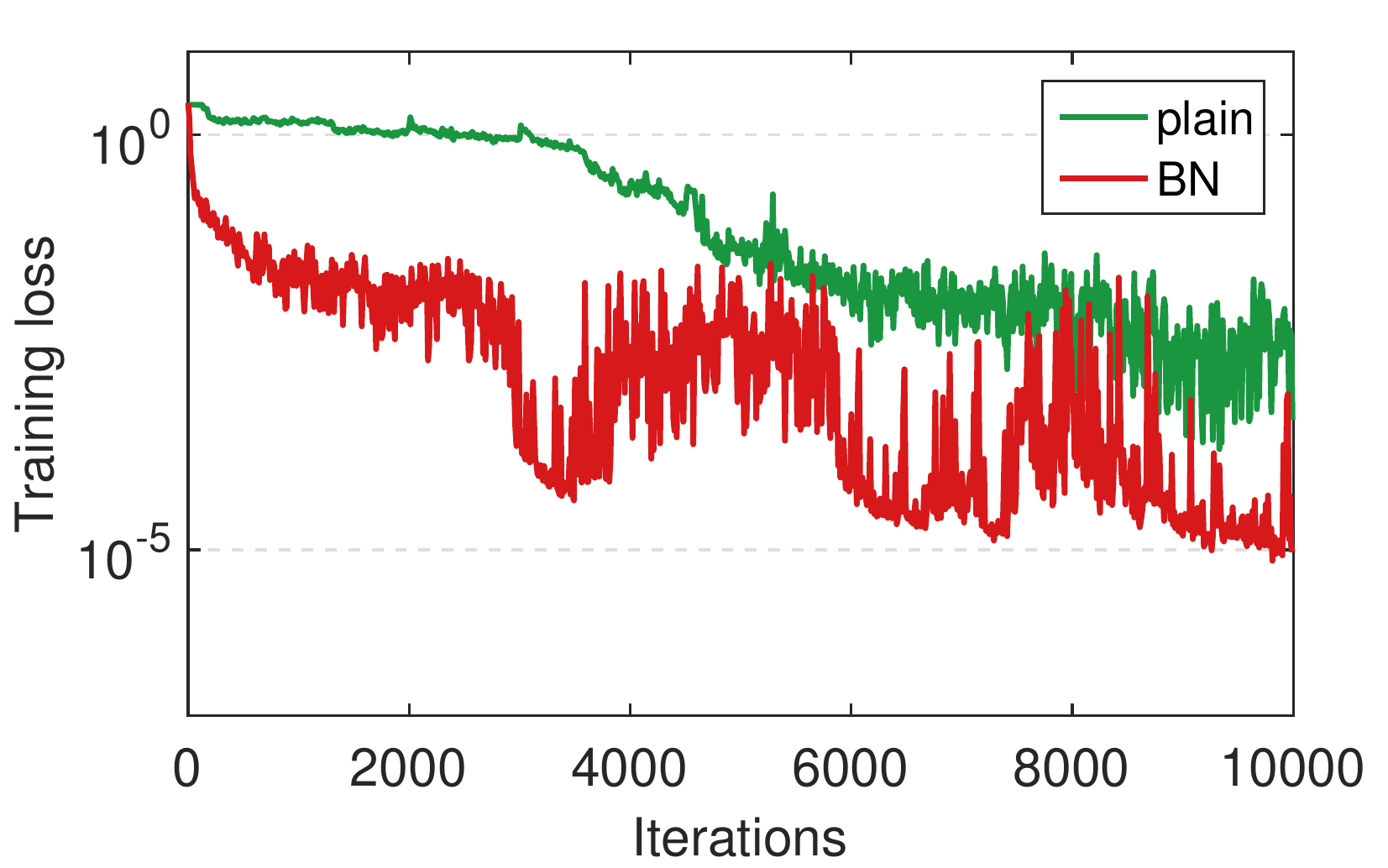}
		\end{minipage}
	}
	\\	
	\hspace{-0.2in}		\subfigure[ $\kappa_{80\%}(\Sigma_x)$]{
		\begin{minipage}[c]{.46\linewidth}
			\centering
			\includegraphics[width=5.0cm]{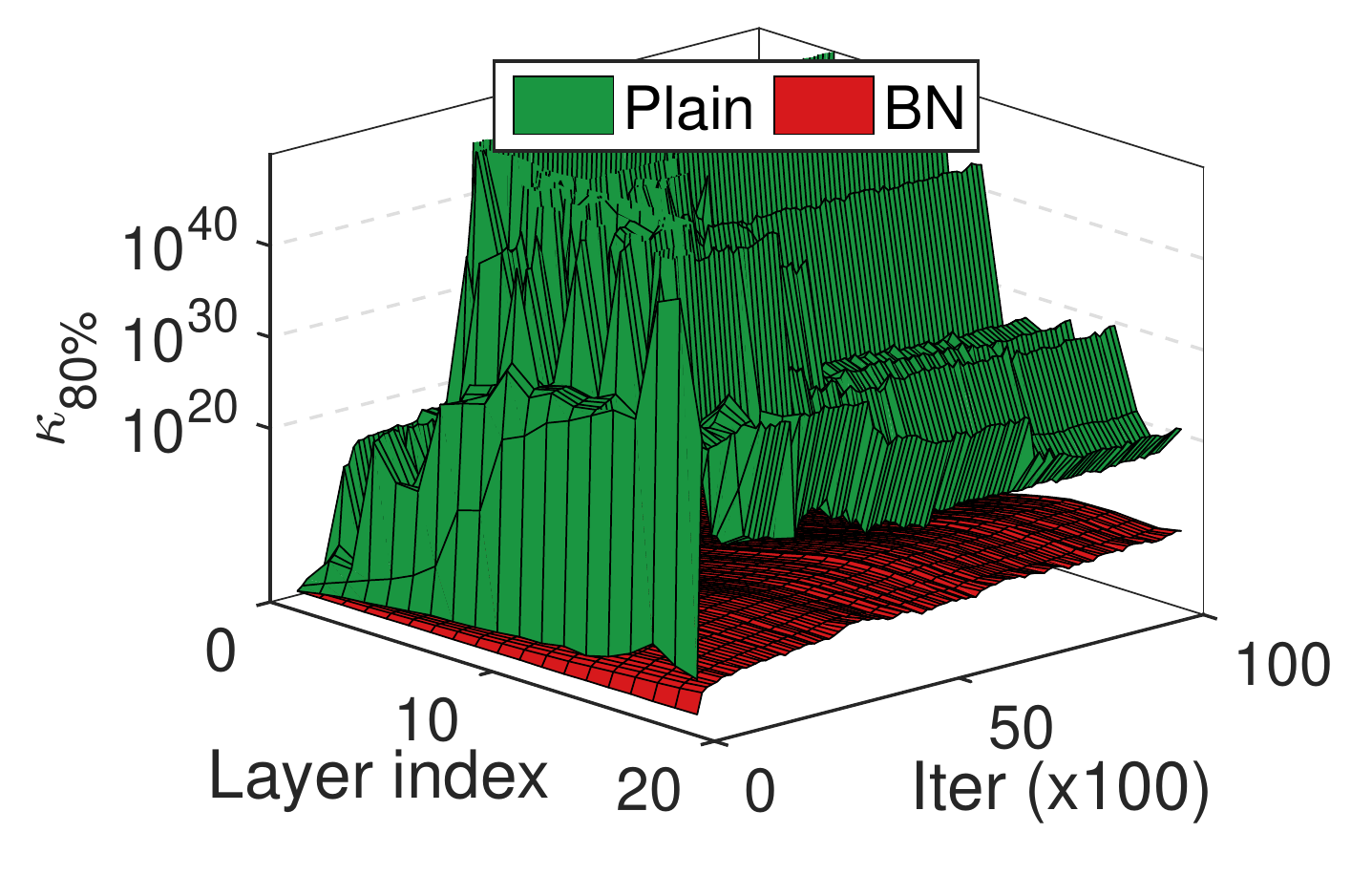}
		\end{minipage}
	}
	\subfigure[$\kappa_{80\%}(\Sigma_{\nabla \mathbf{h}})$]{
		\begin{minipage}[c]{.46\linewidth}
			\centering
			\includegraphics[width=5.0cm]{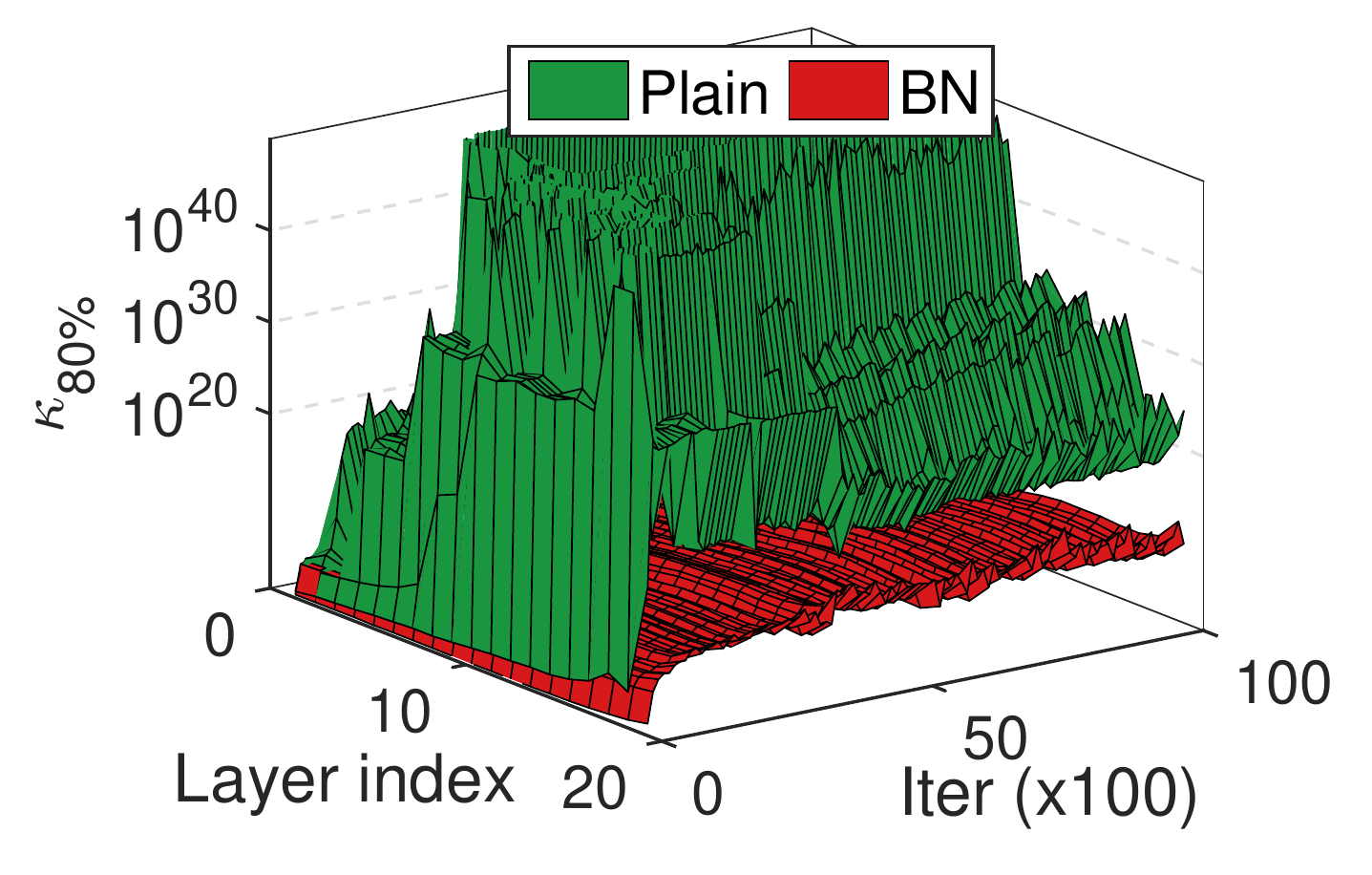}
		\end{minipage}
	}	
	\vspace{-0.15in}
	\caption{Layer-wise conditioning analysis results with Adam optimizer~\cite{2014_CoRR_Kingma}.  We use a batch size of 1024 to train the 20-layer MLP for classification. Figures (a) and (b) show the magnitude of the layer input  and layer output-gradient, respectively. Figure (c) shows the training loss with respect to the epochs. Figures (d) and (e) show the condition number of the layer input  and layer output-gradient, respectively.  }
	\label{fig:BNAdam}
	\vspace{-0.16in}
\end{figure*}

We also conduct experiments using the Adam optimizer~\cite{2014_CoRR_Kingma}.  We again use a batch size of 1024 to train the 20-layer MLP for classification.  We report the best training loss among the learning rates in $\{0.001, 0.0005, 0.0001 \}$. The results under random initialization \cite{1998_NN_Yann} are shown in Figure~\ref{fig:BNAdam}. We observe that: 1) `Plain' with the Adam optimizer can well adjust the magnitude of the layer input (Figure~\ref{fig:BNAdam} (a)) and layer output-gradient (Figure~\ref{fig:BNAdam} (b)) during training, compared to `Plain' with the naive SGD
optimizer (Figure ~\ref{fig-sup:stablizeBN} (a) and (b)); 2) `BN' can better stabilize the training; 3) `BN' has better conditioning than `Plain' during training. 

\begin{figure*}[]
	\centering
	\hspace{-0.15in}	\subfigure[$\kappa_{100\%}(\Sigma_x)$]{
		\begin{minipage}[c]{.3\linewidth}
			\centering
			\includegraphics[width=4.0cm]{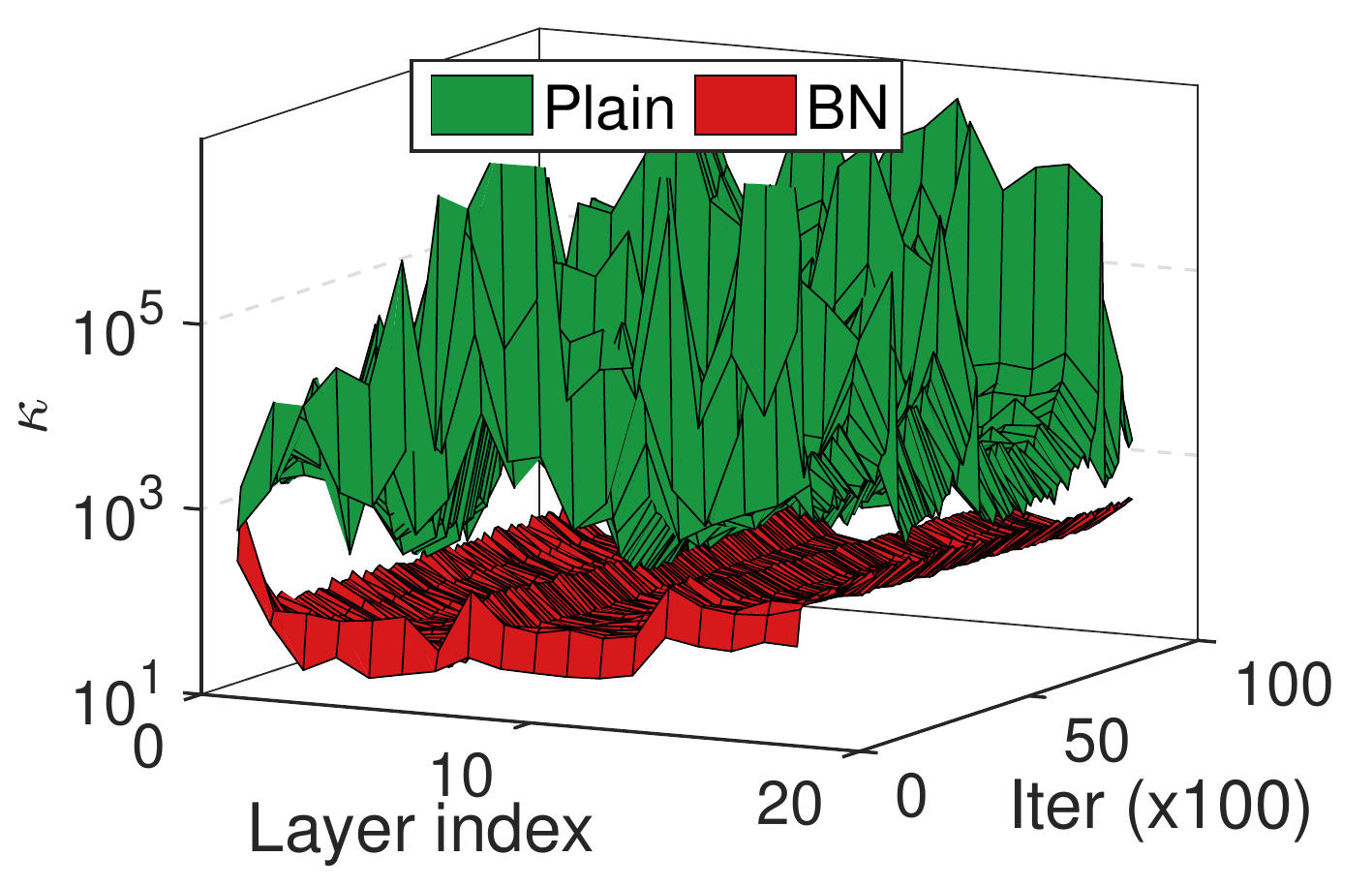}
		\end{minipage}
	}
	\hspace{0.05in}	\subfigure[$\kappa_{100\%}(\Sigma_{\nabla \mathbf{h}})$]{
		\begin{minipage}[c]{.3\linewidth}
			\centering
			\includegraphics[width=4.0cm]{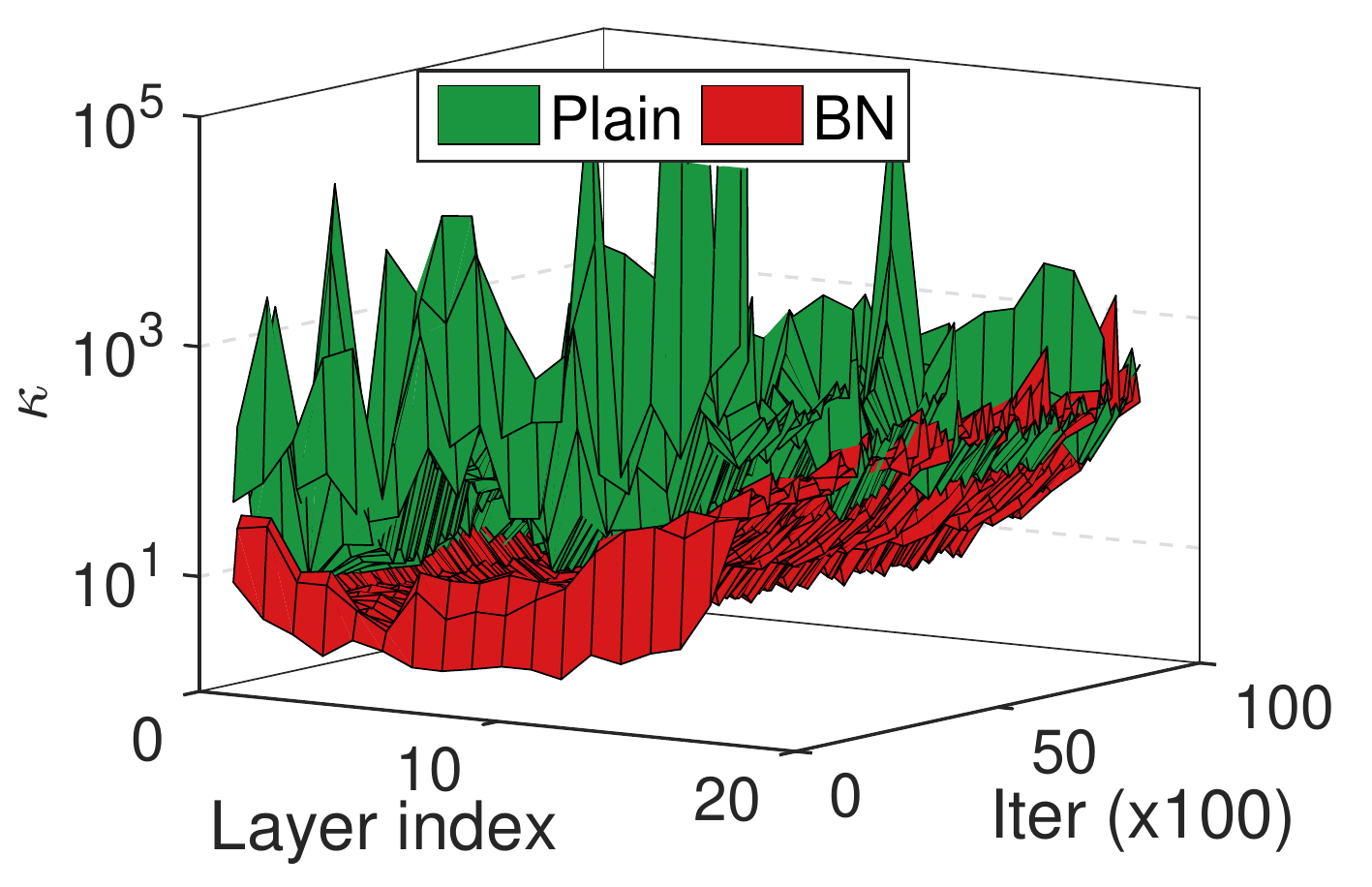}
		\end{minipage}
	}
	\hspace{0.05in}	\subfigure[Training loss]{
		\begin{minipage}[c]{.3\linewidth}
			\centering
			\includegraphics[width=4.0cm]{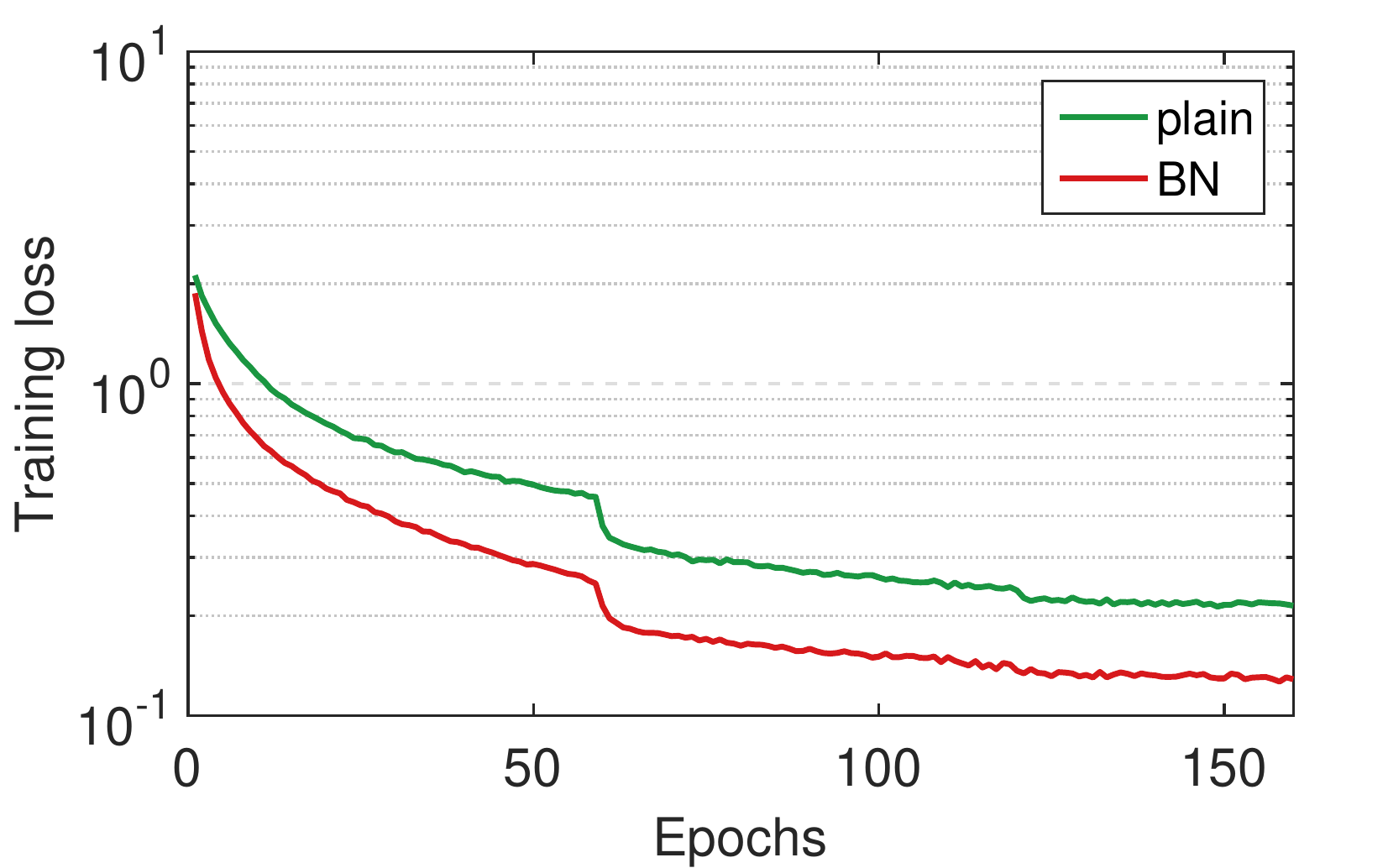}
		\end{minipage}
	}
	\vspace{-0.15in}
	\caption{Layer-wise conditioning analysis on the VGG-style network for CIFAR-10 classification. 
		Figures (a) and (b) show the condition number of the layer input (indicated by $\kappa_{p}(\Sigma_x)$) and layer output-gradient (indicated by $\kappa_{p}(\Sigma_{\nabla \mathbf{h}})$ ), respectively. Figure (c) shows the training loss with respect to the epochs. }
	\label{fig:P0}
	\vspace{-0.2in}
\end{figure*}

\begin{figure*}[]
	\centering
	\hspace{-0.15in}	\subfigure[$\kappa_{100\%}(\Sigma_x)$]{
		\begin{minipage}[c]{.3\linewidth}
			\centering
			\includegraphics[width=4.0cm]{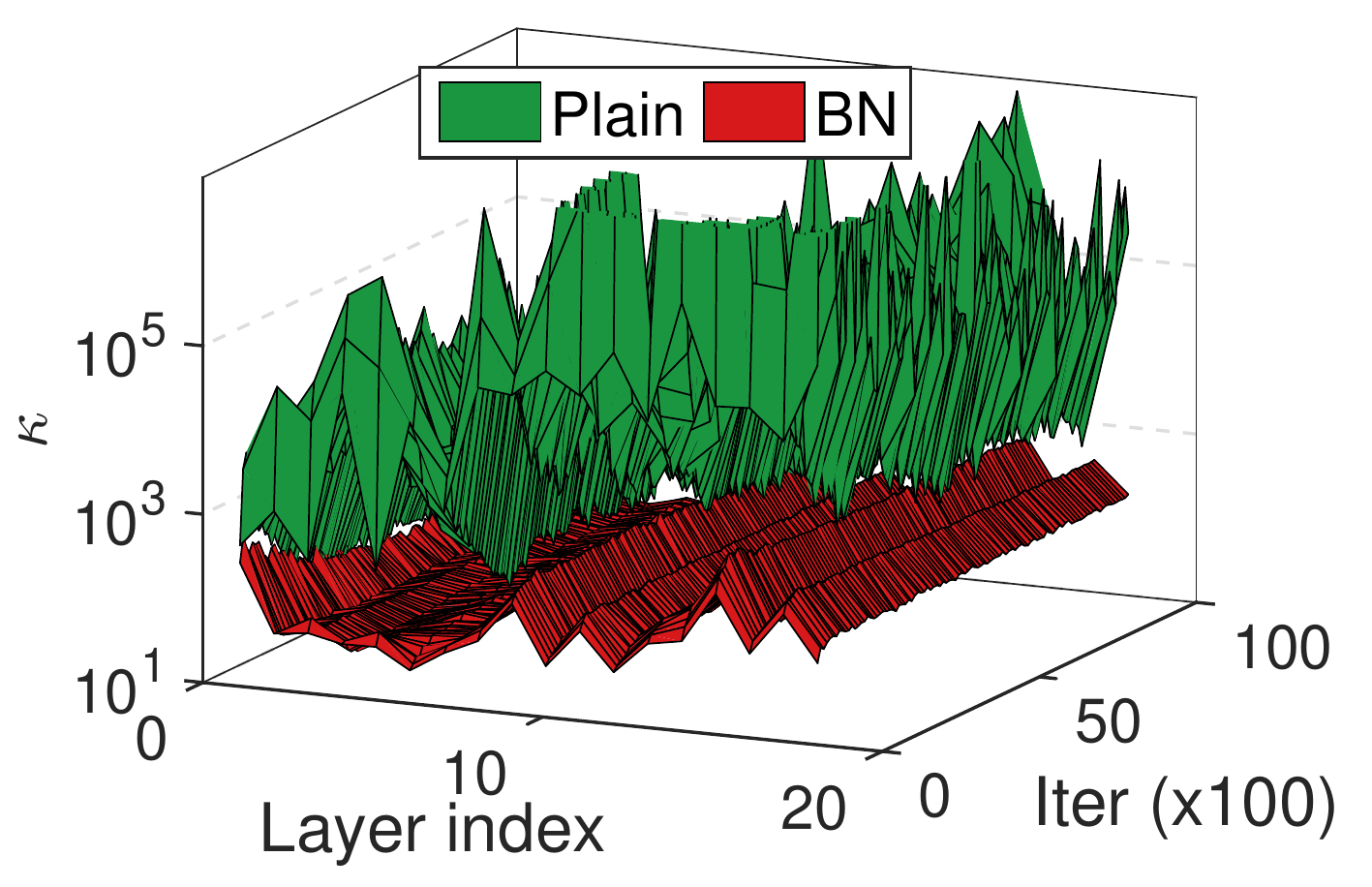}
		\end{minipage}
	}
	\hspace{0.05in}	\subfigure[$\kappa_{100\%}(\Sigma_{\nabla \mathbf{h}})$]{
		\begin{minipage}[c]{.3\linewidth}
			\centering
			\includegraphics[width=4.0cm]{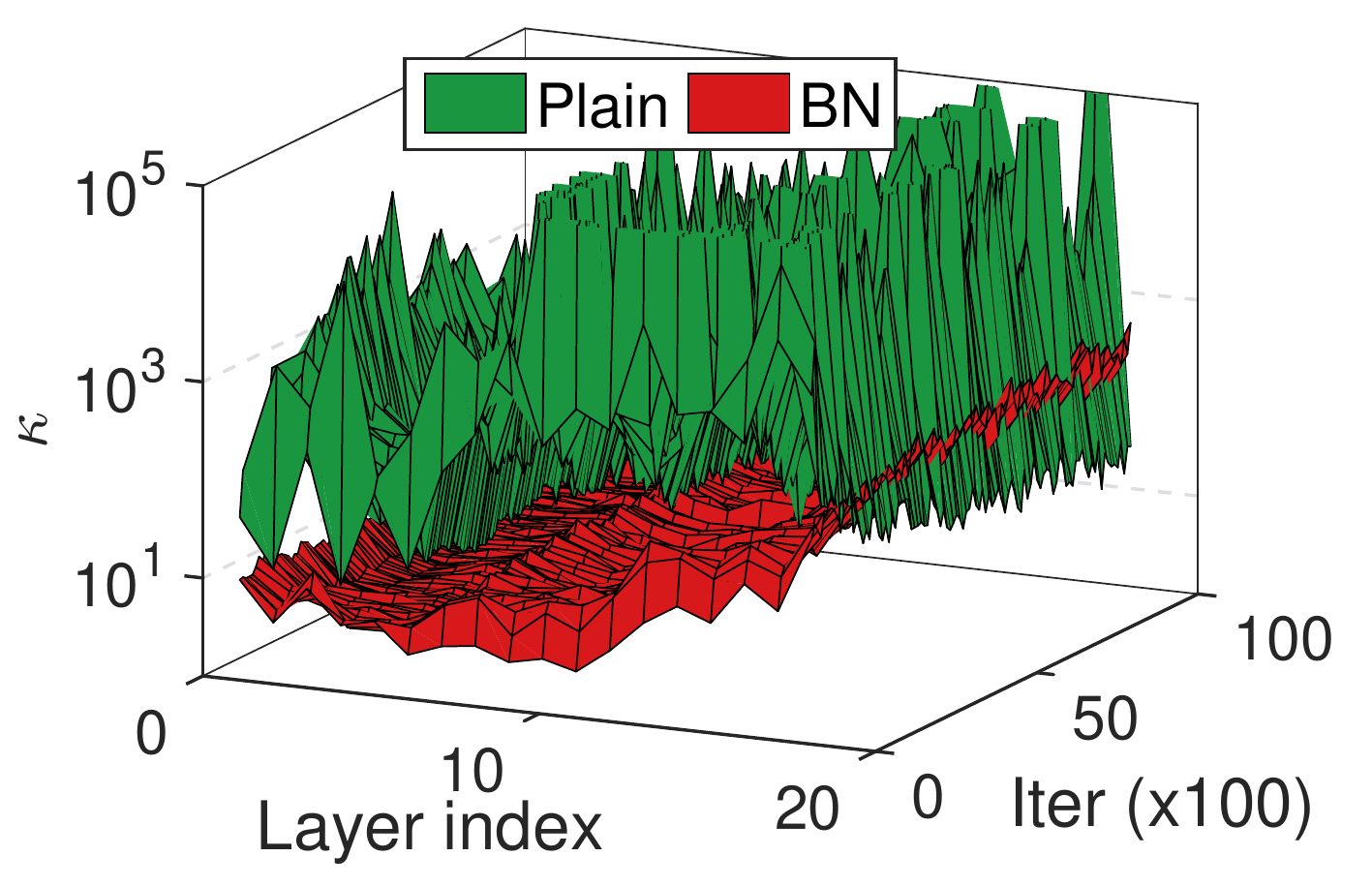}
		\end{minipage}
	}
	\hspace{0.05in}	\subfigure[Training loss]{
		\begin{minipage}[c]{.3\linewidth}
			\centering
			\includegraphics[width=4.0cm]{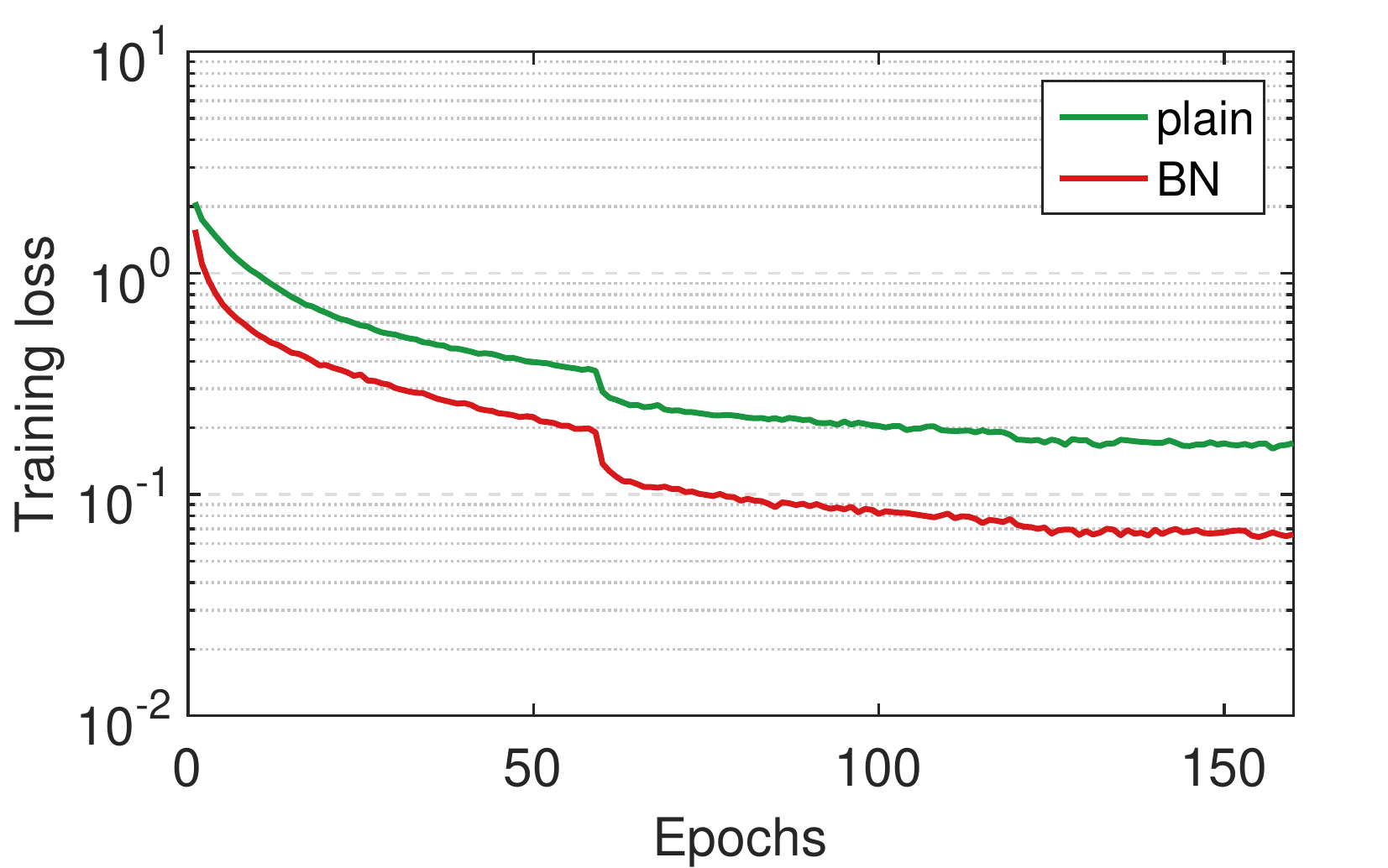}
		\end{minipage}
	}
	\vspace{-0.15in}
	\caption{Layer-wise conditioning analysis on the residual network \cite{2015_CVPR_He} for CIFAR-10 classification.  Figures	(a) and (b) show the condition number of the layer input (indicated by $\kappa_{p}(\Sigma_x)$) and layer output-gradient (indicated by $\kappa_{p}(\Sigma_{\nabla \mathbf{h}})$ ), respectively. Figure (c) shows the training loss with respect to the epochs. }
	\label{fig:Res}
	\vspace{-0.2in}
\end{figure*}

\vspace{-0.1in}
\subsubsection{CNN for CIFAR-10 Classification}
	\vspace{-0.1in}
We perform a layer-wise conditioning analysis on the VGG-style and residual network \cite{2015_CVPR_He} architectures.
Note that we view the activation in each spatial location of the feature map as an independent example,  when calculating the covariance matrix of the convolutional layer input and output-gradient. This process is similar to the process proposed in BN to normalize the convolutional layer \cite{2015_ICML_Ioffe}.

We use the 20-layer residual network  described in the paper \cite{2015_CVPR_He} for CIFAR-10 classification. The VGG-style network is constructed based on the 20-layer residual network, removing the residual connections.

We use the same setups as described in \cite{2015_CVPR_He}, except that we do not use  weight decay in order to simplify the analysis and run the experiments on one GPU.
Since the unnormalized networks (including the VGG-style and residual network)  do not converge with the large learning rate of 0.1, we  run additional experiments with a learning rate of 0.01, and report these results.

Figures \ref{fig:P0} and Figure \ref{fig:Res} show the results for the VGG-Style and residual network, respectively. We obtain the same observations as those made for the MLP for MNIST classification.

	\begin{figure*}[]
		\centering
		\hspace{-0.15in}	\subfigure[$\kappa_{100\%}(\Sigma_x)$]{
			\begin{minipage}[c]{.3\linewidth}
				\centering
				\includegraphics[width=4.0cm]{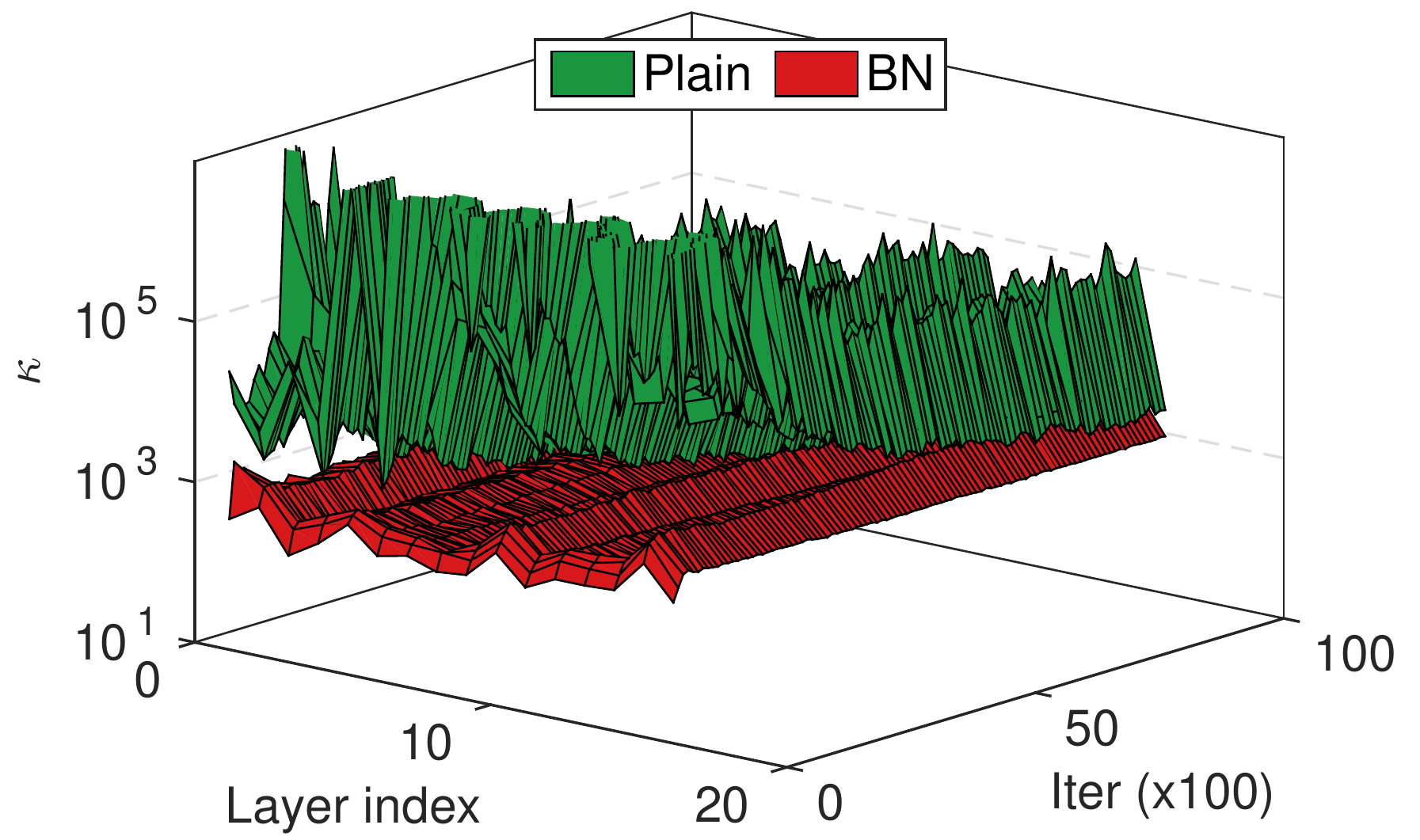}
			\end{minipage}
		}
		\hspace{0.05in}	\subfigure[$\kappa_{100\%}(\Sigma_{\nabla \mathbf{h}})$]{
			\begin{minipage}[c]{.3\linewidth}
				\centering
				\includegraphics[width=4.0cm]{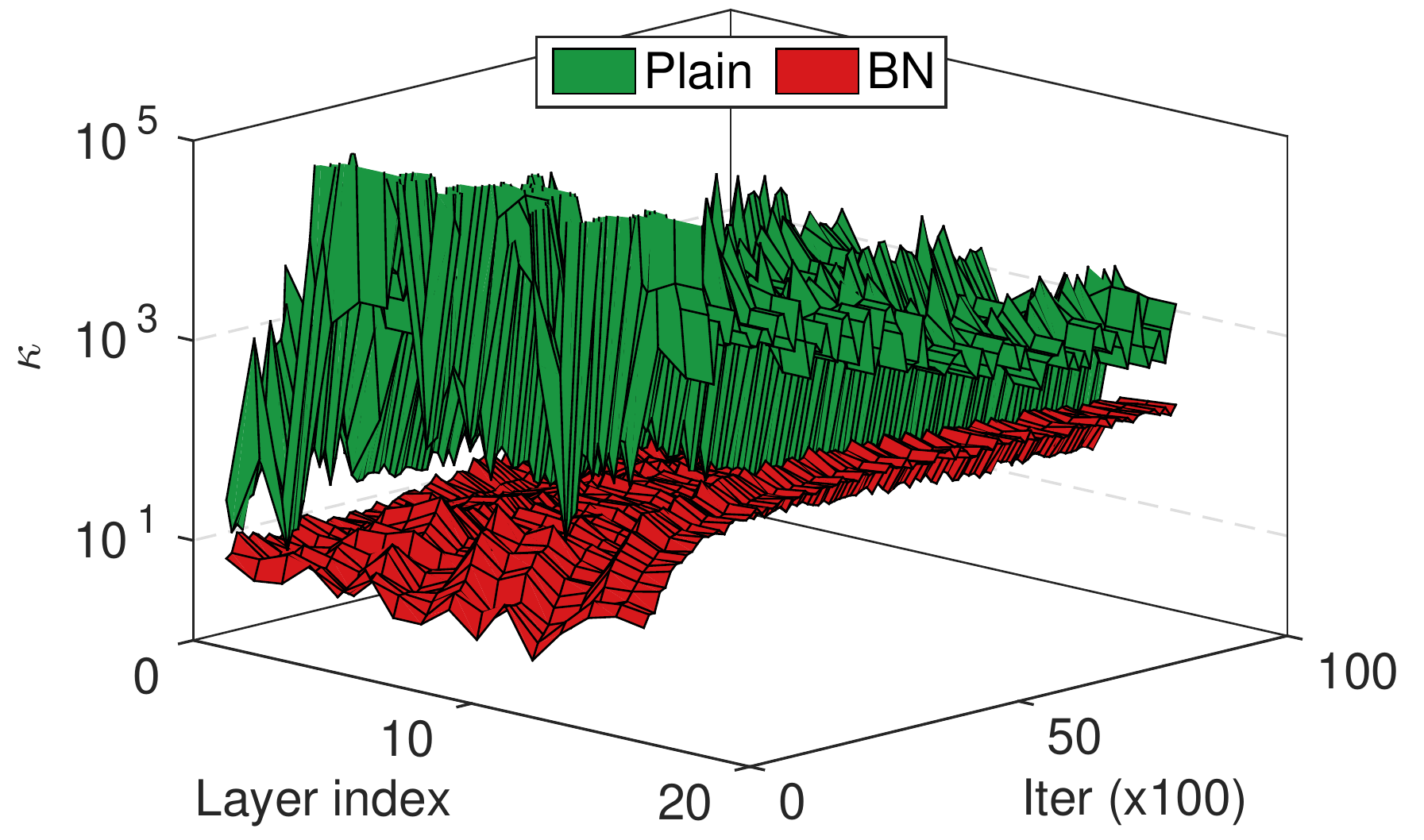}
			\end{minipage}
		}
		\hspace{0.05in}	\subfigure[Training loss]{
			\begin{minipage}[c]{.3\linewidth}
				\centering
				\includegraphics[width=4.0cm]{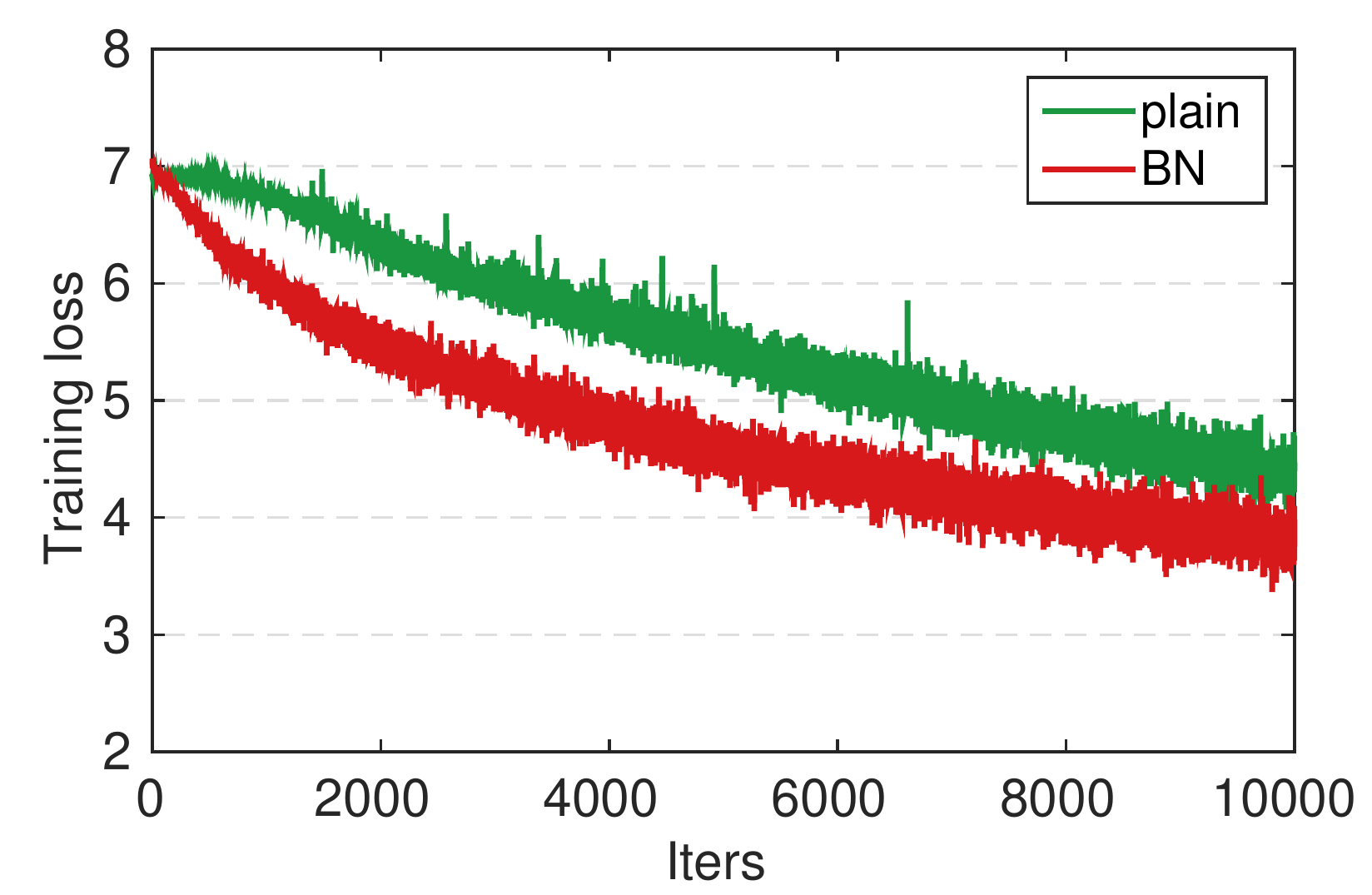}
			\end{minipage}
		}
		\vspace{-0.15in}
		\caption{Layer-wise conditioning analysis on the residual network \cite{2015_CVPR_He} for ImageNet classification.  Figures	(a) and (b) show the condition number of the layer input (indicated by $\kappa_{p}(\Sigma_x)$) and layer output-gradient (indicated by $\kappa_{p}(\Sigma_{\nabla \mathbf{h}})$ ), respectively. Figure (c) shows the training loss with respect to the training iterations. }
		\label{fig:Res-ImageNet-condition}
	\end{figure*}
	
	\revise{	
		\subsubsection{CNN for ImageNet Classification}
		We also perform a layer-wise conditioning analysis on ImageNet using a 18-layer residual network \cite{2015_CVPR_He}.
		We use the same setups as described in \cite{2015_CVPR_He}, except that we run the experiments on one GPU.
		Figure \ref{fig:Res-ImageNet-condition} show the results. We also obtain the same observations as those made for the MLP for MNIST classification.
	}

	\vspace{-0.1in}
\subsection{Experiments Relating to Weight Domination}
	\vspace{-0.1in}
\label{Sec-sup-ExpBNWD}
\begin{figure}[]
	\centering
	\hspace{-0.3in}\subfigure[100-layer MLP]{
		\begin{minipage}[c]{.4\linewidth}
			\centering
			\includegraphics[width=5.0cm]{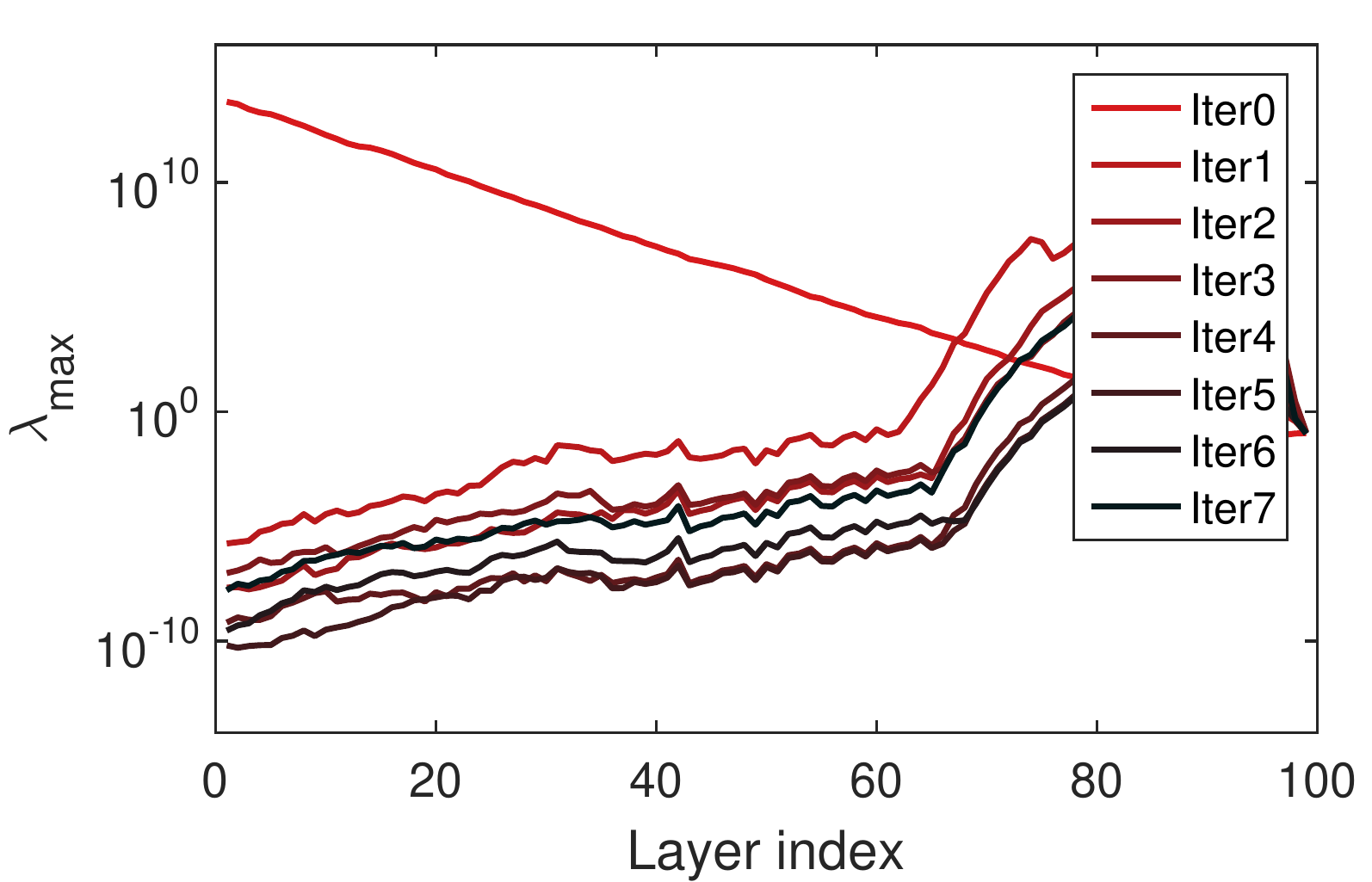}
		\end{minipage}
	}
	\hspace{0.2in}	\subfigure[110-layer VGG-style CNN]{
		\begin{minipage}[c]{.4\linewidth}
			\centering
			\includegraphics[width=5.0cm]{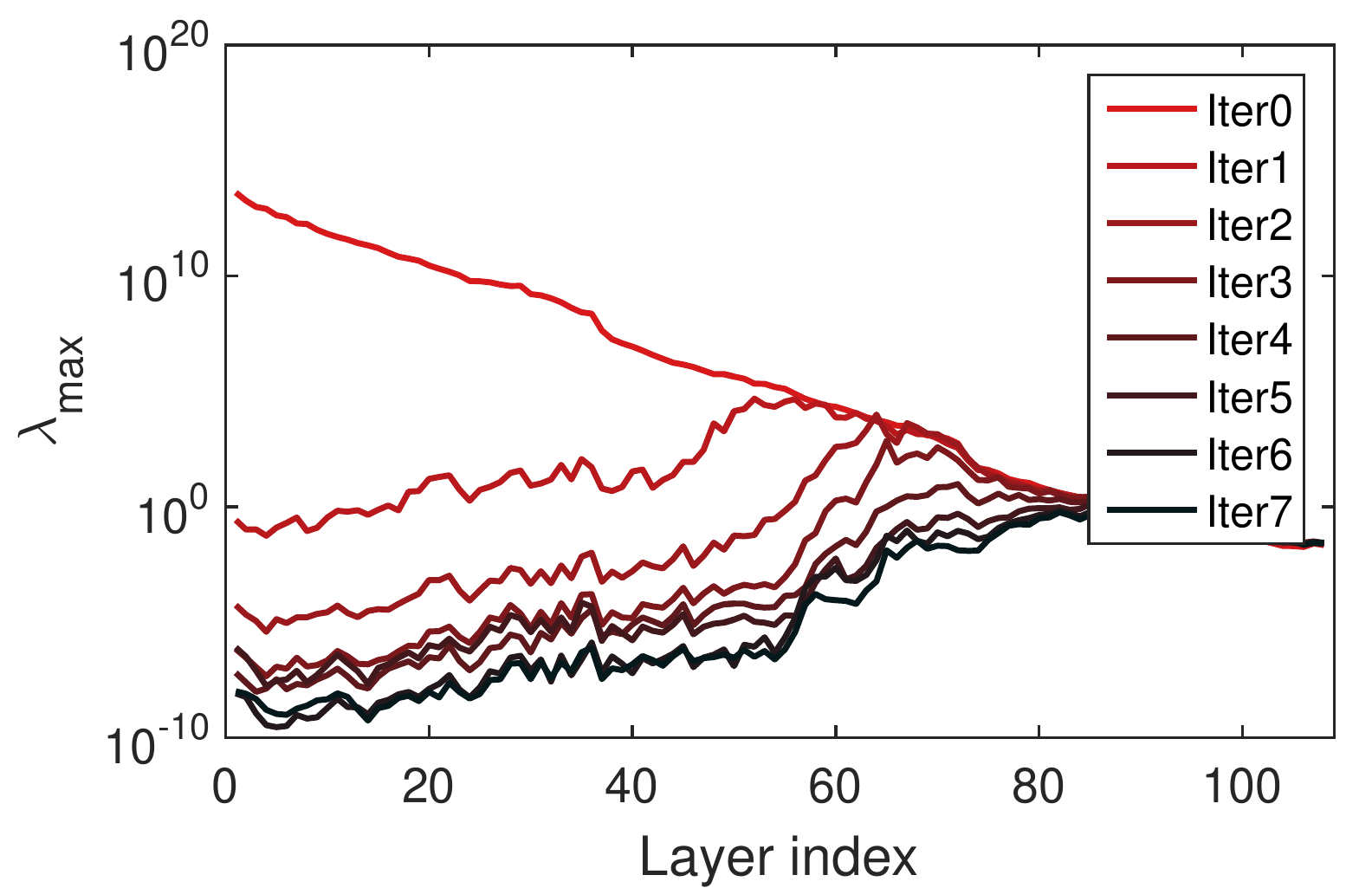}
		\end{minipage}
	}	
	\vspace{-0.15in}
	\caption{Experiments relating to gradient explosion of BN in deep networks without residual connections. We show the results of (a) a 100-layer MLP for MNIST classification and (b) a 110-layer VGG-style CNN for CIFAR-10 classification. }
	\label{fig:Explosion}
	\vspace{-0.15in}
\end{figure}

\begin{figure}[]
	\centering
	\subfigure[Training loss]{
		\begin{minipage}[c]{.3\linewidth}
			\centering
			\includegraphics[width=4.0cm]{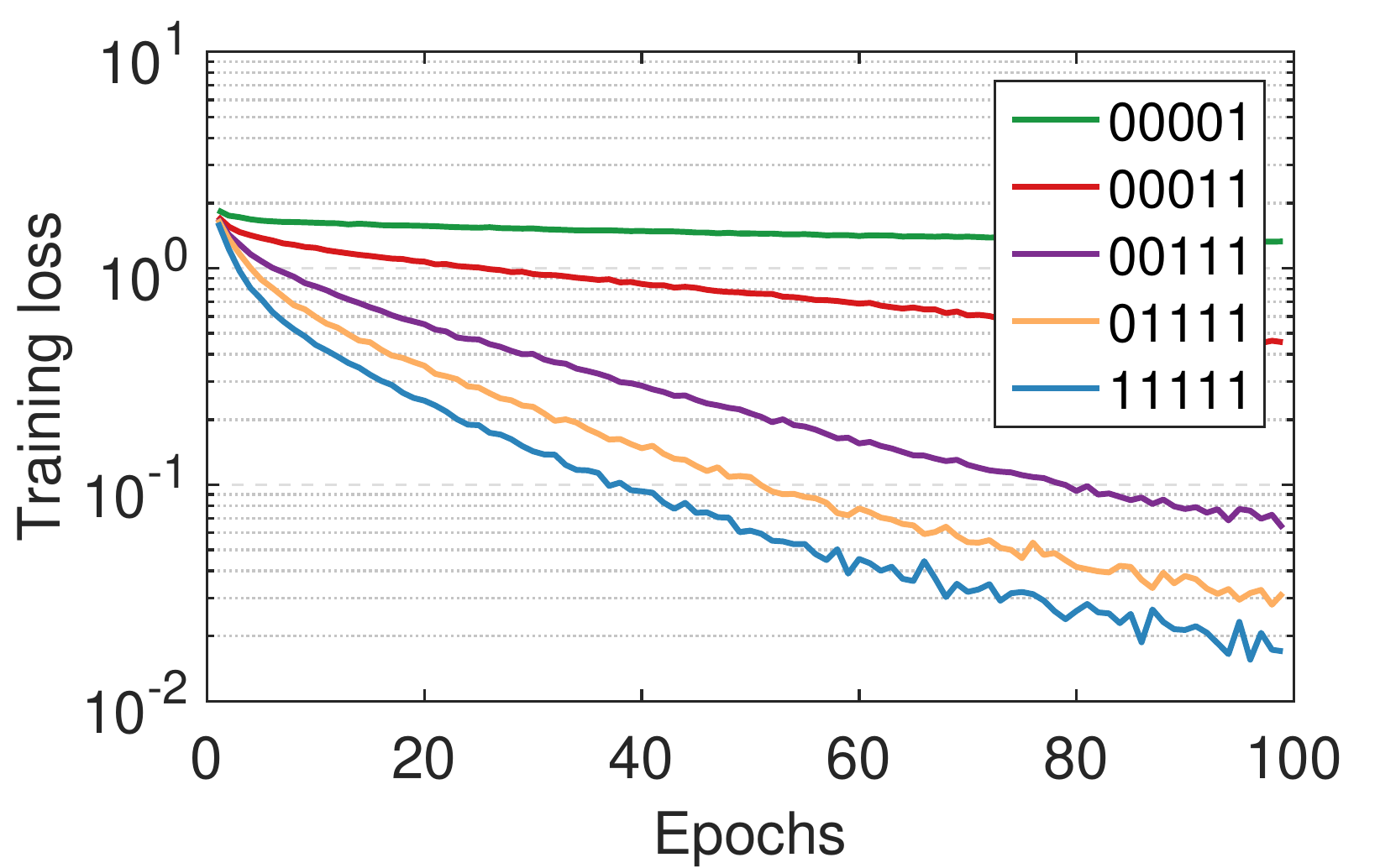}
		\end{minipage}
	}
	\subfigure[Training error]{
		\begin{minipage}[c]{.3\linewidth}
			\centering
			\includegraphics[width=4.0cm]{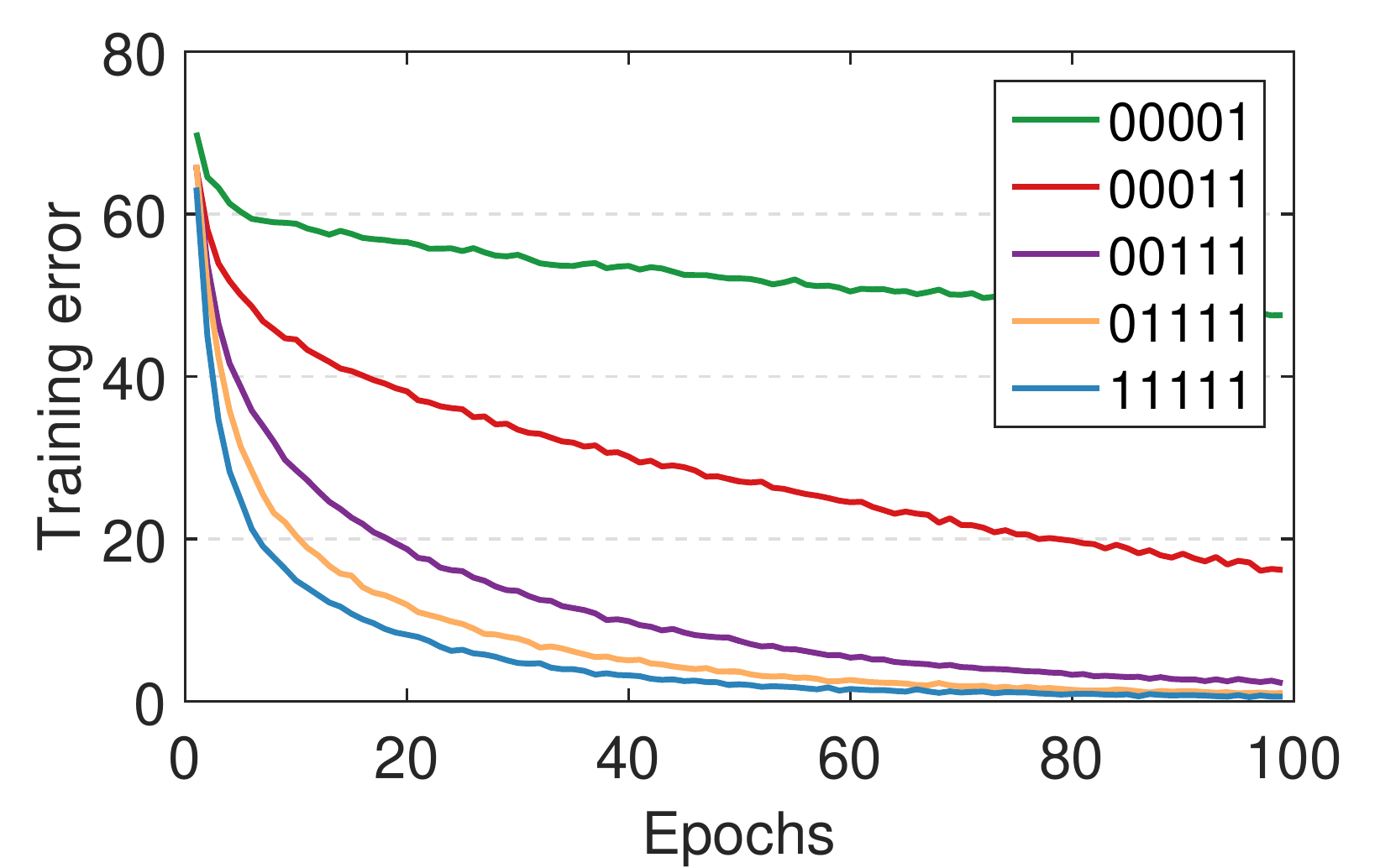}
		\end{minipage}
	}	
	\subfigure[Test error]{
		\begin{minipage}[c]{.3\linewidth}
			\centering
			\includegraphics[width=4.0cm]{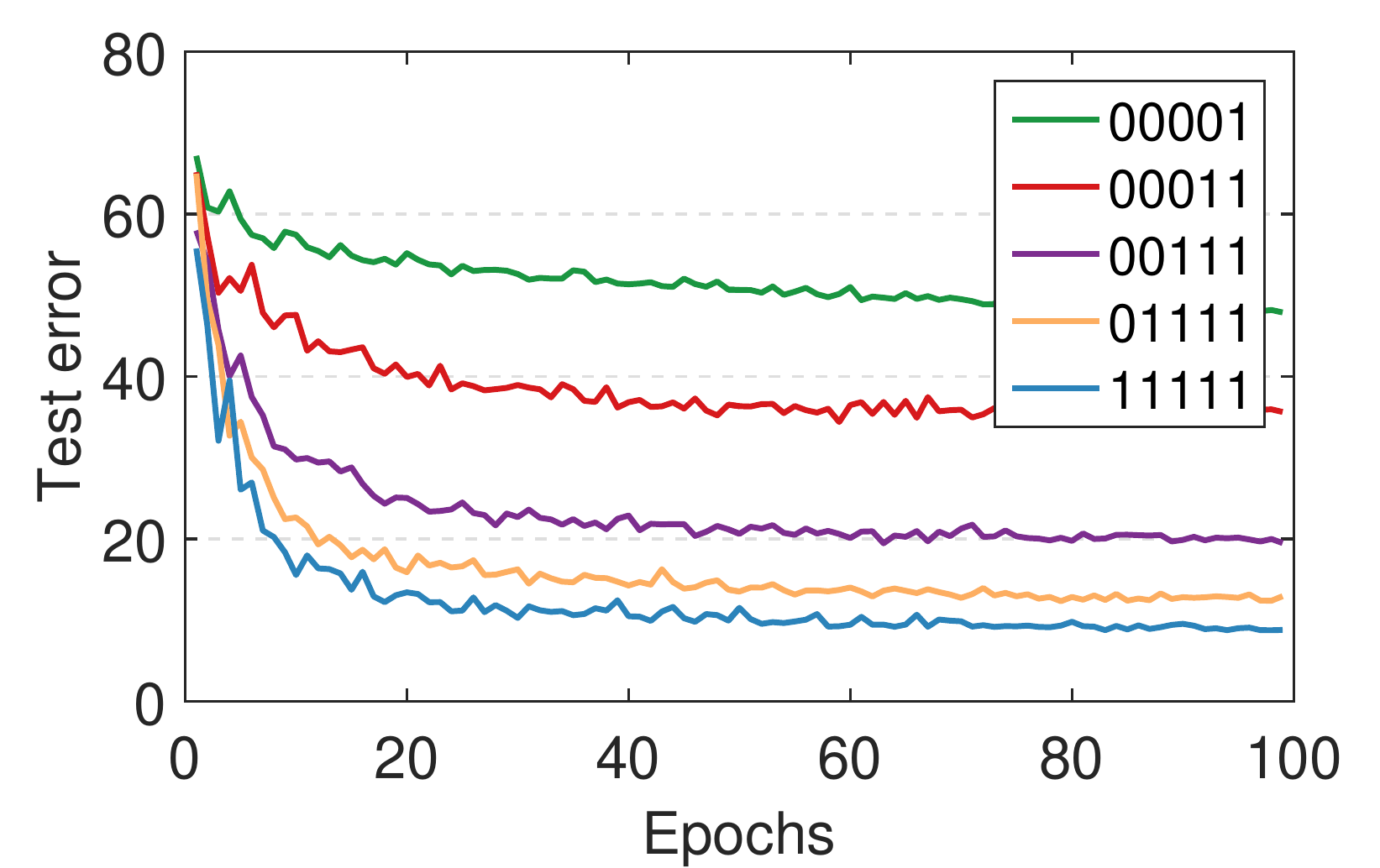}
		\end{minipage}
	}	
	\vspace{-0.15in}
	\caption{Exploring the effectiveness of  weight domination on a 16-layer VGG network with BN for CIFAR-10 classification.  We simulate  weight domination
		in a specific layer by blocking its weight updates. We denote
		`0' in the legend as the state of weight dominant (the first digit represents the first three consecutive convolutional layers). }
	\label{fig:Control-CNN}
\end{figure}
\begin{figure}[]
	\centering
	\begin{minipage}[c]{.96\linewidth}
		\centering
		\includegraphics[width=8.0cm]{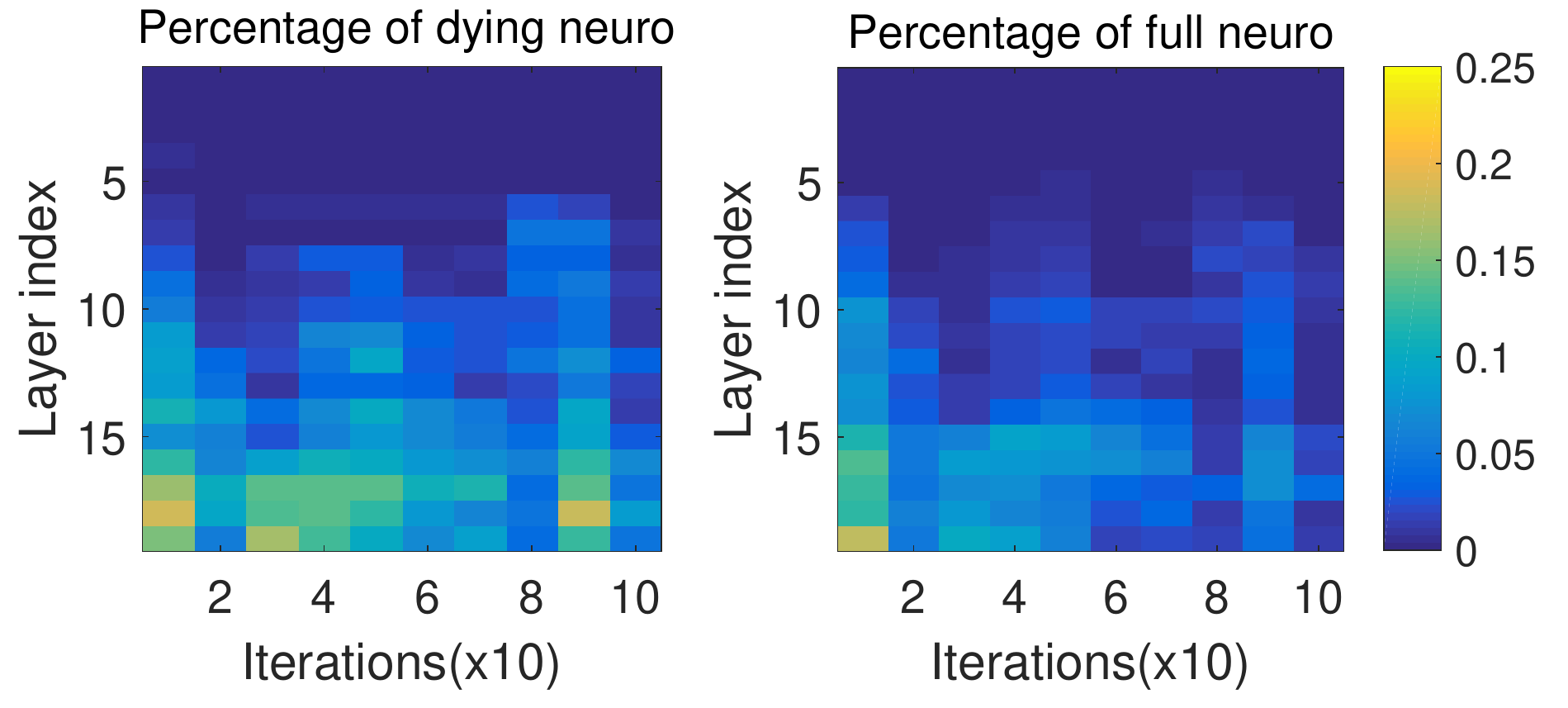}
	\end{minipage}
	\vspace{-0.14in}
	\caption{Dying and full neurons during training. The experiments are performed on a 20-layer MLP with 256 neurons in each layer, for MNIST classification. We show the results corresponding to the He-initialization \cite{2015_ICCV_He}.}
	\label{fig:dyingNode}
	\vspace{-0.2in}
\end{figure}
	\vspace{-0.1in}
\subsubsection{Gradient Explosion of BN}
	\vspace{-0.1in}
In Section~\ref{Sec:BN_stablization} of the  paper, we mention that, even for the network with BN, there is still the possibility that the  magnitude of the weight in certain layers is significantly increased.
Here, we provide the experimental results.

We conduct experiments on a 100-layer batch normalized MLP with 256 neurons in each layer for MNIST classification. We calculate the maximum eigenvalues of the sub-FIMs, and provide the results for the first seven iterations in Figure~\ref{fig:Explosion} (a). We observe that the weight-gradient has exponential
explosion at initialization (`Iter0'). After a single step, the first-step gradients dominate the
weights due to gradient explosion in lower layers, hence the exponential growth in the magnitude of the weight. This increased magnitude of weight leads to small weight gradients (`Iter1' to `Iter7'), which is caused by BN, as discussed in Section~\ref{Sec:BN_stablization} of the paper. Therefore, some layers (especially the lower layers) of the network enter the state of \emph{weight domination}.   We obtain similar observations on the 110-layer VGG-style network for CIFAR-10 classification, as shown in Figure ~\ref{fig:Explosion} (b).
	\vspace{-0.1in}
\subsubsection{Investigation of Weight Domination}
	\vspace{-0.1in}
	 Weight domination sometimes harms the learning of the network, because this state limits the representational ability  of the corresponding layer.  We conducted experiments on a five-layer MLP and provided the results in Section~\ref{Sec:BN_stablization} of the paper.
Here, we also conduct experiments on CNNs for CIFAR-10 datasets, shown in Figure ~\ref{fig:Control-CNN}.
We observe that the network with certain layers being in states of \textit{weight domination} can still decrease the loss, but with degenerated performance.

	\vspace{-0.1in}
\subsection{Experiments Relating to  Dying Neurons}
	\vspace{-0.1in}
\label{Sec-sup-dyingNode}
In Section~\ref{Sec-BN-Accelerate} of the  paper,  we mentioned that  `Plain' has dying/full neurons, and the number of dying/full neurons increases as the layer number increases. Figure~\ref{fig:dyingNode} shows the details of this phenomenon.

\begin{figure}[]
	\centering
	\subfigure[18-layer]{
		\begin{minipage}[c]{.40\linewidth}
			\centering
			\includegraphics[width=5.0cm]{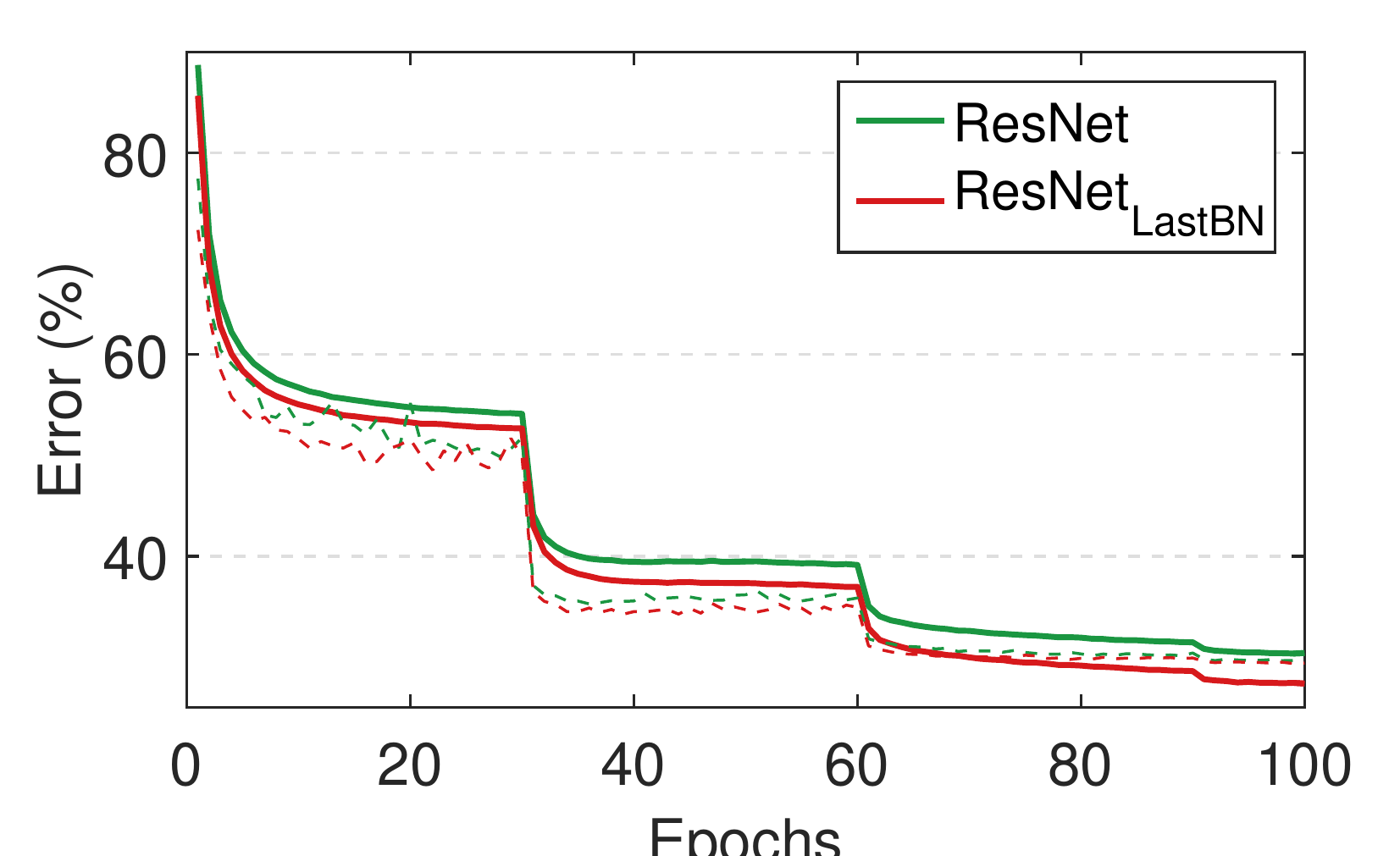}
		\end{minipage}
	}
	\subfigure[50-layer]{
		\begin{minipage}[c]{.40\linewidth}
			\centering
			\includegraphics[width=5.0cm]{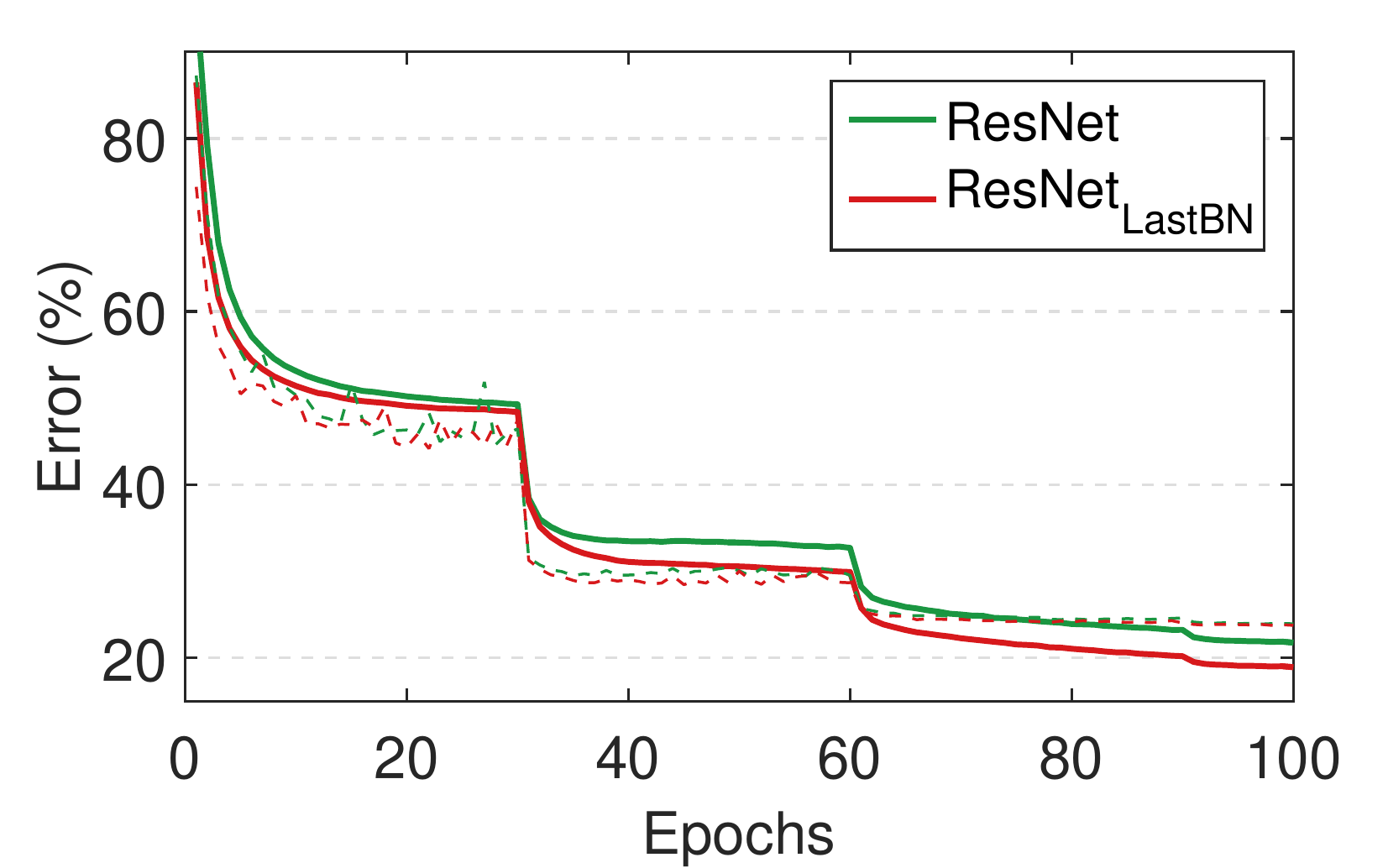}
		\end{minipage}
	}	
	\caption{Comparison of top-1 training errors (solid lines) and test errors (dashed lines) of  ResNet and  ResNet$_{LastBN}$ on ImageNet. }
	\label{fig:ImageNet}
\end{figure}

\section{More  Results for Deep Residual Network}

\subsection{Results on ImageNet classification}
\label{Sec-sup-ExpRes}
As mentioned in Section~\ref{Sec:veryDeep} of the  paper, we validate the effectiveness of ResNet$_{LastBN}$ on the large-scale ImageNet classification, with 1000 classes \cite{2009_ImageNet}. Here, we provide the details.

We use the official 1.28M training images as the training set, and evaluate the top-1 classification errors on the validation set with 50k images. We perform the experiments using the 18-layer and 50-layer networks. We follow the same setup as described in \cite{2015_CVPR_He}, except that 1) we train over 100 epochs with an extra lowered learning rate at the 90th epoch; 2) we use one GPU for the 18-layer network and four GPUs for 50-layer network.

Figures \ref{fig:ImageNet} (a) and (b) show  the training results for the 18-layer and 50-layer residual networks, respectively.
We find ResNet$_{LastBN}$ has a better optimization efficiency than ResNet, at both depths. Table \ref{Table:ImageNet} shows the validation errors. ResNet$_{LastBN}$ has a slightly improved  performance, under the standard hyper-parameters configuration.  Due to the improved optimization efficiency of ResNet$_{LastBN}$,
the advantage can be further amplified if we add the magnitude of the regularization, \eg, when we  use a weight decay (WD) of 0.0002 or add a dropout (DR) of 0.3, ResNet$_{LastBN}$ achieves better performance.

\begin{table}[]
	\caption{Comparison of top-1 validation errors ($\%$, single model and
		single-crop) on the 18- and 50-layer residual networks
		for ImageNet classification.}
	\label{Table:ImageNet}
	\centering
	\begin{small}
		\begin{tabular}{p{0.8in} |p{0.61in}<{\centering} p{0.63in}<{\centering} p{0.61in}<{\centering}|p{0.61in}<{\centering} p{0.63in}<{\centering} p{0.61in}<{\centering}}
			\toprule
			& \multicolumn{3}{c|}{depth-18} &    \multicolumn{3}{c}{depth-50}   \\
			Method   & standard    & WD=0.0002  & DR=0.3   &  standard    & WD=0.0002  & DR=0.3   \\
			\hline
			ResNet    & 29.78 &  30.07   & 30.62 & 23.97 & 24.37 & 23.81  \\
			ResNet$_{LastBN}$  & \textbf{29.38} & \textbf{28.96} & \textbf{29.32} & \textbf{ 23.76}  & \textbf{23.49}  & \textbf{23.47}\\
			\toprule
		\end{tabular}
	\end{small}
	\vspace{-0.12in}
\end{table}

\subsection{More Results of Putting a BN Layer before the Last Linear Layer}
\revise{
	We also observe that the method of putting a BN layer before the last linear layer is useful in other architectures, in which the last linear layer’s input $\x{}$ has significantly varying magnitude (indicated by $\lambda_{\Sigma_{\x{}}}$) during training (\eg, the ResNets case).
	We conduct experiments on the VGG-style networks (without BN) and ResNets (without BN) for CIFAR-10 classification. The setup is the same as in Section~\ref{Sec-sup-ExpBNSGD}.   We vary the depth ranging in 20, 44, 56, 110. We observe that for all depths, putting a BN layer before the last linear layer improves the performance (Table~\ref{Tabe:Sup-C10}).   
}

\begin{table*}[t]
	\caption{Experiments of putting a BN layer before the last linear layer  on the VGG-style networks (without BN) and ResNets (without BN) for CIFAR-10 classification. We report the test error ($\%$).}
	\label{Tabe:Sup-C10}
	\centering
	\begin{tabular}{p{1in} p{0.8in}<{\centering}  p{0.8in}<{\centering}  p{0.8in}<{\centering} p{0.8in}<{\centering} p{1in}<{\centering}}
		\toprule
		method     & depth-20     & depth-44  & depth-56 & depth-110 \\
		\midrule
		VGG     &  14.88 $\pm$  0.47  & 33.90 $\pm$  7.1 & 89.93 $\pm$  0.12 & 90 $\pm$ 0  \\
		VGG$_{LastBN}$    &  \textbf{12.89 $\pm$  0.20}  & \textbf{18.40 $\pm$ 0.90 } & \textbf{23.40 $\pm$  1.74} & \textbf{73.99 $\pm$ 16}  \\
		\midrule
		ResNet  &  11.21 $\pm$ 0.14  & 9.90 $\pm$ 0.31 & 9.48 $\pm$ 0.07 & 8.93 $\pm$ 0.22  \\
		ResNet$_{LastBN}$  &  \textbf{10.31 $\pm$ 0.25} &  \textbf{8.99 $\pm$ 0.25} & \textbf{8.61 $\pm$ 0.05}&  \textbf{8.37 $\pm$ 0.07}  \\
		\toprule[1pt]
	\end{tabular}
	\vspace{-0.14in}
\end{table*}

\end{document}